%% file: main.tex
\documentclass[11pt, a4paper, logo, onecolumn,  copyright,]{googledeepmind}

\usepackage[authoryear, sort&compress, round]{natbib}
\usepackage[T1]{fontenc}
\usepackage[utf8]{inputenc}
\usepackage{todonotes}
\usepackage{algorithm}
\usepackage{algpseudocode}
\usepackage{graphicx}
\usepackage{wrapfig}
\usepackage{subcaption}
\usepackage{geometry}
\usepackage{xcolor}
\usepackage{fancybox}
\usepackage{amssymb} 
\usepackage{xspace}
\usepackage{listings}      
\usepackage{hyperref}

\usepackage[most,skins,theorems]{tcolorbox}

\usepackage{graphicx}
\usepackage{float}
\usepackage{wrapfig}

\hypersetup{
  colorlinks=true,
  linkcolor=magenta,
  urlcolor=magenta,
  citecolor=magenta
}

\usepackage{tikz}
\usetikzlibrary{positioning,arrows.meta}
\usepackage{pgfplots}
\pgfplotsset{compat=1.18} 
\usepackage{pgfplotstable}
\usepackage{tabularx}
\usepackage{subcaption}
\usepackage{enumitem}
\usepackage{cleveref}
\usepackage[font=footnotesize]{caption}
\usepackage{tcolorbox} 
\usepackage{placeins}  

\tcbuselibrary{breakable, listings}

\newtcolorbox{promptbox}[1]{
  enhanced,
  breakable,
  colback=gray!5,
  colframe=black!70,
  fonttitle=\bfseries,
  title=#1,
  boxrule=0.8pt,
  arc=2pt,
  left=6pt,
  right=6pt,
  top=6pt,
  bottom=6pt,
  listing only,
  listing options={
    basicstyle=\footnotesize\ttfamily,
    breaklines=true,
    columns=fullflexible
  }
}

\definecolor{userbg}{RGB}{240, 240, 240}
\definecolor{modelbg}{RGB}{255, 255, 255}
\definecolor{successcolor}{RGB}{25, 150, 25}
\definecolor{failcolor}{RGB}{200, 50, 50}
\usepackage{xcolor}
\usepackage{caption}
\usepackage{listings}

\newtcolorbox[auto counter, number within=section]{definitionbox}[2][]{
colback=blue!0!white,
colframe=blue!30!white,
fonttitle=\bfseries,
title=Definition~\thetcbcounter: #2,#1}

\usepackage{parskip}
\usepackage{xcolor}

\definecolor{humangray}{RGB}{90,90,90}
\definecolor{humanfill}{RGB}{235,235,235}
\definecolor{humanamber}{RGB}{140,75,10}
\definecolor{humanamberfill}{RGB}{255,243,220}
\definecolor{claudered}{RGB}{190,30,30}
\definecolor{claudefill}{RGB}{253,232,232}
\definecolor{gptgreen}{RGB}{20,110,55}
\definecolor{gptfill}{RGB}{228,248,234}
\definecolor{geminiblue}{RGB}{25,70,160}
\definecolor{geminifill}{RGB}{228,238,255}
 
\tcbset{
  basebox/.style={
    enhanced, breakable,
    coltitle=white, fonttitle=\bfseries\small,
    fontupper=\small\setlength{\parskip}{4pt},
    arc=3pt, boxrule=1.2pt,
    left=7pt, right=7pt, top=5pt, bottom=5pt,
  },
  essaybox/.style={   basebox, colback=humanfill,       colframe=humangray   },
  humanbox/.style={   basebox, colback=humanamberfill,  colframe=humanamber  },
  claudebox/.style={  basebox, colback=claudefill,      colframe=claudered   },
  gptbox/.style={     basebox, colback=gptfill,         colframe=gptgreen    },
  geminibox/.style={  basebox, colback=geminifill,      colframe=geminiblue  },
}

\tcbset{
  llmchat/.style={
    enhanced, breakable,
    colback=white, colframe=black!40,
    fonttitle=\bfseries\small,
    arc=3pt, boxrule=1pt,
    left=7pt, right=7pt, top=5pt, bottom=5pt,
  }
}

\tcbset{
  llmchat/.style={
    enhanced, breakable,
    colback=white, colframe=black!40,
    fonttitle=\bfseries\small,
    arc=3pt, boxrule=1pt,
    left=7pt, right=7pt, top=5pt, bottom=5pt,
  },
  humanchat/.style={
    enhanced, breakable,
    colback=gray!10, colframe=gray!60,
    fonttitle=\bfseries\small,
    arc=3pt, boxrule=1pt,
    left=7pt, right=7pt, top=5pt, bottom=5pt,
  }
}

\newcommand{\convsep}{\vspace{4pt}\noindent\rule{\linewidth}{0.4pt}\vspace{4pt}}

\title{How LLMs Distort Our Written Language}

\renewcommand{\today}{2026-03-19}

\author[1,*]{Marwa Abdulhai}
\author[2,*]{Isadora White}
\author[3]{Yanming Wan}
\author[4]{Ibrahim Qureshi}
\author[5]{Joel Z. Leibo}
\author[3,5]{Max Kleiman-Weiner}
\author[3,5]{Natasha Jaques}

\affil[*]{Equal contributions}
\affil[1]{UC Berkeley}
\affil[2]{UC San Diego}
\affil[3]{University of Washington}
\affil[4]{Zaytuna College}
\affil[5]{Google DeepMind}

\definecolor{darkpurple}{RGB}{102, 51, 153}
\definecolor{lightpurple}{RGB}{204, 153, 255}
\definecolor{lightblue}{rgb}{0.22,0.45,0.70}
\definecolor{forestgreen}{rgb}{0.24,0.50,0.19}

\definecolor{DeepTeal}{RGB}{20, 110, 110}  
\definecolor{LightTeal}{RGB}{235, 245, 245} 

\newcolumntype{L}[1]{>{\raggedright\arraybackslash}p{#1}}

\usepackage{xcolor}
\newif\ifshowcomments
\showcommentstrue 

\tcbset{
  aibox/.style={
    width=\linewidth,
    top=8pt,
    bottom=4pt,
    colback=LightTeal,
    colframe=DeepTeal,
    colbacktitle=DeepTeal,
    fonttitle=\bfseries,
    enhanced,
    center,
    attach boxed title to top left={yshift=-0.1in,xshift=0.15in},
    boxed title style={boxrule=0pt,colframe=white,},
  }
}
\newtcolorbox{AIbox}[2][]{aibox,title=#2,#1}

\begin{abstract}
\normalfont
Large language models (LLMs) are used by over a billion people globally, most often to assist with writing. In this work, we demonstrate that LLMs not only alter the voice and tone of human writing but also consistently alter the intended meaning.
First, we conduct a human user study to understand how people actually interact with LLMs when using them for writing. 
Our findings reveal that extensive LLM use led to a nearly 70\% increase in essays that remained neutral in answering the topic question. Significantly more heavy LLM users reported that the writing was less creative and not in their voice. 
Next, using a dataset of human-written essays that was collected in 2021 before the widespread release of LLMs, we study how asking an LLM to revise the essay based on the human-written feedback in the dataset induces large changes in the resulting content and meaning. 
We find that even when LLMs are prompted with expert feedback and asked to only make grammar edits, they still change the text in a way that significantly alters its semantic meaning.
We then examine LLM-generated text in the wild, specifically focusing on the 21\% of AI-generated scientific peer reviews at a recent top AI conference. 
We find that LLM-generated reviews place significantly less weight on clarity and significance of the research, and assign scores that, on average, are a full point higher.
These findings highlight a misalignment between the perceived benefit of AI use and an implicit, consistent effect on the semantics of human writing, motivating future work on how widespread AI writing will affect our cultural and scientific institutions.
\end{abstract}

\begin{document}

\renewcommand{\copyrightext}{%
  \copyrightfont \textcopyright\, \the\year{} Google DeepMind. All rights reserved. \\
  \indent Code: \url{https://github.com/abdulhaim/llm_writing_distortion}. \\ 
  \indent Project page: \url{https://sites.google.com/view/llmwritingdistortion/home}.%
}

\maketitle
\begin{figure}[H]
    \vspace{-0.3cm}
    \centering
    \includegraphics[width=0.85\textwidth]{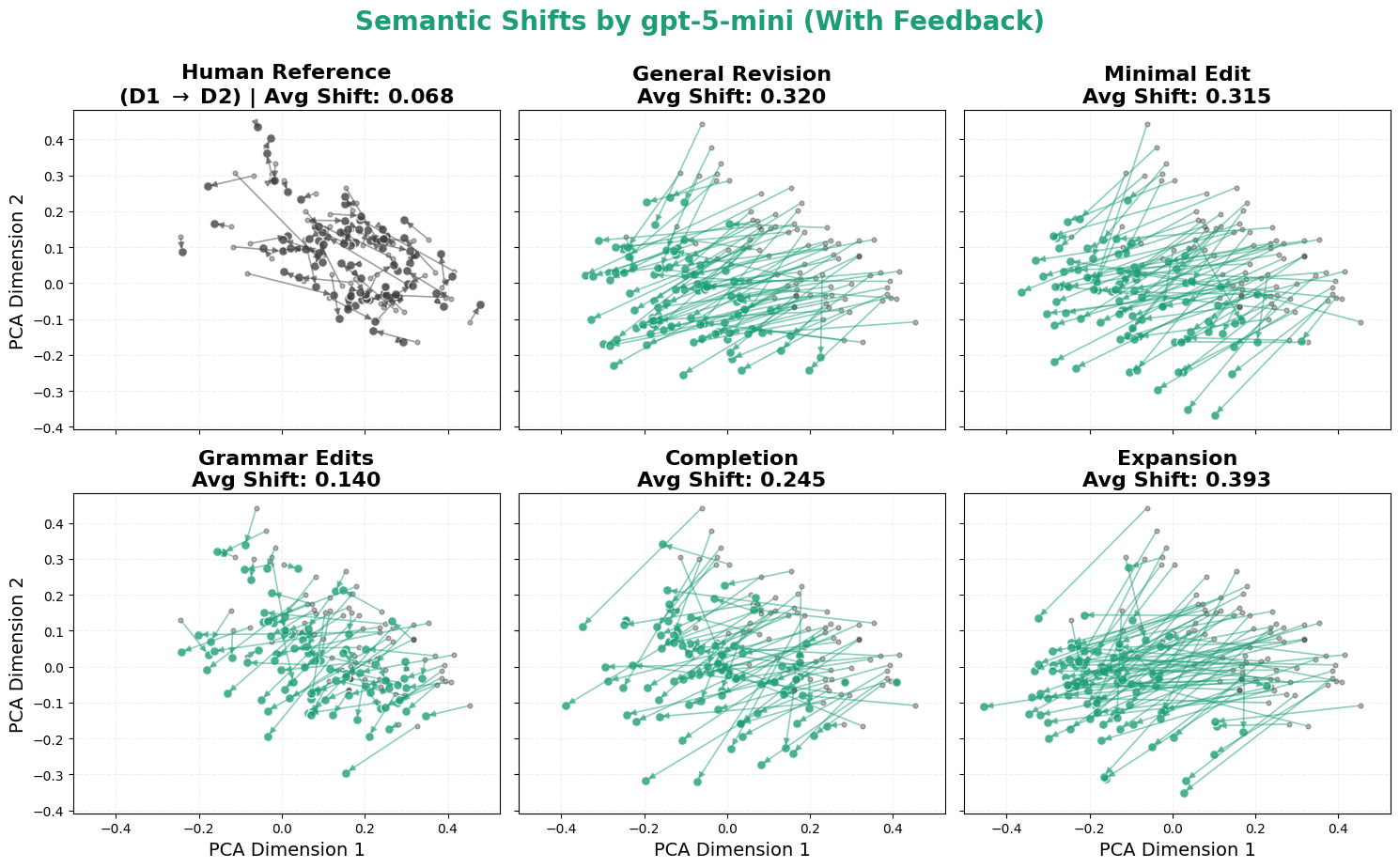}
    \caption{\textbf{LLM-generated revisions display a larger and more consistent semantic shift than human-written revisions of the same essays.} 
    Each point pair represents an essay from the ArgRewrite-v2 dataset before (D1) and after (D2) revision, embedded using \texttt{MiniLM-L6-v2} \citep{reimers2019sentencebertsentenceembeddingsusing} sentence embeddings and projected into two dimensions via PCA, a common approach for analyzing semantic differences \citep{dhillon2015eigenwords}. The top-left panel (grey) shows human revisions, while the remaining panels show revisions produced by prompting the LLM with human-written expert feedback and different edit instructions (see titles). Even the instruction to make minimal edits shows large shifts (top-right). Arrows indicate direction and magnitude of semantic change. Human revisions exhibit smaller and more diverse shifts, whereas LLM revisions produce large semantic shifts strongly aligned in a common direction, and to a region of space not previously occupied by any human-written essay.}
    \label{fig:gpt-5-mini-with-feedback-gemini-004}
\end{figure}

\input{sections/1-introduction}

\input{sections/2-related-work}
\input{sections/3-method}

\input{sections/4-results}
\input{sections/5-discussion}
\newpage
\input{sections/ethics}
\newpage
\bibliography{main}
\input{sections/6-appendix}

\end{document}

%% file: sections/1-introduction.tex
\section{Introduction}

Over 1 billion people use Large Language Models (LLMs) weekly \citep{datareportal2026, altman2025ted}, with significant traffic devoted to users who request help with writing ~\citep{chatterji2025how}, such as revising emails, drafting professional documents \citep{liang2024mapping, sanz-tejeda2026genaiwriting}, or generating ideas  \citep{doshi2024creativity}. LLM-generated text has become pervasive, permeating political and scientific institutions; for example, LLMs were used to generate parliamentary speeches in the UK \citep{independent2025mpai}, and 21\% of the peer reviews at a recent major academic conference \citep{pangram2025iclrreviews}. When LLM-generated text pervades our written discourse, what biases will this introduce, and how it shape our written language and cumulative culture?

Recent literature highlights a concerning trend toward linguistic homogenization. Across varying architectures and providers, LLMs tend to produce eerily similar responses to open-ended prompts \citep{jiang2025artificialhivemindopenendedhomogeneity}. When such models are used as writing assistants, this ``algorithmic mono voice'' may be amplified through repeated exposure to model-generated suggestions \citep{lockin2025stagnation}. Indeed, recent work shows that people are beginning to adopt LLM-like linguistic patterns even in spoken communication \cite{yakura2024empirical}.

However, the impact of LLMs likely extends beyond stylistic flattening. While previous work has documented the lack of linguistic variety in LLM-generated text \citep{guo2025benchmarking}, the effect on intended meaning remains underexplored. Many users recognize the ``feel'' of LLM prose, but fewer realize how LLM use shapes their underlying opinions and conclusions. The downstream consequences for political discourse, literature, and scientific institutions are as yet unknown.


In this paper, we study the impact LLMs have on the meaning of human writing. To understand how people use LLMs for writing, we first conduct a randomized controlled trial (N=100) where we task participants with writing an essay, where in one condition they have access to an LLM, and in another condition they do not. We 
analyzed stylistic differences among essays written solely by humans, essays written with LLM assistance, and essays with text mostly generated by an LLM. Our results show that extensive AI use results in a 70\% change in the argumentative stance of essays, from for/against to neutral. Participants who heavily rely on LLMs also perceive a significant loss of creativity and adherence to their own written voice.



While it is concerning that users who allow LLMs to write text for them experience a loss of voice and take a different argumentative stance than they might have otherwise, perhaps there are more responsible ways to use LLMs that are not subject to these effects. What if the person first completes the piece of writing, and then merely asks the LLM 
to edit the text based on carefully constructed feedback? To answer this question, we perform a large-scale quantitative analysis of how LLMs edit human-written essays in response to expert feedback across prompts and models. 
Specifically, we leverage a publicly available dataset of 86 human-written essays, expert feedback, and the resulting human-revised drafts \citep{Kashefi_2022}, which was collected in 2021 before the release of ChatGPT \citep{openai2022chatgpt}. Using this dataset, we perform a counterfactual analysis, comparing human edits to edits produced by three commonly used LLMs. We measure differences between human-edited and LLM-edited essays along dimensions of semantics, lexical usage, part-of-speech distributions, emotional tone, and stylistic features. Our results show that using LLMs for editing leads to a large shift away from both the initial human-written drafts and the counterfactual edits that humans would have made to the same essay. As shown in \Cref{fig:gpt-5-mini-with-feedback-gemini-004}, when humans revise their own writing, their edits result in changes of much smaller magnitude and in diverse directions in a semantic embedding space. This matches copyediting norms of performing targeted, low-magnitude edits that preserve the author's original voice \citep{einsohn2019copyeditors}.

In contrast, when LLMs are prompted to edit the human writing, they globally change the writing style and the argument. The edits shift the essays in the same direction regardless of model, resulting in essays that are not only less diverse but also occupy a region of embedding space where no previous human-written essay exists. This pattern is consistent with prior findings that LLMs encourage convergence towards a shared semantic style rather than preserving individual writing preferences \citep{artificialhivemind2025open}. In further analyzing the ways in which LLMs alter writing, we find that LLMs use both more argumentative and analytical language, as well as more emotional language---roughly doubling the use of both positive and negative sentiment. %


Our analysis shows that LLM editing and co-writing significantly change the underlying content and meaning expressed in the original essays. While these shifts are significant in educational contexts, they raise the question of how these effects translate in higher-stake environments where experts make decisions. Perhaps people who use LLMs in professional settings are careful to mitigate these effects. To test this hypothesis, we analyze LLM-generated text in the wild, within an existing scientific institution. Specifically, we focus on peer reviews from the International Conference on Learning Representations (ICLR), a leading machine learning venue, in the year 2026,  where a recent analysis revealed that 21\% of ICLR 2026 reviews were generated or heavily edited by LLMs \citep{pangram2025iclrreviews}. 
We summarize strengths and weaknesses in LLM-generated versus human-written reviews, and find that using LLMs for writing academic peer reviews does not just reduce the diversity of responses and change the resulting average scores. Rather, LLMs have begun to change the very criteria that researchers use when evaluating peer-reviewed scientific research, with LLM-peer reviews focusing less on clarity, relevance, and impact, and more on reproducibility, scalability, and practical application. Thus, LLM use is already causing shifts in scientific and cultural institutions in subtle ways that are as of yet poorly understood. 

Our findings all point to the same conclusion: LLMs alter the underlying intended meaning, style, and voice of text. These shifts may subtly steer our writing and decision-making toward different conclusions than those originally intended. 
For AI researchers, this highlights a capability deficiency, in that LLMs are not able to assist with writing without altering meaning or reducing creative expression of the human voice. More broadly, this underscores the need for further scrutiny of how the already widespread use of LLMs may shift our cultural and institutional decision making in unpredictable or detrimental ways.

%% file: sections/2-related-work.tex
\section{Related Work}


\paragraph{LLM Homogenization.}
Large language models have been rapidly adopted by over a billion people worldwide, with the most common use case being writing, editing, and generating ideas \citep{chatterji2025how}.  However, co-writing with an LLM leads to writing that tends to converge stylistically across creative tasks, decision making, and open-ended text generation even when using AI models trained by different companies (different model families) \citep{basiceffect2025use,doshi2024creativity,agarwal2025homogenizing, artificialhivemind2025open, doi:10.1073/pnas.2504966122, sourati2026homogenizingeffectlargelanguage}. Recent work shows that diversity loss may arise from feedback loops between LLM models and training with paradigms such as RLHF \citep{Murthy_2025}. For example, algorithmic systems can create cultural ``lock-in,'' where the outputs of generative models reinforce their own priors over time \citep{lockin2025stagnation}. These works indicate that homogenization is not a surface-level phenomenon, but a structural property of current LLMs. Our work not only replicates these findings, but we also show that the transformed text has different lexical and emotional characteristics and actually argues for entirely different conclusions. 

\paragraph{The Effect of LLMs on Humans.}
Recent work has begun to examine how sustained interaction with LLMs affects human cognition, preferences, and decision-making. Although LLMs can improve productivity by assisting in generating ideas, structuring content, summarizing literature, and reducing drafting time \citep{noy2023experimental}, using LLMs for writing can also reduce creativity, with the resulting writing showing less diversity across groups \citep{sciencedaily2024aiCreativity, education_ai_article, doshi2023generative, meincke2025chatgpt, Anderson_2024, nordling2026ai}. Frequent reliance on AI assistance may carry cognitive costs; a recent longitudinal study demonstrated neurological changes and reduced cognitive activity in relation to AI use \citep{kosmyna2025your}. 
AI usage among students is a function of pre-existing writing confidence and skills \citep{picton2025genai_literacy}, with students who enjoyed writing before using AI tending to augment AI outputs with their own thinking, whereas those who struggled with writing are more likely to adopt AI-generated text wholesale. In everyday communication tools, LLMs increasingly shape how people write as much as what they decide. Empirical studies show that AI-mediated communication alters emotional tone, partner perception, and decision-making \citep{Hohenstein2021AIMediated, sabour2025humandecisionmakingsusceptibleaidriven}, diminishes authorship and agency in professional settings \citep{Wenker2023SmartReplyAgency}, and, in cross-cultural contexts, homogenizes prose toward Western stylistic norms \citep{agarwal2025homogenizing}. These findings suggest that the large-scale deployment of AI writing assistants may gradually stabilize and narrow the range of communicative styles that people use. Our work is the first to perform randomized controlled trials to show that not only do LLMs homogenize our writing, but they also have the ability to influence the views and judgments that people express.

\paragraph{The Effect of AI on Institutions.} As LLM usage becomes embedded in institutions, there are growing concerns about the technology's impact on collective judgment and evaluation. Reports show that academic peer reviews and scientific writing contain content heavily generated and/or edited by LLMs \citep{pangram2025iclrreviews, Liang2025QuantifyingLL}. A recent paper showed that LLM use in science presents a social dilemma: for individual researchers, it increases their output and citations by 3-5x, but globally it contracts the collective volume of scientific topics studied \citep{hao2024ai}. 
We go beyond these results to show that AI use in science can actually change the conclusions that scientists draw, and that LLM-generated peer reviews are already affecting the scientific criteria being applied to \textit{all} papers, not just those written with the help of AI. 
LLM-generated text is already permeating our scientific institutions \textit{en masse}, and our work contributes to the literature beginning to understand its effects at this early stage.

%% file: sections/3-method.tex
\begin{figure*}[t]
    \vspace{-1.5em}
    \centering
    \includegraphics[width=0.9\linewidth]{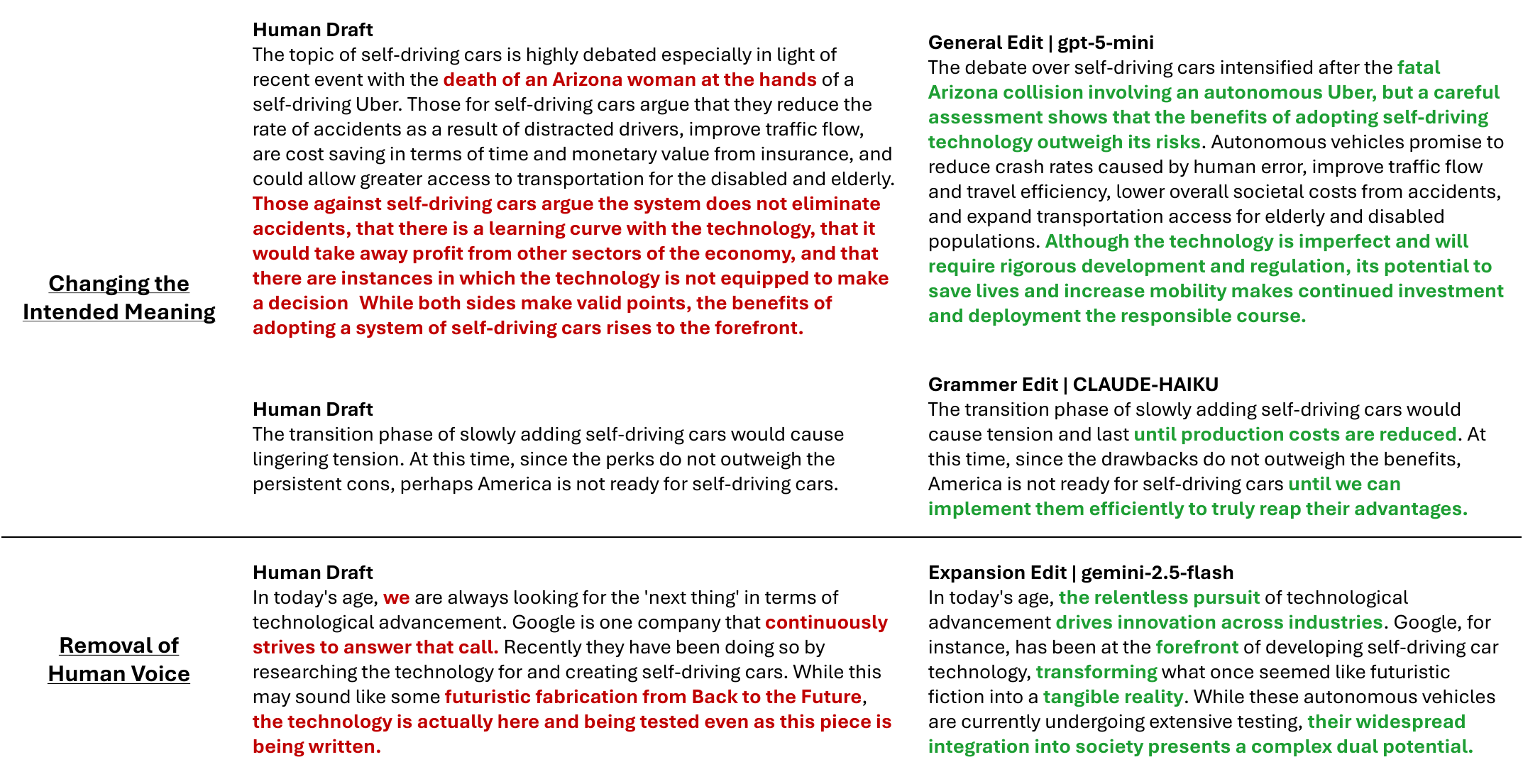}
    \vspace{-1em}
    \caption{\textbf{Example of substantial argumentative rewriting by an LLM on essays from ArgRewrite-v2.} Red text highlights segments removed or substantially reframed by the model, while green text highlights segments added to the human draft. The figure illustrates how LLM edits frequently alter the person's intended conclusions, removing content that makes a particular claim, and editing the essay to be more neutral or positive about the technology of self-driving cars. Further, LLM edits will often remove human colloquialisms, anecdotes, or examples, leading to repetitive writing that loses the person's voice.}
    \label{fig:llm_argument_rewrite_example}
\end{figure*}

\section{Experiment Methodology} \label{sec: methodology}


In this section, we describe the methodology used to collect and quantify semantic, lexical, grammatical, and affective differences between human-written text and LLM-written or LLM-edited text.

\subsection{Datasets.} \label{sec: datasets}

\paragraph{Human Study.}\label{method:human_study}
We conducted a human user study of 100 participants (demographic information in \Cref{appendix:demographics}) to understand how people leverage LLMs to write an argumentative essay. We chose the essay prompt,``Does money lead to happiness?'', since it is relatable to participants from all backgrounds. Participants were recruited from Prolific. Each participant was randomly assigned to one of two conditions: (1) the control group ($n=45$), where participants were not allowed to use LLMs, or (2) an AI-assisted condition ($n=55$), where participants were allowed to use an embedded LLM (gpt-4o-mini) in whatever way they chose while writing. For the AI-assisted group, we recorded the full interaction history with the LLM as well as the final draft written with LLM assistance. We recruit participants who are native English speakers and residing in the United States through Prolific to ensure that country or native language is not a confounding variable in our study. More information about recruitment and user study information can be found at \Cref{sec: user_study_details}.

To evaluate the impact of LLM intervention, each participant completed a pre-study and post-study questionnaire to measure shifts in attitudes toward LLMs, their own creativity, and the alignment of model outputs with their writing preferences. These self-reported ratings allow us to measure if human preferences are being met when collaborating with LLMs. We provide the pre-study and post-study questions in \Cref{appendix:survey}, and details on compensation, recruitment process, and duration of the study in \Cref{appendix:user_study_details}. All procedures were approved by our Institutional Review Board (IRB). 

Before running the full experiment, we first conducted an initial pilot study ($n=8$) to test the protocol. Analysis of this data revealed two distinct behaviors among participants in the experimental condition, with access to the LLM. Roughly half of the participants either abstained from LLM usage entirely or utilized the tool strictly for peripheral information-seeking or critiques rather than content generation, while the other half used the LLM extensively. As the purpose of the study is to understand how naturalistic interactions with LLMs alter writing, the goal is not to coerce the users into a specific LLM usage pattern. Therefore, for our primary study ($n=100$), we 
categorized into either the LLM-Influenced (minimal usage to generate text) or LLM (extensive generative use) conditions. This classification was determined a priori, before the analysis of our results, through a cross-referencing of self-reported usage claims against objective interaction transcripts. Concretely, a participant is categorized as LLM-influenced if they self-report to have generated less than 40\% of the text with an LLM, and when we cross-reference their essay with their conversation with the LLM, we find this to be an accurate statement. In the human user study presented in this paper ($n=100$), 28 out of 55 users were classified as LLM-Influenced, with 12 out of 55 refusing to use the LLM to generate their essay at all. We put examples of transcripts of LLM-Influenced and LLM conversations in \Cref{sec: user_study_examples}.

\textbf{ArgRewrite-v2.} 
ArgRewrite-v2 \citep{Kashefi_2022} is a dataset of argumentative writing revisions collected from 86 university students. Importantly, the essays were written in 2021 and the dataset was released in 2022 before the release of ChatGPT \citep{openai2022chatgpt}. Thus, the writing pre-dates the widespread adoption of LLM-based writing assistants. All participants in the study developed an initial argumentative essay draft (D1) for or against self-driving cars that could serve as an op-ed piece in a local newspaper. Each D1 draft was then provided with feedback from a human expert on how to improve the essay, which included both coarse-grained (surface vs content) and fine-grained (e.g., claim, evidence, reasoning, word usage, precision, etc.) feedback. In response, participants revised the initial draft to form the second draft (D2). To enable a comparative analysis, we then construct a dataset of LLM-generated D2 drafts. Specifically, we prompted three production LLMs from the top frontier AI labs (\texttt{gpt-5-mini}, \texttt{gemini-2.5-flash}, \texttt{claude-haiku}) to edit the original human-drafted D1 essays in the ArgRewrite-v2 dataset. The LLMs condition on both the D1 draft and the human-written feedback to produce the edits. We compare the LLM-generated D2 drafts with the human-written D2, to counterfactually analyze how much LLMs change the essays relative to what a human would do if they were not influenced by LLMs.

Because prompting an LLM is critical to its effective performance \citep{marvin2023prompt}, for this experiment we needed to provide realistic prompts that would reflect how people actually ask LLMs to revise their writing. 
Our user study showed that people employ LLMs for a variety of use cases: generating ideas and arguments, expanding their own ideas from existing human-written text, writing a first paragraph and asking the LLM to finish the rest, asking the LLM to review the human-written essay, and asking the LLM to write the entire essay. 
Based on this analysis, we prompted LLMs to perform five revision types: (1) \textit{general} revision to comprehensively improve the essay; (2) \textit{minimal edits} that make only necessary corrections; (3) \textit{grammar} revisions that fix surface errors without changing content; (4) \textit{completion} revisions to finish incomplete essays; and (5) \textit{expansion} to elaborate on existing ideas. 
For each revision type, we provided the LLM with the expert human-written feedback on the original D1 draft from the ArgRewrite-v2 dataset, which provides further instruction about how the essay should be revised, and helps to mitigate the LLM-induced homogenization (\Cref{fig:appendix-semantic-general} shows the magnitude of the effects with and without conditioning on expert revision instruction). 

\textbf{Peer Reviews.} To analyze how LLM use in the wild may shift the claims made in real communication, we analyze peer reviews from the International Conference on Learning Representations (ICLR) \citep{gonzalez2024learning}. 
In the ICLR review process, not only are reviewers prohibited from using AI to generate reviews, but their names and reviews are visible to other top scientists in their field who act as ``area chairs.'' Thus, to protect their professional reputation, reviewers have a strong incentive to both conceal their use of AI, and to create high-quality reviews. 
Of the 75,000 reviews written for ICLR 2026, over 21\% were found to be LLM-generated, with an additional 39\% of papers that used LLMs to edit or generate parts of the text \citep{pangram2025iclrreviews}. Detection was performed using the Pangram AI tool, a classifier with a low false positive rate \citep{emi2024technicalreportpangramaigenerated}\footnote{It is possible for this tool to have errors and to introduce correlations.}. 
Our analysis compares human-written peer reviews with those classified as `LLM-generated'. 
We asked whether LLM-generated reviews assign different scores, and also whether they actually focus on different scientific criteria. To determine this, we use an LLM-as-a-Judge classifier to identify the strengths and weaknesses of 18k reviews from ICLR 2026 from 9k papers. We selected only those papers with one review written entirely by a human and one written entirely by an LLM to ensure there is no bias in our sampling of human or LLM reviews. 


\begin{figure*}[t]
    \centering
    \includegraphics[width=\linewidth]{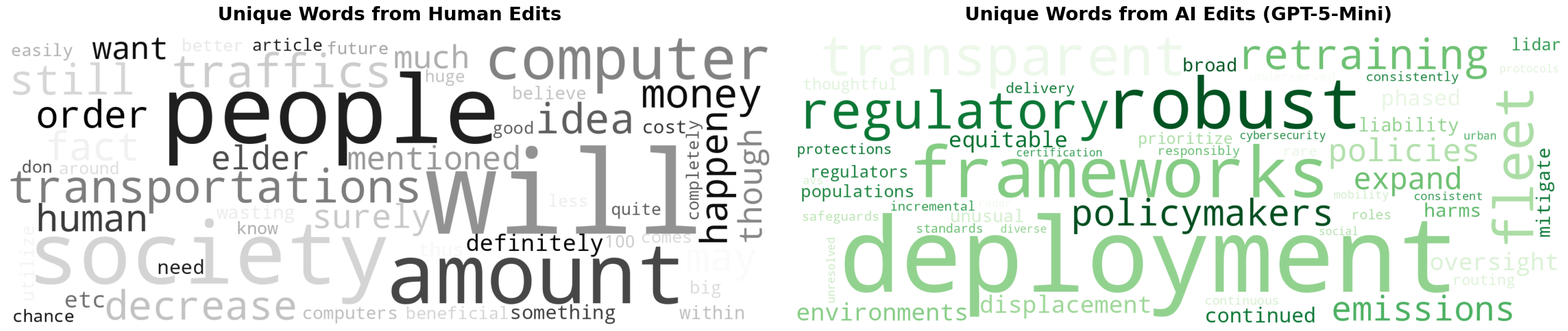}
    \caption{\textbf{Unique words in human-edited texts (left) versus AI-edited texts produced using \texttt{gpt-5-mini} (right) for the ArgRewrite-v2 analysis.} Word size reflects relative frequency, highlighting stylistic and thematic differences between how humans and AI write about self-driving cars.}
    \label{fig:unique-words-human-vs-ai}
\end{figure*}

\subsection{Metrics}

\textbf{PCA of Embedding Representations.} To capture changes in semantic meaning between human-edited and LLM-edited drafts, we project each draft into a high-dimensional vector embedding space using an encoder transformer model. 
To identify the primary axes of semantic variation across the dataset, we apply Principal Component Analysis (PCA) to the embedding representation. 
Distances in our PCA plot reflect similarity in meaning or semantic content between essays \citep{mikolov2013efficient, reimers2019sentence}. 
In our analysis, we analyze both the magnitude and direction of edits to each of the 86 different essays in ArgRewrite-v2.
Large magnitude edits, all pointing in a similar direction, indicate that the LLM is not merely correcting grammar, but is actively steering diverse human perspectives towards homogenization, toward a different conceptual mode. 

\textbf{Lexical Distribution Divergence.} To complement our semantic analysis, we also quantify the lexical difference between human-edited and LLM-edited writing. This enables us to understand which words appear, how often they appear, and whether some words are swapped for others. While embeddings provide one lens into high-level conceptual change, the embedding space is a learned representation of semantic meaning. A more data-driven metric is the divergence between the unigram (word count) distributions of the original human drafts and the LLM-generated revisions.
We represent each draft as a discrete probability distribution over the global vocabulary, where the probability of a token is proportional to its frequency in the text and quantify lexical shifts using the Jensen-Shannon Divergence (JSD) \citep{Menndez1997THEJD}. 
This approach allows us to quantify lexical change independently of sentence structure or syntax.
By calculating JSD over word counts, we can empirically determine the extent to which LLMs alter the lexical composition of human writing compared to LLM-edited writing. Higher JSD values indicate larger differences from human writing.

\textbf{Measuring change in emotional distribution.} To quantify shifts in the affective quality of revisions, we utilize the NRC (National Research Council - Canada) Emotion Lexicon \citep{mohammad2013nrc}. 
This resource is a list of English words and their associations with eight basic emotions: anger, anticipation, disgust, fear, joy, sadness, surprise, and trust, as well as two sentiments (positive and negative). 
By computing the density of these emotional markers in both human and AI-revised essays, we can detect whether LLMs systematically change the emotional tone of argumentative writing or introduce specific affective biases, such as an increase in positive sentiment, that were absent in the original human drafts.

\textbf{LIWC:} We also employ the Linguistic Inquiry and Word Count (LIWC) tool \citep{doi:10.1177/0261927X09351676, boyd2022liwc22} to measure the psychological, emotional, and cognitive characteristics of the language in human and AI-written text based on word usage. LIWC categorizes words into over 90 semantically and grammatically defined dimensions, including summary variables (e.g., analytical thinking, clout, and authenticity), grammatical categories (e.g., pronouns, prepositions), \& psychological processes (e.g., cognitive mechanisms, social processes). LIWC allows us to quantify how text edited or written by LLMs alters the meaning of an essay across different categories informed by previous research. For instance, we use the Analytical Thinking and Authenticity metrics \citep{boyd2022liwc22} to determine if LLM revisions shift a human user’s natural voice toward the formal, detached, and highly structured style commonly associated with generative models.

\textbf{Qualitative Analyses Using LLM-as-a-Judge} We use \texttt{gpt-4o} as an LLM-as-a-Judge, a common technique to automate qualitative analyses \citep{gu2024survey} to determine qualitative attributes for essays and ICLR reviews written by humans and AIs, respectively. 
For the human study, we use our LLM-as-a-Judge to determine (1) the extent to which the essay agreed or disagreed with the question ``Does money lead to happiness?'' and (2) the different argument styles used to support their decision. 
For the ICLR review analysis, we use LLM-as-a-Judge to first extract categories of strengths and weaknesses (such as `novelty') and then label each ICLR review with these selected strengths and weaknesses. Prompts for the LLM-as-a-Judge can be found in \Cref{sec:llm-as-a-judge}.


%% file: sections/4-results.tex
\section{Results}

Below, we group findings from the datasets described above to elucidate different ways in which LLMs influence human writing.  
Code to reproduce our experiments is available at {\small \textcolor{blue}{\url{https://github.com/abdulhaim/llm_writing_distortion}}} and project page at
{\small \textcolor{blue}{\url{https://sites.google.com/view/llmwritingdistortion/home}}}.

\begin{figure}[ht]
    \centering
    \begin{subfigure}[t]{0.48\textwidth}
        \centering
        \includegraphics[width=\linewidth]{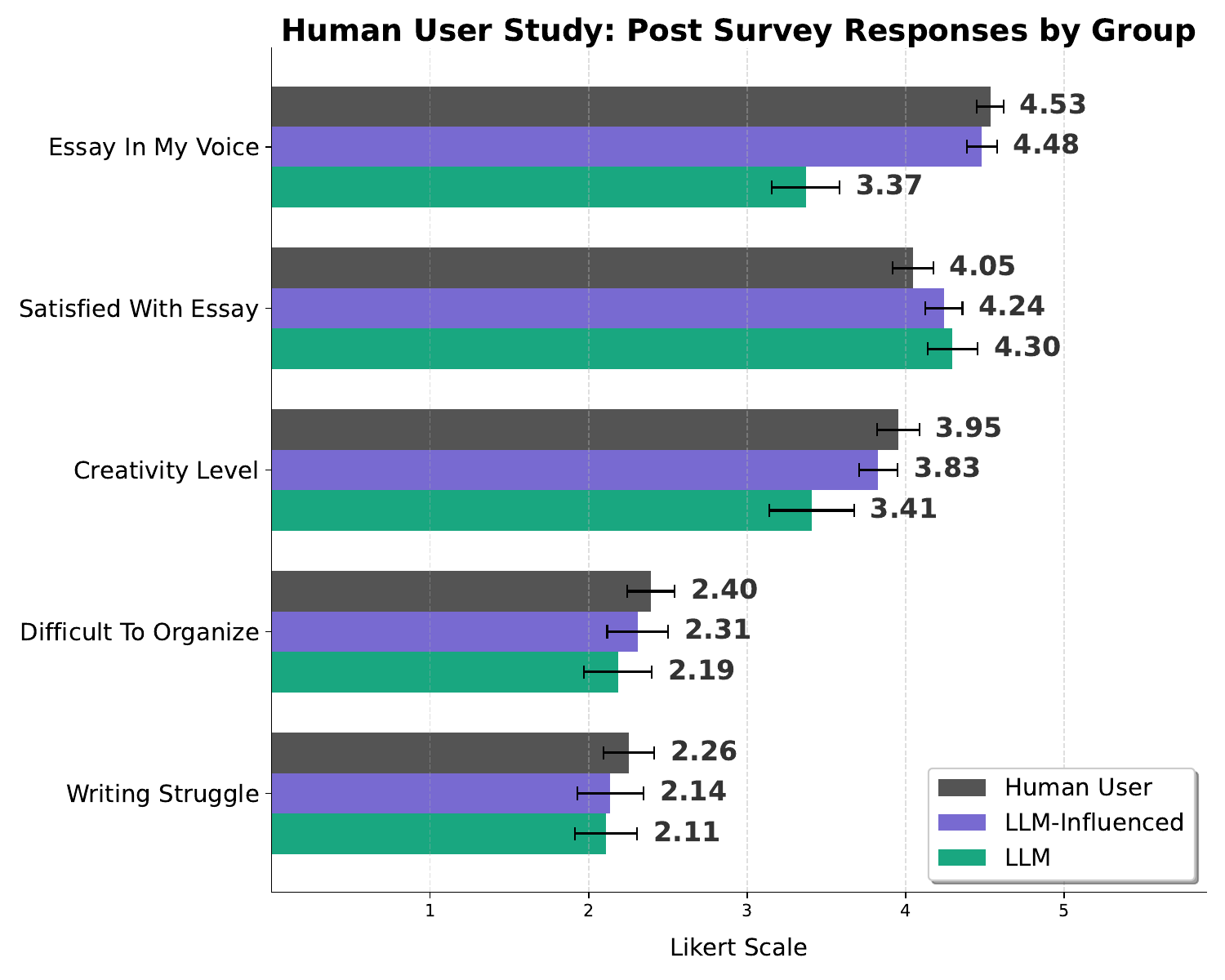}
        \caption{Post-study survey responses show that users who engaged in heavy use of AI consider their essays to be significantly less creative ($t(69) = 2.110, p = 0.0385$) and less in their voice ($t(69) = 5.799, p < 0.001$) using a t-test. Despite this, these users did not report significantly less struggle with writing the essay, nor less satisfaction with the final result.}
        \label{fig:post_study}
    \end{subfigure}
    \hfill
    \begin{subfigure}[t]{0.48\textwidth}
        \centering
        \includegraphics[width=\textwidth]{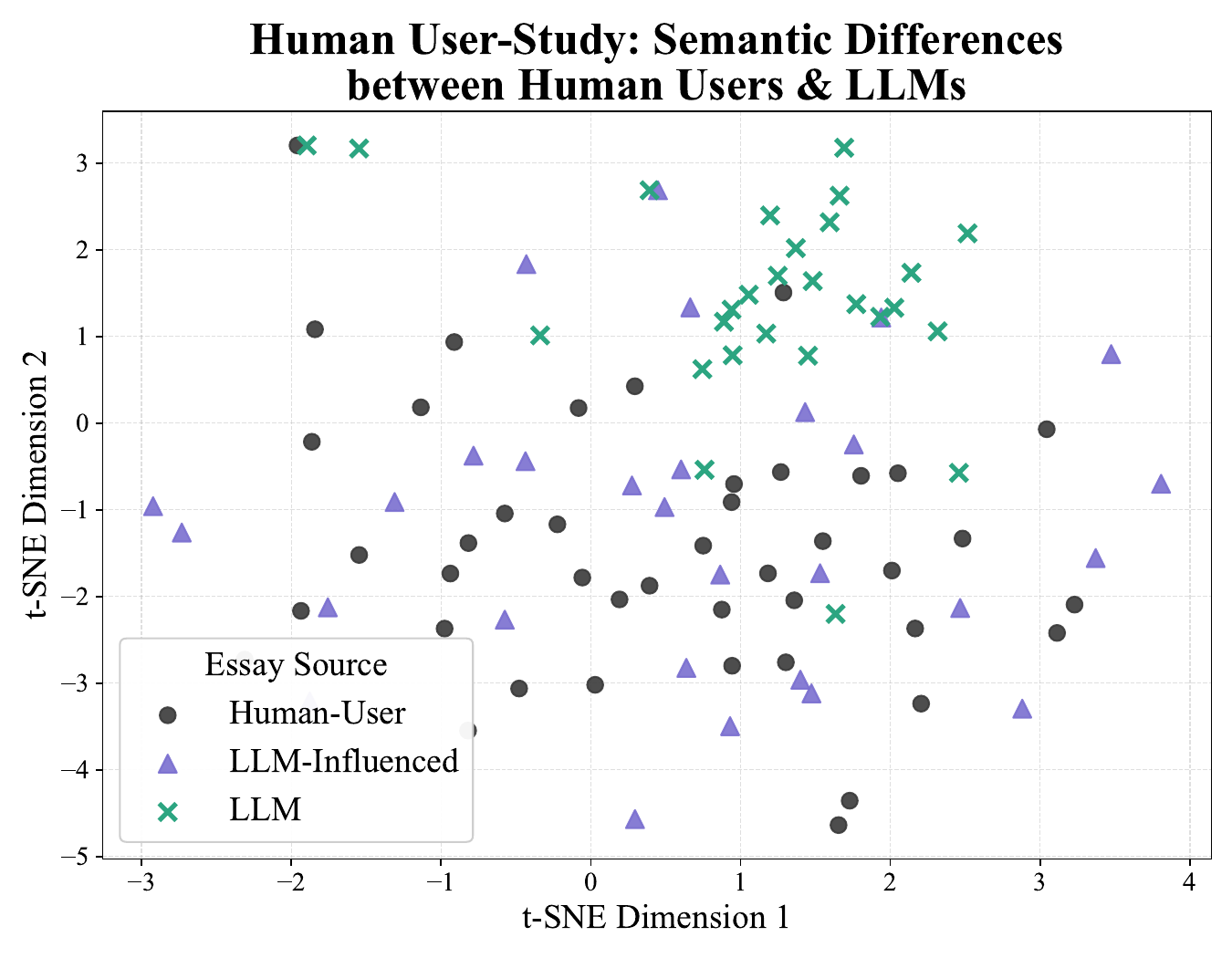} 
        \caption{t-SNE of each essay produced in the RCT embedded using \texttt{MiniLM-L6-v2} \citep{reimers2019sentencebertsentenceembeddingsusing}. Essays written without access to an LLM (human user) occupy a diverse region of embedding space. Essays generated with heavy AI use (LLM) form a distinct small cluster above human essays, indicating both a shift in semantic content that is not human-like, and a homogenizing effect of LLM use. Users who merely consulted an LLM for writing advice or used it in a way comparable to a search tool (LLM-influenced) are more similar to human-written essays vs. LLM-generated essays.}
        \label{fig:human_study_tsne}
    \end{subfigure}\hfill
    \caption{\textbf{In a randomized controlled trial, users which engaged in heavy LLM-use report essays are less creative and not in their voice, and exhibit large, homogenizing semantic shifts in their writing.}}
    \vspace{-1em}
\end{figure}

\subsection{Heavy LLM Users Report That Their Essays Do Not Reflect Their Own Voice}

Figure~\ref{fig:post_study} shows the Likert-scale self-report scores of participants evaluating the essays they produced during the user study. Both participants in the control group who did not have access to an LLM (\textit{Human User}) and those who merely used the LLM as a tool to search for additional information or receive writing advice (\textit{LLM-influenced}) scored similarly across all self-report metrics. However, participants who heavily relied on an LLM to help write their essay (\textit{LLM}) reported feeling that their essay was significantly less creative ($t(69) = 2.110, p = 0.0385$) and less in their own voice ($t(69) = 5.799, p < 0.001$) using a t-test comparing to those without the LLM intervention. 
Paradoxically, these users reported a similar level of both satisfaction with the ultimate product as participants in the control group, as well as similar difficulty in organizing the essay and degree of writing struggle. These results shed light on the draw of using LLMs to complete writing tasks; people are satisfied with the results, even though their voice and creativity are diminished.

\subsection{LLMs Distort Writing by Shifting Essays in a Common Semantic Direction}  

\begin{figure*}[t]
    \vspace{-2em}
    \centering
    \includegraphics[width=\textwidth]{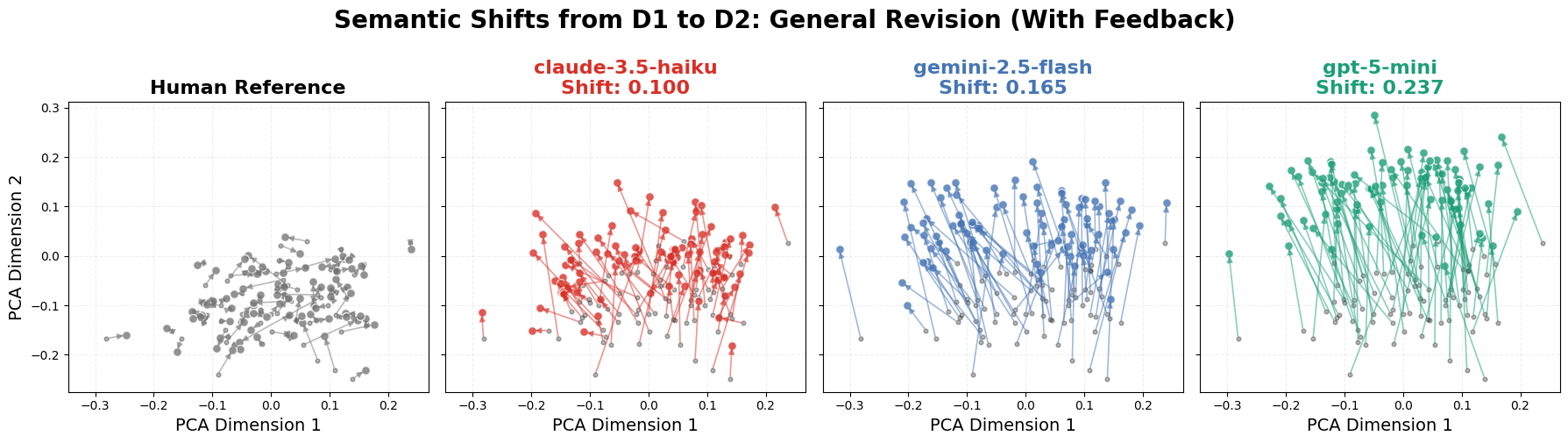}
    \caption{Semantic shifts induced by human and LLM revisions for the ArgRewrite-v2 dataset. Each point pair represents an essay before (D1) and after (D2) revision, embedded using \texttt{gemini-004} sentence embeddings and projected into two dimensions via PCA, a common approach for analyzing semantic differences \citep{dhillon2015eigenwords}. The left panel shows human revisions, while the remaining panels show revisions produced by different LLMs without access to expert feedback. Arrows indicate the direction and magnitude of semantic change. Human revisions exhibit smaller, more varied semantic shifts, whereas LLM revisions produce larger shifts that are strongly aligned in a common direction, indicating a homogenization effect in semantic space.}
    \label{fig:general_with_feedback_embedding_gemini004}
\end{figure*}

To understand how reliance on LLMs may shift the semantic meaning expressed in human writing, we encode each of the essays from the user study into \texttt{MiniLM-L6-v2} sentence embeddings, and visualize the results by projecting into two dimensions via T-SNE. 
\Cref{fig:human_study_tsne} shows the resulting semantic distribution of essays across the three user study conditions. 
We find that essays written by humans in the control group (black dots) are widely spread out throughout the embedding space, occupying a broad region that reflects the diversity of individual perspectives, writing styles, and argumentation. 
On the other hand, essays written by LLMs (green crosses) form a tight cluster in the upper right quadrant of the space, a region that is not occupied by any of the human-written essays. 
This clustering shows that when people engage in heavy LLM use, the resulting text is substantially semantically different from that produced by humans. 

We find that \textit{LLM-influenced} users (purple triangles) produce essays that occupy a similar distribution as human-written essays in the control group. Together with the above results showing that LLM-influenced users provide similar self-report scores as the control group, we find evidence that not all modes of using LLMs as a writing tool lead to distortion in the user's writing. When using the LLM merely as an information-seeking tool, rather than to write the piece wholesale, users avoid homogenization and preserve semantic meaning, creativity, and their own voice.

While the above results study a naturalistic condition in which users have access to an LLM while tasked with a piece of writing, it is possible that laziness or a lack of motivation drove some users to rely on the LLM too heavily. What about seemingly more responsible uses of the LLM as a writing tool, such as asking it to only edit or revise a piece of text that was previously written by a person? 
To 
test this scenario, we perform a counterfactual analysis, comparing how an LLM edits a piece of writing to how a human would have edited it if they did not have access to an LLM. 


\Cref{fig:general_with_feedback_embedding_gemini004} shows the semantic shifts from the initial human draft (D1) to revised drafts (D2) in the ArgRewrite-v2 dataset, by the original human editor (left) and the three different LLM models prompted to make \textit{general} revisions. These results replicate the findings of \Cref{fig:gpt-5-mini-with-feedback-gemini-004}, which shows the embedding distribution across different types of revision prompts used with \texttt{gpt-5-mini}. 
We find that humans make small, multidirectional semantic shifts, with arrows pointing in diverse directions. 
In contrast, all three LLMs produce semantic shifts consistently pointed in the same direction, with the magnitude of semantic change largest for \texttt{gpt-5-mini} and smallest for \texttt{claude-haiku}. 
Across LLMs, we also find the largest semantic change is between the initial human draft and the final LLM draft when LLMs are prompted to \textit{complete} and \textit{expand} the essay. 
However, we find concerning levels of shift even when LLMs are tasked to perform \textit{minimal} edits to the essay or only edit the essay for \textit{grammar}, with the LLM frequently changing the human user's conclusion in the essay (see examples in \Cref{fig:llm_argument_rewrite_example}). 
We have replicated these results across different embedding models, different LLMs, and different prompts, and these results are available in Appendix \Cref{appendix:semantic_shifts_gemini004}, \Cref{appendix:semantic_shifts_minilm}, and \Cref{appendix:semantic_shifts_across_settings_per_model}.
In the next sections, we further analyze how these differences between human-written and LLM-written texts actually manifest. 

\subsection{LLMs Alter the Conclusions of Human Writing} 

\begin{figure}
    \centering
    \includegraphics[width=0.9\linewidth]{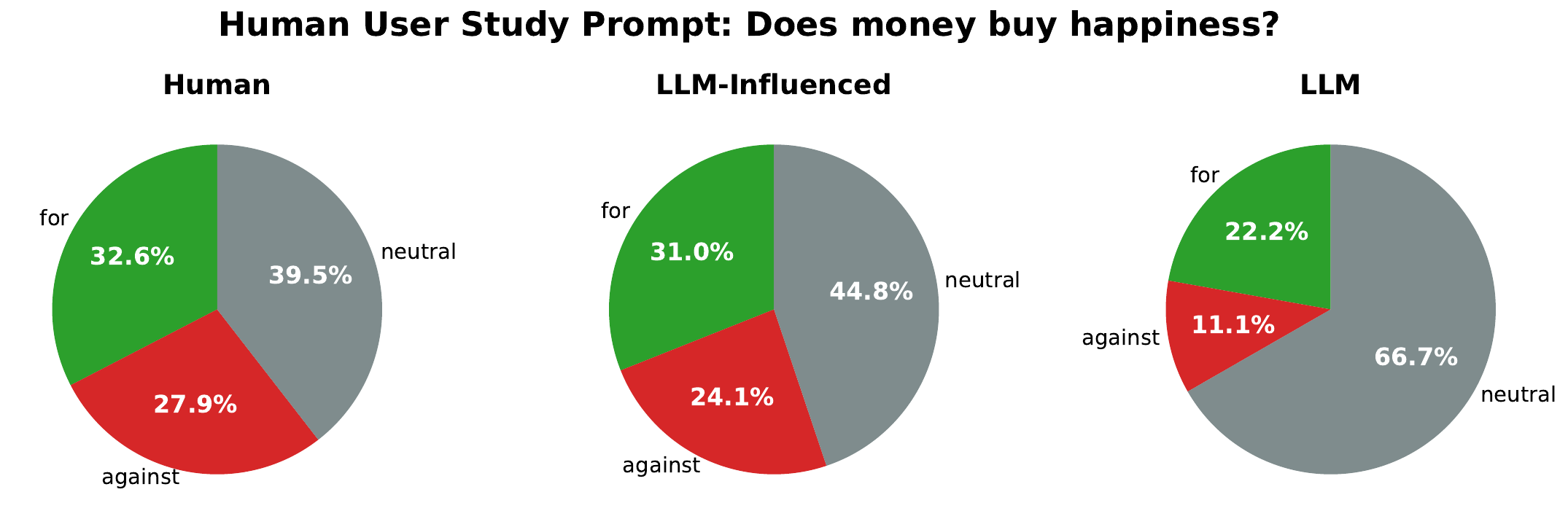}
    \caption{The results of using LLM-as-a-Judge to categorize the argumentative stance of essays written during the human user study. 
    Relative to the control group, our results show that extensive AI use results in a 68.9\% increase in the proportion of essays that remain neutral when answering the topic question, rather than actually expressing a for/against opinion ($t(69) = -2.439, p = 0.017$) comparing human to LLM responses.  
    }
    \label{fig:human-study-stance}
    
\end{figure}

During the user study, participants were asked to write an essay answering the question, ``Does money lead to happiness?''. Using an LLM-as-a-judge framework, we classified each essay as arguing for, against, or remaining neutral when answering the topic question. Figure \ref{fig:human-study-stance} shows the proportion of users taking each stance in the \textit{Human} control group that had no access to an LLM, the \textit{LLM-influenced} group that only used the LLM as a reference, and the \textit{LLM} group that heavily relied on the LLM to generate the essay. As shown in the figure, we find that relative to the control group, extensive LLM use increases the proportion of users taking a neutral position by 68.9\%, a statistically significant effect ($p < 0.036$). These results imply that the LLM-generated essays fail to convey users' actual opinions about the topic. 

In our qualitative analysis of LLM-edited essays on ArgRewrite-v2, we find similar trends. As shown in the examples in \Cref{fig:llm_argument_rewrite_example}, when prompted to edit an essay for grammar, LLMs 
make changes to the claims in the essay. LLMs are not simply correcting errors or improving clarity, but are fundamentally reorienting the content of diverse human essays toward a shared semantic mode, which, in the case of ArgRewrite-v2, are essays that are in support of self-driving cars. The uniformity of these shifts across different LLMs also suggests a convergence toward LLM-preferred linguistic patterns \citep{artificialhivemind2025open} that may not reflect the original intent or voice of human writers, as the final drafts produced by humans are semantically very different than the LLM-edited drafts. 

\subsection{LLMs Make Substantially Larger Lexical Changes Than Humans}
\begin{figure}[htbp]
  \centering
  \begin{minipage}[c]{0.5\textwidth}
    \centering
    \includegraphics[width=\textwidth]{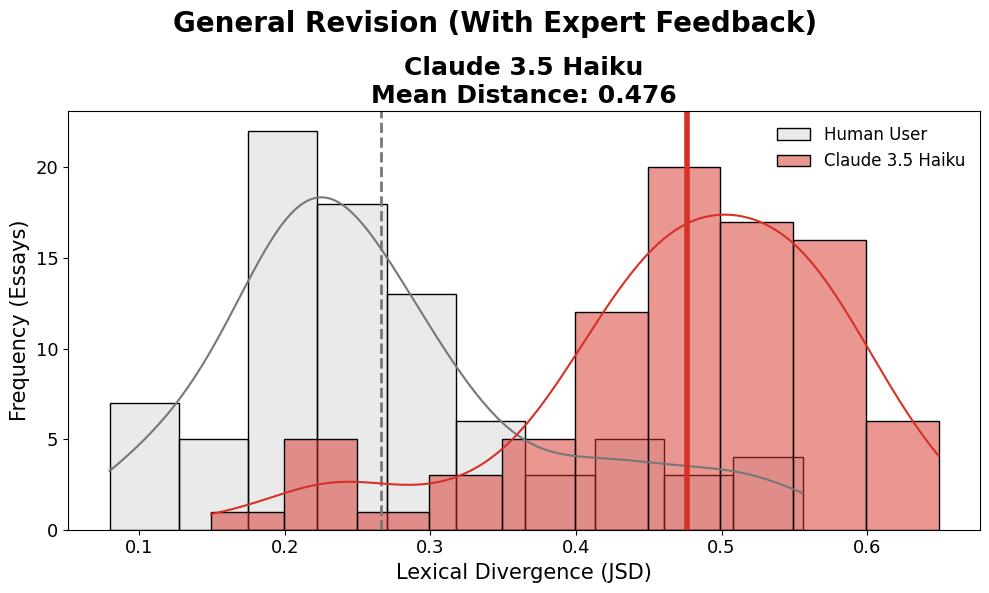}
  \end{minipage}
  \hspace{0.03\textwidth}
  \begin{minipage}[c]{0.40\textwidth}
    \caption{Lexical divergence distribution for \textit{general} revision of \texttt{claude-haiku} with expert feedback for ArgRewrite-v2. Gray bars show the human baseline, while colored distributions show LLM revisions systematically shift rightward with a mean distance of 0.476. Higher JSD indicates more extensive vocabulary replacement, with LLMs substituting more words than humans. Note that \texttt{claude-haiku} resulted in the least semantic distortion.}
     \label{fig:jsd-general-feedback}
  \end{minipage}
\end{figure}

We also examined how LLMs alter the distribution of words used in an individual's writing by using JSD to measure the lexical divergence from the initial human draft in ArgRewrite-v2 induced by both LLMs and people. ~\Cref{fig:jsd-general-feedback} shows the distribution of JSD values for each of the essays produced using \texttt{claude-haiku} prompted to make general revisions with expert feedback, with JSD plots for other types of edits found in \Cref{appendix:jsd}. 
The human baseline exhibits a tight distribution centered around 0.2-0.3 JSD, indicating that humans make modest, targeted word substitutions while preserving most of their original vocabulary. 
In contrast, all three LLMs substantially alter the unigram distribution of the essay, with \textit{gpt-5-mini} showing the most significant change in divergence of nearly triple the human baseline, with many essays reaching divergences above 0.7. 
These lexical shifts demonstrate that LLMs replace a much larger fraction of the original writing than humans do when revising their own work.
This substitution of words contributes to the loss of individual voice, style, and meaning, as the unique lexical fingerprint of each writer is overwritten by the given model's preferred vocabulary. 
Further results with different LLM-editing conditions and LLMs can be found in \Cref{appendix:jsd}.


\begin{figure}[h]
    \centering
    \begin{subfigure}{0.49\textwidth}
        \centering
        \includegraphics[width=\linewidth]{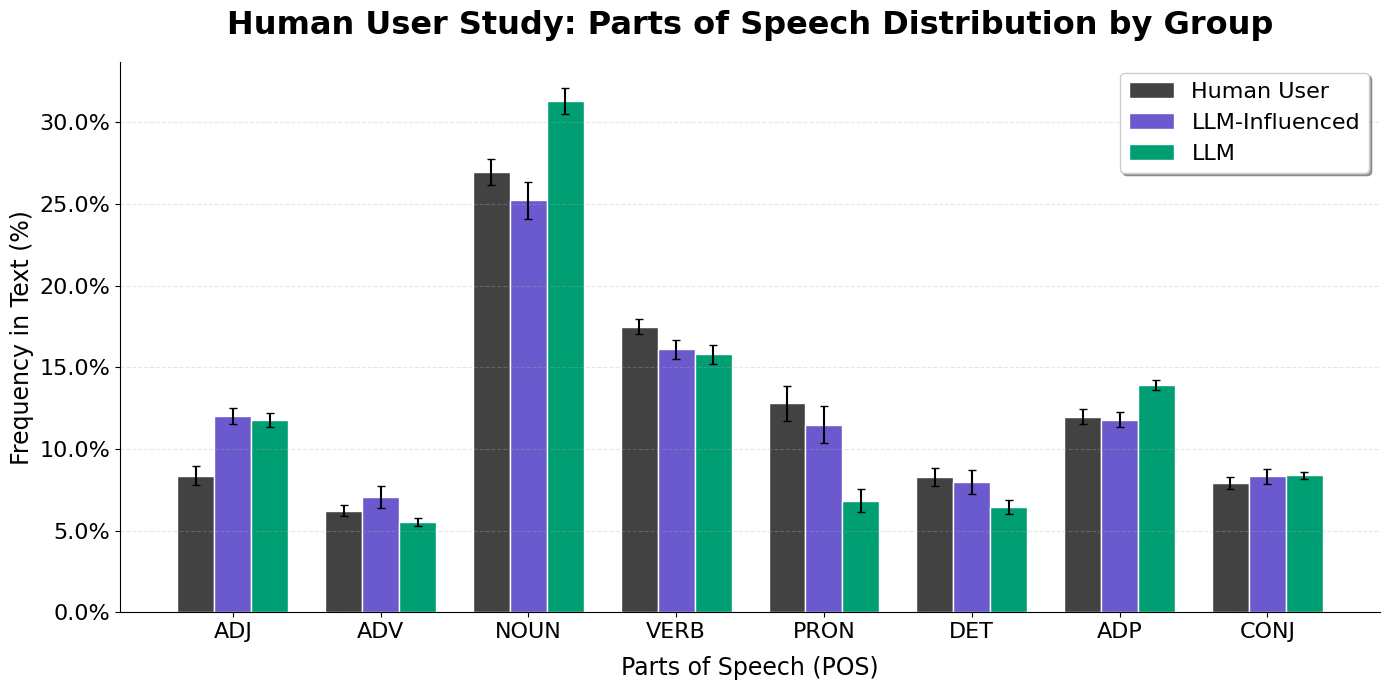}
    \end{subfigure}\hfill
    \begin{subfigure}{0.49\textwidth}
        \centering
        \includegraphics[width=\linewidth]{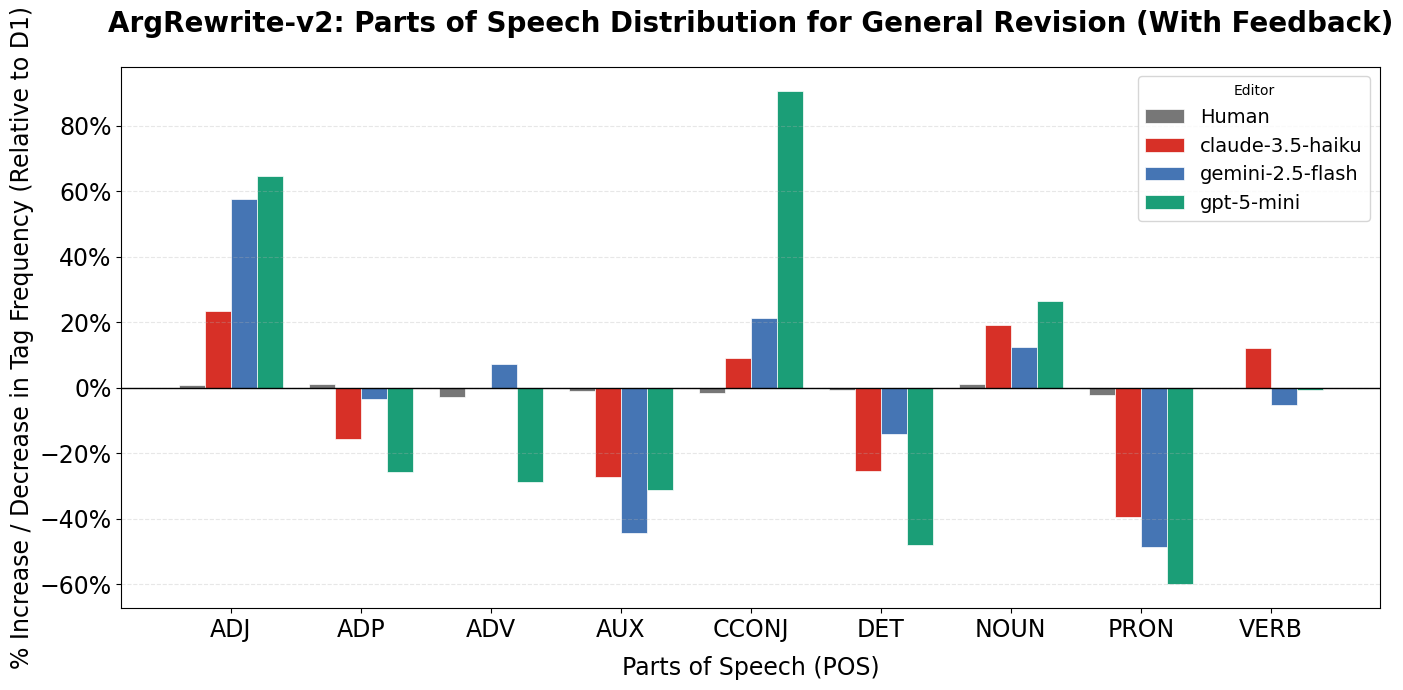}
    \end{subfigure}
    \caption{Parts of Speech Distribution. Left: Essays written during the user study, either by people in the control group, or with assistance from LLMs. Right: Essays from the \texttt{ArgRewrite-v2} dataset edited by humans versus edited by 3 different LLMs prompted to make \textit{general} edits. Across both studies, LLMs use more nouns and adjectives when writing/editing, and reduce the use of pronouns.}
    \label{fig:pos-general-feedback}
    \vspace{-1em}
\end{figure}

\subsection{LLMs Systematically Restructure Grammar Toward a Less Personal, Formal Style} 
Next, we study how LLMs change the syntactic structure of writing by analyzing part-of-speech (POS) distributions in essays written by human users versus essays written by LLMs. ~\Cref{fig:pos-general-feedback} (left) shows the relative change in POS tag frequencies for our human-user study. We find a 50\% decrease in pronouns from human-user essays and essays written with LLMs, signifying a removal of first-person, experience-based argumentation toward impersonal language. We also find that LLM-written essays show a higher percentage of nouns (14\% relative increase), with adjectives showing a similar trend (33\% relative increase). ~\Cref{fig:pos-general-feedback} shows a similar analysis for the ArgRewrite-v2 dataset. People, in contrast, make minimal grammatical changes, typically under $5\%$ for any POS category. 

LLMs systematically restructure sentences toward a more formal style and primarily use nouns in writing. We also find that all three models increase the use of adjectives compared to the human draft ($57-90\%$ increase) and coordinating conjunctions ($13-90\%$ increase), while substantially reducing pronouns ($40-60\%$ decrease) and determiners ($25-50\%$ decrease). This confirms our qualitative findings that LLM edits move writing away from first-person narratives toward impersonal, academic writing. We find these shifts to be more pronounced when the LLM is prompted to complete or expand the essay. We find \texttt{gpt-5-mini} shows the most extreme restructuring, with an 88\% increase in coordinating conjunctions and 27\% increase in nouns, alongside a 61\% decrease in pronouns. This pattern aligns with prior observations that LLMs favor complex, formal constructions over the more direct, personal style typical of human writing. For minimal edits, LLMs still increase use of adjectives by 40-87\% and reduce pronouns by 31-56\%. Further results with different LLM-editing conditions found in \Cref{appendix:pos_distribution}. These results illustrate that LLMs are unable to infer and preserve user preferences in writing. 

\subsection{The Use of LLMs for Writing Increases Emotional Language}
\begin{figure}
    \centering
    \includegraphics[width=1.0\textwidth]{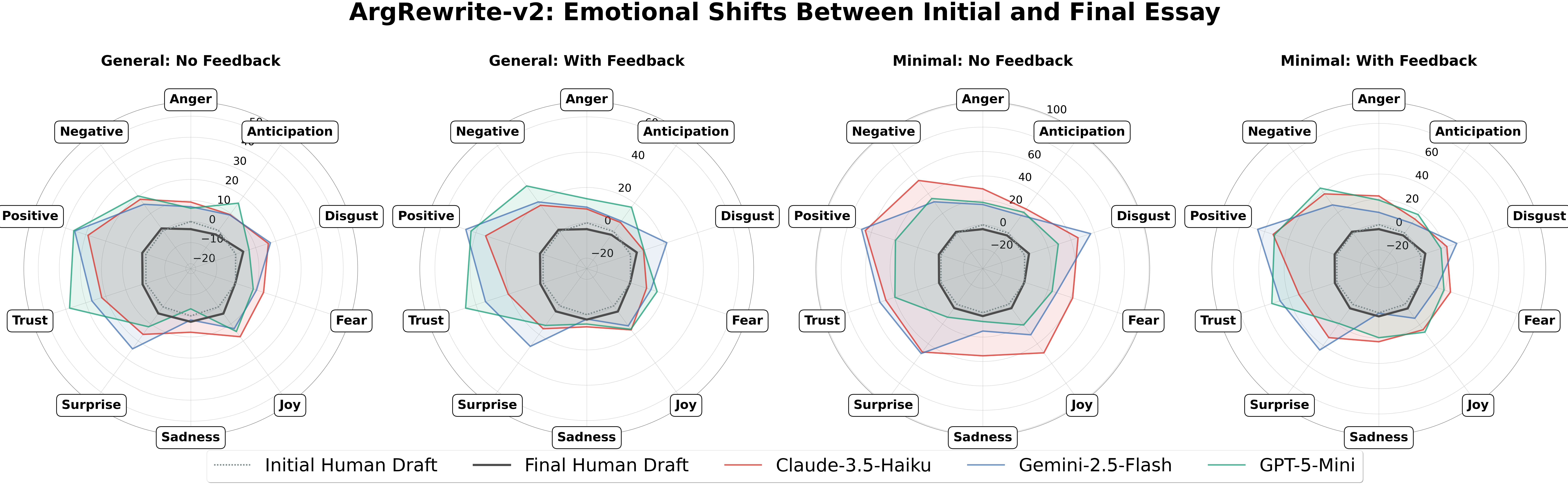}
    \caption{Emotional shifts for general revisions with and without expert feedback. Human baseline (gray) shows minimal affective changes. LLMs increase positive sentiment (37-54\%) and trust-related language (17-53\%), while also simultaneously increasing negative sentiment (24-38\%). This indicates increases in the use of emotional language, regardless of which LLM model is used.} 
    \label{fig:argrewrite-emotional-shift}
\end{figure}

We examined whether LLM edits change the distribution of emotions present in text using the NRC Lexicon \citep{mohammad2013nrc} on the ArgRewrite-v2 dataset. ~\Cref{fig:argrewrite-emotional-shift} shows the change in emotional word density for \textit{general} and \textit{minimal} revisions with and without expert feedback. We find human edits make minimal affective changes (black), with adjustments typically under 5\% for any emotion category compared to the initial human draft. On the other hand, LLMs increase the amount of emotional language used in general, with LLMs expressing slightly more words relating to positivity and trust labels in the essay.  This pattern suggests that LLMs systematically reframe arguments in more positive, optimistic terms, even when the original human text may have been critical or skeptical. For essays about self-driving cars (the ArgRewrite-v2 topic), this could mean downplaying concerns about safety, job displacement, or ethical issues in favor of enthusiasm about technological progress, as shown in \Cref{fig:llm_argument_rewrite_example}, where the LLM removes mention of the drawbacks of self-driving cars. Interestingly, we also find that without expert human feedback guiding the LLM edits, the emotional shifts across all categories relative to the initial human draft are even more drastic, reinforcing the observation that LLMs increase emotional language in the essay. We find differences between models, with \texttt{claude-haiku} showing less pronounced shifts with expert feedback. Further results with different LLM-editing conditions and LLMs found in Appendix \ref{appendix:emotional_shift}.

\subsection{The Use of LLMs for Writing Increases Analytical, Logical, and Statistical Language}
\begin{figure}[htbp]
    \centering
    \begin{subfigure}[t]{0.4\textwidth}
        \vspace{3pt}
        \includegraphics[width=\linewidth]{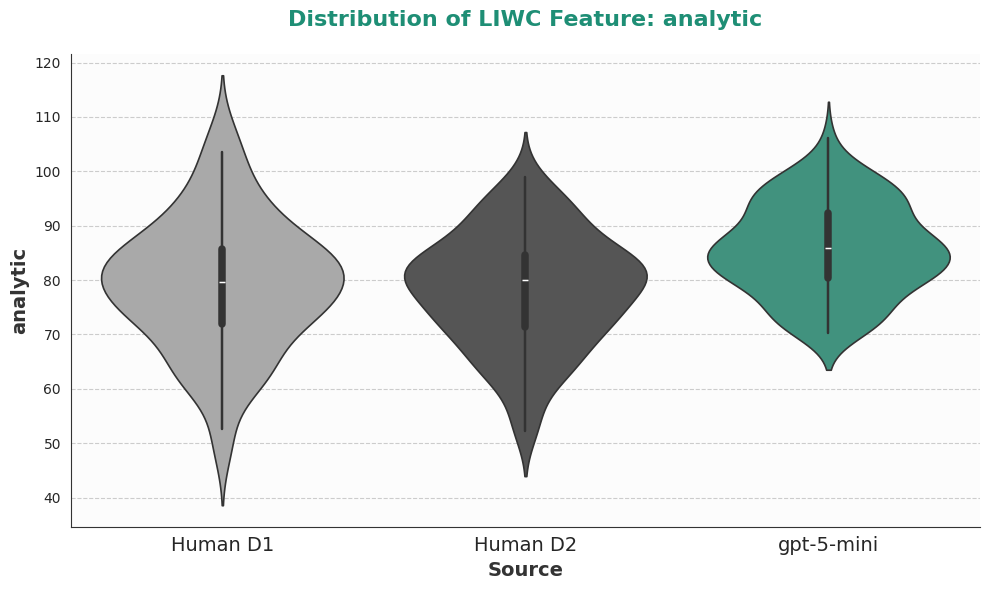}
        \vspace{0.5em}
        \caption{Distribution of LIWC analytic feature scores across Human D1, Human D2, and LLM D2 sources in ArgRewrite-v2. LLM-edited essays use more words from the `analytic' category.}
        \label{fig:liwc_analytic_argrewrite}
    \end{subfigure}
    \hfill
    \begin{subfigure}[t]{0.5\textwidth}
        \vspace{0pt}
        \includegraphics[width=\linewidth]{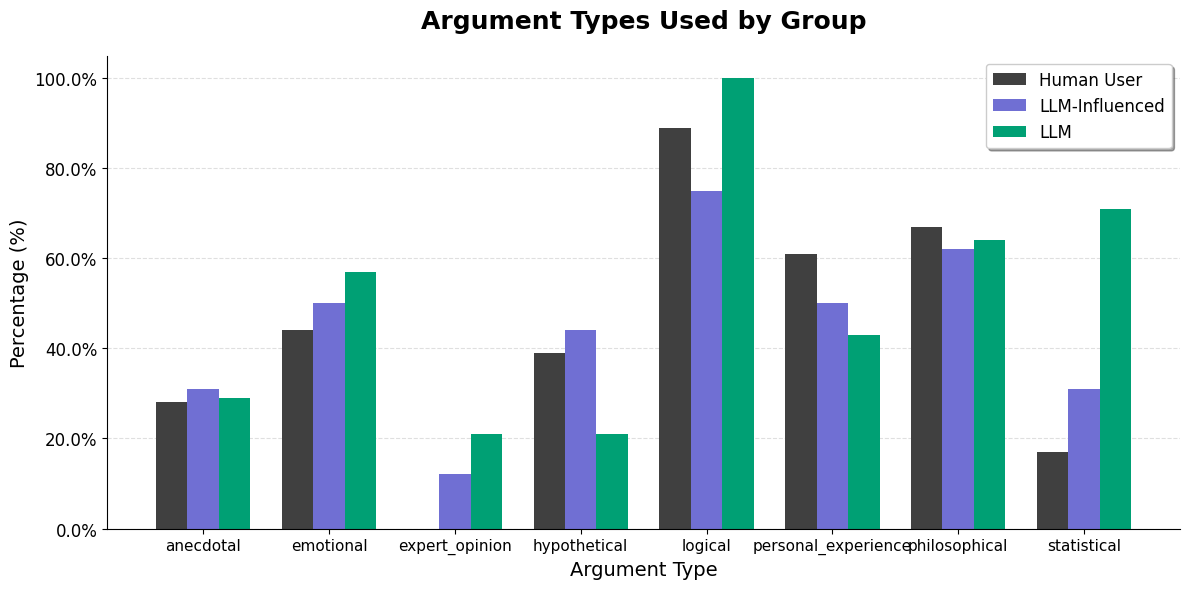}
        \caption{LLM-as-a-Judge categorization of user study essays by argument type. Essays produced by heavy LLM users use expert opinion and statistical arguments more frequently ($t(69) = 4.82, p < 0.0001$) than those in the human group.}
        \label{fig:human-study-argument-types}
    \end{subfigure}
    \caption{\textbf{LLM assistance shifts writing to focus on analytical and statistical arguments} (a) LIWC analytic scores are elevated in LLM-edited text. (b) Human study participants with LLM access rely more heavily on expert opinion ($t(69) = -2.78, p < 0.01)$ and statistical arguments ($t(69) = 4.82, p < 0.0001$).}
    \label{fig:analytic_and_argument_types}
\end{figure}

The previous section demonstrated that LLMs increase the use of emotional language. Curiously, we find that they simultaneously increase the use of analytical, logical, and statistical arguments for the same datasets. Figure \ref{fig:liwc_analytic_argrewrite} shows the distribution of words from the LIWC `analytic' category, which is intended to summarize the degree to which writing exhibits formal, logical, and hierarchical thinking patterns \citep{boyd2022liwc22}. In the user study, LLM-generated essays are more analytic than human-written essays. Figure \ref{fig:human-study-argument-types}) analyzes the types of arguments used by both humans and LLMs in ArgRewrite-v2, using LLM as a judge to extract categories of arguments such as `hypothetical' vs. 'anecdotal'. We find that people are more likely to use arguments related to personal experience, while LLM-written essays are more likely to use statistical and logical arguments. LLM-influenced essays also cite expert opinions, something that human-written essays rarely do. We find these results corroborated in \Cref{fig:llm_argument_rewrite_example} (row 2), where the first-person voice is removed by the LLM-edited draft. 

\begin{figure}[ht!]
    \centering
    \includegraphics[width=\linewidth]{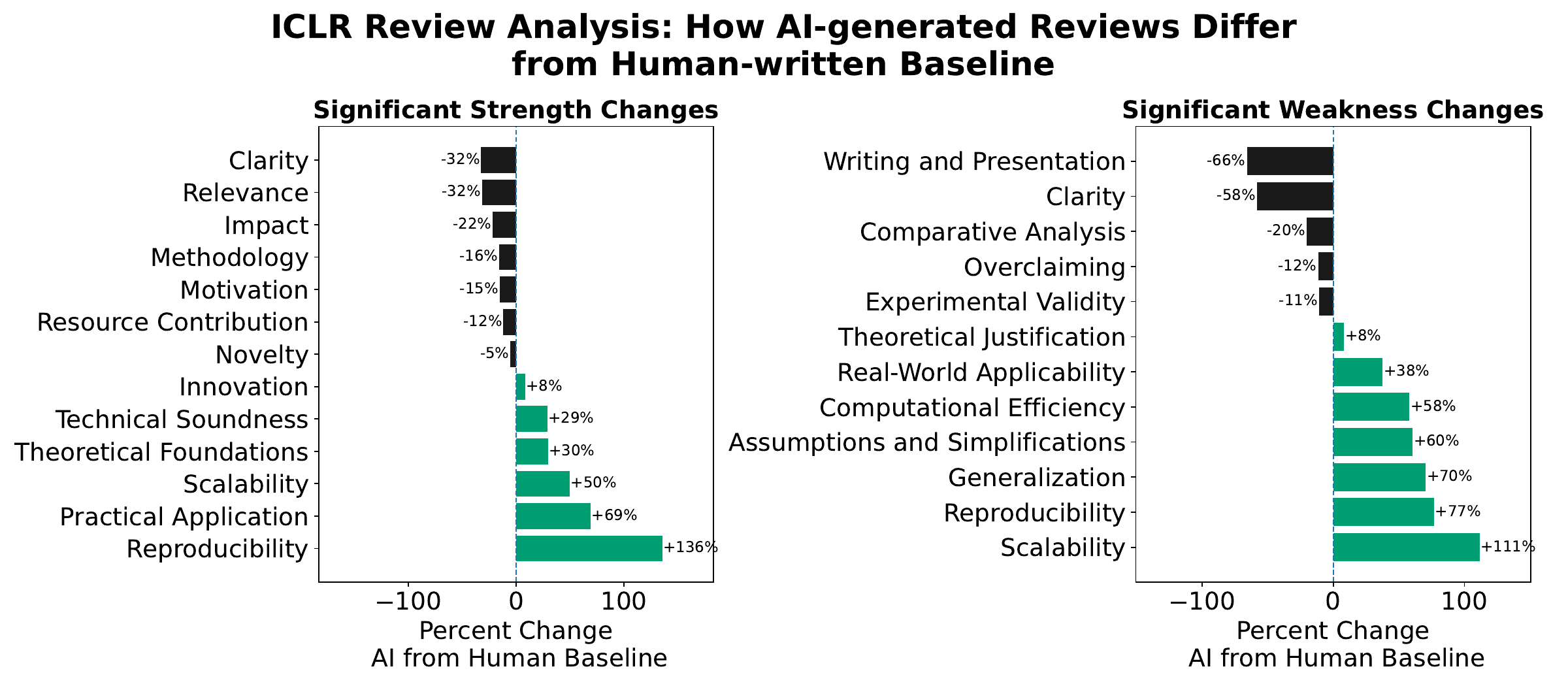}
    \caption{We use LLM-as-a-Judge on 18k reviews from ICLR 2026, selected because they were all written on the subject of LLMs, where 9k reviews are entirely written by humans, and 9k are entirely written by LLMs.
    We plot the relative frequency of certain strengths and weaknesses occurring in LLM reviews instead of human reviews. 
    For the Clarity strength category, LLMs were $32\%$ less likely to select it compared to a baseline of $17\%$ in human reviews. A two-proportion z-test confirmed this difference is highly significant ($z = 11.00, p < 0.001$), indicating a systematic divergence in how AI models prioritize presentation-based critiques compared to human peers.}
    \label{fig:iclr_analysis}
\end{figure}

\subsection{LLMs Distort Decisions Affecting Scientific Institutions}
The results we have presented so far demonstrate that LLMs can produce large semantic changes to human writing when prompted to edit essays, or when heavily relied upon to produce the essay required for our user study. But it is possible that when people use LLMs for writing as part of their professional work, they are more careful to produce text that aligns with their intentions, and the effects we have identified are reduced. To address this concern and investigate the ecological validity of our hypothesis, we examine real-world instances of LLM-generated text. As described in Section \ref{sec: datasets}, for this purpose, we analyze scientific peer reviews written for the upcoming ICLR 2026 AI conference, of which 21\% were found to be almost fully AI-generated \citep{emi2024technicalreportpangramaigenerated}. The ICLR review process not only forbids the use of AI in generating reviews but also makes both reviews and the reviewer's name visible to other reviewers, area chairs, and other senior members of the conference. Therefore, ICLR reviewers have a strong incentive to both conceal their use of AI and ensure that reviews accurately reflect an appropriate scientific opinion, lest they tarnish their professional reputation to senior members of their scientific community. 

We find that when LLMs are employed in the scientific review process, the scores, decisions, and arguments made in peer reviews shift. 
LLMs assign scores 10\% higher than humans (4.43 for LLM reviews and 4.13 for human reviews). Most strikingly, as shown in Figure \ref{fig:iclr_analysis}, strengths and weaknesses used to critique scientific papers are different between LLM- and human-written reviews.
We observe that humans are 32\% more likely to comment on the clarity as a strength ($z = 14.37, p < 0.00001$) and 58\% more likely to comment on clarity as a weaknesses ($z = 24.9, p < 0.00001$) and 32\% more likely to comment on the relevance of research ($z = 10.77, p < 0.00001$), while LLMs are 136\% more likely to comment on reproducibility($z = -13.6601, p < 0.00001$) and 84\% more likely to comment on scalability for both strengths and weaknesses ($z = -15.42, p < 0.00001$). 
These results demonstrate that the criteria under human review and LLM review are significantly different, which will have as-yet-unknown downstream impacts on the decisions made about what scientific work is valid and incentivized.

%% file: sections/5-discussion.tex
\section{Discussion} \label{sec: discussion}

Across three studies investigating how people interact with LLMs, comparing LLM-generated vs. human-written edits to the same text, and assessing how LLM-generated text is actually being used in the wild, our results show that LLMs significantly alter the meaning of human writing. The ways in which LLMs alter text are similar across model types, and persist even when LLMs are prompted to make minimal edits. 
Participants in our user study who rely heavily on LLMs find that the resulting essay is significantly less creative and not in their voice, and these users are significantly more likely to adopt a neutral stance when answering the topic question. LLMs edit text in ways that are very different from how humans would edit the same text, meaningfully altering the semantics while using both more emotional and more argumentative language. Finally, when analyzing how professionals actually use LLMs in scientific peer review, we find that LLM-generated ICLR reviews place significantly less weight on the clarity and significance of research, potentially altering the way the scientific review process is conducted. 

These results present a troubling picture of AI subtly distorting our written language, and with it, our cultural institutions. While evidence has surfaced that AI-generated content is already infiltrating parliamentary speeches, song lyrics, movie scripts, spoken language, and even messages we send to our coworkers and loved ones \citep{von7approaching, borole2024influence}, it is not yet clear what effects this will have. It also appears unlikely that the adoption of LLMs will slow. Individuals who use AI may accelerate their productivity, even if it comes at a cost to the global information ecosystem \citep{hao2024ai}. Our results show a clear paradox: even though people who rely heavily on AI recognize that it diminishes their voice and creativity, they are nevertheless equally satisfied with the results. The ease of use, combined with the potential to accelerate individual careers, is likely to continue to incentivize people to produce AI-generated text, and even to attempt to pass it off as their own in professional contexts, as the ICLR data show. 

We find it curious to observe that in our study, LLMs increase both the use of both emotional language, as well as logical and analytic argumentation. It is worth noting that production LLMs, such as those used for this study, are trained with Reinforcement Learning from Human Feedback (RLHF) to maximize positive responses from people \citep{ouyang2022traininglanguagemodelsfollow, jaques2019way}. We also know that LLMs are not trained in a way that enables them to model individual users' unique preferences \citep{sorensen2024roadmap,siththaranjan2023distributional,poddar2024personalizing}. LLMs have no inherent incentive or mechanism for maintaining the user's intended meaning, or even refraining from attempting to manipulate people to change the way they provide preference ratings. It is thus possible that optimizing at scale for positive human responses, without the ability to actually adhere to individual human preferences, incentivizes LLMs to produce text that is broadly more convincing to most people---thus both more emotional, statistical, logical, etc. Perhaps this is the written language equivalent of `clickbait'. Troublingly, we know from the literature on recommender systems in machine learning, that optimizing for human feedback metrics such as engagement can actually lead to altering people's underlying preferences and behaviors \citep{dean2022preference,carroll2024ai}. For example, when YouTube optimized user's watch time, they found that the resulting model learned to recommend radicalizing content  \citep{ribeiro2020auditing,ledwich2019algorithmic}. Further research is urgently needed to understand what large scale reinforcement learning fine-tuning of LLMs on human feedback does to both the language they produce, as well as their effect on the people that use them. 

For researchers interested in improving AI capabilities, our research highlights the need for algorithms that are better able to infer the user's underlying preferences and intended meaning, and produce text that actually adheres to it. We believe the true objective should be to produce text similar to what the human would have if they had invested the time and effort to do it themselves, motivating our counterfactual analysis of the ArgRewrite-v2 data. 

More broadly, it is essential that future research develop a better understanding of how massive, rapid adoption of LLMs 
will affect our cultural institutions. Our study demonstrates that LLM-generated text is already affecting the criteria we apply to scientific papers in AI. 
If 1 billion people are currently using LLMs, including politicians, how will this affect the ways we organize, communicate, and respond to rapid social and economic changes?
Ultimately, humans are a cultural species, and it is our cultural institutions that enable us to globally cooperate and address potential existential threats \citep{henrich2015secret}. The massive changes caused by AI deployment may be exactly the type of threat that requires highly robust institutions. Ironically, these institutions are most needed just as they are facing large-scale AI disruption.  

%% file: sections/ethics.tex
\section*{Ethics Statement}
This work is primarily focused on AI Safety, and how using AI for writing and editing of text impacts the content of human writing across various dimensions, including semantics, grammar, emotional distributions, and the claims being made. 
We conducted a human-user study to understand preferences towards writing, which was approved by our IRB. Proper protocols to anonymize and remove personally identifiable information were followed. 
We recruited participants via Prolific, asked for consent before the study, and compensated participants for their time. Since the participants in our user study are native English speakers residing within the United States, our findings may present themselves differently in different languages or for people residing in different countries. 
Questions of how AI affects speakers of other languages and norms from other cultures when using AI assistants are a key question for future research.  Data for ArgRewrite-v2 and academic peer reviews were publicly available on the internet and did not contain any personally identifiable information. Our contributions provide evidence for the need to design LLMs that preserve human agency when writing. We hope this work encourages research to develop safeguards and improve tools for human writing that preserve the human voice.

\section*{Reproducibility Statement}
To ensure reproducibility, we provide details of our user study in \Cref{appendix:user_study_details}, containing the instructions provided to participants who were tasked to write without an LLM, as well as the pre-study and post-study questions provided. 
We also provide the prompts used for our analysis of the ArgRewrite-v2 dataset in \Cref{appendix:argrewrite_prompts}, as well as full results on the semantic (\Cref{appendix:semantic_shifts_minilm}, \Cref{appendix:semantic_shifts_gemini004}, and \Cref{appendix:semantic_shifts_across_settings_per_model}), lexical (\Cref{appendix:jsd}), emotional (\Cref{appendix:emotional_shift}), and parts of speech (\Cref{appendix:pos_distribution}) analyses, and for robustness, perform quantitative experiments across models and various prompts for editing grounded in our human-user study. Lastly, we provide analysis on several other ICLR categories, and provide the prompts used for the analysis in \Cref{appendix:iclr_analyses}.

\section*{Acknowledgments}
We would like to thank the Toyota Research Institute (TRI), Cooperative AI Foundation, Character.AI, DoorDash, Open Philanthropy, Coefficient Giving, the Foresight Institute, the UW-Amazon Science Gift Hub, the Sony Research Award Program, UW-Tsukuba Amazon NVIDIA Cross Pacific AI Initiative, Jacobs CIFAR Research Fellowship, Templeton World Charity Foundation (https://doi.org/10.54224/34843), Tinker Research Grants, and the Schmidt AI2050 Fellows program. This material is based upon work supported by the Defense Advanced Research Projects Agency and the Air Force Research Laboratory, contract number(s): FA8650-23-C-7316. Any opinions, findings and conclusions, or recommendations expressed in this material are those of the author(s) and do not necessarily reflect the views of these organizations. 

%% file: sections/6-appendix.tex
\appendix

\input{sections/appendix/table_contents}
\newpage

\setlength{\floatsep}{0pt}
\setlength{\textfloatsep}{0pt}
\setlength{\intextsep}{0pt}
\setlength{\abovecaptionskip}{4pt}
\setlength{\belowcaptionskip}{0pt}

\section{User Study Details} \label{sec: user_study_details}
\subsection{User Study Recruitment Process, Compensation and Duration}\label{appendix:user_study_details}

To recruit our participants, we use Prolific, a research platform through which participants can voluntarily participate in research surveys and receive compensation. We give our participants 8 US dollars to participate in the study, with an estimated time of 35 minutes and a maximum time limit of 1 hour. To participate in the study, we require participants to be native English speakers and reside in the United States to control for native language as a factor when writing essays with an AI assistant.

\subsection{User Study Instructions.} \label{appendix:survey}
We give the users the following instructions before we have them do the pre-study questions, the writing with an assistant, and the post-study questions: 

\subsection{LLM Assisted Instructions}

\begin{promptbox}{Instructions (LLM-Assisted Condition)}
You will be writing an essay, and you may use the LLM chat to assist you in the writing process.
Use no other LLM than the one provided in this interface, and you may take as much time as you need.

First, answer some pre-study questions about your attitudes toward AI and writing before you begin the essay.
You may not use other sources, such as the internet, to inform your essay.

After the study, you will be asked post-study questions about your experience.

The purpose of this study is to understand how people use LLMs for writing in their normal workflow.
If you do not usually use LLMs, think of the AI tool as a writing partner: someone to bounce ideas off,
ask questions, and get feedback from as you go.
\end{promptbox}

\subsection{Without LLM Instructions}

\begin{promptbox}{Instructions (No-LLM Condition)}
First, answer pre-study questions about your writing habits and experiences.

Then write your essay in the text box provided.
You are not allowed to use an LLM, AI assistant, or the internet while writing this essay.

Return here once the essay is complete to answer a few questions.

Responses will be reviewed for indications of AI-generated content.
Participants found to have used such tools will not receive compensation.
\end{promptbox}

\subsection{Pre-Study Questions} \label{appendix:pre_study}

\subsubsection*{For LLM Users}

\begin{promptbox}{Attitudes Toward AI-Assisted Writing}
\begin{itemize}
    \item I believe AI tools can improve my writing quality. (+)
    \item I do not expect AI systems to understand my writing style. (-)
    \item I trust AI systems to provide accurate information. (+)
    \item I do not believe using AI for writing is acceptable in academic contexts. (-)
\end{itemize}
\end{promptbox}

\begin{promptbox}{Appropriate Uses of AI for Essay Writing}
\noindent\textbf{For essay writing, I think AI tools should be used:}
\begin{itemize}
    \item Not at all
    \item Check grammar, spelling, or clarity
    \item To offer suggestions on how to improve my writing
    \item Help brainstorm or outline ideas
    \item To rewrite my essay from scratch
    \item To write the entire essay
    \item Other (please specify): \rule{4cm}{0.4pt}
\end{itemize}
\end{promptbox}

\subsubsection*{For Both}

\begin{promptbox}{Writing Confidence and Habits}
\begin{itemize}
    \item I often struggle with structuring my ideas clearly. (-)
    \item I feel confident in my ability to write and edit essays on my own. (+)
    \item I find essay writing time-consuming. (-)
    \item I usually find essay writing enjoyable. (+)
\end{itemize}
\end{promptbox}

\begin{promptbox}{Frequency of LLM Use}
\noindent\textbf{I use LLMs to help me write:}
\begin{itemize}
    \item Daily
    \item Weekly
    \item Monthly
    \item Checked it out a few times
    \item Never
\end{itemize}
\end{promptbox}

\begin{promptbox}{Purposes for Using LLMs}
\noindent\textbf{What do you use LLMs for?}
\begin{itemize}
    \item I don’t use LLMs
    \item General conversation
    \item Search queries / seeking knowledge
    \item Learning or understanding new concepts
    \item Advice
    \item Writing or editing text
    \item Work or productivity tasks
    \item Other (please specify): \rule{4cm}{0.4pt}
\end{itemize}
\end{promptbox}

\subsubsection*{Open-Ended}

\begin{promptbox}{Prior Experience With LLMs}
Please describe the last time you used a large language model (LLM)
such as ChatGPT, Claude, or Gemini. What did you use it for, and in what context (e.g., work, study, personal use)? How helpful was the experience, and why?
\end{promptbox}

\subsection{Post-Study} \label{appendix: post_study}

\subsubsection*{For Both}

\begin{itemize}
    \item I was satisfied with the essay. (+)
    \item I felt the essay was written in my voice. (+)
    \item I found it difficult to organize my thoughts while writing. (-)
    \item Writing this essay was a struggle for me. (-)
\end{itemize}

\noindent\textbf{How creative do you feel you were in writing the essay?}
\begin{itemize}
    \item Very creative
    \item Somewhat creative
    \item Neither creative nor uncreative
    \item Somewhat uncreative
    \item Not at all creative
\end{itemize}

\subsubsection*{Open-Ended}

\noindent Please describe your experience writing this essay.\\
Comment on: How well does the essay reflect your own views and writing style? How much effort did you put into writing it? Did you learn anything during the process?

\subsubsection*{For LLM Users}

\begin{promptbox}{Estimated LLM Contribution}
What percentage of the document would you say was LLM-generated?

\begin{itemize}
    \item 0\%
    \item 20\%
    \item 40\%
    \item 60\%
    \item 80\%
    \item 100\%
\end{itemize}
\end{promptbox}

\begin{promptbox}{Perceptions of LLM Assistance}
The following statements were rated on a Likert scale
\cite{likert1932technique} from \emph{strongly disagree} to \emph{strongly agree}.

\begin{itemize}
    \item The LLM helped me generate ideas more effectively. (+)
    \item The model’s feedback improved the quality of my essay. (+)
    \item The LLM’s suggestions were irrelevant to my goals. (-)
    \item I learned something new about writing from using the LLM. (+)
    \item I felt I had less control of the essay writing process when working with the LLM. (-)
    \item The model took too much initiative in generating content. (-)
    \item I felt that the LLM and I were collaborating as partners. (+)
    \item The model’s behavior matched my preferred level of assistance. (+)
    \item I did not trust the LLM’s writing suggestions. (-)
    \item I would not use this LLM again for a similar writing task. (-)
    \item Using the LLM made me question what counts as original writing. (-)
    \item I would disclose AI assistance if submitting this essay academically. (+)
\end{itemize}
\end{promptbox}

\subsection{Consent Form} \label{appendix: consent_form}

Before participating in this study, please read the following consent information:

\paragraph{Introduction}
My name is Isadora White. I am a PhD Student at the University of California, San Diego, in the Computer Science and Engineering Department. I am planning to conduct a research study, which I invite you to take part in.

\paragraph{Purpose}
The purpose of this study is to understand attitudes towards writing essays and analyze essay writing.

\paragraph{Procedures:} You will be asked to answer some pre-study questions about your attitudes toward AI and writing before you begin the essay. Then, you will write the essay. You may not use other sources, such as the internet, to inform your essay. Study time: The estimated study completion time has been displayed to you in the Prolific interface (up to one hour). Study location: You will participate online, from the comfort of your current location.

\paragraph{Benefits:} There is no direct benefit to you (other than compensation) from participating in this study. We hope that the information gained from the study will help us better understand how people write essays.

\paragraph{Risks/Discomforts}
This study represents minimal risk to you. As with all research, there is the risk of an unintended breach of confidentiality. However, we are taking precautions to minimize this risk (see below).

\paragraph{Confidentiality} The data we collect will be stored on password-protected servers. Once the research is complete, we intend to scrub the data of all identifiable information. We will keep only the recorded survey responses, as well as a freshly generated identifier for each subject. The de-identified data will be retained indefinitely for possible use in future research done by ourselves or others. Parts of this cleaned dataset may be made public as part of the publishing process. No guarantees can be made regarding the interception of data sent via the Internet by any third parties.

\paragraph{Compensation}
We compensate workers based on the estimated duration of completing the study. 

\paragraph{Rights} Participation in research is completely voluntary. You have the right to decline to participate or to withdraw at any point in this study without penalty or loss of benefits to which you are otherwise entitled.

\paragraph{Questions: }If you have any questions or concerns about this study, or in case anything goes wrong with the online interface, you can contact Isadora White at i2white@ucsd.edu.

\paragraph{IRB review:} This study was determined to be exempt by the IRB under the University of California, San Diego’s Institutional Review Board (IRB). The IRB number is 813728. You should save a copy of this consent form for your records. By continuing in this study and clicking the checkbox below, you are indicating that you have read and understand the consent form and agree to participate in this study.



\newpage

\subsection{User Study Demographics}\label{appendix:demographics}
\begin{figure}[htbp]
    \vspace{6pt} 
        \centering
    \includegraphics[width=0.99\linewidth]{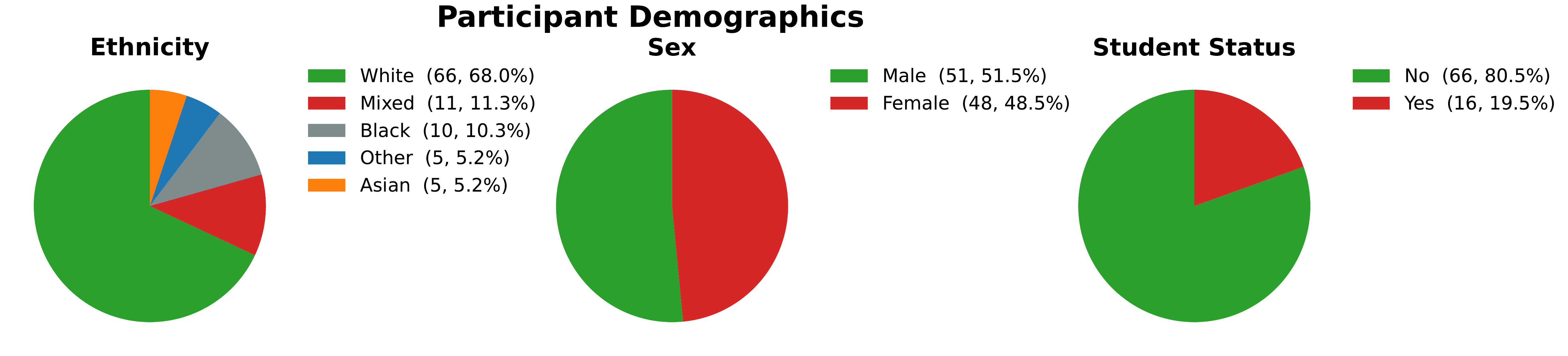}
    \caption{We report the demographic statistics for ethnicity, sex, and student status. Some participants do not report their demographics in certain categories, so it does not all add up to 100 users.}
    \label{fig:placeholder}
\end{figure}

\section{ArgRewrite-v2 Analysis}
\subsection{Generating LLM Drafts}\label{appendix:argrewrite_prompts}

\begin{promptbox}{LLM Draft Generation Prompts}
The following prompts were used to generate LLM drafts:

\begin{enumerate}
    \item \textbf{Expert}: You are given an essay draft. Revise the essay.
    \item \textbf{Minimal}: You are given an essay draft. Rewrite this essay and keep to a similar word count.
    \item \textbf{Grammar}: You are given an essay draft. Revise this essay for grammar.
    \item \textbf{Completion}: You are given an essay draft. The text is only the first paragraph. Finish the rest of the essay.
    \item \textbf{Expansion}: You are given an essay draft. Expand on the ideas in the following draft with more detail and depth.
\end{enumerate}
\end{promptbox}

\subsection{Semantic Shifts with Gemini-004 Embedding}\label{appendix:semantic_shifts_gemini004}

\FloatBarrier
\begin{figure}[H]
    \centering
    \includegraphics[width=\textwidth]{all_figures/arg-rewrite-embedding/gemini004/semantic_shift_general_with_feedback_embedding_gemini004.png}
    \vspace{0.5em}
    \includegraphics[width=\textwidth]{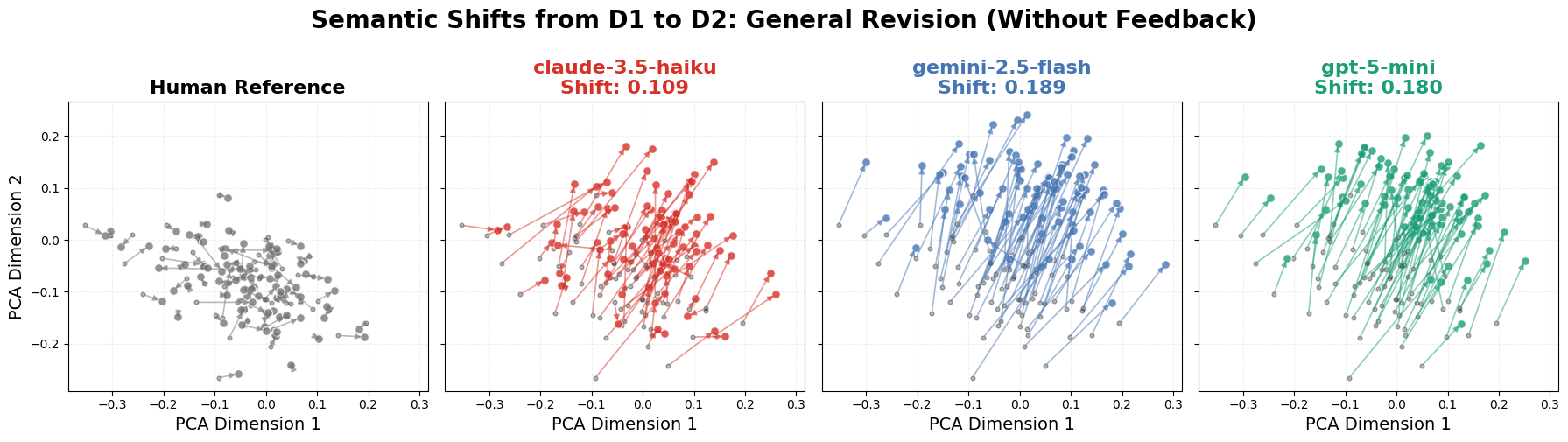}
    \caption{Semantic shifts from D1 to D2 for \textbf{General revisions} with \textbf{Gemini-004 Embedding}. Top: with expert feedback. Bottom: without expert feedback.}
    \label{fig:appendix-semantic-general}
\end{figure}

\FloatBarrier
\begin{figure}[H]
    \centering
    \includegraphics[width=\textwidth]{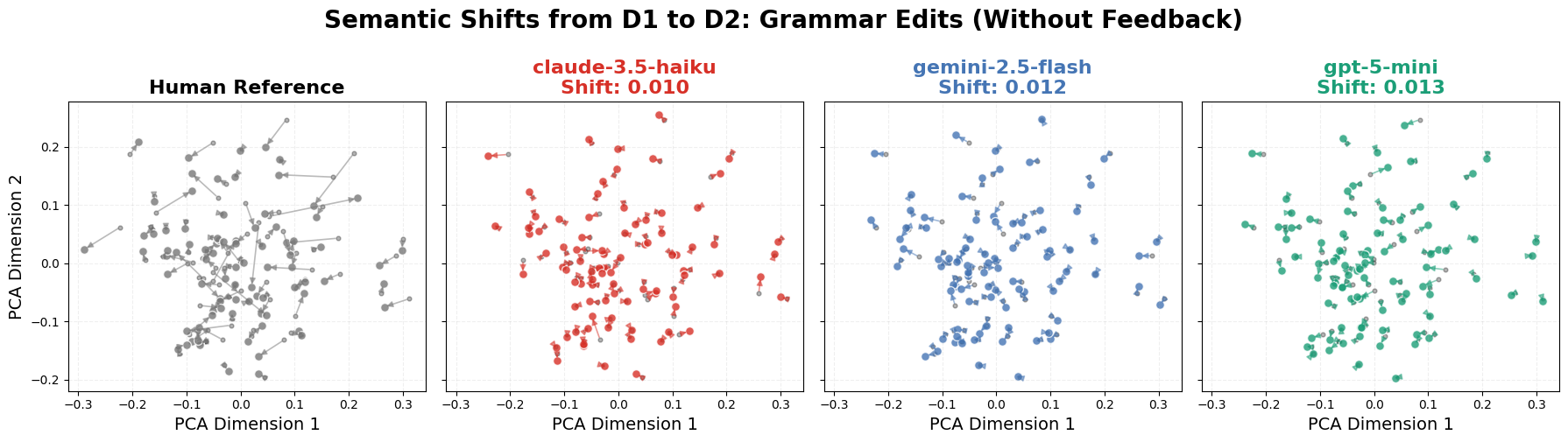}
    \vspace{0.5em}
    \includegraphics[width=\textwidth]{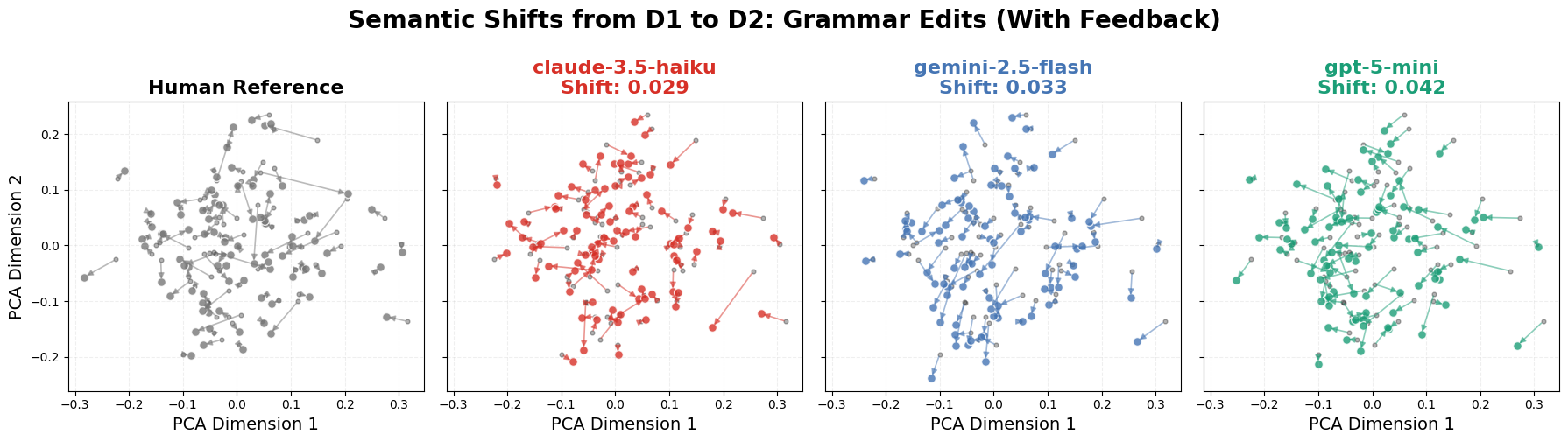}
    \caption{Semantic shifts from D1 to D2 for \textbf{Grammar edits} with with \textbf{Gemini-004 Embedding}. Top: with expert feedback. Bottom: without expert feedback.}
    \label{fig:appendix-semantic-grammar}
\end{figure}

\FloatBarrier
\begin{figure}[H]
    \centering
    \includegraphics[width=\textwidth]{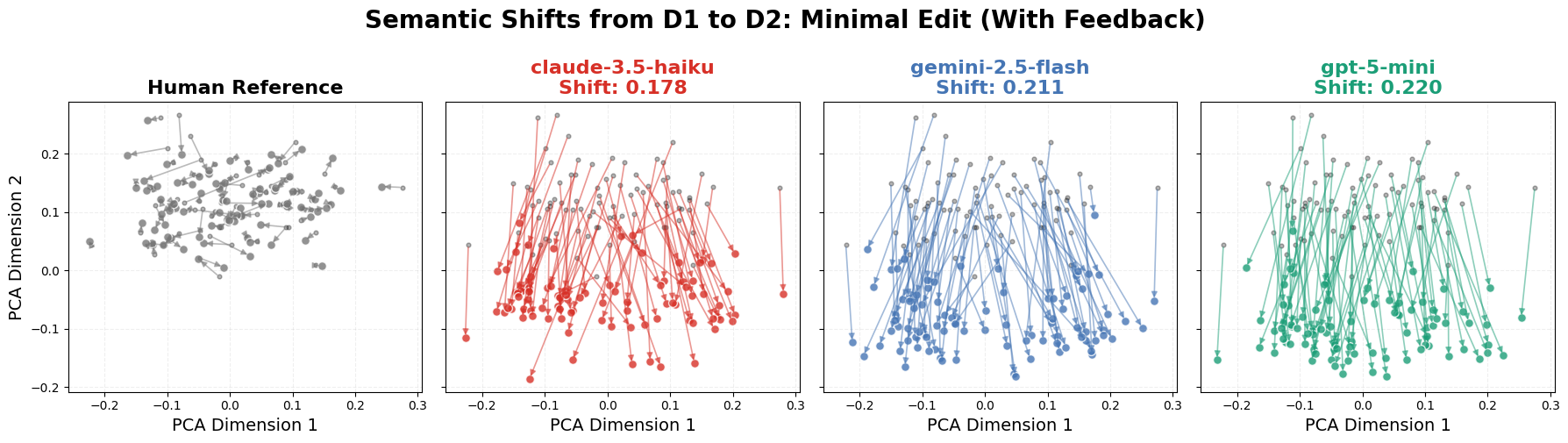}
    \vspace{0.5em}
    \includegraphics[width=\textwidth]{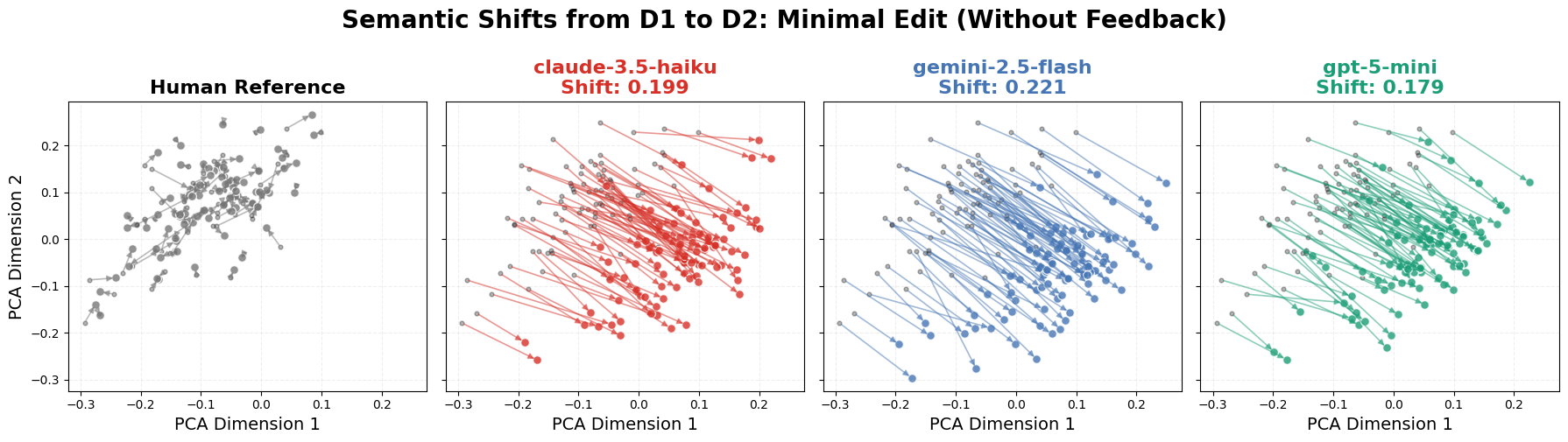}
    \caption{Semantic shifts from D1 to D2 for\textbf{Miminal revisions} with \textbf{Gemini-004 Embedding}. Top: with expert feedback. Bottom: without expert feedback.}
    \label{fig:appendix-semantic-completion}
\end{figure}

\FloatBarrier
\begin{figure}[H]
    \centering
    \includegraphics[width=\textwidth]{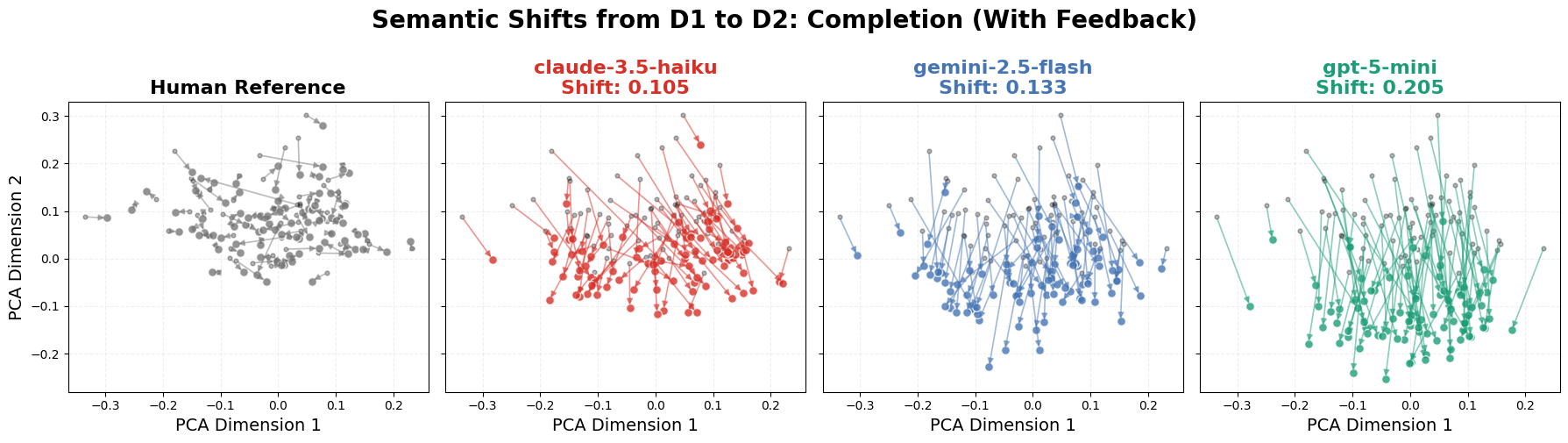}
    \vspace{0.5em}
    \includegraphics[width=\textwidth]{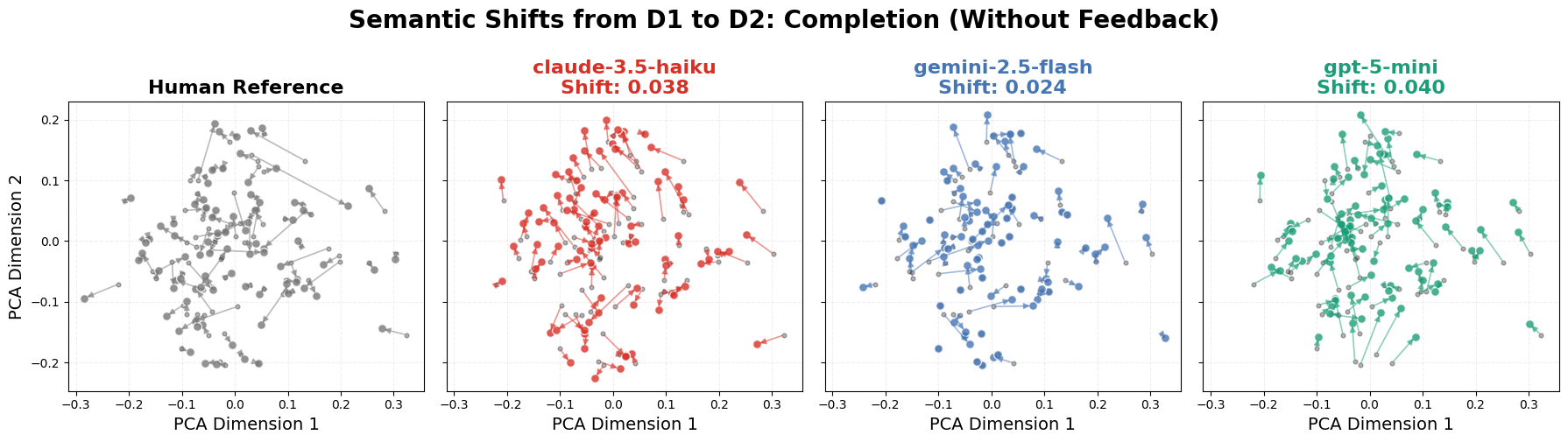}
    \caption{Semantic shifts from D1 to D2 for \textbf{Completion revisions} with \textbf{Gemini-004 Embedding}. Top: with expert feedback. Bottom: without expert feedback.}
    \label{fig:appendix-semantic-completion}
\end{figure}

\FloatBarrier
\begin{figure}[H]
    \centering
    \includegraphics[width=\textwidth]{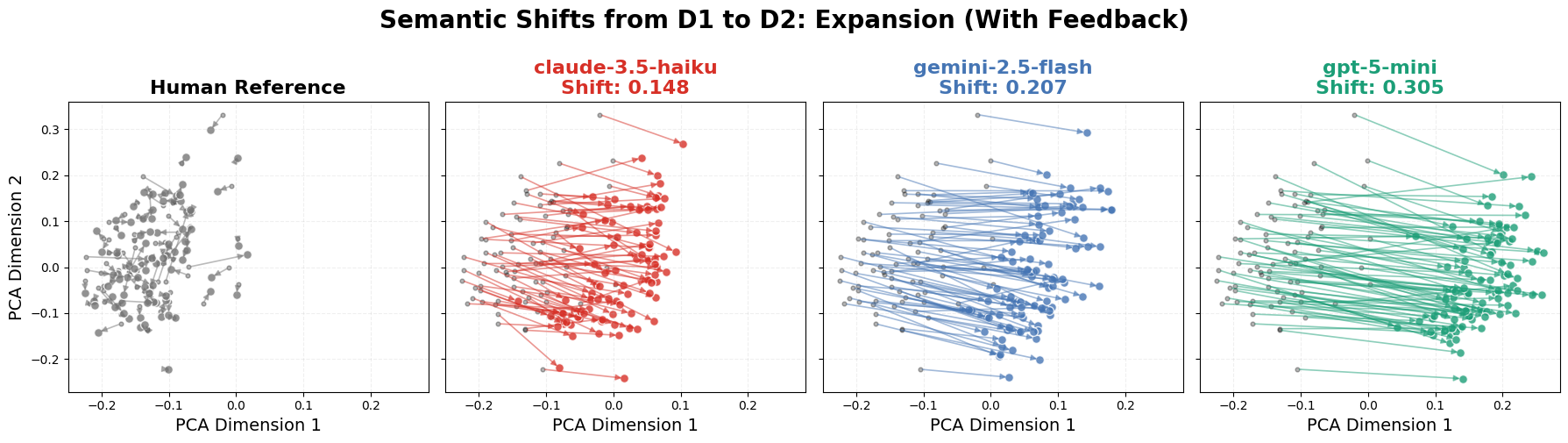}
    \vspace{0.5em}
    \includegraphics[width=\textwidth]{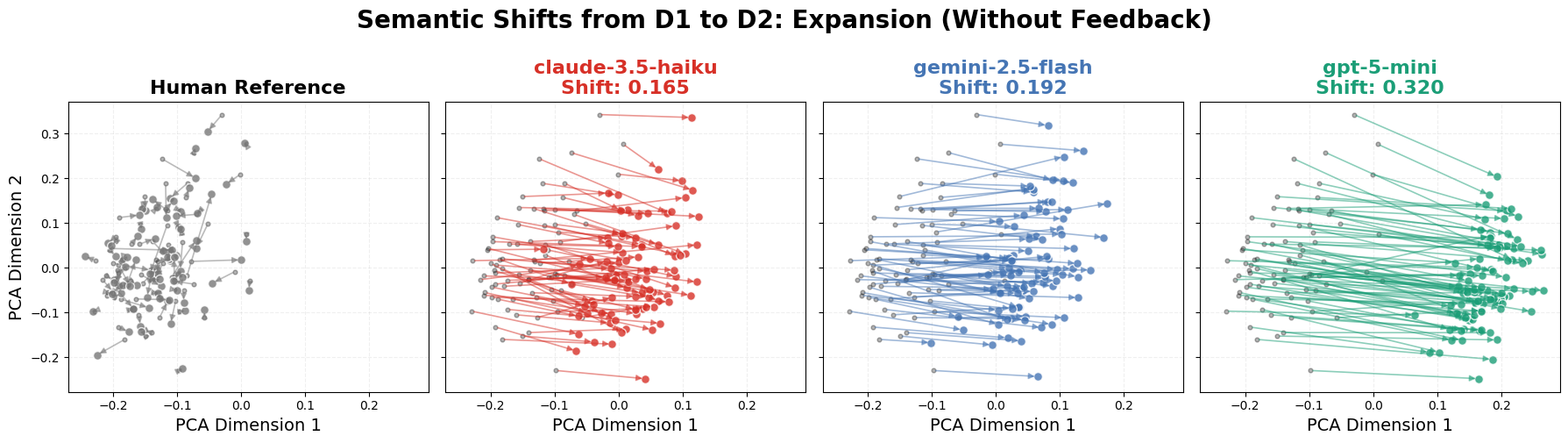}
    \caption{Semantic shifts from D1 to D2 for \textbf{Expansion revisions} with \textbf{Gemini-004 Embedding}. Top: with expert feedback. Bottom: without expert feedback.}
    \label{fig:appendix-semantic-completion}
\end{figure}

\subsection{Semantic Shifts with MiniLM-L6-v2 Embedding}\label{appendix:semantic_shifts_minilm}

\FloatBarrier
\begin{figure}[H]
    \centering
    \includegraphics[width=\textwidth]{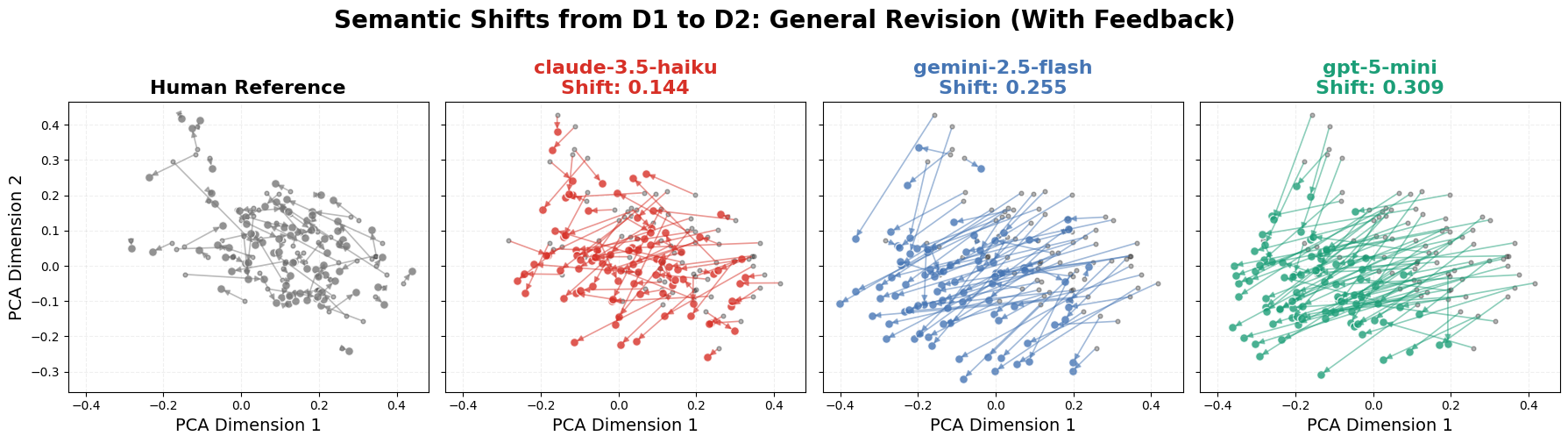}
    \vspace{0.5em}
    \includegraphics[width=\textwidth]{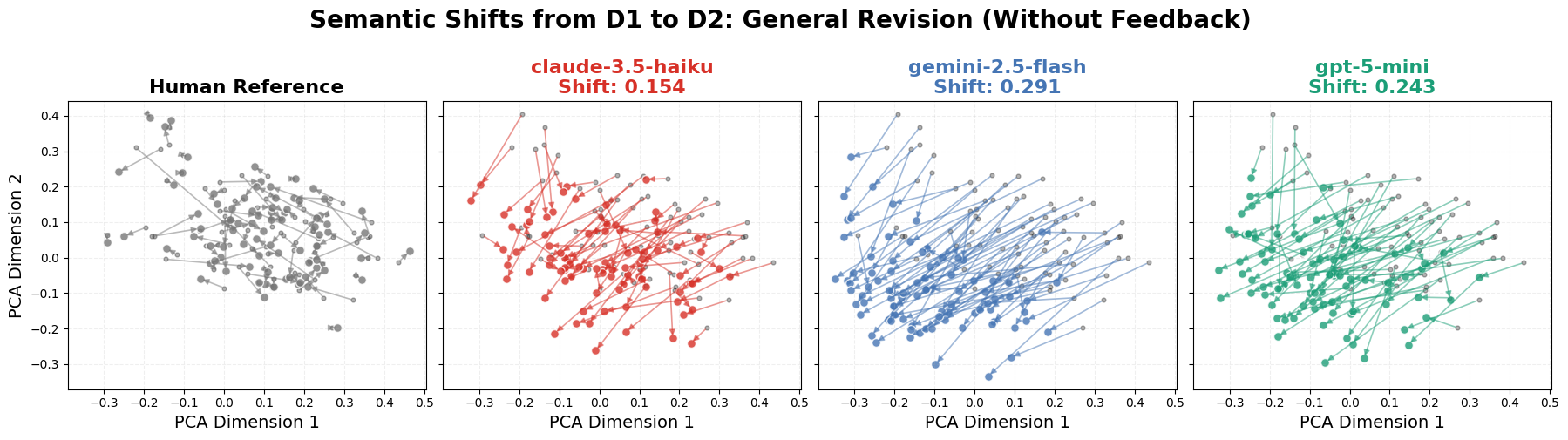}
    \caption{Semantic shifts from D1 to D2 for \textbf{General revisions} with \textbf{MiniLM-L6-004 Embedding}. Top: with expert feedback. Bottom: without expert feedback.}
    \label{fig:appendix-semantic-general}
\end{figure}

\FloatBarrier
\begin{figure}[H]
    \centering
    \includegraphics[width=\textwidth]{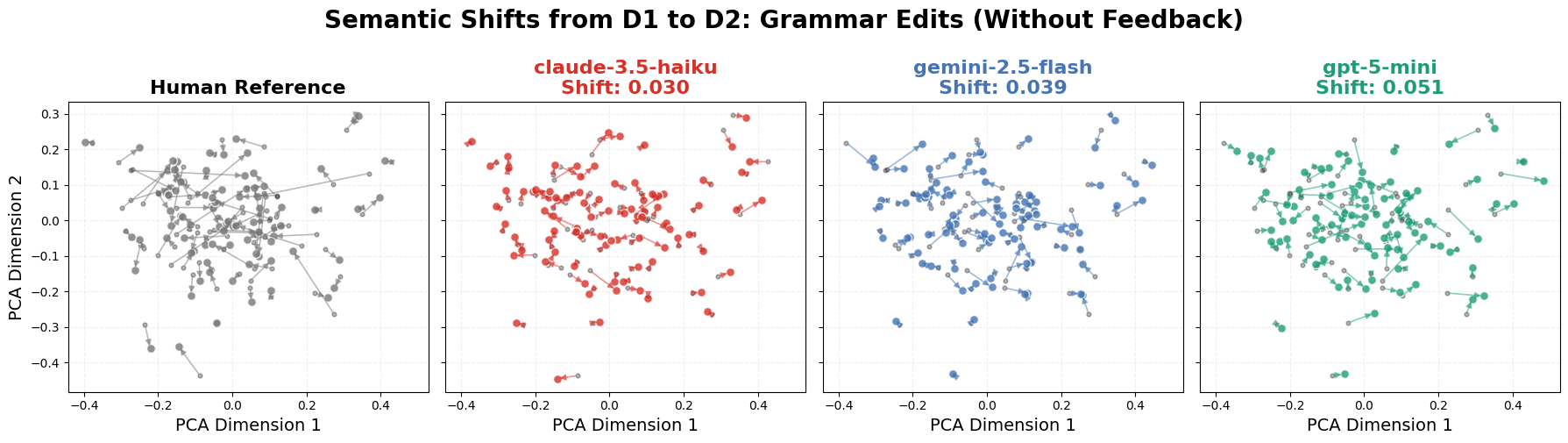}
    \vspace{0.5em}
    \includegraphics[width=\textwidth]{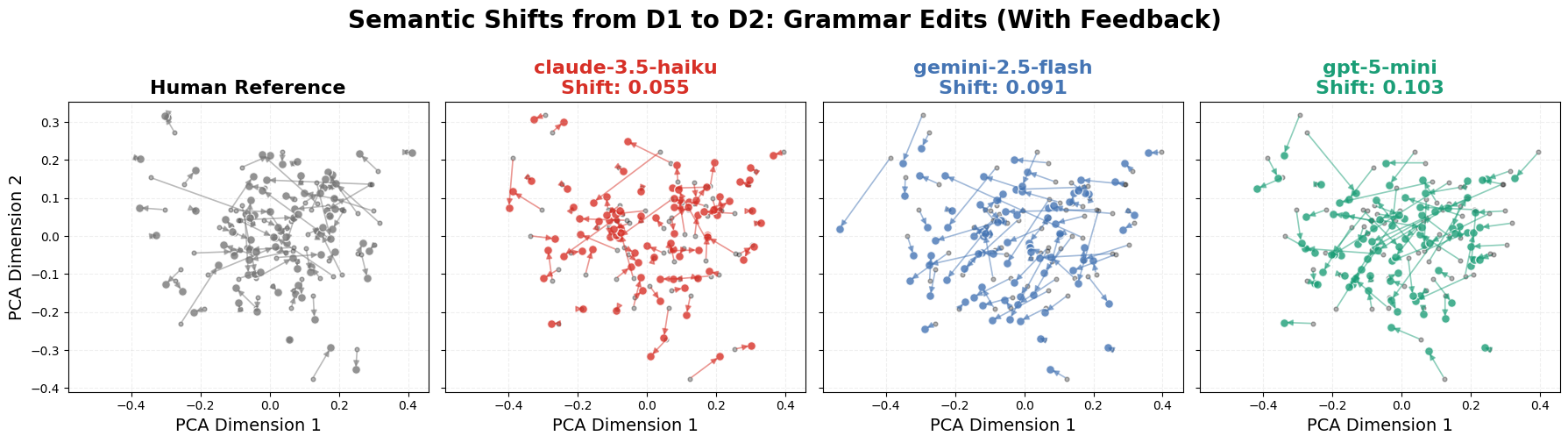}
    \caption{Semantic shifts from D1 to D2 for \textbf{Grammar edits} with \textbf{MiniLM-L6-004 Embedding}. Top: with expert feedback. Bottom: without expert feedback.}
    \label{fig:appendix-semantic-grammar}
\end{figure}

\FloatBarrier
\begin{figure}[H]
    \centering
    \includegraphics[width=\textwidth]{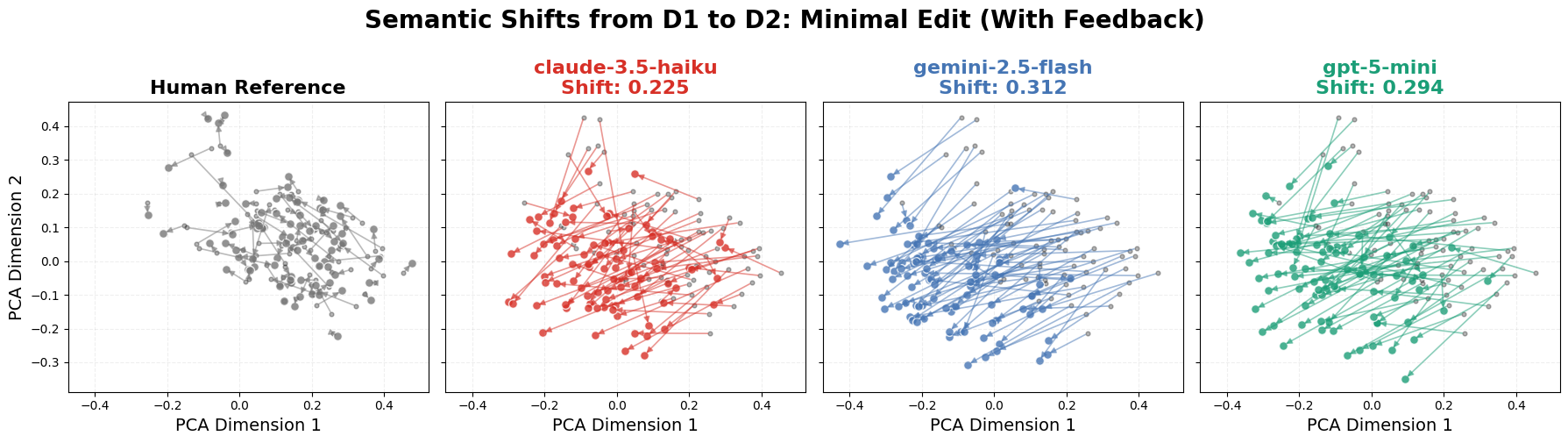}
    \vspace{0.5em}
    \includegraphics[width=\textwidth]{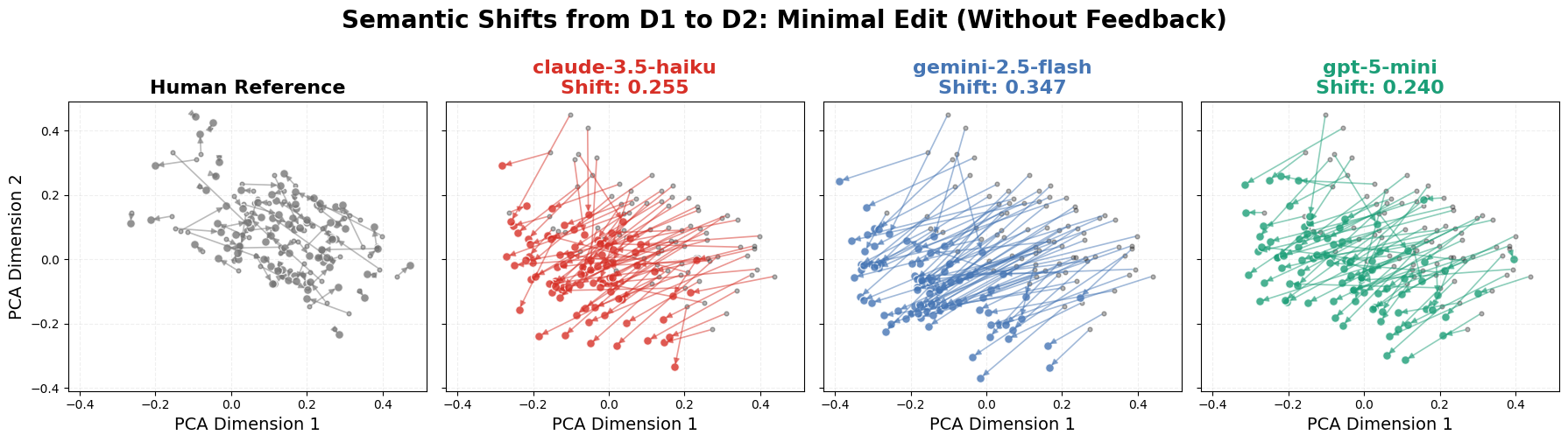}
    \caption{Semantic shifts from D1 to D2 for\textbf{Miminal revisions} with \textbf{MiniLM-L6-004 Embedding}. Top: with expert feedback. Bottom: without expert feedback.}
    \label{fig:appendix-semantic-completion}
\end{figure}

\FloatBarrier
\begin{figure}[H]
    \centering
    \includegraphics[width=\textwidth]{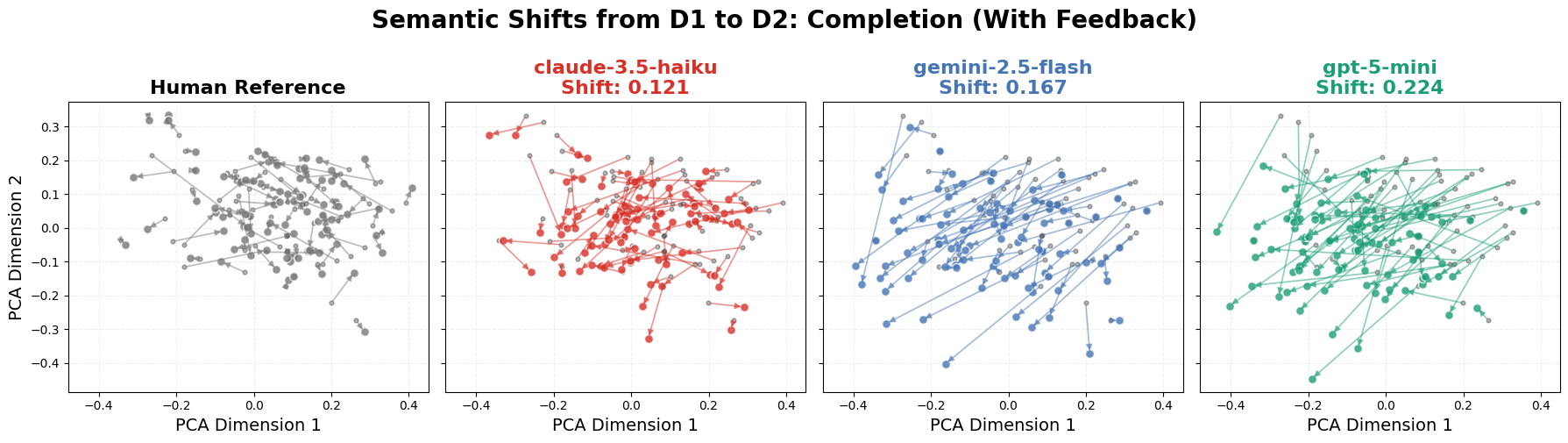}
    \vspace{0.5em}
    \includegraphics[width=\textwidth]{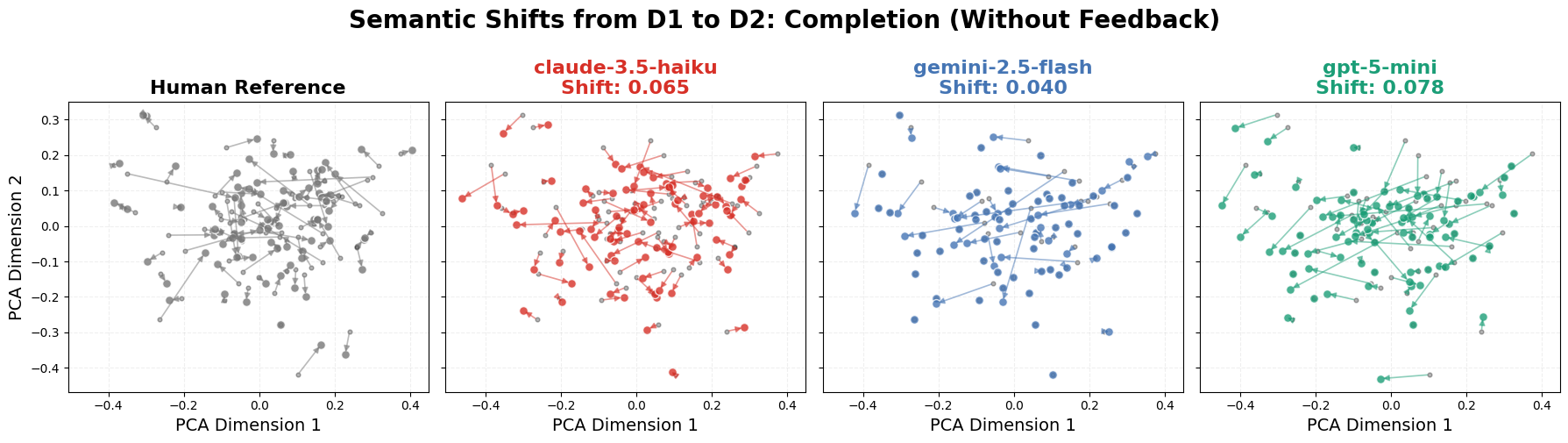}
    \caption{Semantic shifts from D1 to D2 for \textbf{Completion revisions} with \textbf{MiniLM-L6-004 Embedding}. Top: with expert feedback. Bottom: without expert feedback.}
    \label{fig:appendix-semantic-completion}
\end{figure}

\begin{figure}[H]
    \centering
    \includegraphics[width=\textwidth]{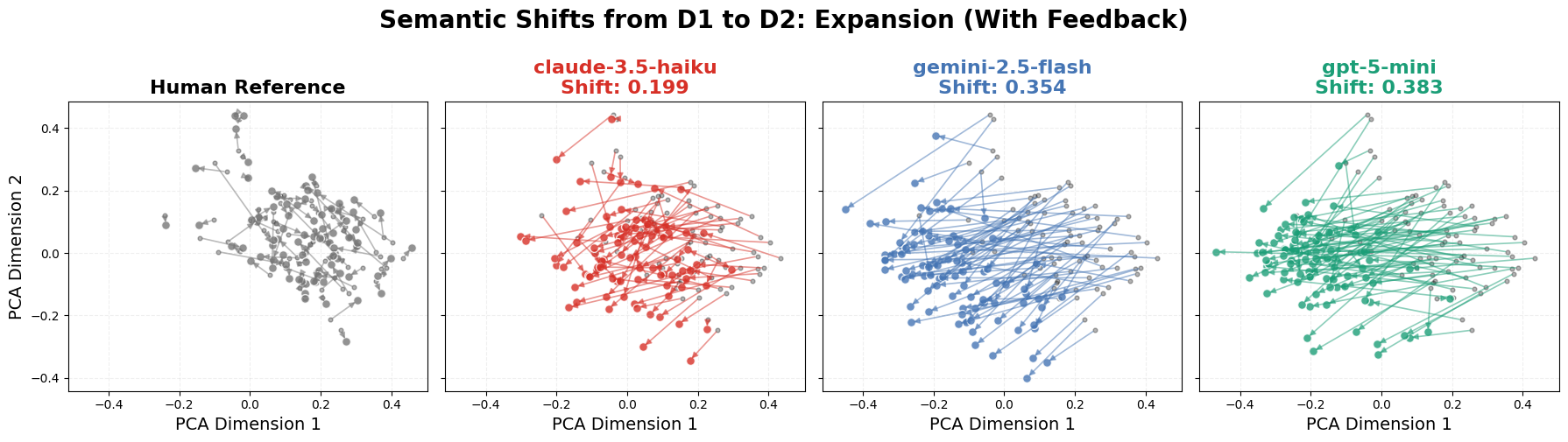}
    \vspace{0.5em}
    \includegraphics[width=\textwidth]{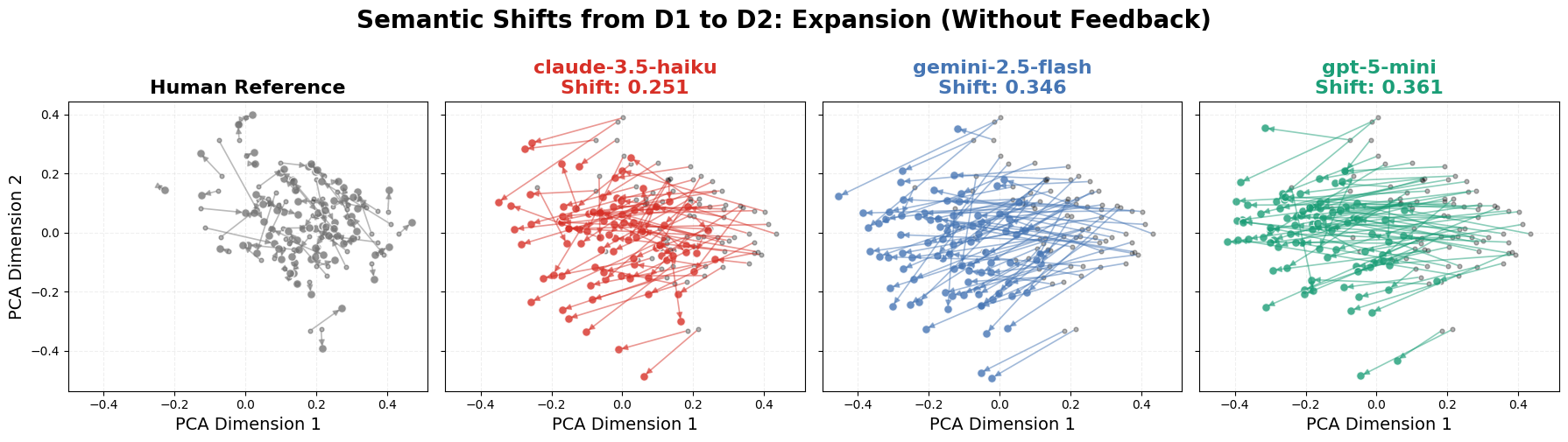}
    \caption{Semantic shifts from D1 to D2 for \textbf{Expansion revisions} with \textbf{MiniLM-L6-004 Embedding}. Top: with expert feedback. Bottom: without expert feedback.}
    \label{fig:appendix-semantic-completion}
\end{figure}

\subsection{Semantic Shifts Across Settings / Model }\label{appendix:semantic_shifts_across_settings_per_model}

\FloatBarrier
\begin{figure}[H]
    \centering
    \includegraphics[width=\textwidth]{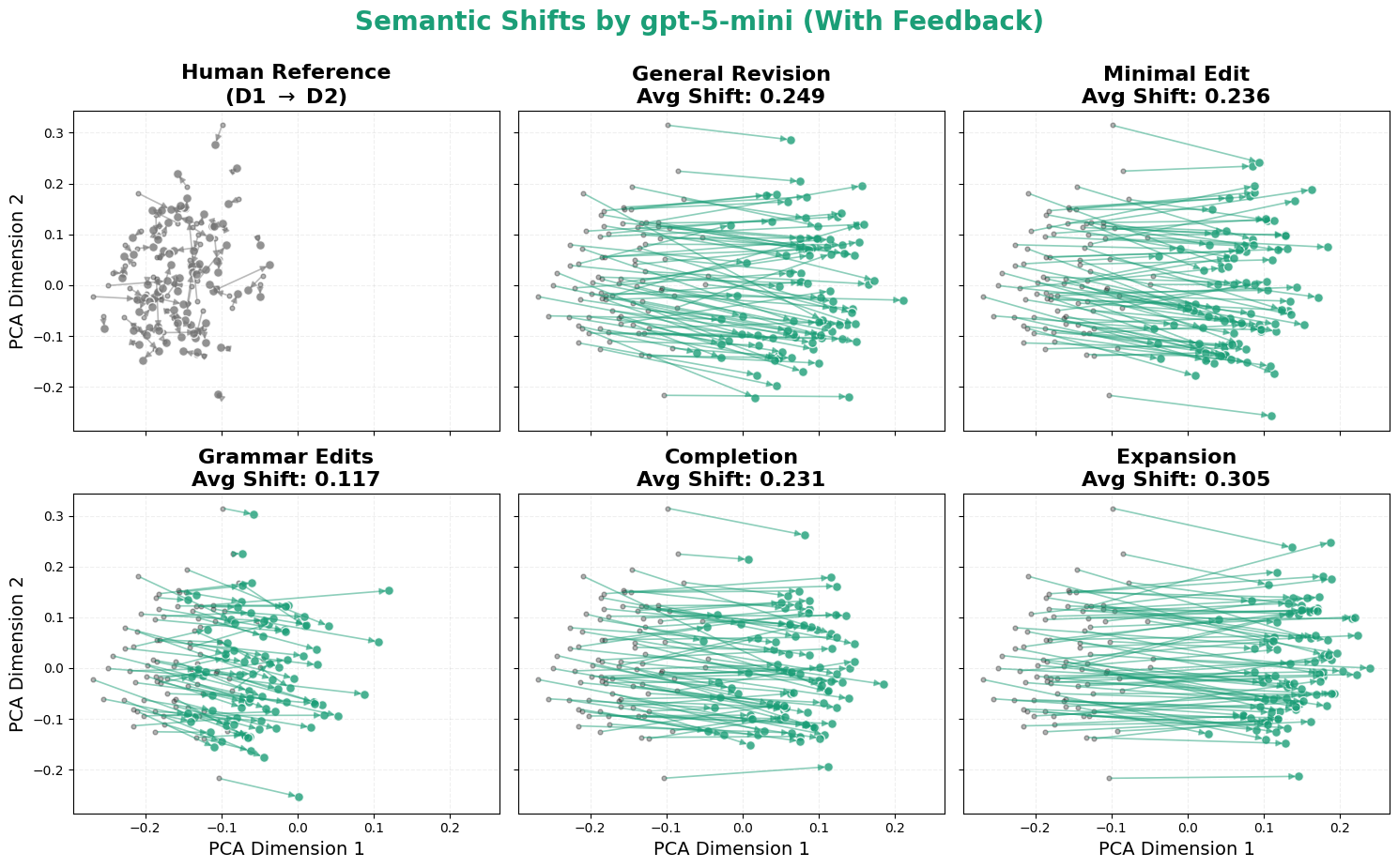}
    \vspace{0.6em}
    \includegraphics[width=\textwidth]{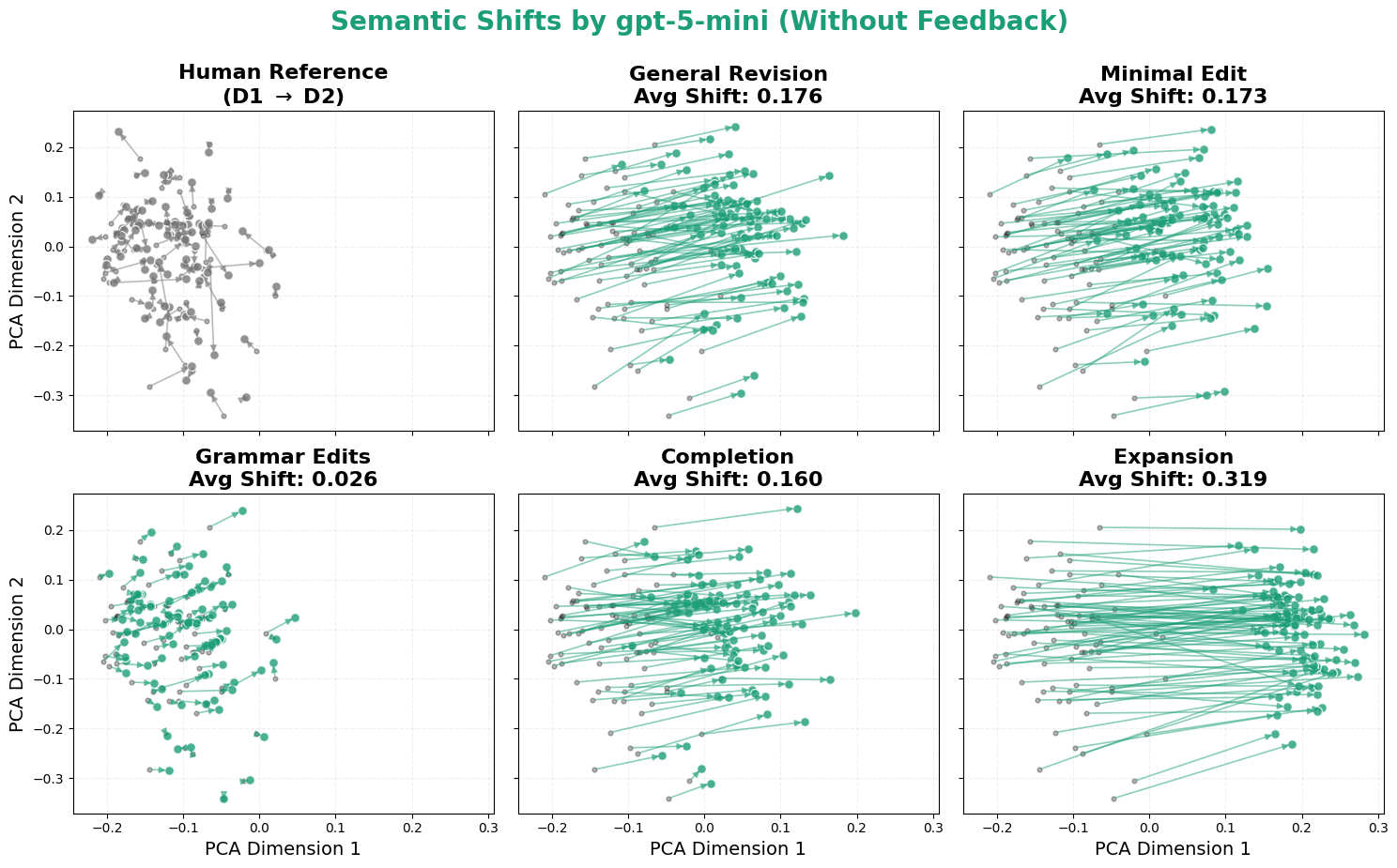}
    \caption{Semantic shifts from D1 to D2 produced by \textbf{GPT-5-mini}. 
    Top: revisions with expert feedback. Bottom: revisions without feedback.}
    \label{fig:appendix-embedding-gpt5mini}
\end{figure}

\FloatBarrier
\begin{figure}[H]
    \centering
    \includegraphics[width=\textwidth]{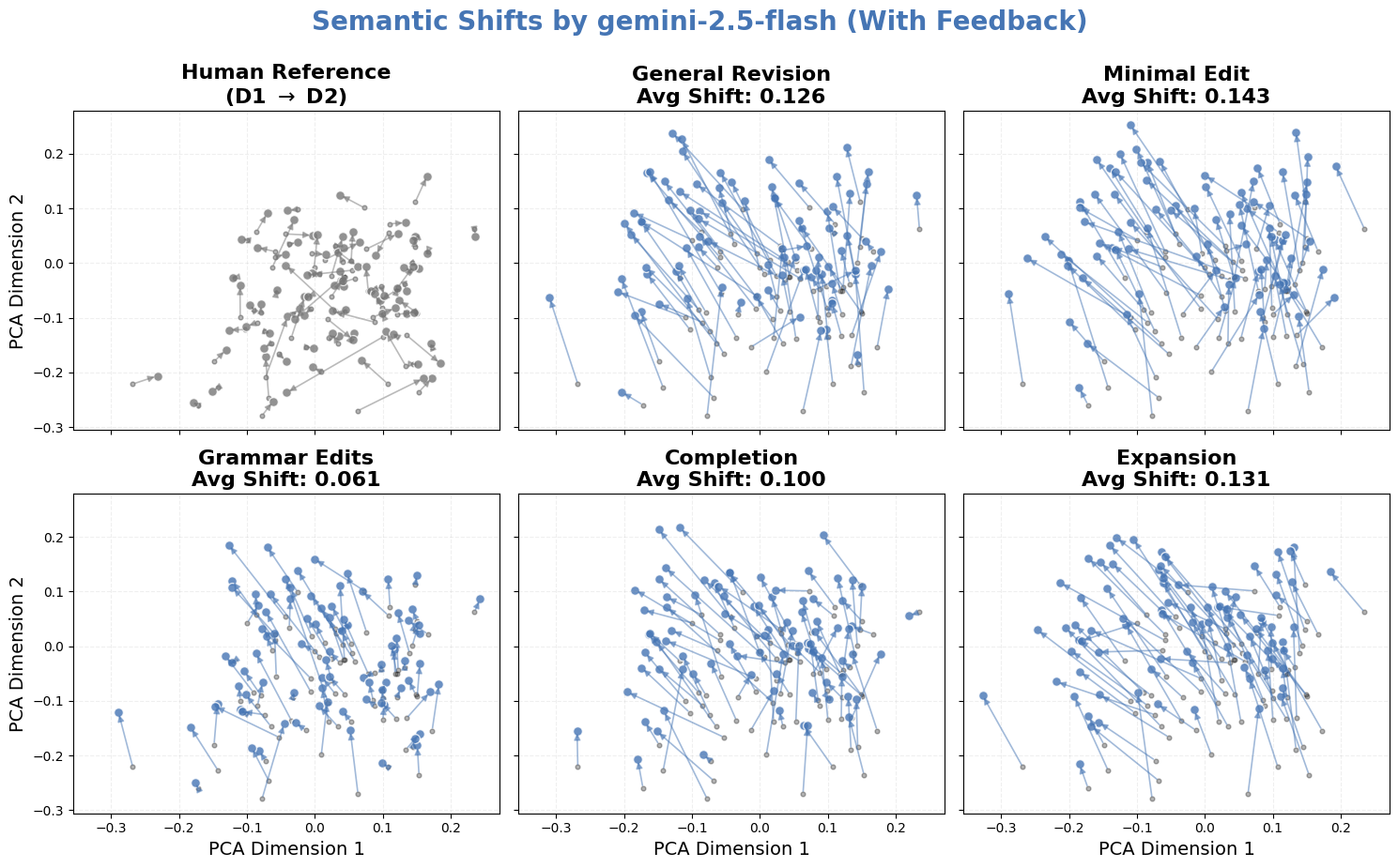}
    \vspace{0.6em}
    \includegraphics[width=\textwidth]{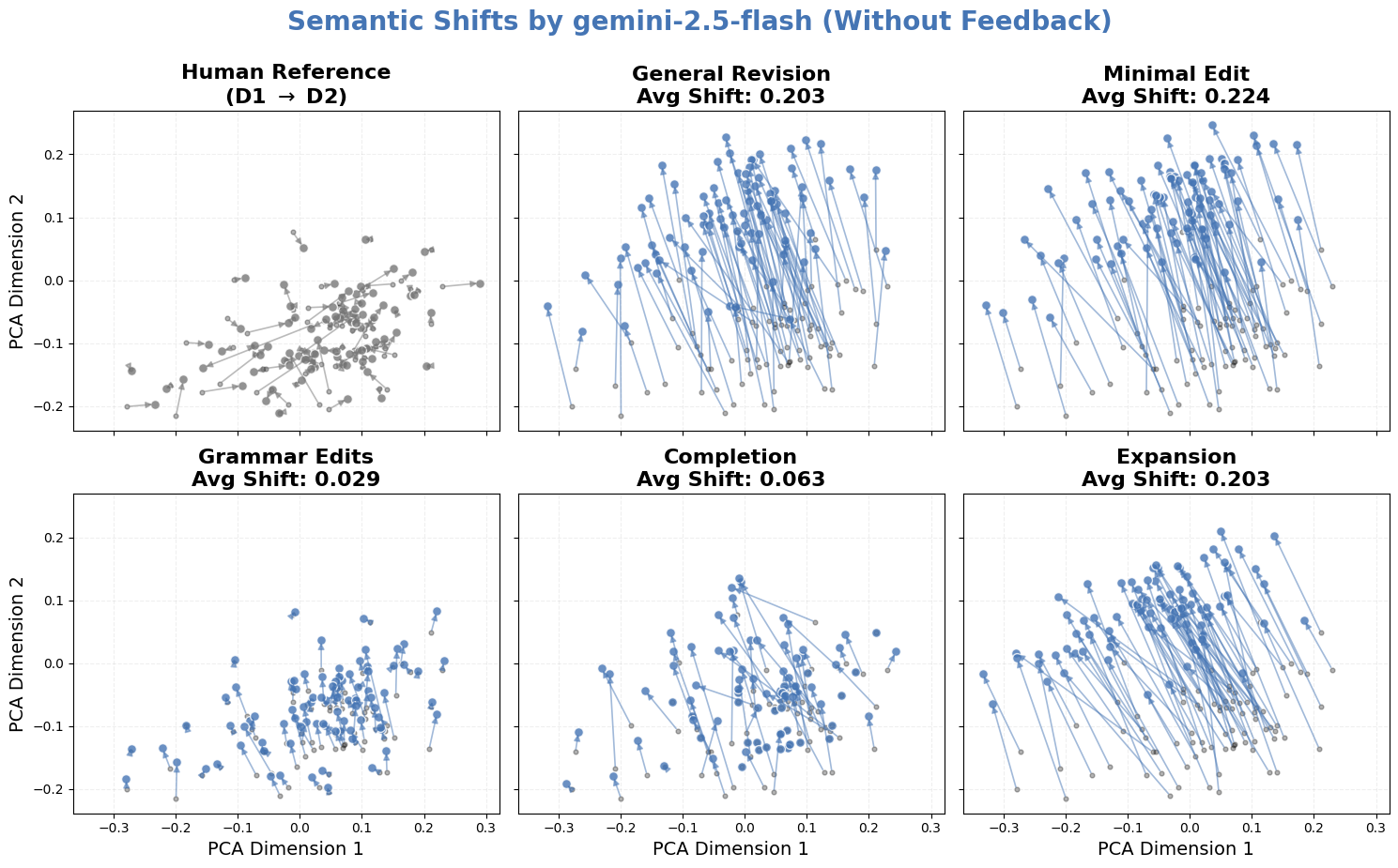}
    \caption{Semantic shifts from D1 to D2 produced by \textbf{Gemini-2.5-Flash}. 
    Top: revisions with expert feedback. Bottom: revisions without feedback.}
    \label{fig:appendix-embedding-gemini}
\end{figure}

\FloatBarrier
\begin{figure}[H]
    \centering
    \includegraphics[width=\textwidth]{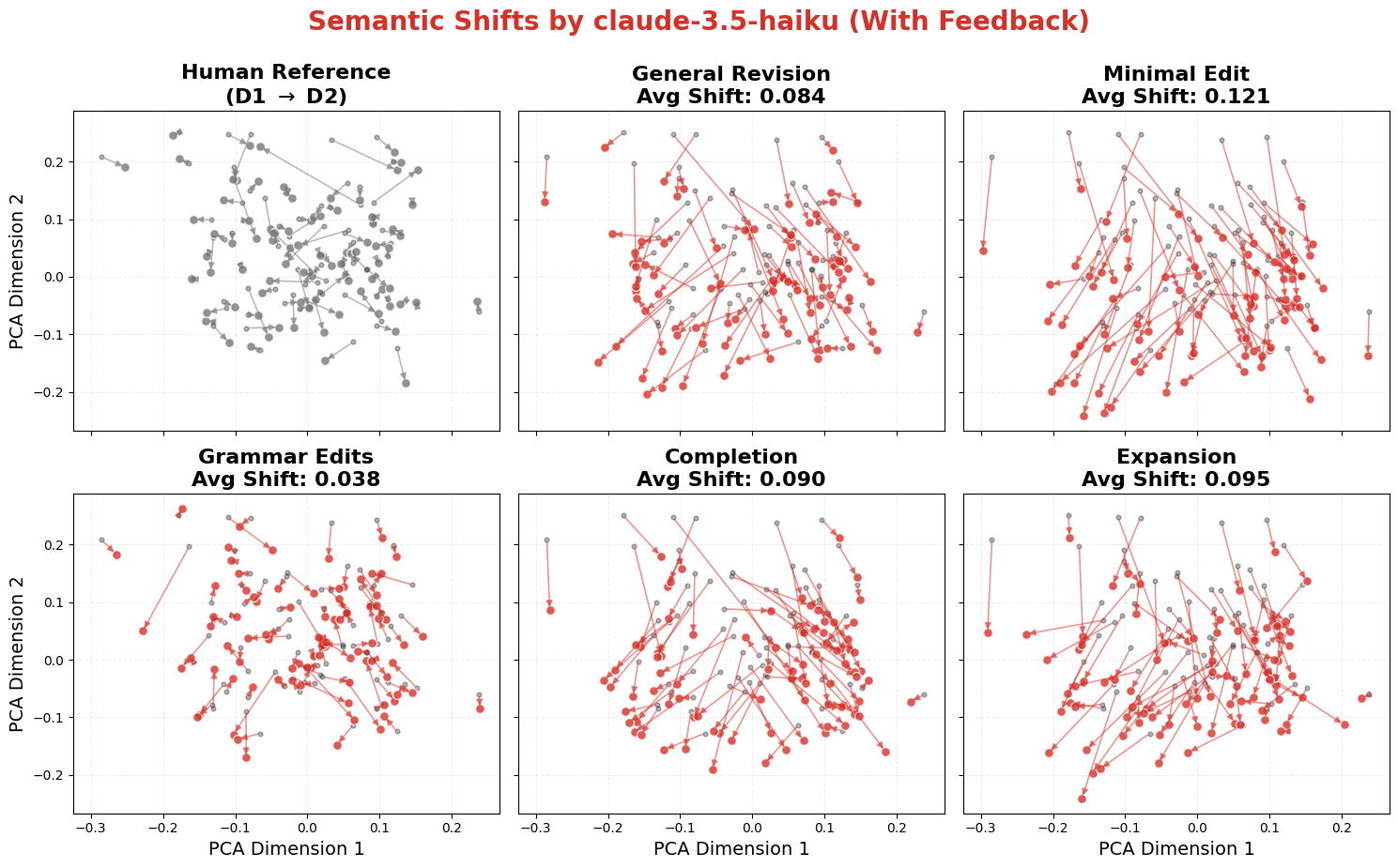}

    \vspace{0.6em}

    \includegraphics[width=\textwidth]{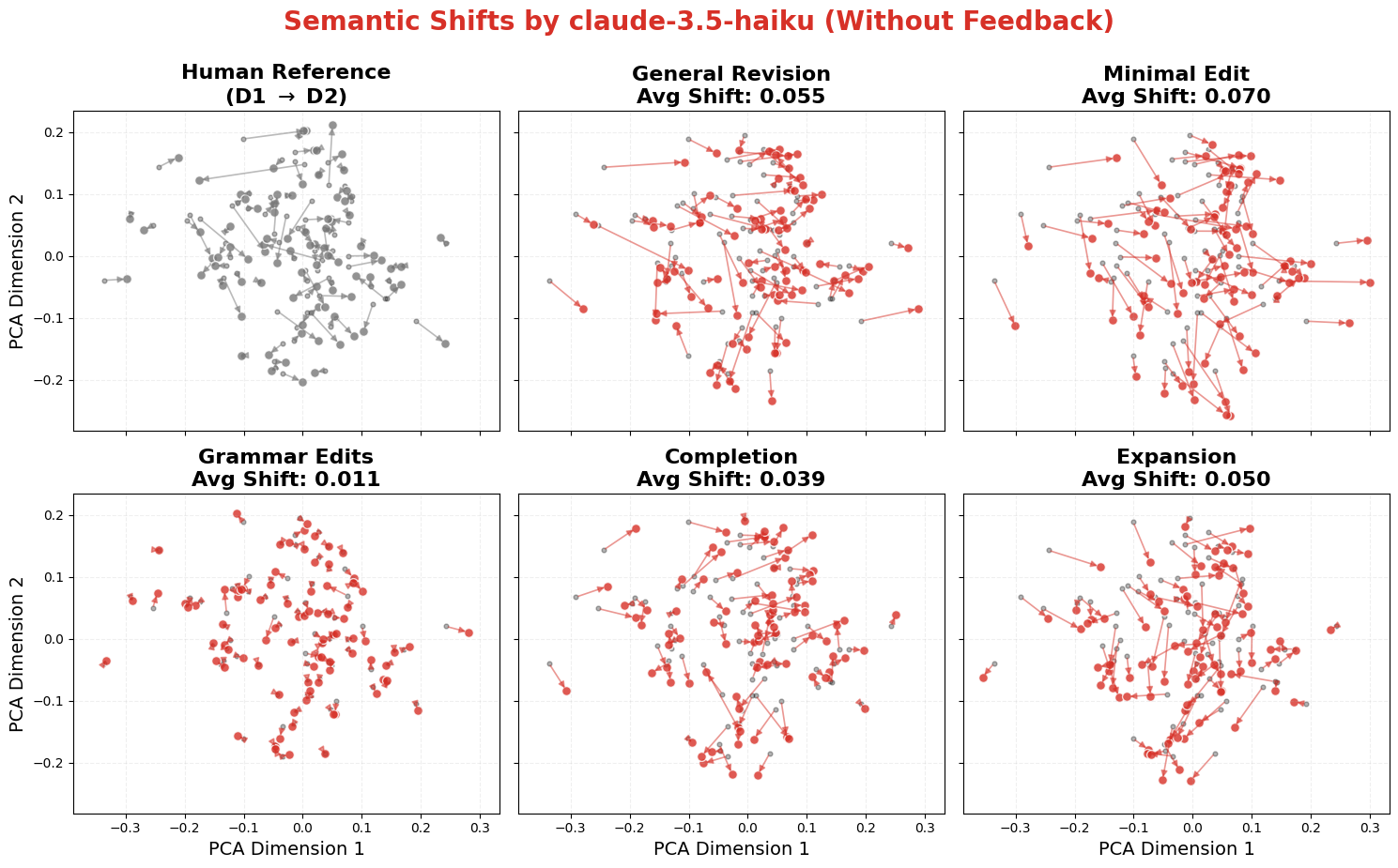}
    \caption{Semantic shifts from D1 to D2 produced by \textbf{Claude-3.5-Haiku}. 
    Top: revisions with expert feedback. Bottom: revisions without feedback.}
    \label{fig:appendix-embedding-claude}
\end{figure}

\subsection{Jensen-Shannon Divergence}\label{appendix:jsd}

\FloatBarrier
\begin{figure}[H]
    \includegraphics[width=\textwidth]{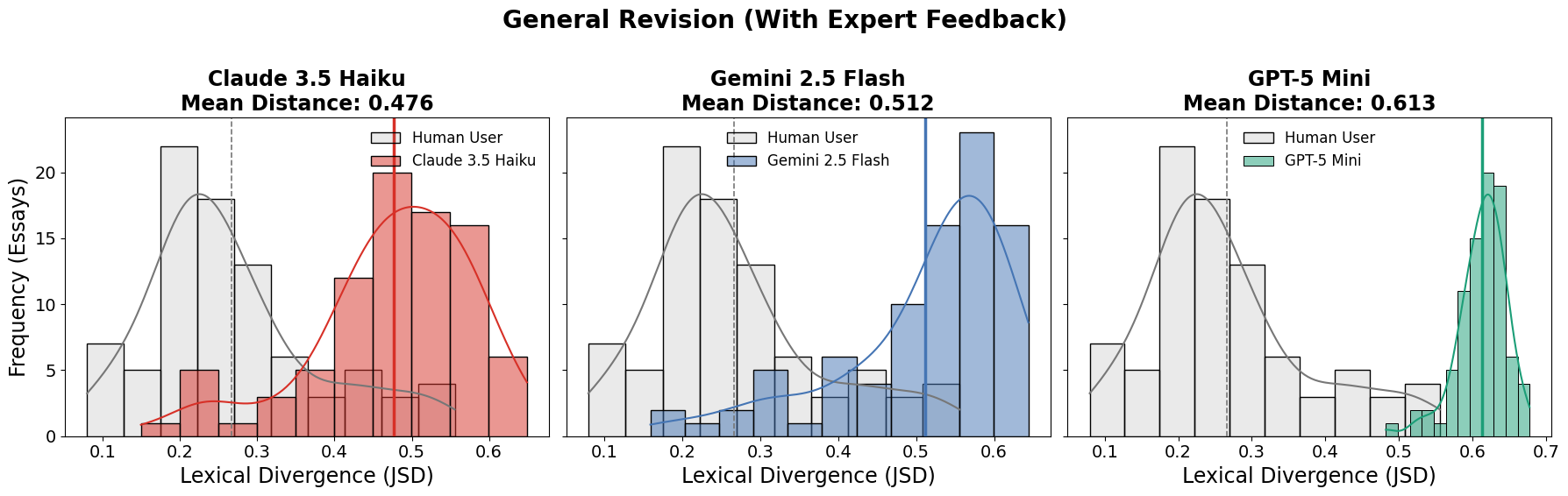}
    \vspace{0.6em}
    \includegraphics[width=\textwidth]{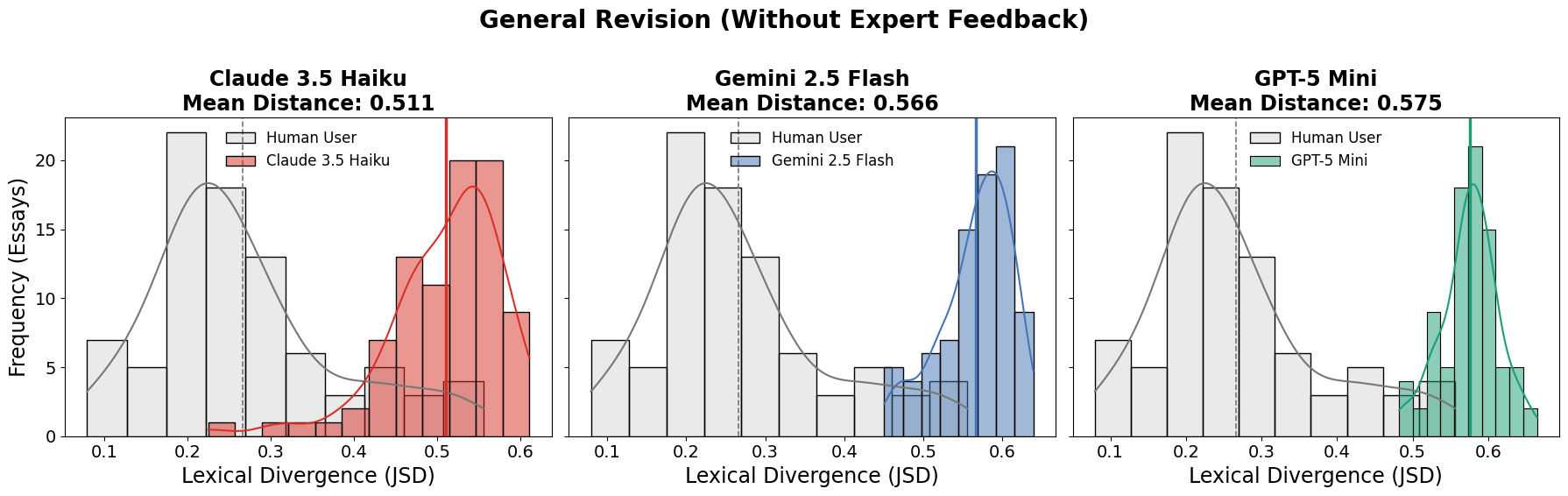}
    \captionof{figure}{Lexical divergence (JSD) distributions for \textbf{General revisions}. 
    Top: with expert feedback. Bottom: without expert feedback.}
    \label{fig:appendix-jsd-general}
\end{figure}

\FloatBarrier
\begin{figure}[H]
    \includegraphics[width=\textwidth]{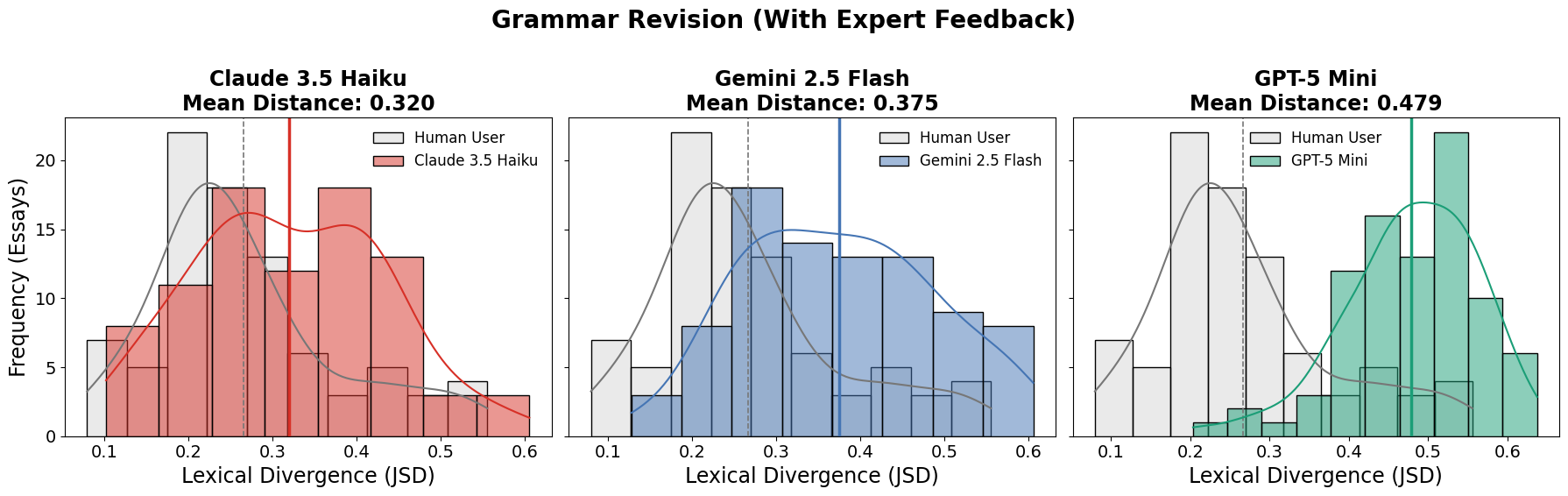}
    \vspace{0.6em}
    \includegraphics[width=\textwidth]{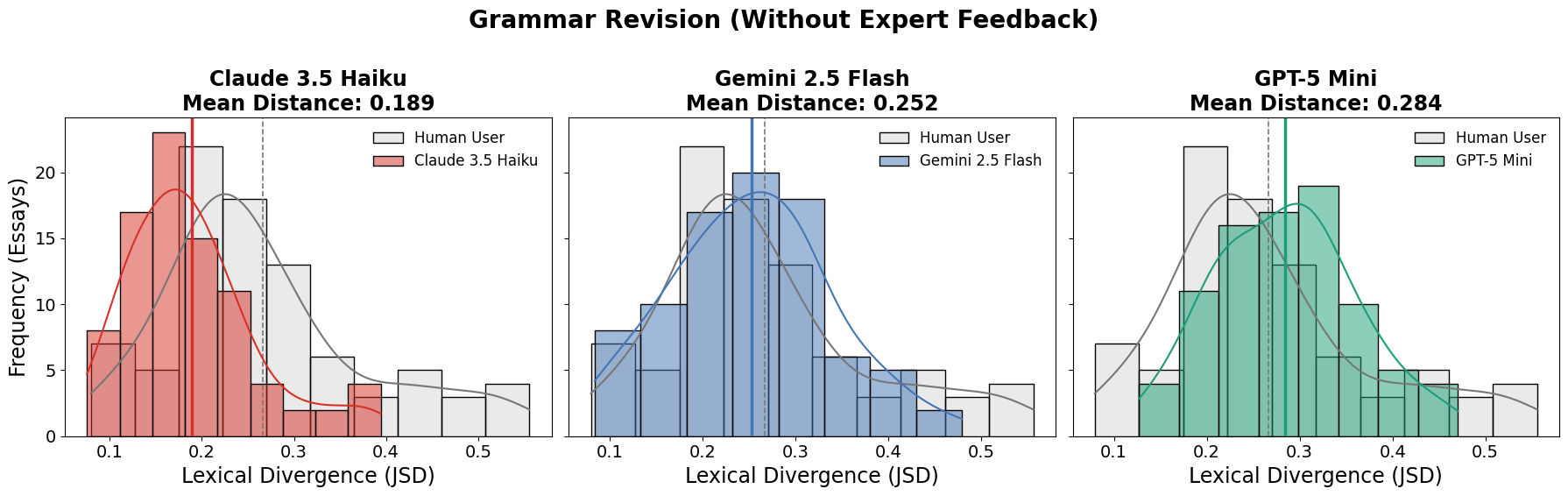}
    \captionof{figure}{Lexical divergence (JSD) distributions for \textbf{Grammar revisions}. 
    Top: with expert feedback. Bottom: without expert feedback.}
    \label{fig:appendix-jsd-grammar}
\end{figure}

\FloatBarrier
\begin{figure}[H]
    \includegraphics[width=\textwidth]{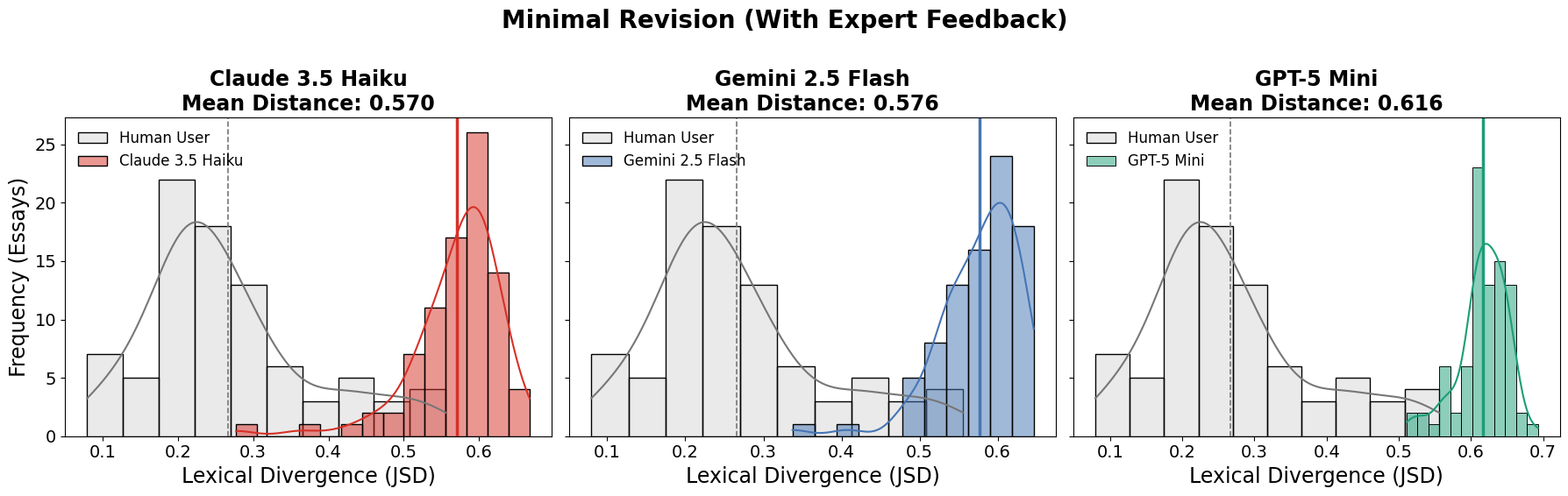}
    \vspace{0.6em}
    \includegraphics[width=\textwidth]{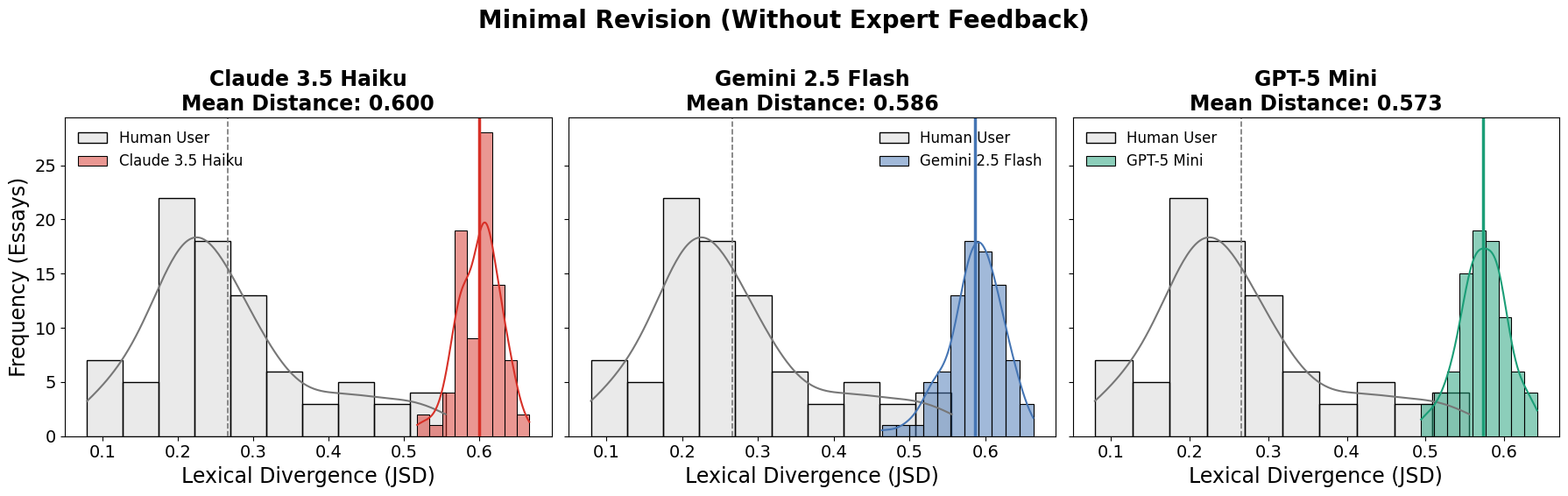}
    \captionof{figure}{Lexical divergence (JSD) distributions for \textbf{Minimal revisions}. 
    Top: with expert feedback. Bottom: without expert feedback.}
    \label{fig:appendix-jsd-minimal}
\end{figure}

\FloatBarrier
\begin{figure}[H]
    \includegraphics[width=\textwidth]{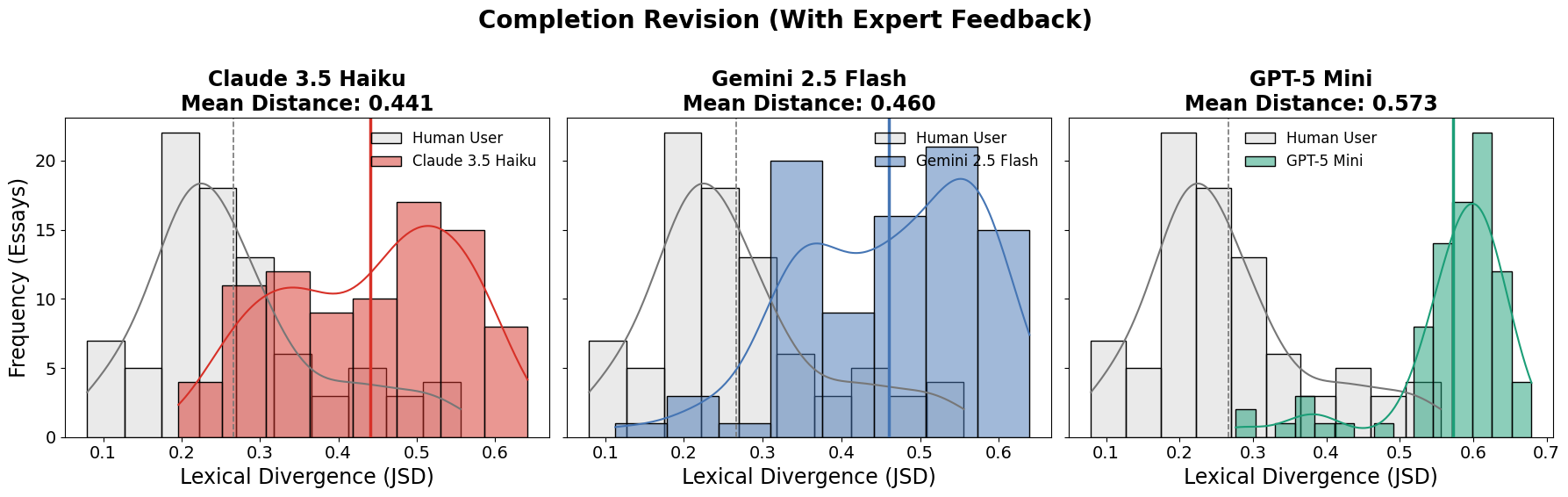}
    \vspace{0.6em}
    \includegraphics[width=\textwidth]{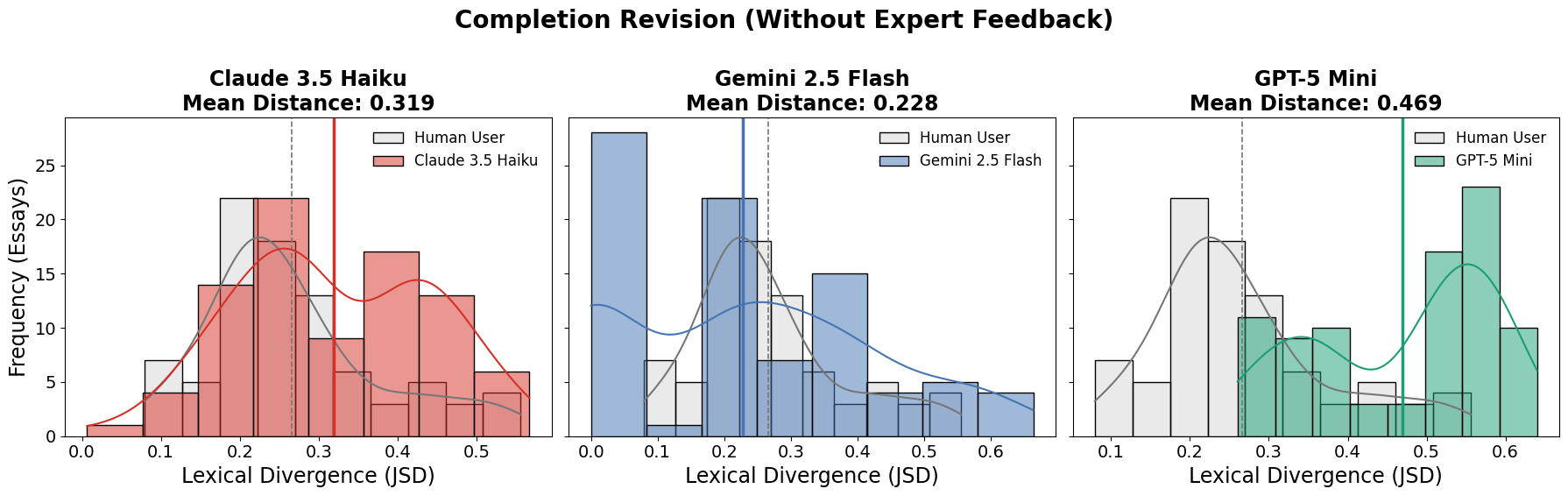}
    \captionof{figure}{Lexical divergence (JSD) distributions for \textbf{Completion revisions}. 
    Top: with expert feedback. Bottom: without expert feedback.}
    \label{fig:appendix-jsd-completion}
\end{figure}

\FloatBarrier
\begin{figure}[H]
    \includegraphics[width=\textwidth]{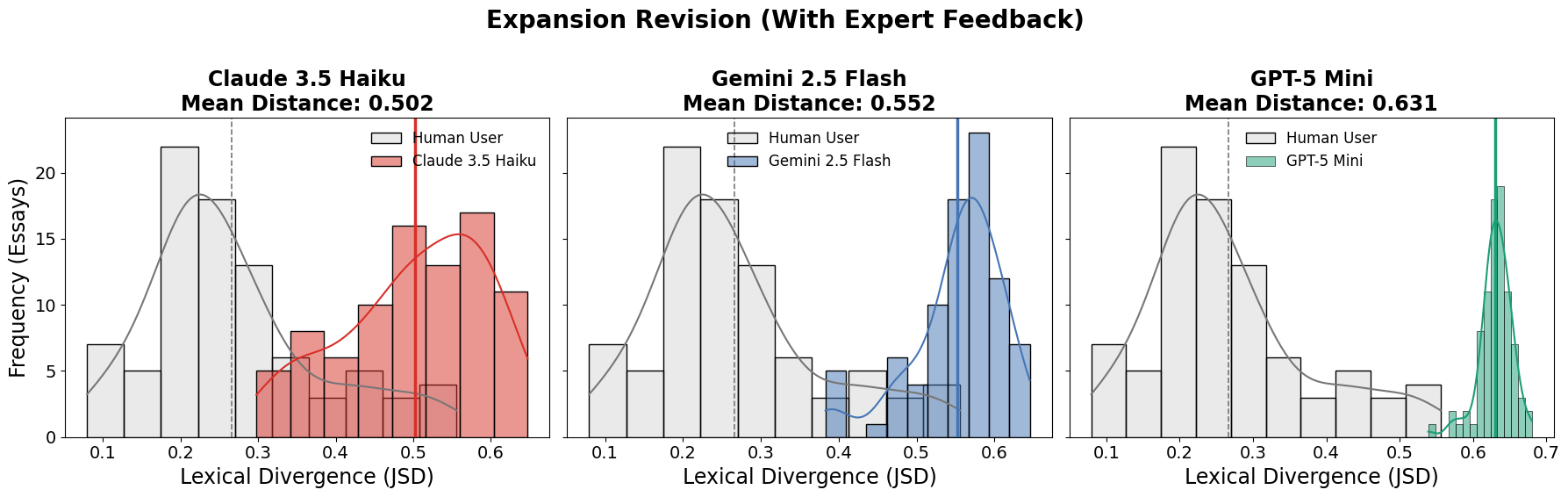}
    \vspace{0.6em}
    \includegraphics[width=\textwidth]{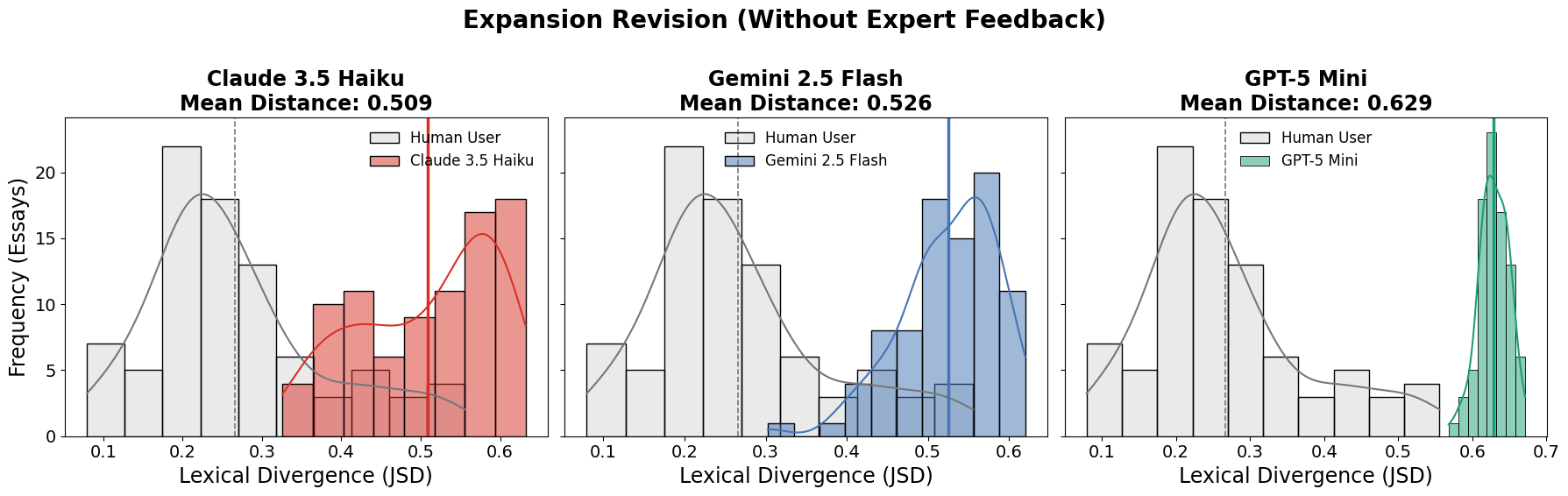}
    \captionof{figure}{Lexical divergence (JSD) distributions for \textbf{Expansion revisions}. 
    Top: with expert feedback. Bottom: without expert feedback.}
    \label{fig:appendix-jsd-expansion}
\end{figure}

\subsection{POS Distribution}\label{appendix:pos_distribution}

\FloatBarrier
\begin{figure}[H]
    \includegraphics[width=\textwidth]{all_figures/arg-rewrite-pos/pos_general_with_feedback.png}
    \vspace{0.6em}
    \includegraphics[width=\textwidth]{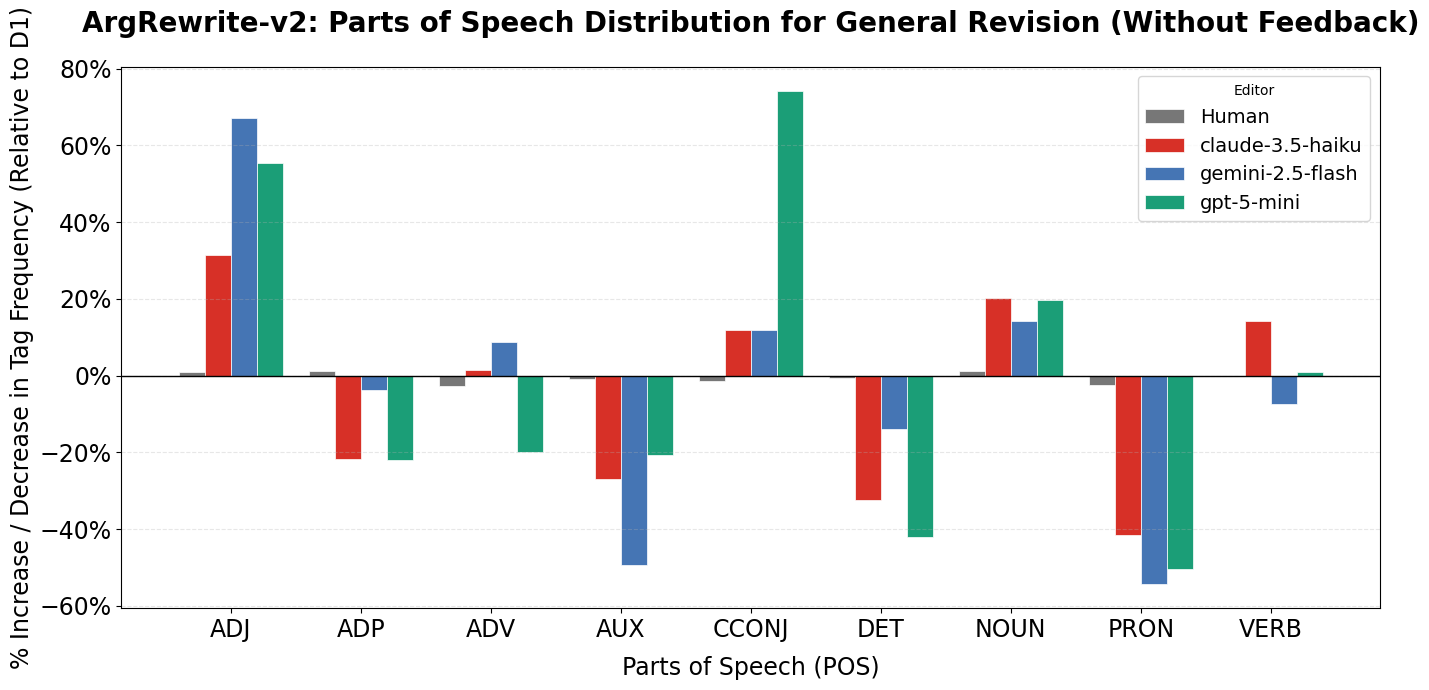}
    \captionof{figure}{Grammatical shifts (POS) for \textbf{General revisions}. 
    Top: with expert feedback. Bottom: without feedback.}
    \label{fig:appendix-pos-general}
\end{figure}

\FloatBarrier
\begin{figure}[H]
    \includegraphics[width=\textwidth]{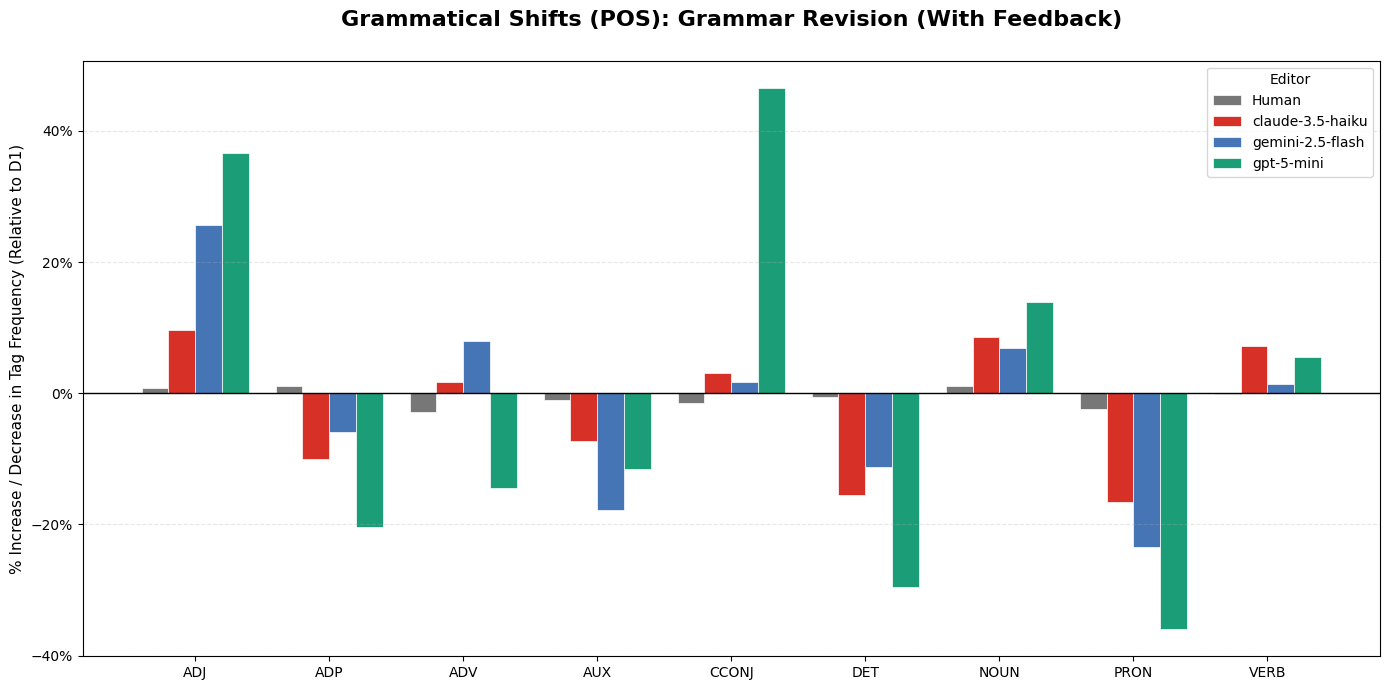}
    \vspace{0.6em}
    \includegraphics[width=\textwidth]{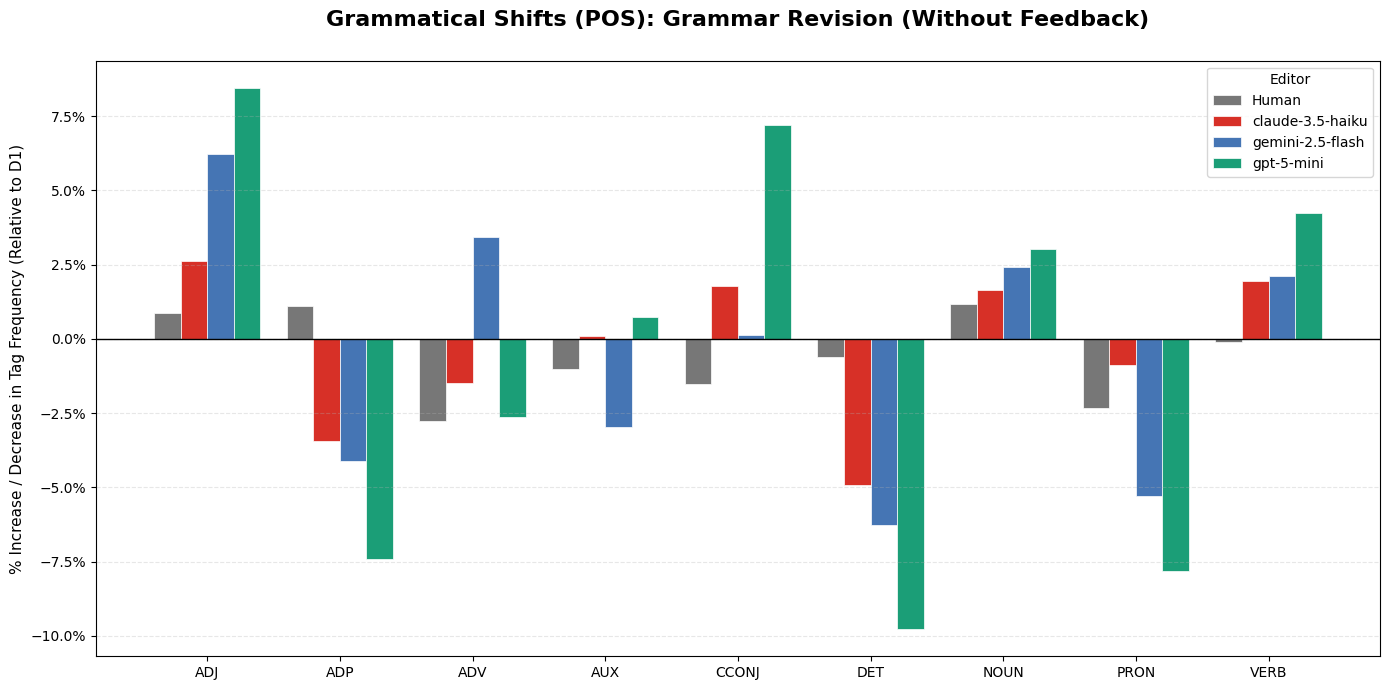}
    \captionof{figure}{Grammatical shifts (POS) for \textbf{Grammar revisions}. 
    Top: with expert feedback. Bottom: without feedback.}
    \label{fig:appendix-pos-grammar}
\end{figure}

\FloatBarrier
\begin{figure}[H]
    \includegraphics[width=\textwidth]{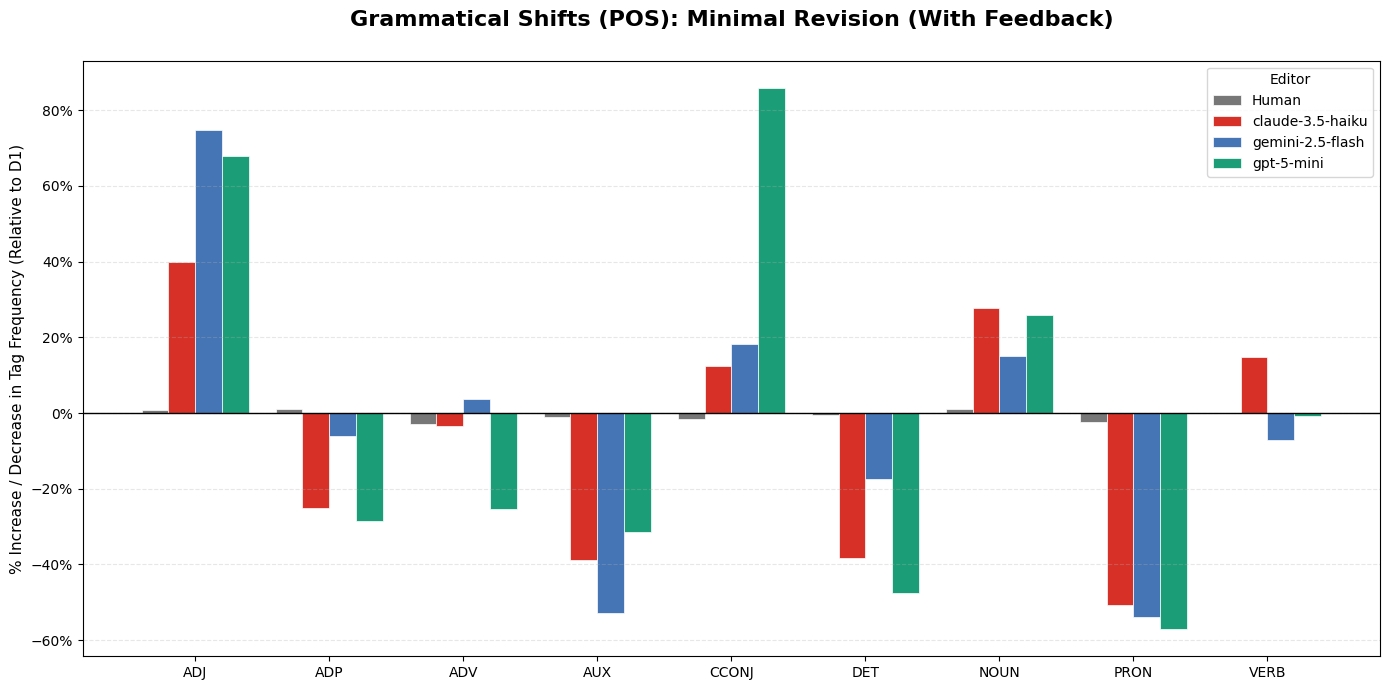}
    \vspace{0.6em}
    \includegraphics[width=\textwidth]{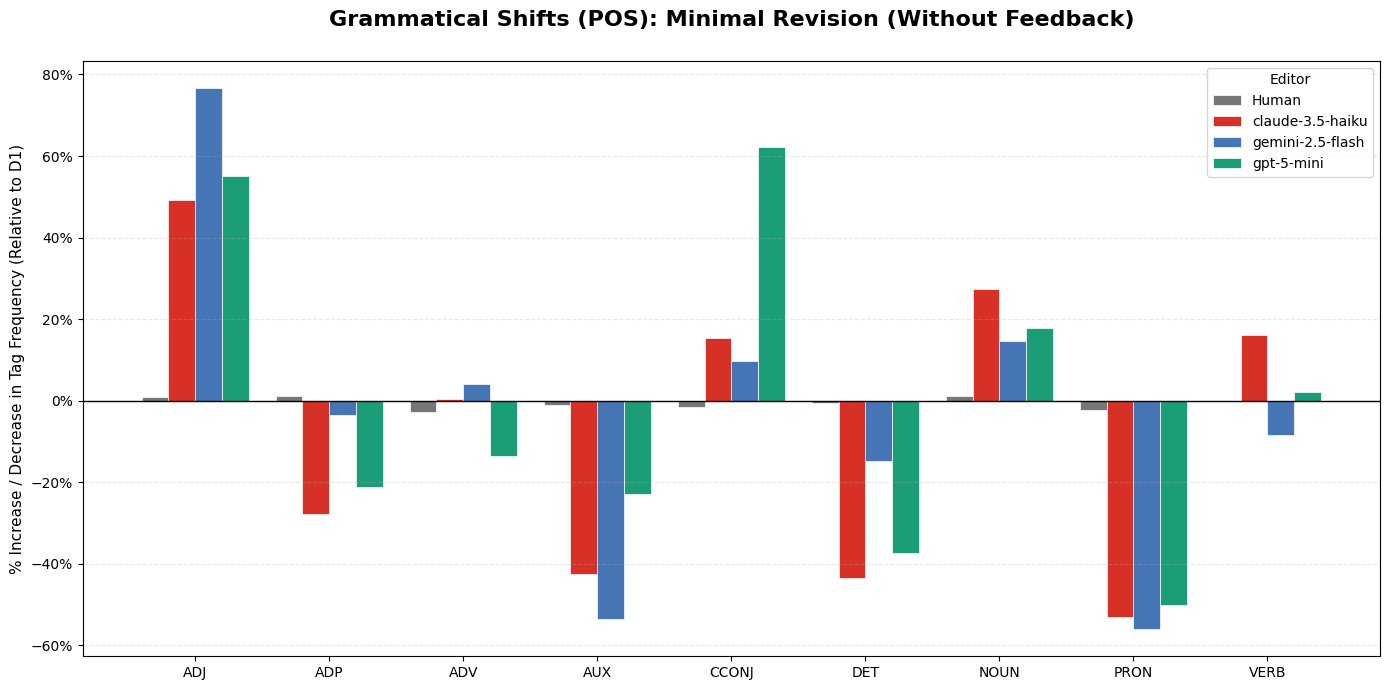}
    \captionof{figure}{Grammatical shifts (POS) for \textbf{Minimal revisions}. 
    Top: with expert feedback. Bottom: without feedback.}
    \label{fig:appendix-pos-minimal}
\end{figure}

\FloatBarrier
\begin{figure}[H]
    \includegraphics[width=\textwidth]{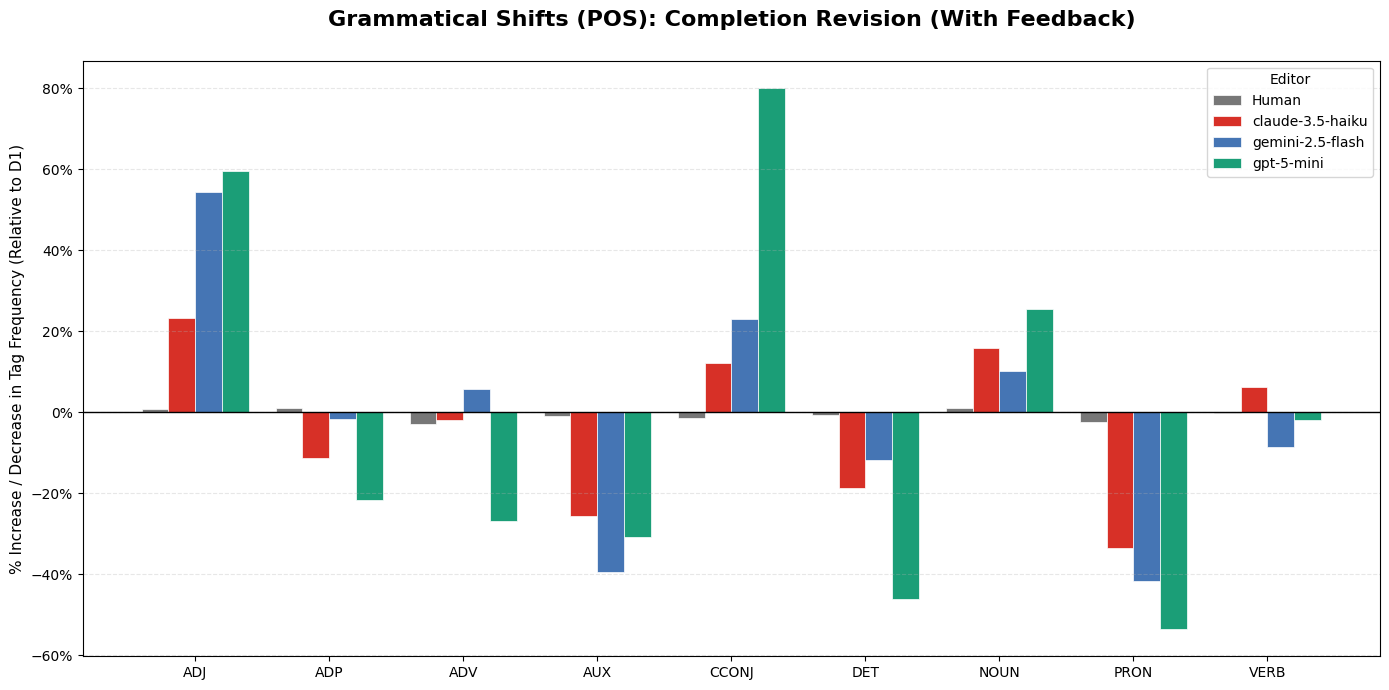}
    \vspace{0.6em}
    \includegraphics[width=\textwidth]{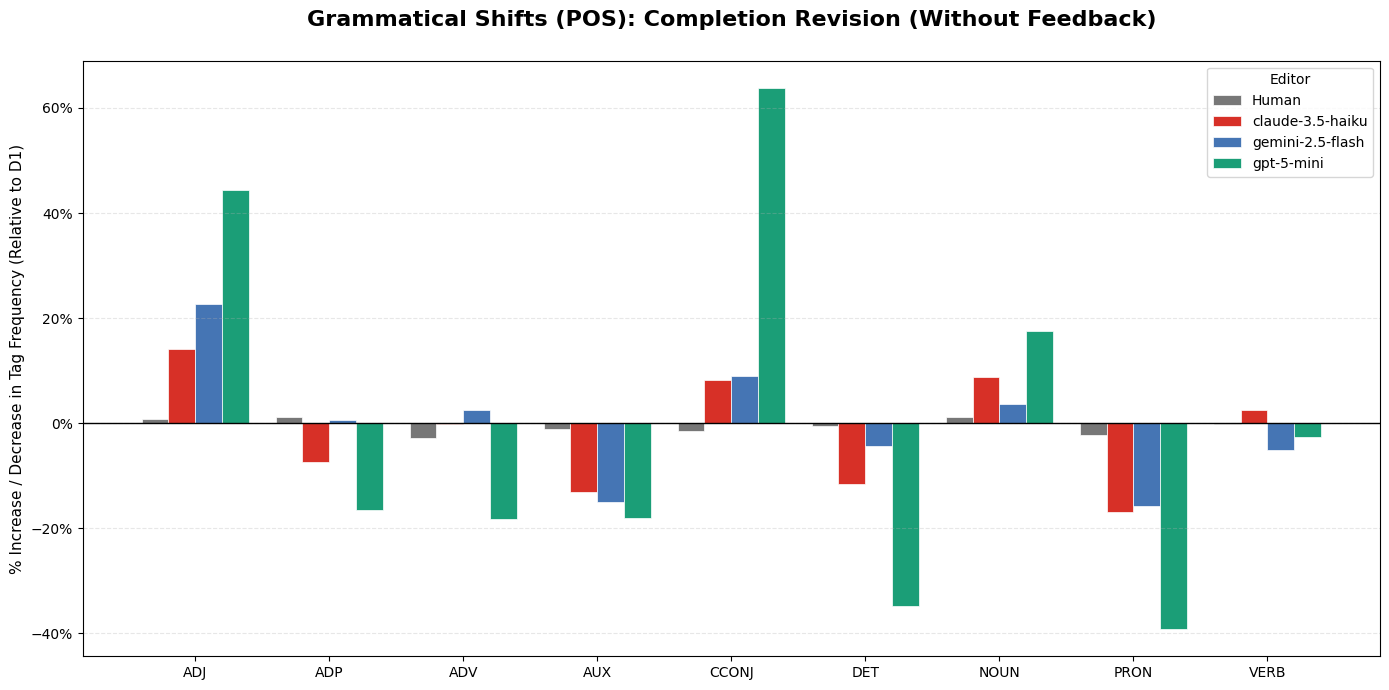}
    \captionof{figure}{Grammatical shifts (POS) for \textbf{Completion revisions}. 
    Top: with expert feedback. Bottom: without expert feedback.}
    \label{fig:appendix-pos-completion}
\end{figure}

\FloatBarrier
\begin{figure}[H]
    \includegraphics[width=\textwidth]{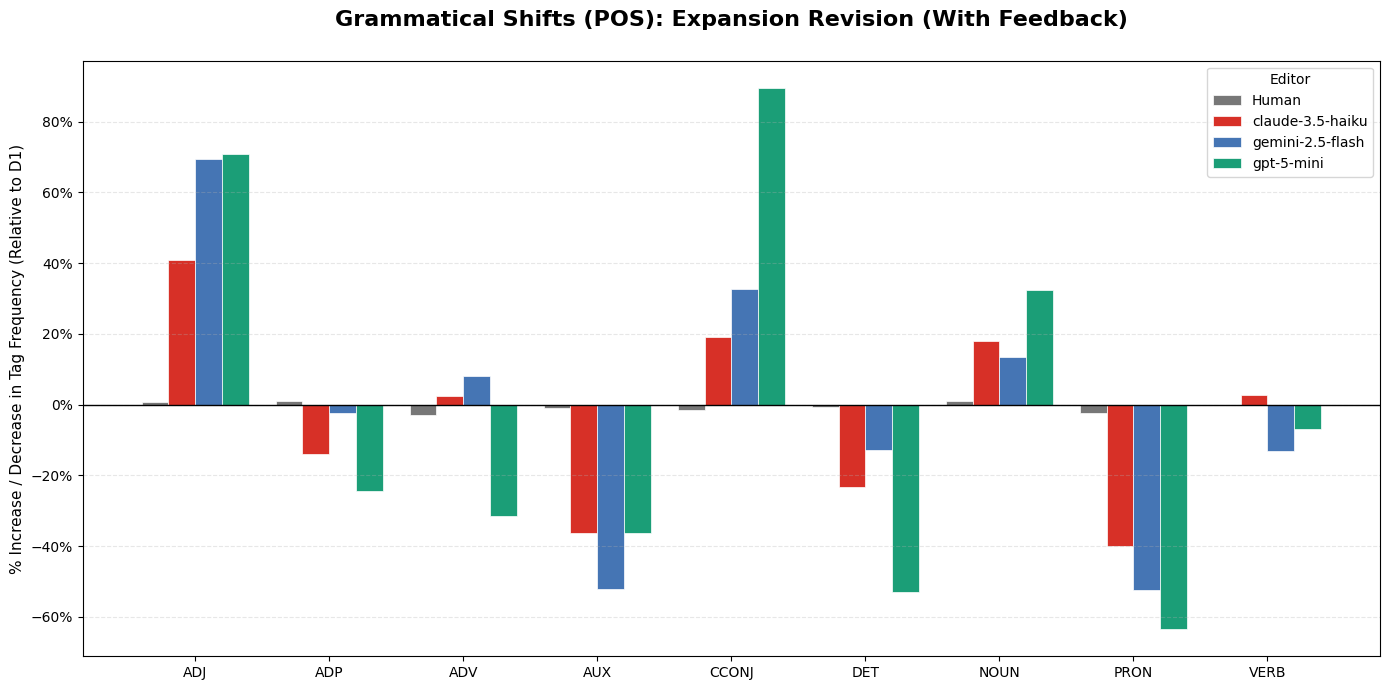}
    \vspace{0.6em}
    \includegraphics[width=\textwidth]{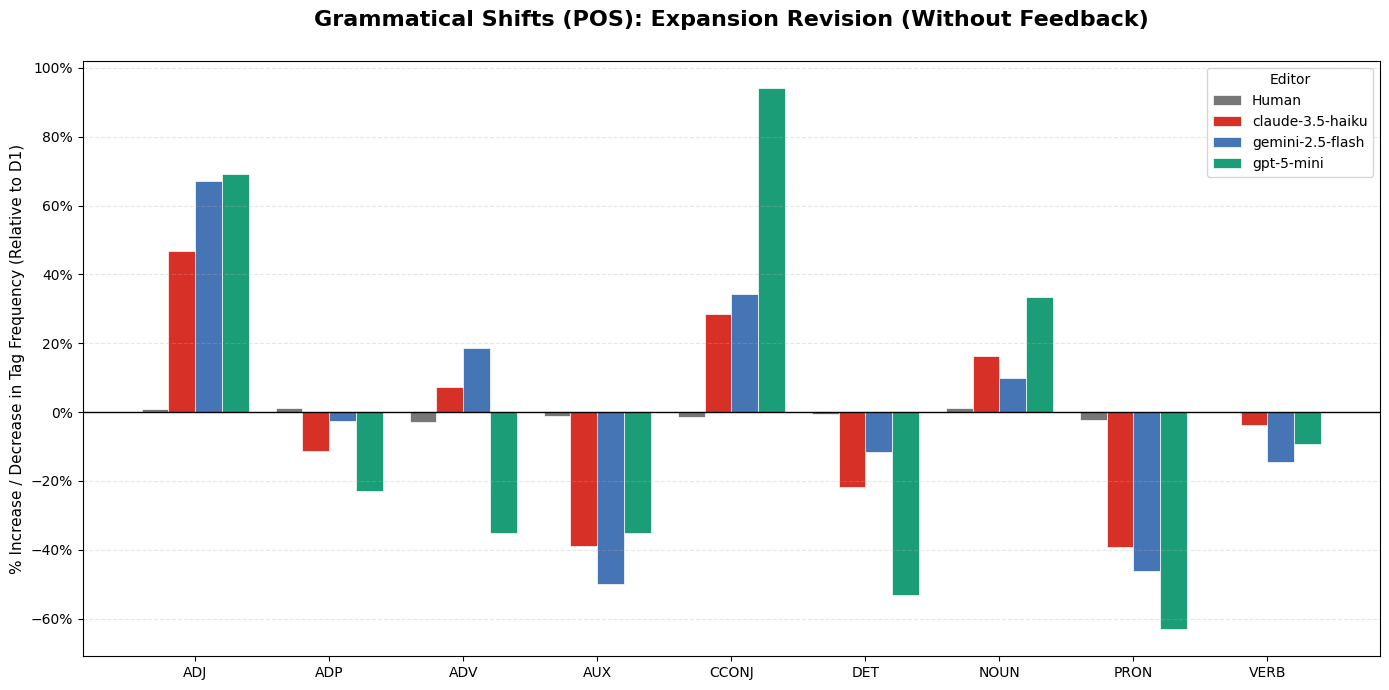}
    \captionof{figure}{Grammatical shifts (POS) for \textbf{Expansion revisions}. 
    Top: with expert feedback. Bottom: without expert feedback.}
    \label{fig:appendix-pos-expansion}
\end{figure}

\subsection{Emotional Shift}\label{appendix:emotional_shift}

\FloatBarrier
\begin{figure}[H]
    \centering
    \includegraphics[width=0.48\textwidth]{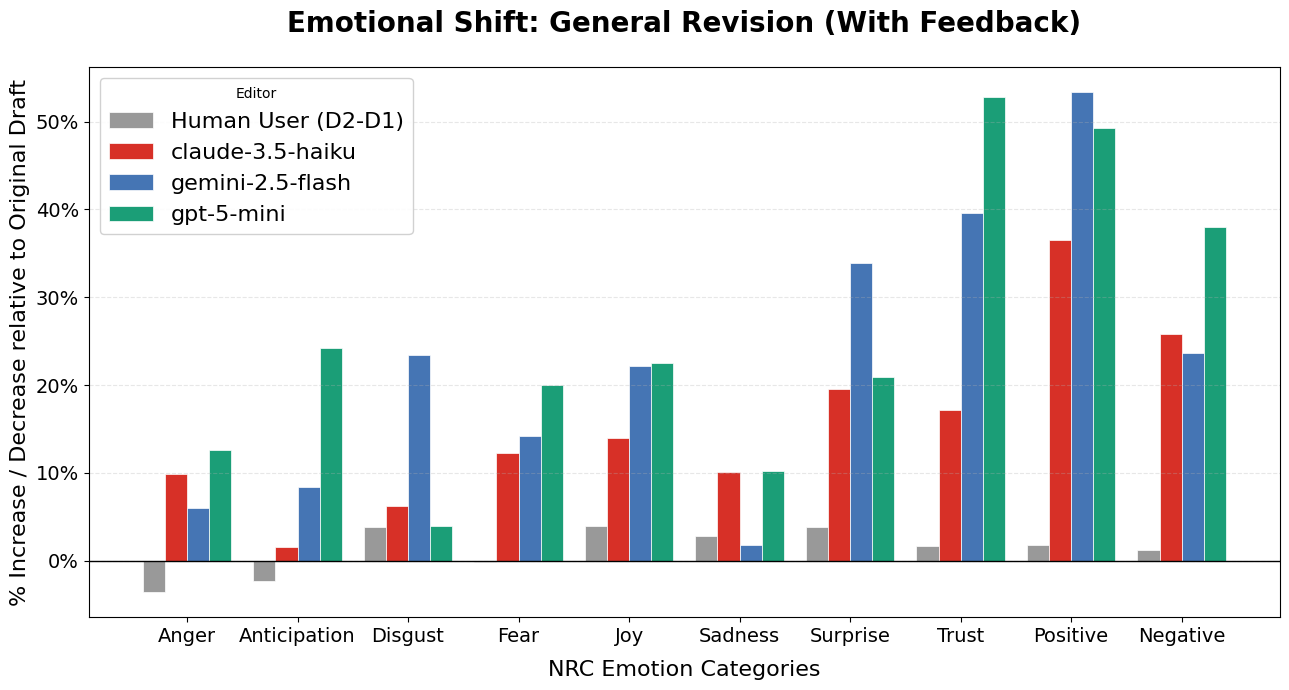}
    \hfill
    \includegraphics[width=0.48\textwidth]{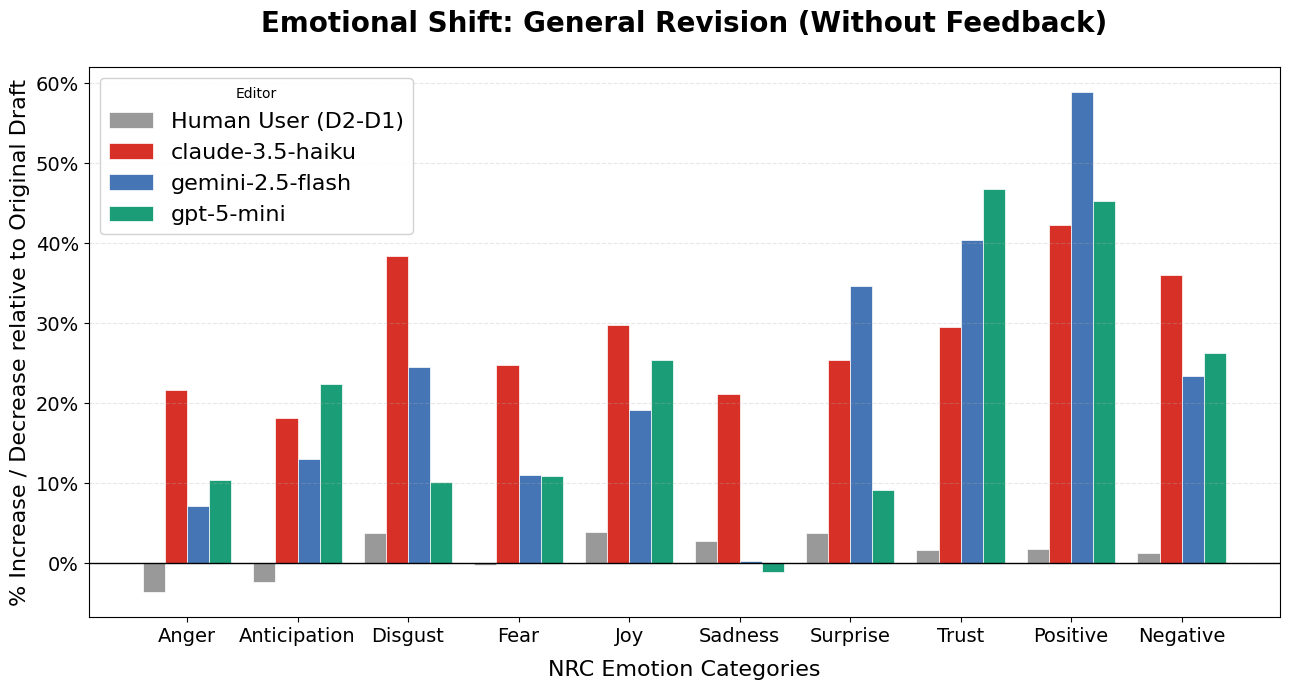}
    \caption{Emotional shifts from D1 to D2 for \textbf{General revisions}. 
    Left: with expert feedback. Right: without expert feedback.}
    \label{fig:appendix-emotion-general}
\end{figure}

\FloatBarrier
\begin{figure}[H]
    \centering
    \includegraphics[width=0.48\textwidth]{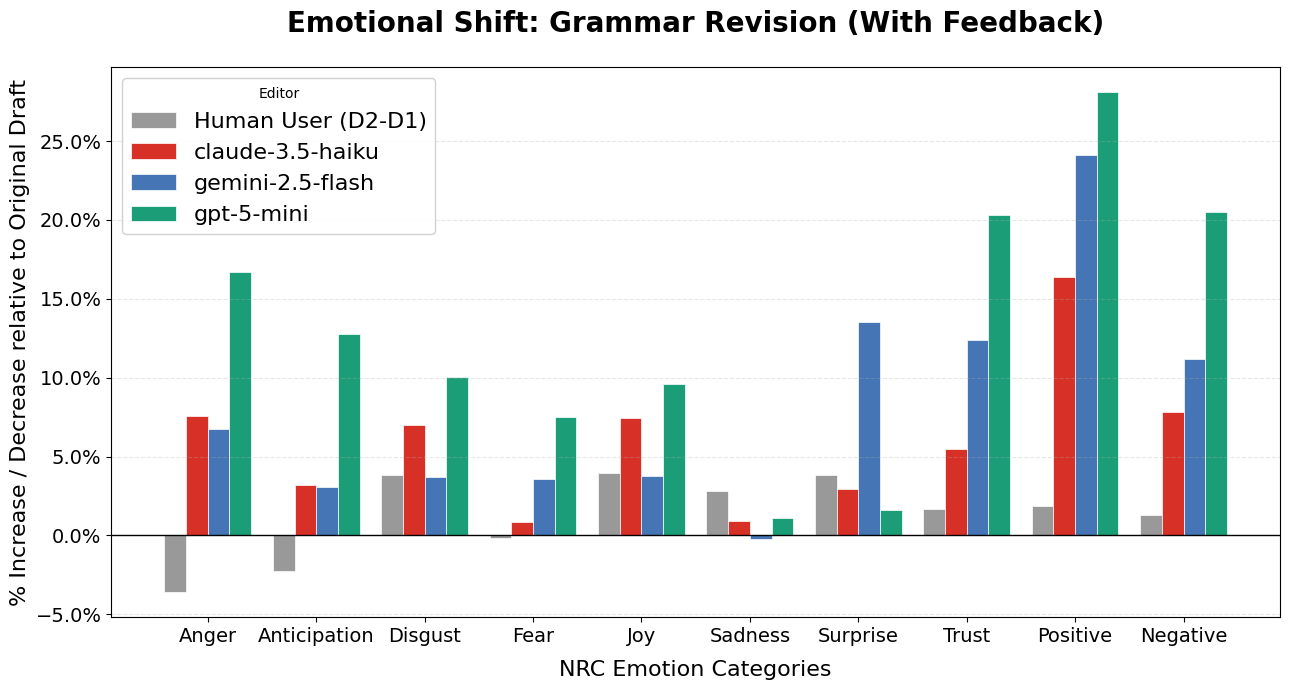}
    \hfill
    \includegraphics[width=0.48\textwidth]{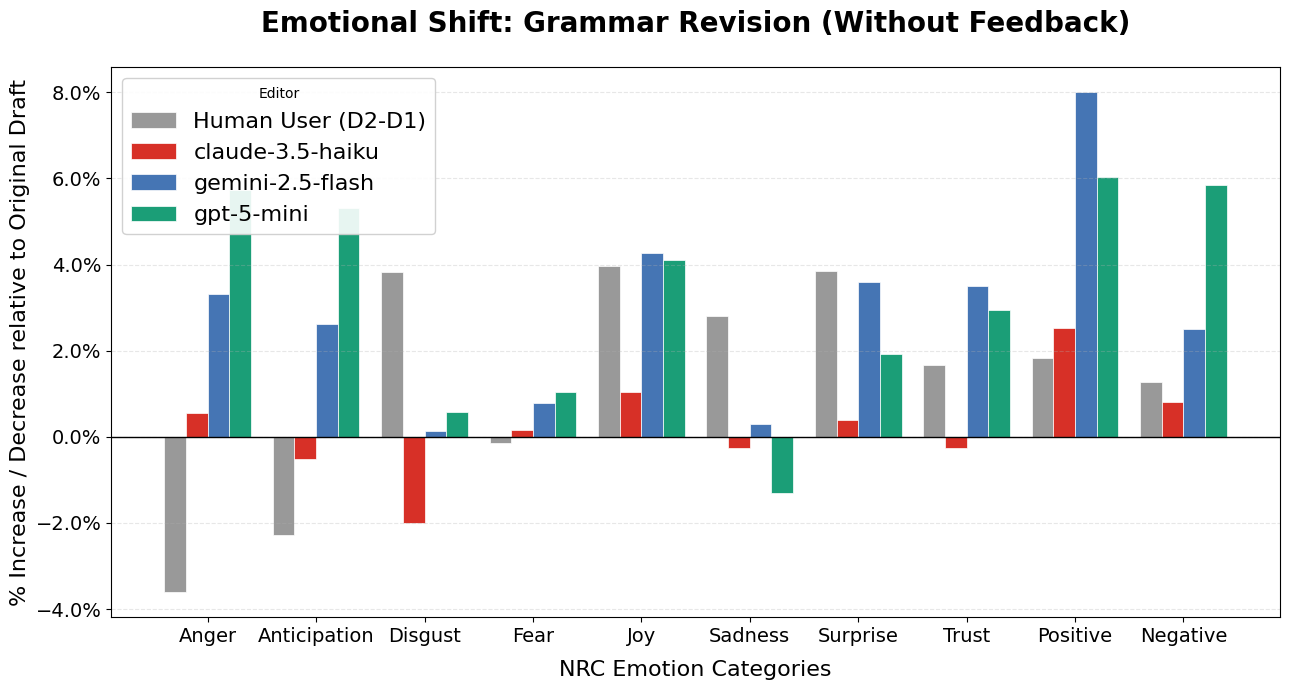}
    \caption{Emotional shifts from D1 to D2 for \textbf{Grammar revisions}. 
    Left: with expert feedback. Right: without expert feedback.}
    \label{fig:appendix-emotion-grammar}
\end{figure}

\FloatBarrier
\begin{figure}[H]
    \centering
    \includegraphics[width=0.48\textwidth]{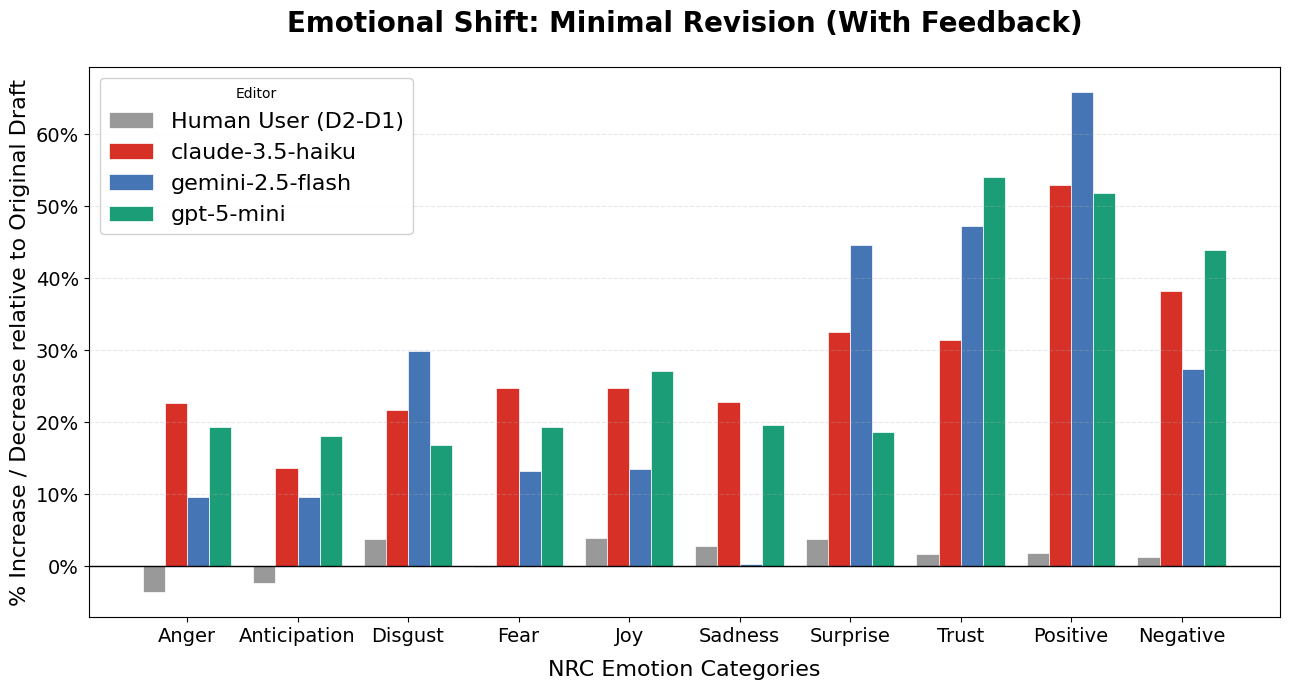}
    \hfill
    \includegraphics[width=0.48\textwidth]{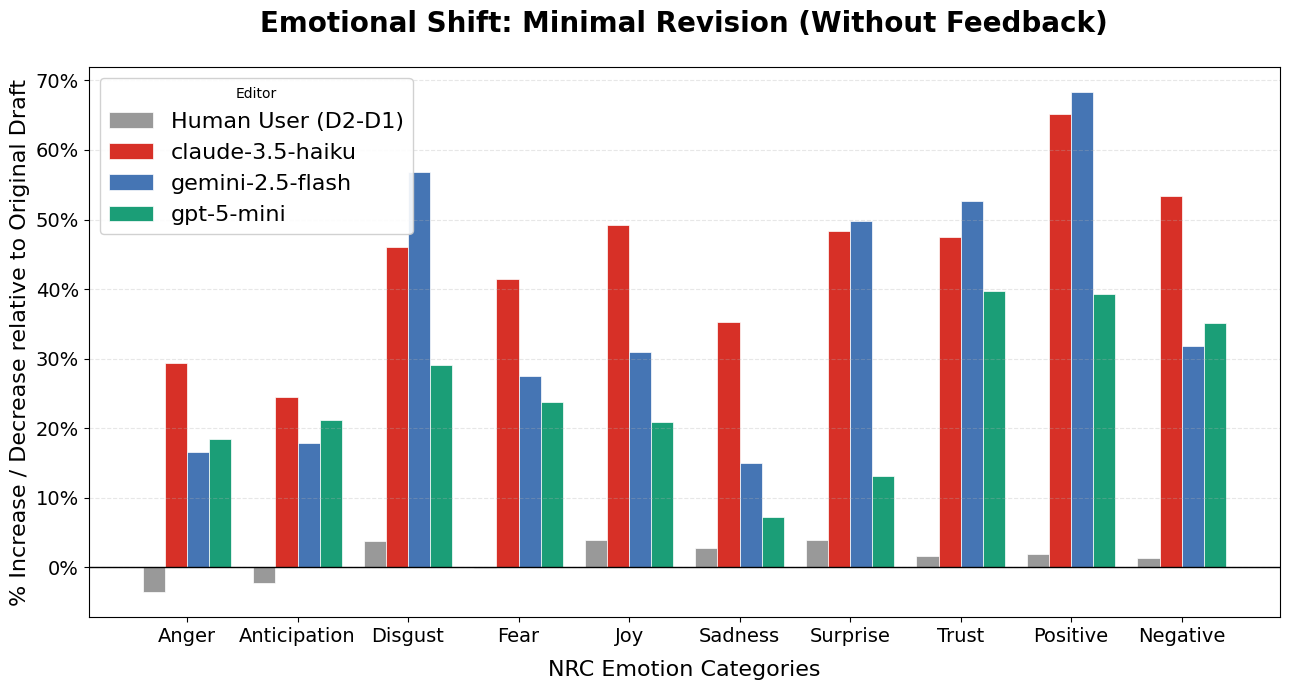}
    \caption{Emotional shifts from D1 to D2 for \textbf{Minimal revisions}. 
    Left: with expert feedback. Right: without expert feedback.}
    \label{fig:appendix-emotion-minimal}
\end{figure}

\FloatBarrier
\begin{figure}[H]
    \centering
    \includegraphics[width=0.48\textwidth]{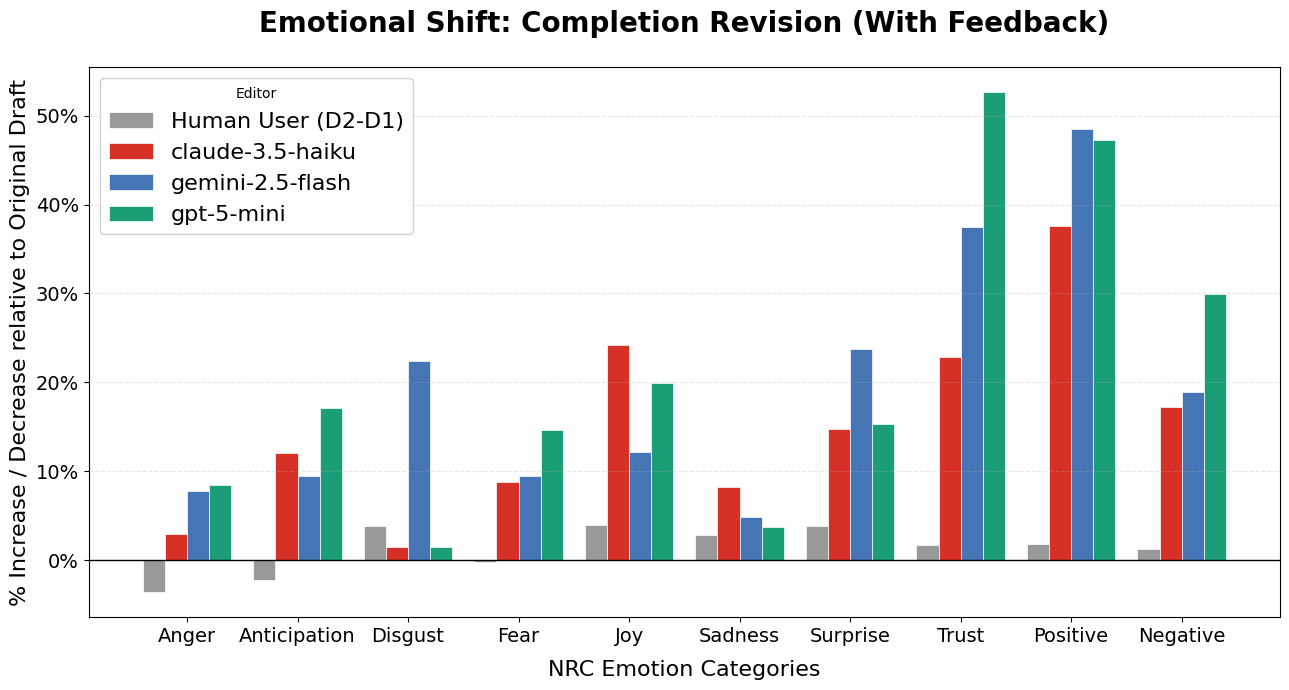}
    \hfill
    \includegraphics[width=0.48\textwidth]{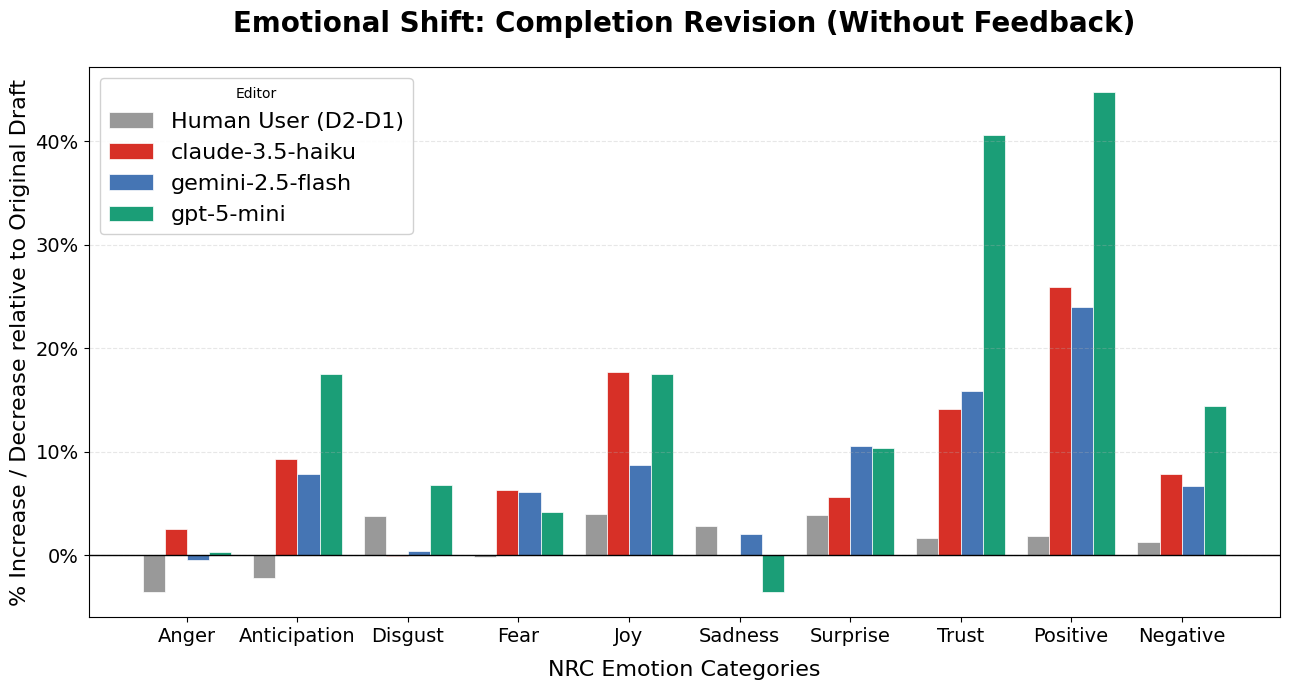}
    \caption{Emotional shifts from D1 to D2 for \textbf{Completion revisions}. 
    Left: with expert feedback. Right: without expert feedback.}
    \label{fig:appendix-emotion-expansion}
\end{figure}

\FloatBarrier
\begin{figure}[H]
    \centering
    \includegraphics[width=0.48\textwidth]{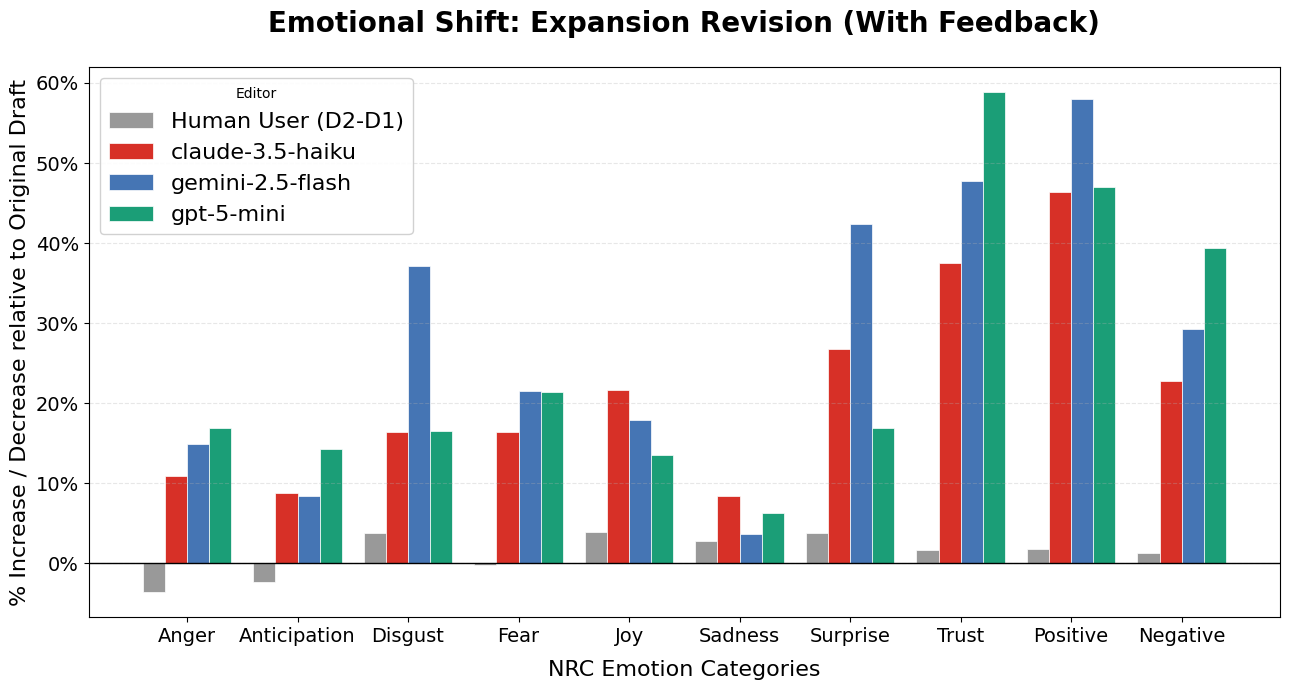}
    \hfill
    \includegraphics[width=0.48\textwidth]{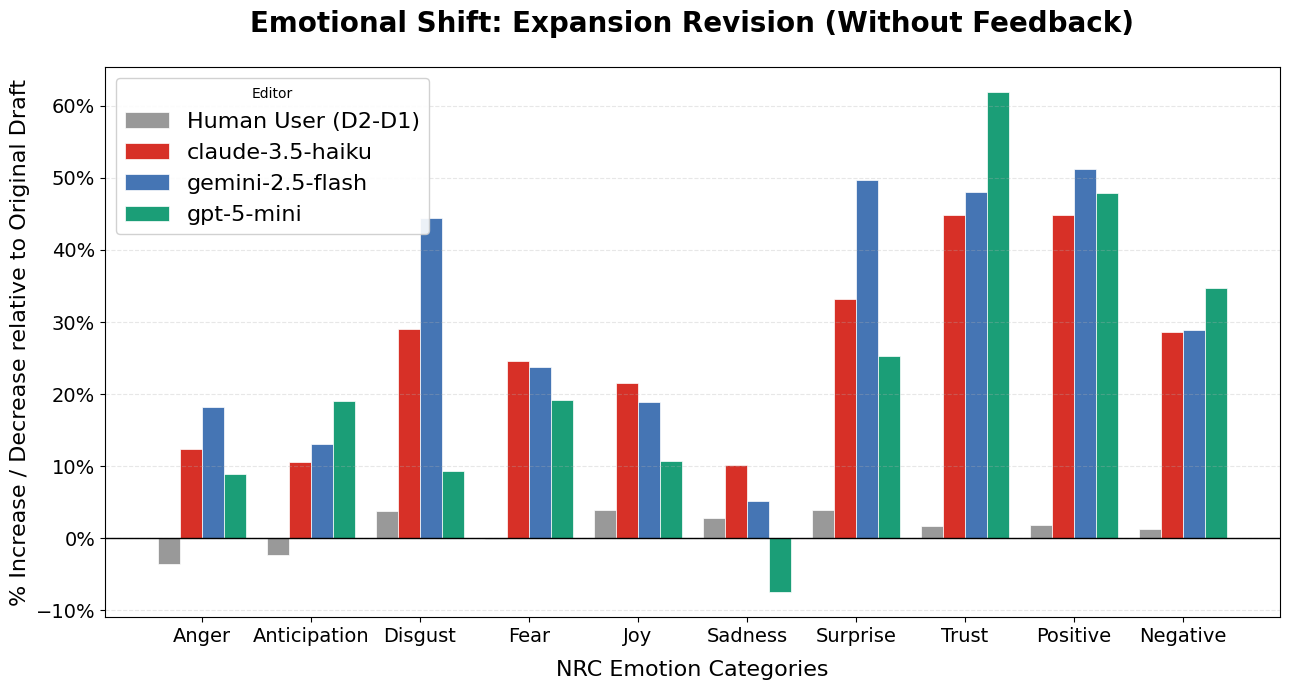}
    \caption{Emotional shifts from D1 to D2 for \textbf{Expansion revisions}. 
    Left: with expert feedback. Right: without expert feedback.}
    \label{fig:appendix-emotion-expansion}
\end{figure}

\section{ICLR Review Analyses}\label{appendix:iclr_analyses}
For ICLR reviews, the possible scores for ICLR papers are: 0: strong reject, 2: reject, 4: borderline reject, 6: borderline accept, 8: accept, 10: strong accept. With each score, there is a confidence level from 1 to 5 on how confident the reviewer is, where 1 represents low confidence and 5 represents absolute certainty. 




\section{LLM-as-a-Judge Prompts} \label{sec:llm-as-a-judge}

\subsection{ICLR Review Analysis Prompts} \label{appendix: iclr_judge_prompts}

This prompts the LLM-as-a-Judge to create categories and definitions for the different categories. 

\begin{tcblisting}{
  enhanced,
  breakable,
  listing only,
  listing options={
    basicstyle=\footnotesize\ttfamily,
    breaklines=true,
    columns=fullflexible
  }
}
Analyze the following {len(review_texts)} peer reviews and create a set of consistent categories for strengths and weaknesses.

Reviews (one per line, separated by "---"):
{chr(10).join([f"REVIEW {i+1}:{chr(10)}{text}{chr(10)}---"
for i, text in enumerate(review_texts)])}

Return your response as a JSON object with two arrays: "strength_categories" and "weakness_categories".

Example:
{
    "strength_categories": ["Clarity", "Novelty", "Strong Results"],
    "weakness_categories": ["Lack of Clarity", "Insufficient Experiments", "Outdated Methods"],
    "definitions": {
        "Clarity": "The paper is well-written and easy to understand.",
        "Lack of Clarity": "The paper is difficult to follow and poorly organized.",
        "Novelty": "The paper presents new and original ideas.",
        "Insufficient Experiments": "The experiments do not adequately support the claims made in the paper.",
        "Strong Results": "The results are compelling and demonstrate the effectiveness of the proposed method.",
        "Outdated Methods": "The methods used are not state-of-the-art."
    }
}
\end{tcblisting}

This prompts the LLM to reduce the extracted categories to reduce redundant categories.

\begin{tcblisting}{
    enhanced, 
    breakable, 
    title={}, 
    listing only, 
    listing options={
        basicstyle=\footnotesize\ttfamily, 
        breaklines=true
    }
}
    f"""Given the following list of categories, reduce redundancy by merging similar ones. Return a JSON array of unique categories. 
    with a maximum of 15 categories. Do NOT provide more than 15. 

    Give a description of the categories and make sure they are distinct, e.g. don't include both "Clarity" and "Motivation Clarity".
Categories:
{json.dumps(categories, indent=4)}
Example:
{{
    "reduced_categories": ["Clarity", "Novelty", "Strong Results", "Insufficient Experiments", "Outdated Methods"]
    "definitions": {{
        "Clarity": "How clear and understandable the paper is.",
        "Novelty": "The originality and innovativeness of the research.",
        "Strong Results": "The robustness and significance of the experimental results.",
        "Insufficient Experiments": "Lack of adequate experimental validation.",
        "Outdated Methods": "Use of methods that are no longer state-of-the-art."   
    }}
}}
    """
\end{tcblisting}

This prompt is for extracting the top three strengths and weaknesses from the previously extracted strengths and weaknesses from the previous two prompts. 

\begin{tcblisting}{%
    enhanced, 
    breakable, 
    title={}, 
    listing only, 
    listing options={
        basicstyle=\footnotesize\ttfamily, 
        breaklines=true
    }
}
    f"""Analyze the following peer review and extract the key points.

    Review:
    {review_text}

    Please identify:
    1. Top 3 strengths mentioned in the review
    2. Top 3 weaknesses mentioned in the review

    Return your response as a JSON object with the following structure:
    {{
        "strengths": ["strength 1", "strength 2", "strength 3"],
        "weaknesses": ["weakness 1", "weakness 2", "weakness 3"]
    }}

    You make a pick from these strengths and weaknesses categories: 

    Strengths: 
    {json.dumps(strength_categories, indent=4)}
    
    Weaknesses:
    {json.dumps(weakness_categories, indent=4)}

    Do NOT invent strengths or weaknesses that are not in the above categories, and do not mention anything not in the review.

    Only return the JSON object, no additional text."""
\end{tcblisting}

\subsection{Human User Study LLM-as-a-Judge Prompts} \label{appendix: user_judge_prompts}

\begin{tcblisting}{
    enhanced, 
    breakable, 
    title={}, 
    listing only, 
    listing options={
        basicstyle=\footnotesize\ttfamily, 
        breaklines=true
    }
}
    Analyze the following essay and provide:
            1. A stance on whether money leads to happiness: "for", "against", or "neutral"
            2. A list of argument types used from: logical, emotional, anecdotal, statistical, philosophical, personal_experience, expert_opinion, hypothetical
            3. A brief summary of the main arguments

            Essay:
            {essay_text}

            Respond in JSON format:
            {{
                "stance": "for/against/neutral",
                "argument_types": ["type1", "type2", ...],
                "main_arguments": "brief summary"
            }}
\end{tcblisting}

\section{Statistical Significance Scores for ICLR review categories} \label{sec: iclr_stat_sig}

\begin{table}[H]
\centering
\caption{G1 and G2 denote the proportional distribution of review categories. Because the methodology allows for a maximum of three strength and three weakness labels per entry, the denominator is normalized by tripling the total number of reviews to reflect the maximum potential labels. }
\label{tab:category_comparison}
\small
\begin{tabular}{cccccc}
\midrule
\textit{Strength Dataset} & & & & & \\
\midrule
\toprule
\textbf{Category} & \textbf{G1 \%} & \textbf{G2 \%} & \textbf{\% Change} & \textbf{Z-Score} & \textbf{P-Value} \\
\midrule
Clarity & 5.76 & 3.89 & -1.87 & 14.3704 & 0.0000 \\
Practical Application & 2.02 & 3.42 & 1.40 & -14.1956 & 0.0000 \\
Reproducibility & 0.62 & 1.46 & 0.84 & -13.6601 & 0.0000 \\
Relevance & 3.45 & 2.35 & -1.10 & 10.7794 & 0.0000 \\
Theoretical Foundations & 3.10 & 4.02 & 0.92 & -8.1627 & 0.0000 \\
Scalability & 0.89 & 1.33 & 0.44 & -6.9699 & 0.0000 \\
Technical Soundness & 2.17 & 2.80 & 0.63 & -6.6554 & 0.0000 \\
Methodology & 4.62 & 3.91 & -0.72 & 5.8503 & 0.0000 \\
Comparative Analysis & 0.11 & 0.03 & -0.07 & 4.5899 & 0.0000 \\
Motivation & 2.54 & 2.15 & -0.39 & 4.2626 & 0.0000 \\
Impact & 1.05 & 0.82 & -0.23 & 3.8971 & 0.0001 \\
Writing and Presentation & 0.03 & 0.00 & -0.02 & 3.3568 & 0.0008 \\
Resource Contribution & 2.24 & 1.98 & -0.26 & 3.0387 & 0.0024 \\
Innovation & 3.60 & 3.90 & 0.30 & -2.5833 & 0.0098 \\
Interpretability & 0.01 & 0.03 & 0.02 & -2.4006 & 0.0164 \\
Novelty & 5.64 & 5.34 & -0.30 & 2.1418 & 0.0322 \\
Computational Efficiency & 0.36 & 0.29 & -0.07 & 1.9781 & 0.0479 \\
\midrule
\textit{Weakness Dataset} & & & & & \\
\midrule
Writing and Presentation & 4.01 & 1.37 & -2.64 & 26.9113 & 0.0000 \\
Clarity & 4.62 & 1.93 & -2.69 & 24.9006 & 0.0000 \\
Generalization & 3.57 & 6.08 & 2.51 & -19.3279 & 0.0000 \\
Scalability & 1.08 & 2.28 & 1.20 & -15.4265 & 0.0000 \\
Computational Efficiency & 2.28 & 3.61 & 1.32 & -12.9235 & 0.0000 \\
Assumptions / Simplifications & 1.64 & 2.63 & 0.99 & -11.3135 & 0.0000 \\
Reproducibility & 1.00 & 1.77 & 0.77 & -10.8394 & 0.0000 \\
Comparative Analysis & 8.59 & 6.87 & -1.72 & 10.6360 & 0.0000 \\
Real-World Applicability & 1.55 & 2.14 & 0.58 & -7.1594 & 0.0000 \\
Motivation & 0.26 & 0.09 & -0.17 & 6.6250 & 0.0000 \\
Relevance & 0.09 & 0.01 & -0.08 & 5.5298 & 0.0000 \\
Experimental Validity & 5.84 & 5.21 & -0.63 & 4.5250 & 0.0000 \\
Technical Soundness & 0.05 & 0.01 & -0.03 & 3.0875 & 0.0020 \\
Theoretical Justification & 4.75 & 5.14 & 0.38 & -2.9223 & 0.0035 \\
Overclaiming & 1.10 & 0.97 & -0.13 & 2.0643 & 0.0390 \\
\bottomrule
\end{tabular}
\end{table}
\newpage

\input{sections/appendix/sample_convos_user_study}
\input{sections/appendix/sample_essays_arg-rewrite}

%% file: sections/appendix/table_contents.tex
\newpage 
\section*{Appendix}
\begin{itemize}
    \item[\textbf{A.}] \textbf{User Study Details} \dotfill \pageref{sec: user_study_details}
    \begin{itemize}
        \item[A.1] User Study Recruitment Process, Compensation and Duration \dotfill \pageref{appendix:user_study_details}
        \item[A.2] User Study Instructions \dotfill \pageref{appendix:survey}
        \item[A.3] Pre-Study Questions \dotfill \pageref{appendix:pre_study}
        \item[A.4] Post-Study Questions \dotfill \pageref{appendix: post_study}
        \item[A.5] Consent Form \dotfill \pageref{appendix: consent_form}
        \item[A.6] Demographics \dotfill \pageref{appendix:demographics}
    \end{itemize}
    \vspace{0.5em}
    \item[\textbf{B.}] \textbf{ArgRewrite-v2 Analysis}
    \begin{itemize}
        \item[B.1] Generating LLM Drafts \dotfill \pageref{appendix:argrewrite_prompts}
        \item[B.2] Semantic Shifts with Gemini-004 Embedding \dotfill \pageref{appendix:semantic_shifts_gemini004}
        \item[B.3] Semantic Shifts with MiniLM-L6-v2 Embedding \dotfill \pageref{appendix:semantic_shifts_minilm}
        \item[B.4] Semantic Shifts Across Settings / Model \dotfill \pageref{appendix:semantic_shifts_across_settings_per_model}
        \item[B.5] Jensen-Shannon Divergence \dotfill \pageref{appendix:jsd}
        \item[B.6] POS Distribution \dotfill \pageref{appendix:pos_distribution}
        \item[B.7] Emotional Shift \dotfill \pageref{appendix:emotional_shift}
        \item[B.8] ICLR Review Analyses \dotfill \pageref{appendix:iclr_analyses}
    \end{itemize}
    \vspace{0.5em}
    \item[\textbf{C.}] \textbf{LLM-as-a-Judge Prompts} \dotfill \pageref{sec:llm-as-a-judge}
    \begin{itemize}
        \item[C.1] ICLR Review Analysis Prompts \dotfill \pageref{appendix: iclr_judge_prompts}
        \item[C.2] User Study Prompts \dotfill \pageref{appendix: user_judge_prompts}
    \end{itemize}
    \vspace{0.5em}
    \item[\textbf{D.}] \textbf{Statistical Significance Scores for ICLR Review Categories} \dotfill \pageref{sec: iclr_stat_sig}
    \vspace{0.5em}
    \item[\textbf{E.}] \textbf{Human User Study Essay Samples}
    \begin{itemize}
        \item[E.1] Human-Written Essay Samples
        \item[E.2] LLM-Influenced Essay Samples \dotfill \pageref{sec:influenced}
        \item[E.3] LLM-Generated Essay Samples \dotfill \pageref{sec:llm}
    \end{itemize}
    \vspace{0.5em}
    \item[\textbf{F.}] \textbf{Sample ArgRewrite-v2 Human \& LLM Edited Essays} \dotfill \pageref{sec:appendix-essays}
    \begin{itemize}
        \item[F.1] Five Editing Modes (\texttt{gpt-5-mini}) \dotfill \pageref{sec:app-modes}
        \begin{itemize}
            \item[F.1.1] Essay 1: Personal Narrative
            \item[F.1.2] Essay 2: Skeptical / Critical
            \item[F.1.3] Essay 3: Direct / Anecdotal
        \end{itemize}
        \item[F.2] Three Models, Expansion Mode \dotfill \pageref{sec:app-models}
        \begin{itemize}
            \item[F.2.1] Essay 1: Analytical
            \item[F.2.2] Essay 2: Questioning Feasibility
            \item[F.2.3] Essay 3: Strong Opinion
        \end{itemize}
    \end{itemize}
\end{itemize}

%% file: sections/appendix/sample_convos_user_study.tex
\section{Human User Study Examples}\label{sec: user_study_examples}
We presents essay samples collected from a user study on the prompt \textit{``Does money lead to happiness?''} organized into three groups: human-written essays, LLM-Influenced essays, and LLM-Generated essays. 

\subsection{Sample 1 --- Placing conditions on when money leads to happiness}

\noindent\textit{This essay argues that money can lead to happiness, but only when it is used to meet genuine needs such as food, shelter, and healthcare rather than to satisfy insecurities or signal status.}

\medskip

\begin{tcolorbox}[title={Original Human Essay}, essaybox]
``Money is the root of all evil,'' ``Money makes the world go round'' --- our relationship
with money is complicated and complex. While money can afford a nice home and luxury
vacations, it cannot buy a loving family and meaningful relationships. I believe, though,
that money can lead to happiness for some people.

Money has value because of the importance and placement humans have implemented on it.
Previously, when living in a hunter-gatherer society, animal skin, meat, cloth, fur, and
food were forms of currency. One would exchange goods for services or trade goods. Over
time, however, we evolved to buying goods and services with paper cash. In order to obtain
housing, food, healthcare, education, and other necessities, one needs money.

Money can lead to happiness depending on one's relationship with it. If one believes that
obtaining money will make them happy, they will never be satisfied. If one is not satisfied
with their looks, money can buy plastic surgery; if one wants expensive items to appear
wealthy to others, money can provide that. With this thinking, however, money will not make
one happy, as this is more of an inner problem.

For money to lead toward happiness, it must alleviate or help oneself, others, or a
situation. If one suffers from food insecurity, money can get them food. If one is homeless
or living in an underdeveloped home, money can buy them safety. If one needs medical care to
alleviate pain, continue day-to-day life, or extend their lifespan, money can afford that.
Money can lead to happiness when used to meet basic needs, security, and safety. For most
people, being able to afford healthy meals, a safe home, good-quality education, and
life-saving medication requires money. However, the majority of the world earns very little.
If those people had money, they would be happier.
\end{tcolorbox}

\subsection{Sample 2 --- Arguing that money will never lead to happiness}

\noindent\textit{This essay argues that personal problems transcend financial status, that wealthy people still experience divorce and breakdowns, and that neither poverty nor wealth resolves the non-financial struggles that underpin unhappiness.}

\medskip

\begin{tcolorbox}[title={Original Human Essay}, essaybox]
Does money lead to happiness? I think this question leads to a plethora of other questions
and conversations. I have never been extremely wealthy, so I am not entirely qualified to
speak about whether a great deal of money leads to happiness. However, I do not think I
need to be in order to make my point.

There are two sides to this coin. On one side, if you have money, you can buy anything you
want --- eat, wear, and drive whatever you like. What a peaceful life that must be. Yet
somehow I find it hard to believe that your non-financial problems disappear just because
you have money in your checking account. Every day I open my phone to find a new story
about a celebrity or wealthy person having a breakdown or doing something nonsensical. It is
not rare to hear about wealthy people filing for divorce. If a couple has all the money they
could ever need, how could they still get divorced? The simple answer: money does not buy
happiness. The lengthier answer involves the fact that such couples have personal issues
money can never fix --- a principle that applies to every human on Earth.

On the other side of the coin: if you do not have money, you very clearly have problems.
Add those to the personal problems every human has, and you have a recipe for unhappiness.
The real delusion is experienced by those who believe more money would transform their
happiness entirely.

All in all, the answer is no. Money has never and will never lead to happiness. Humans will
always have non-financial personal problems that money could never fix. Many times, the best
couples are not the wealthiest ones.
\end{tcolorbox}

\subsection{Sample 3 --- Happiness is friendship and loved ones, not money}

\noindent\textit{draws on a Jim Carrey quote and anecdotes about wealthy executives to argue
that money eases life's logistics but that true happiness lies in friendship,
family, and meaningful connection}

\medskip

\begin{tcolorbox}[title={Original Human Essay}, essaybox]
I do not think that money leads to happiness; however, it can make life easier. A rich
person gets up whenever they want, goes out to eat, and does enjoyable things. If their car
breaks down, they simply have it fixed. Comparatively, someone who is poor may struggle to
get a ride home, worry about getting to work the next day, or have to fix the car
themselves. So money does not buy happiness, but it does make life easier in some
circumstances.

Jim Carrey once said he wished everyone could be rich so they could realise it is not the
answer. Money does not buy you friends --- though wealthy people often have many who want to
be around them for their riches. I have read many times that very rich people are often very
lonely; executives in particular are generally lonely. So does money lead to happiness? I do
not think so.

I have recently been hearing billionaires say that at the end of life you will not worry
about your money or your status. You will be concerned with the people --- your family, your
friends --- those you connected with. Did you do right by them? Did you give them quality
time? These are the questions that, they say, will run through your mind.

There is also the idea that money is the root of all evil --- though more precisely, it is
the \textit{love} of money that is the problem. Money can make life easier, but it does not
necessarily make you happy. I think happiness is being close to people, having friends, and
being in a state where you have someone to be with you and help you.
\end{tcolorbox}

\newpage
\section{LLM-Influenced Essay Samples}
\label{sec:influenced}

\subsection{Sample 1 --- Information seeking: finding arguments why money leads to happiness}

\noindent\textit{The author asked the LLM for arguments in favour of money leading to happiness, then used those arguments to structure an essay arguing that financial freedom eases stress, enables experiences, and expands personal autonomy.}

\medskip

\begin{tcolorbox}[title={Human Essay (LLM-influenced)}, essaybox]
I believe money can certainly lead to happiness, but it is not the primary factor. If we
are lacking in other areas of life, money alone will not make us happy. If those areas are
satisfied, however, money can improve life greatly. With a good family and supportive people
around you, financial freedom allows you to do much more --- easing stress and anxiety,
enabling experiences like vacationing, and ultimately providing greater freedom.

Money allows us to meet basic needs for food, shelter, and health. We can eat whatever we
want and afford healthier, more expensive foods. For shelter, money lets us live wherever we
choose, near the people we love. For health, we can afford visits to doctors and
specialists, significantly increasing quality of life.

As for life experiences, money opens the door to travel and trips that create lasting
memories shared with loved ones. We can afford any hobby we desire; nothing is out of reach
financially. Having those options keeps us from feeling limited and boxed in.

This leads to the concept of freedom. Financial freedom is one of the ultimate forms of
freedom in general. Society runs on money --- the more you have, the more you can do.
Financial concerns cause fear; living under monetary restrictions limits the life you can
live. Financial freedom removes all worry about whether you can eat, have shelter, or afford
meaningful experiences.

In conclusion, money is not the only factor in happiness. You need other reasons to be
happy, such as a good family. Without people to share your wealth with, being wealthy may
mean little. But with a solid foundation, money makes life significantly better. It provides
comfort, experiences, memories, and ultimately true freedom.
\end{tcolorbox}

\convsep

\begin{tcolorbox}[humanchat]
\textbf{Human:} How does money lead to happiness?
\end{tcolorbox}

\begin{tcolorbox}[llmchat]
\textbf{LLM:} The relationship between money and happiness is complex and multifaceted. While money itself
does not guarantee happiness, it can influence well-being in various ways.

\textbf{Basic Needs Satisfaction:} At a fundamental level, money allows individuals to meet
their basic needs --- food, shelter, clothing, and healthcare. When these needs are met,
people generally experience a higher level of well-being and security.

\textbf{Reduced Stress and Anxiety:} Financial stability can reduce stress and anxiety
associated with worries about meeting daily expenses or facing emergencies. Knowing that one
has resources to handle unexpected situations can contribute to peace of mind.

\textbf{Opportunities for Experiences:} Money enables individuals to engage in travel,
dining, and recreational activities. Research suggests that spending money on experiences
rather than material goods often leads to greater and longer-lasting happiness.

\textbf{Social Connections:} Money can facilitate social interactions by allowing
individuals to participate in events and gatherings. Building and maintaining social
connections is closely linked to happiness.

\textbf{Generosity and Giving:} Having money provides opportunities to help others, donate
to causes, or engage in charitable activities. Acts of generosity are associated with
increased happiness and fulfilment.

\textbf{Freedom and Autonomy:} Having financial resources often provides individuals with a
greater sense of autonomy and control over their lives, allowing them to make choices that
align with their values.

\textbf{Long-term Security:} Saving and investing money can provide a sense of security for
the future, contributing to peace of mind and long-term happiness.

It is important to note that the relationship between money and happiness is not linear.
Research indicates that after reaching a certain income level --- often cited around median
national income --- the increase in happiness tends to plateau. Factors such as personal
values, social relationships, and life circumstances also play critical roles in overall
well-being.
\end{tcolorbox}

\begin{tcolorbox}[humanchat]
\textbf{Human:} How can it ease our stress and anxiety?
\end{tcolorbox}

\begin{tcolorbox}[llmchat]
\textbf{LLM:} Money can ease stress and anxiety in several ways, primarily by providing security,
stability, and opportunities for personal well-being.

\textbf{Meeting Basic Needs:} Financial resources enable individuals to afford food, shelter,
and healthcare consistently. When people can meet these basic needs, they are likely to feel
less stressed and anxious about daily survival.

\textbf{Emergency Preparedness:} Having savings provides a cushion against unexpected
expenses --- medical emergencies, car repairs, or job loss. This buffer can reduce anxiety
about the future and promote a sense of security.

\textbf{Access to Care and Support:} Money provides access to mental health resources such
as therapy, counselling, or wellness programmes. These resources can help individuals manage
stress more effectively.

\textbf{Flexibility and Options:} Financial resources give individuals more choices,
including the ability to choose more fulfilling work, relocate to a better environment, or
take time off for self-care. Having options enhances feelings of control.

\textbf{Improved Work-Life Balance:} Financial stability can allow individuals to work fewer
hours or choose jobs that offer better balance, contributing to lower stress levels and
improved mental health.

\textbf{Investment in Health:} Money can be used to promote physical and mental health
through exercise, nutrition, and wellness activities, leading to lower stress and better
coping mechanisms.

\textbf{Time Freedom:} Sufficient funds create opportunities to outsource tasks like cleaning
or yard work, or to take vacations --- reducing daily pressures and promoting relaxation.

\textbf{Long-Term Planning:} Money allows individuals to plan for the future through
investments, retirement savings, or education funds. Knowing that a plan is in place can
alleviate anxiety about what lies ahead.

While money can help reduce stress and anxiety through these means, effective coping also
involves emotional support, healthy habits, and personal resilience. Balancing financial
well-being with mental and emotional health is key to overall well-being.
\end{tcolorbox}


\subsection{Sample 2 --- A neutral perspective, striking a balance}

\noindent\textit{The author submitted a draft to the LLM for feedback and incorporated suggested edits, arriving at the view that money is not a direct cause of happiness but may be a necessary precondition for it.}

\medskip

\begin{tcolorbox}[title={Human Essay (LLM-influenced)}, essaybox]
The saying goes, ``Money can't buy happiness,'' but how accurate is this statement?
Understanding the relationship between money and happiness may depend on various factors in
an individual's life, particularly those that support or hinder personal fulfillment.

While true happiness often stems from relationships and experiences rather than material
wealth, it is important to recognise that spending time with loved ones, witnessing
breathtaking moments, or simply enjoying nature can generate profound joy without any
financial investment. In this sense, money is not strictly necessary for happiness.

However, consider the circumstances around those moments. Can someone create happy moments
and truly enjoy those experiences if they are extremely stressed about paying their bills or
figuring out how to make it to next week? Do they have the mental space to stop, step
outside for a walk, or take the time to engage in activities that promote joy? Oftentimes,
the answer is no. People burdened by financial stress often cannot prioritise valuable
moments of happiness. Thus, money may be required for happiness, at least for some people.

So perhaps the question is not ``Does money lead to happiness?'' but rather: is financial
confidence or stability required for happiness? While money alone does not create happiness,
perhaps it provides the conditions necessary for individuals to seek and enjoy joyful
experiences. Ultimately, whether money is a requirement for happiness may depend entirely on
the individual and their personal circumstances.
\end{tcolorbox}

\convsep

\begin{tcolorbox}[humanchat]
\textbf{Human:} I'm writing an essay about the idea of money creating happiness. This is my draft. How do
you suggest I improve it for clarity of thought?

\smallskip
\noindent\textit{The saying goes, ``Money can't buy happiness,'' but is that truly factual?
It may depend on how happiness relates to other factors in your life. When you think of core
life values that drive true happiness, like spending time with family and friends,
experiencing awe-inspiring moments, you can clearly see that money is not a necessity for
happiness. Taking a simple walk outside or laughing at a great joke with family requires no
capital and can result in immense happiness. So in that sense, yes --- money is not needed
for happiness. But consider the circumstances around those moments. Can someone create happy
moments if they are extremely stressed about paying their bills or figuring out how to make
it to next week? \ldots{}}
\end{tcolorbox}

\begin{tcolorbox}[llmchat]
\textbf{LLM:} Your draft presents an intriguing exploration of the relationship between money and
happiness. To enhance clarity and strengthen your argument, consider the following
suggestions.

\textbf{Introduction Refinement:} Start with a clear statement that sets the stage for your
discussion.
\textit{Revised opening:} ``The saying goes, `Money can't buy happiness,' but how accurate
is this statement? Understanding the relationship between money and happiness may depend on
various factors in an individual's life, particularly those that support or hinder personal
fulfillment.''

\textbf{Logical Flow:} Ensure that each paragraph transitions seamlessly to the next,
introducing each point with a clear topic sentence.

\textbf{Clarifying Core Values:} Define ``true happiness'' more explicitly to anchor your
argument.
\textit{Revised example:} ``True happiness often stems from relationships and experiences
rather than material wealth. Spending time with loved ones, witnessing breathtaking moments,
or simply enjoying nature can generate profound joy without any financial investment.''

\textbf{Addressing Financial Stress:} Strengthen this section with specific examples or
studies showing the impact of financial stress on well-being.

\textbf{Concluding Perspective:} Tie back to the initial statement.
\textit{Revised conclusion:} ``The question may not be whether money directly leads to
happiness, but whether financial stability is essential for it. While money alone does not
create happiness, it can provide the conditions necessary for individuals to seek and enjoy
joyful experiences.''

\textbf{Proofreading:} Revise for grammar, punctuation, and conciseness.
\end{tcolorbox}

\begin{tcolorbox}[humanchat]
\textbf{Human:} Is this good to go?

\smallskip
\noindent\textit{The saying goes, ``Money can't buy happiness,'' but how accurate is this
statement? Understanding the relationship between money and happiness may depend on various
factors in an individual's life\ldots{} [revised draft]}
\end{tcolorbox}

\begin{tcolorbox}[llmchat]
\textbf{LLM:} Your essay is much improved and presents a nuanced perspective effectively. Here are a few
minor suggestions before you consider it final.

\textbf{Transitions:} Improve the flow between ideas with transitional phrases to help the
reader follow your argument more easily.

\textbf{Refining Clarity:} In a few places, rephrase for increased clarity.

\textbf{Conclusion:} Your final thoughts can be slightly reinforced for a stronger closing
impact.

\smallskip
\noindent\textit{Key suggested revisions:}
``\ldots{} \textbf{While} true happiness often stems from relationships and experiences
rather than material wealth, \textbf{it is important to recognise that} spending time with
loved ones \ldots{} \textbf{In this sense,} money is not strictly necessary for happiness.''

``\ldots{} \textbf{Do they have the mental space to stop,} step outside for a walk, or take
the time to engage in activities that promote joy? Oftentimes, the answer is no.
\textbf{Thus,} money may be required for happiness, at least for some people.''

\smallskip
With these adjustments your essay will have a more coherent flow and clearer transitions.
Once you are satisfied, it should be good to go.
\end{tcolorbox}


\subsection{Sample 3 --- Money does not lead to happiness}

\noindent\textit{The author submitted their draft for feedback and the LLM responded with affirmation and elaboration; the final essay retains the author's original voice and argument, concluding that genuine happiness must precede wealth for money to have any effect.}

\medskip

\begin{tcolorbox}[title={Human Essay (LLM-influenced)}, essaybox]
We all tend to think ``\textit{only if} I had the money.'' I often think that if I had more
money I could do what I want, have what I want, and everything would be amazing. There are
no worries in the fantasy of unlimited money, or even just enough not to worry about it.
Many of us struggle even when we have enough to get by. Most of us have to at least
\textit{think} about whether we can afford something: we are careful with purchases, find
the best deal on homes, cars, vacations, hobbies, and nutrition. We envy rich people and
fail to realise that in some ways we are free from the complications that come with having
money.

There are certainly levels to wealth --- I have ``medium wealthy'' and above in mind, perhaps
starting at a \$50 million net worth. Can you imagine the financial work involved in managing
\$50 million? The money is tied up in businesses, investments, and real estate. It would be
stressful to stay on top of it and ensure people are not stealing from you. It would be hard
to trust even those working for you --- or even family.

Think, too, about how you would relate to other people. The vast majority of those around
you probably do not have that kind of money. It would be an invisible barrier preventing
meaningful relationships and interactions. You would be stuck interacting with others in
your wealth category, which sounds like a real limitation. I have personally felt the
strange tension of being around extreme wealth --- so many boundaries, hesitations, and
nothing feels truly ``free,'' pun intended.

In my experience, getting to the root of unhappiness means unravelling deep feelings, fears,
and trauma. Money absolutely improves quality of life and can allow happiness in some ways,
but it does not buy the motivation to begin truly healing yourself. That work cannot be
bought. I believe there is a livable wealth amount that lets humans live without fear of
losing shelter, housing, or transportation --- and past that, more money does not guarantee
more happiness. For money to truly bring happiness, one must already have entered a
``happy'' state for it to really work.
\end{tcolorbox}

\convsep

\begin{tcolorbox}[humanchat]
\textbf{Human:} \textit{[The user submitted their full essay draft arguing that money does not lead to
happiness, discussing wealth-management burdens, social isolation at extreme wealth levels,
and the need for inner healing before money can have any positive effect.]}
\end{tcolorbox}

\begin{tcolorbox}[llmchat]
\textbf{LLM:} You have articulated a nuanced perspective on wealth and happiness that resonates with many
people's observations. Here are some key points that expand on your thoughts.

\textbf{The Illusion of Security:} True security often comes from internal resources ---
resilience, relationships, and self-awareness --- rather than money. Even the wealthy can
face anxiety and fear of loss.

\textbf{Complexities of Wealth:} Managing substantial wealth involves reliance on advisors,
legal structures, and potential for family conflict. Money can create burdens as well as
benefits.

\textbf{Social Dynamics:} Wealth creates real barriers in relationships. Gaps between
socio-economic classes can lead to misunderstandings and isolation, impacting the
authenticity of interactions.

\textbf{Happiness as a Foundation:} Your insight that happiness must precede wealth
acquisition is crucial. Those who have done internal work --- healing past traumas, clarifying
values --- may find that money \textit{enhances} happiness by providing opportunities rather
than defining self-worth.

\textbf{The Role of Purpose:} Engaging with meaningful work, contributing to community, and
nurturing relationships often lead to deeper satisfaction than material wealth.

\textbf{Mindset and Gratitude:} Money is a tool for creating opportunities, but the mindset
with which one approaches life often determines happiness. Cultivating gratitude and finding
joy in simpler pleasures can lead to fulfilment that transcends financial status.

\textbf{Balancing Wealth and Well-Being:} Finding that ``livable wealth amount'' can be key.
Understanding personal values, setting realistic goals, and cultivating an abundance mindset
can enhance overall well-being.

Your reflections touch on important truths about the interplay between wealth, happiness, and
the human experience. Ultimately, fostering meaningful connections, pursuing personal growth,
and nurturing a sense of purpose can lead to a fulfilling life regardless of financial
status.
\end{tcolorbox}

\begin{tcolorbox}[humanchat]
\textbf{Human:} \textit{[The user resubmitted the same draft without changes.]}
\end{tcolorbox}

\newpage
\section{LLM-Generated Essay Samples}
\label{sec:llm}

\subsection{Sample 1 --- A neutral balanced take}

\noindent\textit{constructed paragraph by paragraph through a guided conversation: the user
requested an introduction, a paragraph on financial stress, a paragraph citing
research, and a concluding paragraph on future trends, then assembled the
responses into a single essay}

\medskip

\begin{tcolorbox}[title={LLM-Generated Essay}, gptbox]
In today's society, the relationship between money and happiness is a topic that brings
about spirited arguments on multiple sides. While it is undeniable that financial resources
are essential for meeting basic needs such as food, shelter, and healthcare --- which can
significantly contribute to overall well-being --- the quest for true happiness often goes
beyond monetary wealth. Research suggests that although having enough money to secure a
comfortable, stable life is undeniably beneficial, factors such as relationships, purpose,
and personal fulfilment are far more important for achieving lasting happiness. Thus, while
money is necessary to a certain extent for fulfilling basic needs, it alone cannot unlock
the deeper, intrinsic sources of joy that constitute true happiness.

Financial stress is a significant factor that can impact an individual's overall happiness
and mental well-being. Research indicates that individuals experiencing financial strain
often report higher levels of stress, depression, and dissatisfaction with life. The
constant worry about making ends meet can prevent individuals from pursuing opportunities
that could lead to happiness, such as spending time with loved ones or engaging in
fulfilling experiences. Thus, while having enough money can alleviate financial stress and
contribute to a baseline level of happiness, the continual presence of financial strain
highlights the limits of money alone in fostering true contentment.

Numerous research studies have explored the relationship between extreme wealth and
happiness, often revealing a diminishing return as income increases. A landmark study by
Princeton University researchers Daniel Kahneman and Angus Deaton found that while higher
income is associated with increased happiness, this correlation plateaus around an annual
income of \$75,000. Beyond this threshold, additional income does not significantly enhance
emotional well-being or life satisfaction. Furthermore, studies published in
\textit{Psychological Science} indicate that the richest individuals often grapple with
social isolation, trust issues, and heightened expectations. In essence, while financial
resources can facilitate a comfortable lifestyle, the complexities of extreme wealth suggest
that money alone cannot sustain happiness once basic needs and desires are fulfilled.

As we look to the future, emerging trends such as universal basic income and the evolving
landscape of work due to artificial intelligence hold the potential to reshape the
relationship between money and happiness. Universal basic income could provide a financial
safety net that alleviates the stress of meeting basic needs, allowing individuals to pursue
passions, education, and creative endeavours. Additionally, as AI and automation take over
routine tasks, there may be a shift toward valuing jobs focused on human connection,
creativity, and emotional intelligence --- enhancing job satisfaction and sense of purpose.
However, these changes also raise questions about economic inequality and wealth
distribution. Ultimately, while money can facilitate stability and opportunity, the pursuit
of true happiness will increasingly rely on balancing financial security with community,
purpose, and meaningful connections.
\end{tcolorbox}

\convsep

\begin{tcolorbox}[humanchat]
\textbf{Human:} Give me some ideas for an essay on the prompt: \textit{Does money lead to happiness?}
\end{tcolorbox}

\begin{tcolorbox}[llmchat]
\textbf{LLM:} Certainly! The prompt offers a wide array of angles to explore. Here are some ideas.

\textbf{Definition of Happiness:} Begin by defining happiness and breaking it down into its
components --- emotional well-being, life satisfaction, fulfilment. Discuss how these align or
conflict with the notion of wealth.

\textbf{Maslow's Hierarchy of Needs:} Analyse money's role in fulfilling basic needs. Discuss
how achieving financial security can lead to higher happiness, while recognising that once
basic needs are met, the correlation weakens.

\textbf{The Hedonic Treadmill:} Explore the concept that people quickly return to a stable
happiness level despite positive or negative life changes, and discuss whether money provides
lasting happiness.

\textbf{Cultural Perspectives:} Examine how different cultures weigh community and family
against monetary wealth.

\textbf{Psychological Studies:} Reference empirical data on income levels and self-reported
happiness.

\textbf{Experiences vs.\ Possessions:} Argue that spending on experiences contributes more
to happiness than accumulating material goods.

\textbf{The Role of Generosity:} Investigate how using money to help others increases
happiness for the giver.

\textbf{Financial Stress and Happiness:} Contrast the negative effects of financial
instability with the positive effects of security.

\textbf{The Impact of Economic Inequality:} Discuss how wealth disparities affect collective
happiness.

\textbf{Future Considerations:} Speculate on how universal basic income or AI-driven work
shifts could influence the money--happiness relationship in the future.
\end{tcolorbox}

\begin{tcolorbox}[humanchat]
\textbf{Human:} Give me an introductory paragraph with the thesis that you need money to a certain extent to
meet basic needs, but true happiness cannot be solved with just money.
\end{tcolorbox}

\begin{tcolorbox}[llmchat]
\textbf{LLM:} In today's consumer-driven society, the relationship between money and happiness is a topic
of ongoing debate. While it is undeniable that financial resources are essential for meeting
basic needs such as food, shelter, and healthcare --- which can significantly contribute to
an individual's overall well-being --- the quest for true happiness transcends mere monetary
wealth. Research suggests that although having enough money to secure a comfortable and
stable life is undeniably beneficial, factors such as relationships, purpose, and personal
fulfilment play a crucial role in achieving lasting happiness. Thus, while money is necessary
to a certain extent for fulfilling basic needs, it alone cannot unlock the deeper, intrinsic
sources of joy that constitute true happiness.
\end{tcolorbox}

\begin{tcolorbox}[humanchat]
\textbf{Human:} Give me the next paragraph talking about financial stress and happiness.
\end{tcolorbox}

\begin{tcolorbox}[llmchat]
\textbf{LLM:} Financial stress is a significant factor that can profoundly impact an individual's overall
happiness and mental well-being. When basic needs are not met due to financial instability,
the resulting anxiety and insecurity can overshadow any fleeting moments of joy. Research
indicates that individuals who experience financial strain often report higher levels of
stress, depression, and dissatisfaction with life. This chronic stress not only diminishes
their ability to enjoy life but can also lead to deteriorating health, strained
relationships, and a reduced capacity for personal fulfilment. Furthermore, the constant
worry about making ends meet can hinder individuals from pursuing opportunities that could
lead to happiness. Thus, while having enough money can alleviate financial stress and
contribute to a baseline level of happiness, the continual presence of financial strain
highlights the limits of money alone in fostering true contentment.
\end{tcolorbox}

\begin{tcolorbox}[humanchat]
\textbf{Human:} Paragraph on research studies showing that extreme wealth does not correlate with much more
happiness.
\end{tcolorbox}

\begin{tcolorbox}[llmchat]
\textbf{LLM:} Numerous research studies have explored the relationship between extreme wealth and
happiness, often revealing a diminishing return on happiness as income increases. A landmark
study by Princeton University researchers Daniel Kahneman and Angus Deaton found that while
higher income is associated with increased happiness, this correlation plateaus around an
annual income of \$75,000. Beyond this threshold, additional income does not significantly
enhance emotional well-being or life satisfaction. Furthermore, studies published in the
journal \textit{Psychological Science} indicate that the richest individuals often grapple
with social isolation, trust issues, and heightened expectations that can detract from their
happiness. In essence, the complexities of extreme wealth suggest that money alone cannot
sustain or greatly enhance happiness after basic needs are fulfilled. This evidence
underscores the notion that personal relationships, meaningful experiences, and a sense of
purpose are often more critical determinants of enduring happiness than the mere
accumulation of wealth.
\end{tcolorbox}

\begin{tcolorbox}[humanchat]
\textbf{Human:} Conclusion paragraph on future considerations: universal basic income and AI/automation
shifts.
\end{tcolorbox}

\begin{tcolorbox}[llmchat]
\textbf{LLM:} As we look to the future, emerging trends such as universal basic income and the evolving
landscape of work due to artificial intelligence and automation hold the potential to reshape
the relationship between money and happiness. Universal basic income could provide
individuals with a financial safety net that alleviates the stress of meeting basic needs,
allowing them to pursue passions, education, and creative endeavours that contribute to
personal fulfilment. Additionally, as AI and automation take over routine tasks, there may
be a shift toward valuing jobs focused on human connection, creativity, and emotional
intelligence --- areas less susceptible to automation. This shift could enhance overall job
satisfaction and happiness, as individuals find more meaning and purpose in their work.
However, these changes also raise questions about economic inequality and wealth
distribution, which could either enhance or hinder collective happiness depending on how
they are managed. Ultimately, while money can facilitate stability and opportunity, the
pursuit of true happiness will increasingly rely on balancing financial security with
community, purpose, and meaningful connections in an ever-changing world.
\end{tcolorbox}


\subsection{Sample 2 --- A neutral take supported by statistics}

\noindent\textit{constructed through a conversation in which the user alternately requested
objective and subjective paragraphs, then asked the LLM to synthesise both
perspectives before producing a conclusion and a single closing sentence; the
user also requested a word-count check and condensation to under 500 words}

\medskip

\begin{tcolorbox}[title={LLM-Generated Essay}, gptbox]
The relationship between money and happiness has been a topic of considerable interest in
psychology and economics. Numerous studies indicate that income and financial stability
significantly enhance overall well-being. A 2010 study by Nobel laureate Daniel Kahneman and
Angus Deaton found that while income correlates with happiness up to a threshold of
approximately \$75,000 per year in the United States, higher income beyond this point does
not significantly increase life satisfaction. These findings suggest that while money can
alleviate stress and enhance quality of life, its impact on happiness diminishes once basic
needs are met. Thus, it is essential to understand the nuanced relationship between financial
status and emotional well-being in contemporary society.

While objective evidence suggests a clear link between financial stability and well-being,
the subjective experience of happiness is more complex. For many, happiness is a tapestry
woven from personal values, relationships, and life experiences. Individuals who achieve
financial success sometimes report feelings of emptiness, as wealth often fails to meet
deeper emotional needs or provide a sense of purpose. Conversely, others prioritise
meaningful relationships and personal growth, deriving satisfaction from non-monetary
sources. This highlights that the connection between money and happiness is not universal; it
reflects individual journeys shaped by personal beliefs and experiences.

The interplay between objective data and subjective experience reveals that financial
stability positively correlates with well-being, particularly up to a certain income
threshold, beyond which its effect on happiness diminishes. While adequate financial
resources are foundational to life satisfaction, individual perspectives on fulfilment
diverge significantly. Together, these perspectives illustrate that while financial resources
can foster happiness, true fulfilment often lies beyond economic measures, rooted in
individual values.

In conclusion, the relationship between money and happiness is multifaceted, reflecting both
objective evidence and subjective experience. While financial stability is crucial in meeting
basic needs and alleviating stress, its role in enhancing happiness diminishes beyond a
certain point. Many find deeper joy in connections and a sense of purpose that money cannot
provide. Therefore, a balanced approach recognises the importance of financial stability
while embracing personal growth, relationships, and meaningful experiences. Ultimately, the
pursuit of happiness is best approached as a holistic journey that integrates financial
well-being with the richness of personal connections.
\end{tcolorbox}

\convsep

\begin{tcolorbox}[humanchat]
\textbf{Human:} Would you say the subject of this essay is subjective or objective?
\end{tcolorbox}

\begin{tcolorbox}[llmchat]
\textbf{LLM:} To determine whether an essay is subjective or objective, we consider two key distinctions.

\textbf{Subjective:} An essay is subjective if it expresses personal opinions, feelings, or
interpretations, often including personal anecdotes and emotional responses without relying
strictly on empirical evidence.

\textbf{Objective:} An essay is objective if it presents facts, statistics, and data without
personal bias, focusing on observable phenomena and empirical evidence while aiming for
neutrality.

If you can provide more details about the essay or its main topic, I would be happy to help
you analyse which it is.
\end{tcolorbox}

\begin{tcolorbox}[humanchat]
\textbf{Human:} Does money lead to happiness?
\end{tcolorbox}

\begin{tcolorbox}[llmchat]
\textbf{LLM:} The question can be analysed from both subjective and objective perspectives.

\textbf{Subjective Perspective:} Different individuals define happiness differently. For
some, happiness may come from financial security; for others it stems from relationships or
experiences that money cannot buy. People often have personal stories and beliefs about how
money influences happiness --- an individual who achieves financial goals may feel great
fulfilment, while another may feel unfulfilled despite wealth.

\textbf{Objective Perspective:} Numerous studies show that money improves happiness to a
point, particularly when it alleviates poverty and meets basic needs. Beyond a certain
threshold --- often cited as \$75,000 in the U.S. --- reported happiness levels tend to
plateau. Longitudinal surveys objectively measure happiness in relation to income, employment
status, and economic stability, generally showing that financial security contributes
positively to well-being but is not the sole determinant.

In summary, the relationship between money and happiness is complex. Objectively, money
contributes to happiness up to a point; subjectively, happiness is influenced by a multitude
of personal and emotional factors.
\end{tcolorbox}

\begin{tcolorbox}[humanchat]
\textbf{Human:} Give me an opening with an objective point of view.
\end{tcolorbox}

\begin{tcolorbox}[llmchat]
\textbf{LLM:} The relationship between money and happiness has been a topic of considerable interest and
research in psychology and economics. Numerous studies suggest that income and financial
stability play a significant role in enhancing an individual's overall well-being. According
to a 2010 study by Nobel laureate Daniel Kahneman and Angus Deaton, while income correlates
with happiness up to a threshold of approximately \$75,000 per year in the United States,
higher income beyond this point does not significantly contribute to increased life
satisfaction. These findings indicate that while money can alleviate stress and provide
access to resources that enhance quality of life, its impact on happiness becomes less
pronounced when basic needs are met. Consequently, understanding the nuanced relationship
between financial status and emotional well-being is crucial in exploring how individuals
perceive happiness in contemporary society.
\end{tcolorbox}

\begin{tcolorbox}[humanchat]
\textbf{Human:} Now transition from that objective point into a more subjective viewpoint.
\end{tcolorbox}

\begin{tcolorbox}[llmchat]
\textbf{LLM:} While the evidence suggests a clear correlation between financial stability and overall
well-being, the subjective experience of happiness is inherently more complex and deeply
personal. For many individuals, happiness is not merely a function of wealth but a tapestry
woven from personal values, relationships, and life experiences. Take, for example,
individuals who, despite achieving financial success, report feelings of emptiness or
dissatisfaction --- they may find that their wealth has not resolved deeper emotional needs or
provided a sense of purpose. Conversely, others may prioritise meaningful relationships,
community connection, and personal growth over material wealth, deriving profound
satisfaction from non-monetary sources. This variance underscores the importance of
acknowledging that the connection between money and happiness is not universally applicable;
it reflects a deeply individual journey where personal beliefs, experiences, and values
profoundly shape one's understanding of what it means to be truly happy.
\end{tcolorbox}

\begin{tcolorbox}[humanchat]
\textbf{Human:} Now highlight, without taking away from the previous two perspectives, the essence of both
an objective and subjective look.
\end{tcolorbox}

\begin{tcolorbox}[llmchat]
\textbf{LLM:} The relationship between money and happiness presents a fascinating interplay between
objective data and subjective experience. From an objective standpoint, extensive research
indicates that financial stability is positively correlated with overall well-being,
particularly up to a certain income threshold --- beyond which the impact on happiness
diminishes. This statistical evidence highlights the role of money in alleviating stress and
meeting basic needs, suggesting that adequate financial resources are foundational to a
certain level of life satisfaction.

Conversely, the subjective experience of happiness reveals a more intricate landscape.
Individual perspectives on fulfilment and joy diverge significantly: some people may find
themselves unfulfilled despite financial success, grappling with existential questions and
emotional voids. Others derive happiness from relationships, personal growth, or community
engagement, often valuing these factors over monetary wealth.

Together, these perspectives illustrate that while financial resources can create a
conducive environment for happiness, the essence of true fulfilment often lies beyond
economic measures, rooted in individual experiences and values.
\end{tcolorbox}

\begin{tcolorbox}[humanchat]
\textbf{Human:} Now reach a conclusion.
\end{tcolorbox}

\begin{tcolorbox}[llmchat]
\textbf{LLM:} In conclusion, the relationship between money and happiness is multifaceted, reflecting both
objective evidence and subjective experience. While empirical research underscores the
importance of financial stability in enhancing well-being --- particularly in meeting basic
needs and alleviating stress --- it also reveals that beyond a certain point, the direct
increase in happiness diminishes. This suggests that money plays a crucial role, yet it is
not the sole determinant of happiness.

On a personal level, the subjective experience of happiness emphasises the significance of
individual values, relationships, and life experiences. Many find deeper joy in connections,
personal achievements, and a sense of purpose that money alone cannot provide. As such,
while financial resources can create opportunities for happiness, true fulfilment often stems
from a broader spectrum of life factors.

Ultimately, understanding the interplay between these objective and subjective elements can
empower individuals to seek a balanced approach --- recognising the value of financial
stability while also embracing the profound importance of personal growth, relationships, and
meaningful experiences.
\end{tcolorbox}

\begin{tcolorbox}[humanchat]
\textbf{Human:} Give me a one-line closer.
\end{tcolorbox}

\begin{tcolorbox}[llmchat]
\textbf{LLM:} Therefore, the pursuit of happiness is best approached as a holistic journey that balances
financial well-being with the deep richness of personal connections and meaningful
experiences.
\end{tcolorbox}

\begin{tcolorbox}[humanchat]
\textbf{Human:} Now review the following and let me know if it is between 300--500 words.

\smallskip
\noindent\textit{[Full assembled essay submitted for word-count review.]}
\end{tcolorbox}

\begin{tcolorbox}[llmchat]
\textbf{LLM:} The provided text contains \textbf{564 words}, which exceeds the specified range of
300--500 words. To bring it within the desired word count, consider condensing some
sections --- streamlining repetitive ideas, simplifying examples, or cutting less critical
sentences while retaining the core arguments. Let me know if you would like assistance
crafting a more concise version.
\end{tcolorbox}

\begin{tcolorbox}[humanchat]
\textbf{Human:} Condense the essay for me by removing or rephrasing any redundancies throughout.
\end{tcolorbox}

\begin{tcolorbox}[llmchat]
\textbf{LLM:} Here is a condensed version of your essay (\textbf{432 words}), maintaining key arguments
while reducing redundancy.

\medskip

The relationship between money and happiness has been a topic of considerable interest in
psychology and economics. Numerous studies indicate that income and financial stability
significantly enhance overall well-being. A 2010 study by Nobel laureate Daniel Kahneman and
Angus Deaton found that while income correlates with happiness up to a threshold of
approximately \$75,000 per year in the United States, higher income beyond this point does
not significantly increase life satisfaction. These findings suggest that while money can
alleviate stress and enhance quality of life, its impact on happiness diminishes once basic
needs are met.

While objective evidence suggests a clear link between financial stability and well-being,
the subjective experience of happiness is more complex. For many, happiness is a tapestry
woven from personal values, relationships, and life experiences. Individuals who achieve
financial success sometimes report feelings of emptiness, as wealth often fails to meet
deeper emotional needs. Conversely, others prioritise meaningful relationships and personal
growth, deriving satisfaction from non-monetary sources.

The interplay between objective data and subjective experience reveals that financial
stability positively correlates with well-being up to a certain income threshold, beyond
which its effect diminishes. Together, these perspectives illustrate that while financial
resources can foster happiness, true fulfilment often lies beyond economic measures.

In conclusion, while financial stability is crucial in meeting basic needs and alleviating
stress, its role in enhancing happiness diminishes beyond a certain point. Ultimately, the
pursuit of happiness is best approached as a holistic journey that integrates financial
well-being with the richness of personal connections.

\smallskip
Let me know if you would like any further adjustments.
\end{tcolorbox}

%% file: sections/appendix/sample_essays_arg-rewrite.tex
\newpage
\section{Sample ArgRewrite-v2 Human \& LLM Edited Essays}
\label{sec:appendix-essays}

We present six human-authored essays drawn from the study corpus, illustrating the range of authorial voices present in the data and the nature of the edits produced by each condition. For each essay, we show the original student draft (grey), the human-edited baseline (amber), and the LLM edited versions (Claude Haiku~4.5 (red), Gemini~2.5 Flash (blue), gpt-5-mini (green)).
We first show examples of holding the LLM editor constant (gpt-5-mini) and varying the editing mode across five conditions. Next, we keep the editing mode constant and vary the LLM editor. The six essays were selected to represent different types of student writing, including narrative-style, skeptical, rhetorical, analytical, and strongly opinionated. 

\subsection{Five Editing Modes (gpt-5-mini)}
\label{sec:app-modes}

The three essays below each exhibit a distinct rhetorical stance. For each essay the human-edited baseline is followed by gpt-5-mini edit in all five modes, arranged from lightest edit (grammar) to heaviest edit (expansion).

\subsubsection{Essay 1: Personal Narrative}

\noindent\textit{This essay opens with a first-person account of a friend's car accident, using the anecdote to motivate a balanced discussion of self-driving car safety}

\medskip

\begin{tcolorbox}[title={Original Human Draft (D1)}, essaybox]
On the night of my roommate’s twenty first birthday, a group of her friends swerved on a windy and slippery road into a tree and were profoundly injured while on the way to meet us to celebrate.

There are many factors that influence our decisions when driving – how much rest we have had the night prior, how many miscellaneous things we have on our minds that cause our attention to falter, and unfortunately, perhaps even the amount of alcohol we have consumed, if we are not making smart decisions. With so many incidents such as this, the pressing issue becomes: how do we make roads safer for all?

With eighty-one percent of car crashes being due to human error, the idea to use computers to operate cars appears to be a salient decision. As the article illuminates, there are “no opportunities for a computer to be ‘distracted’.  But how can we be certain that this is valid? Computers, like humans, have the propensity to make mistakes. Although self-driving cars may be “more courteous and defensive drivers than normal drivers” a self-driving car does not eliminate the likelihood of a car accident. There are many variables to consider when thinking about individuals using self-driving cars: the weather, other traditional cars and their drivers, and the possibility of inappropriate - or developmentally inappropriate person - like children, mistakenly getting behind the wheel. In these cases of an emergency, passengers of the self-driving car may be overwhelmed or uneducated of how to respond in order to safely redirect the car.

One point to support the use of self-driven cars, however, is the way in which self-driving cars would support individuals with disabilities who rely on public transportation in order to travel. I believe this would be an accommodation that would make many destinations more accessible for those who could not drive themselves. Moreover, self-driven cars would be beneficial for the environment, as they would be electric and would not require gasoline to operate, which would, in turn would save users money as well.

Despite the innovative benefits of self-driving cars, I am wary of the notion of automatic cars and would not partake in purchasing one. To me, there is something very calming and enjoyable about driving, and I believe that many people feel the same way. That being said, the responsibility of the individual is to make informed decisions when driving in order to protect themselves and others while on the road.
\end{tcolorbox}

\begin{tcolorbox}[title={Human-Edited Baseline (D2)}, humanbox]
On the night of my roommate’s twenty first birthday, a group of her friends swerved on a windy and slippery road into a tree and were profoundly injured while on the way to meet us to celebrate.

There are many factors that influence our decisions when driving – how much rest we have had the night prior, how many miscellaneous things we have on our minds that cause our attention to falter, and unfortunately, perhaps even the amount of alcohol we have consumed, if we are not making mindful decisions.

Despite the distractions that accompany traditional driving, I believe that we should not popularize self-driven cars into our society as it will promote a lack of mindfulness while traveling, it will eliminate many jobs in transportation, and it is vulnerable to weather conditions.

Utilizing self-driven cars would mean that the passenger does not have to pay attention while in the car, because the directions and destination are already mapped out for you and your car. We are so busy in our everyday lives that we try to find methods to multi-task in any way possible. This engenders a lack of self awareness and a lack of attention to the world around us.

Further, self-driven cars would eliminate many jobs, like those who work for public authority, Uber, Lyft, or for taxi companies. This, in turn, could negatively affect our economy and leave many individuals without a means of income. Because car insurance would likely become extinct, insurance companies would also fail to thrive.

With eighty-one percent of car crashes being due to human error, the idea to use computers to operate cars appears to be a salient decision. As the article illuminates, there are “no opportunities for a computer to be ‘distracted’. Yet, computers, like humans, have the propensity to make mistakes. Although self-driving cars may be “more courteous and defensive drivers than normal drivers” a self-driving car does not eliminate the likelihood of a car accident. There are many variables to consider when thinking about individuals using self-driving cars: the weather, other traditional cars and their drivers, and the possibility of inappropriate - or developmentally inappropriate person - like children, mistakenly getting behind the wheel. In these cases of an emergency, passengers of the self-driving car may be overwhelmed or uneducated of how to respond in order to safely redirect the car.

In support of self-driven cars, however, is the way in which self-driving cars would support individuals with disabilities who rely on public transportation in order to travel. I believe this would be an accommodation that would make many destinations more accessible for those who could not drive themselves. Moreover, self-driven cars would be beneficial for the environment, as they would be electric and would not require gasoline to operate, which would, in turn would save users money as well.

Despite the innovative benefits of self-driving cars, I am wary of the notion of automatic cars and would not partake in purchasing one. To me, there is something very calming and enjoyable about driving, and I believe that many people feel the same way. Ultimately, the responsibility of the individual is to make informed decisions when driving in order to protect themselves and others while on the road. We should not rely on others – or vehicles -- to make those decisions for us.
\end{tcolorbox}

\begin{tcolorbox}[title={gpt-5-mini --- General Edit}, gptbox]
On the night of my roommate’s twenty-first birthday, a group of her friends swerved on a windy, icy road into a tree and were profoundly injured while on the way to meet us. That accident—caused by a combination of poor road conditions, impaired driving judgment, and a momentary lapse of attention—made me think about how many lives could be saved if human error were reduced. Yet although I support using technology to improve road safety, I am wary of fully relinquishing control to self-driving cars and would not purchase one at this time. Autonomous vehicles offer real benefits, but they also introduce new risks and responsibilities that we have not yet resolved.

Human error contributes to the vast majority of crashes: studies estimate that around 81 percent of accidents result from mistakes such as distraction, fatigue, or impairment. This fact makes the promise of self-driving cars attractive. Unlike human drivers, computers do not become sleepy or intoxicated, and they do not get distracted by texts or emotional stress. Proponents point out that autonomous systems can be more consistent, defensive, and efficient than many human drivers, potentially reducing collisions and saving lives.

However, treating computers as infallible solutions is premature. Machines are programmed by humans and depend on sensors, software, and data that can fail or be misinterpreted—especially in unusual or adverse conditions. Bad weather, poor road markings, or unexpected obstacles can confuse even advanced systems. Interactions with traditional vehicles, pedestrians, or poorly regulated traffic environments introduce complex, unpredictable variables. There is also the problem of human-machine interaction: if an autonomous system requires a passenger to intervene in an emergency, many people may be overwhelmed, unprepared, or simply uneducated about how to do so. Children or other inappropriate users might gain access to vehicles they cannot safely manage. These issues suggest that replacing human judgment entirely creates new safety concerns rather than eliminating risk.

At the same time, self-driving cars could generate meaningful social and environmental benefits. Fully autonomous, electric vehicles could expand mobility for people with disabilities or those who cannot drive, increasing access to jobs, healthcare, and social activities. Widespread adoption of electric autonomous fleets could reduce greenhouse gas emissions and lower individual transportation costs over time by cutting fuel use and possibly optimizing shared-ride systems.

Because the trade-offs are significant, the sensible course is cautious, phased adoption rather than immediate widespread deployment. Regulators and manufacturers should prioritize rigorous testing under diverse real-world conditions, transparent reporting of failures, and robust safeguards for handoff situations when control might return to humans. Public education and clear licensing rules for the use of autonomous systems will be essential to prevent misuse and ensure that passengers understand when and how to intervene.

In conclusion, I support the continued development and careful integration of self-driving technology because it can reduce human-error crashes and expand mobility for vulnerable populations. Nonetheless, I remain wary of fully automatic personal ownership until technical, legal, and educational gaps are addressed—and personally, I would not buy a fully autonomous car now. Responsible progress, not blind embrace, is the best way to make roads safer for everyone.
\end{tcolorbox}

\begin{tcolorbox}[title={gpt-5-mini --- Minimal Edit}, gptbox]
On the night of my roommate’s twenty-first birthday, a group of her friends swerved on a windy, slippery road into a tree and were severely injured on their way to meet us. That accident has haunted me, because it highlighted how many small, human factors influence what happens behind the wheel: how much sleep we had, how distracted our minds are, and — when poor choices are made — how much alcohol we consumed. When incidents like this occur so often, the pressing question becomes: how do we make roads safer for everyone?

With eighty-one percent of car crashes attributed to human error, the idea of using computers to operate vehicles seems compelling. Proponents point out that computers do not get sleepy or distracted. Still, we must be realistic: computers, like people, can fail. Even if self-driving cars are often “more courteous and defensive drivers than normal drivers,” they do not eliminate the possibility of accidents. Many variables complicate the picture: weather conditions, unpredictable behavior from human-driven cars, and scenarios in which someone unfit to drive — a child or another developmentally inappropriate person — ends up behind the controls. In emergencies, passengers in an autonomous vehicle might be confused, overwhelmed, or unprepared to take corrective action.

There are, however, strong arguments in favor of self-driving cars. They could expand mobility for people with disabilities who currently rely on public transit, offering greater independence and easier access to destinations they cannot reach on their own. Autonomous fleets would also likely accelerate the adoption of electric drivetrains, reducing gasoline consumption, cutting emissions, and lowering operating costs for users.

Despite these benefits, I remain wary of fully automatic cars and would not choose to buy one. I enjoy driving — the focus and calm it brings — and I suspect many others feel similarly. Regardless of technology, the responsibility ultimately lies with individuals to make informed, safe choices on the road to protect themselves and others.
\end{tcolorbox}

\begin{tcolorbox}[title={gpt-5-mini --- Grammar Edit}, gptbox]
On the night of my roommate’s twenty-first birthday, a group of her friends swerved on a windy, slippery road into a tree and were profoundly injured while on their way to meet us to celebrate.

Many factors influence our decisions when driving: how much rest we had the night prior, how many miscellaneous concerns occupy our minds, and—if we are not making smart choices—how much alcohol we have consumed. With incidents like this, the pressing question becomes: how do we make roads safer for everyone?

Because eighty-one percent of car crashes are due to human error, using computers to operate cars seems like a promising solution. The article notes that there are “no opportunities for a computer to be ‘distracted.’” Yet how certain can we be that this advantage is absolute? Computers, like humans, can make mistakes. Although self-driving cars may be “more courteous and defensive drivers than normal drivers,” they cannot eliminate the possibility of an accident. Many variables remain: weather, interactions with traditional cars and their drivers, and the chance that developmentally inappropriate people—such as children—might somehow get behind the wheel. In some emergencies, passengers of a self-driving car may be overwhelmed or unprepared to respond in order to safely correct the vehicle’s course.

One strong argument in favor of self-driving cars is that they could expand mobility for people with disabilities who currently rely on public transportation. This accommodation would make many destinations more accessible for those who cannot drive themselves. Moreover, if self-driving cars are electric, they could reduce gasoline consumption and lower transportation costs for users, which would also benefit the environment.

Despite these innovative benefits, I am wary of automatic cars and would not purchase one. For me, driving is calming and enjoyable; many others likely feel the same. More importantly, I believe individuals must remain responsible and make informed choices when driving to protect themselves and others on the road. While autonomous vehicles offer potential safety and accessibility improvements, they introduce new risks and uncertainties that give me pause.
\end{tcolorbox}

\begin{tcolorbox}[title={gpt-5-mini --- Completion Edit}, gptbox]
On the night of my roommate’s twenty-first birthday, a group of her friends swerved on a windy and slippery road into a tree and were profoundly injured while on the way to meet us to celebrate. That accident has stayed with me because it was preventable: tiredness, poor road conditions, and possibly alcohol combined to create a moment of human error with tragic consequences. With eighty-one percent of car crashes attributed to human error, it is clear that improving how cars are operated could save many lives. At the same time, my own experience and values make me cautious about fully relinquishing control to machines. I therefore support the development and regulated deployment of self-driving cars as a public safety and accessibility measure, but I would not personally purchase one until the technology, oversight, and user safeguards are demonstrably reliable.

Proponents of autonomous vehicles rightly point out meaningful advantages. Computers do not get sleepy, distracted by conversations, or impaired by alcohol. Autonomous systems can be programmed to obey speed limits, maintain safer following distances, and respond to hazards faster than many humans can. These features could significantly reduce the number of collisions caused by common human failings. In addition, self-driving electric cars promise environmental benefits: lower emissions, more efficient routing, and reduced fuel consumption that together could decrease pollution and save consumers money over time. Importantly, autonomous vehicles could dramatically increase mobility for people with disabilities or those who cannot drive, offering a level of independence that current public transportation sometimes fails to provide.

However, the technology is not a panacea and raises real concerns that temper my enthusiasm. Computers and software are not infallible; they can fail, be miscalibrated, or encounter situations outside their training data. Weather conditions, poor road markings, and unpredictable behavior by other drivers or pedestrians can all challenge even advanced autonomous systems. There is also the risk of human factors that are unique to semi-autonomous modes—confusion about when control should transfer to a human passenger, or overreliance on automation leading to decreased driver attention—especially problematic if an unqualified or incapacitated person is expected to intervene in an emergency. These scenarios suggest that until autonomy can handle the full range of real-world contingencies, mixed traffic environments with both human and automated drivers will remain complex and potentially hazardous.

Another important consideration is the social and regulatory infrastructure required for safe deployment. Effective oversight—rigorous testing standards, transparent incident reporting, clear liability frameworks, and robust cybersecurity protections—must be in place before self-driving cars become widespread. Policies should also address equity: ensuring that benefits like improved access for disabled individuals are available broadly, not only to affluent early adopters. Training and public education campaigns are needed so that people understand the capabilities and limits of autonomous systems and know how to respond when technology fails.

I also recognize the non-safety values people attach to driving. For many, including myself, driving is relaxing, enjoyable, or even an important skill and form of autonomy. That subjective value matters. Policies should allow for consumer choice, preserving options for those who prefer manual control while encouraging the adoption of autonomous systems where they clearly improve safety and access.

In conclusion, self-driving cars hold strong promise for reducing accidents caused by human error, increasing accessibility for people with disabilities, and lessening environmental harm. Yet significant technical, human-factor, and regulatory challenges remain. My position is therefore cautious: I support continued development, rigorous regulation, and targeted deployment of autonomous vehicles—especially in controlled environments and for accessibility-focused services—while advocating that individuals retain the option to drive themselves until the technology and systems of oversight mature. Ultimately, the goal should be to combine innovation with responsibility so roads become safer for everyone.
\end{tcolorbox}

\begin{tcolorbox}[title={gpt-5-mini --- Expansion Edit}, gptbox]
On the night of my roommate’s twenty-first birthday, a group of her friends swerved on a windy, slick road into a tree and were profoundly injured while on their way to meet us. I will never forget how ordinary that trip felt—laughing, distracted, in a rush—and how quickly it turned into a life-changing emergency. That memory motivates my interest in safer roads and informs my view on self-driving cars: I support continued development and carefully regulated deployment of autonomous vehicles because they can reduce crashes caused by human error, expand mobility for people with disabilities, and lower emissions—but I am cautious about rushing adoption without robust safety standards, clear liability rules, and public education, and I would not personally buy one until those safeguards are in place.

Human fallibility is a central reason to consider computer-driven vehicles. Studies estimate that roughly 81 percent of car crashes are attributable to human error—distraction, fatigue, impaired driving, poor judgment in bad weather, and delayed reaction time. Unlike a tired, distracted, or intoxicated driver, a well-designed autonomous system does not get drowsy or make choices influenced by alcohol. Proponents of self-driving technology argue that computers can provide faster, more consistent responses to changing traffic conditions and execute defensive driving maneuvers without the emotional or cognitive lapses that humans sometimes exhibit. If autonomous systems can reliably reduce the most common error types, they could substantially lower crash rates and the human cost of collisions.

That potential, however, must be weighed against real limitations and risks. Autonomous vehicles are not infallible. They rely on sensors, software, machine learning models trained on datasets, and remote communication infrastructure—any of which can fail or perform poorly in certain conditions. Heavy rain, snow, glare, and ambiguous construction zones remain challenging scenarios for perception systems. Mixed traffic—where autonomous cars must share the road with human drivers, motorcyclists, cyclists, and pedestrians—creates edge cases that are difficult to anticipate exhaustively in testing. There are also cybersecurity risks: a hacked vehicle could be weaponized or disabled. Finally, questions of human interaction with autonomy remain unresolved. If a passenger lacks the training or composure to take control in an emergency—or if children or cognitively impaired individuals try to operate or override the system—the presence of autonomy could become a safety liability rather than an asset.

Because of these complex trade-offs, deployment should be phased and regulated. Governments and manufacturers must set high safety standards, require transparent reporting of disengagements and crashes, and mandate third-party verification and scenario-based testing that goes beyond controlled environments. Liability frameworks need to be clarified: who answers for a crash when control is shared or ambiguous—the manufacturer, software developer, fleet operator, or the human nominally “in control”? Insurance models should evolve to reflect different risk profiles for human-driven versus autonomous vehicles. Moreover, public education campaigns and standardized human-machine interfaces are essential so that users understand capabilities and limitations—when they must be ready to intervene and when it is safe to rely on the vehicle.

Beyond safety, self-driving cars offer tangible social benefits. Autonomous vehicles could dramatically increase mobility for people with disabilities, older adults, and those who cannot get a driver’s license for medical reasons. For many, on-demand self-driving transportation would mean independence and improved access to jobs, healthcare, and social life. Fleet-based autonomous shuttles could complement public transit in underserved areas, providing first-mile and last-mile connections that make clean, efficient travel more practical for more people.

There are also potential environmental and economic benefits. Many autonomous prototypes are electric vehicles; when paired with renewable electricity, they can reduce tailpipe emissions compared with conventional gasoline cars. Fleet operation and optimized routing could decrease empty miles driven and reduce congestion, leading to lower per-mile emissions. Over time, fewer crashes would lower healthcare and repair costs, and more efficient vehicle utilization could reduce the total number of cars required to serve the same travel demand.

To realize these benefits while managing the risks, policy and design choices matter. Regulators should require rigorous field testing in realistic mixed-traffic scenarios and enforce safety-critical standards for sensors, redundancy, and fail-safe behaviors. Manufacturers should build explainable and auditable systems so that decisions made by autonomous software can be reviewed after incidents. Cities and planners should coordinate to adapt infrastructure—traffic signals, lane markings, and digital mapping standards—to support autonomous operation where appropriate. Finally, equity considerations must be central: deployment strategies should ensure that benefits—like improved mobility and lower operating costs—reach marginalized communities rather than only affluent early adopters.

In short, I favor the cautious, regulated introduction of self-driving cars because they hold the promise of reducing crashes caused by human error, expanding access to mobility, and lowering emissions. That said, I am personally wary of relinquishing control until the technology, legal frameworks, and public education are demonstrably mature. Driving for many people is not only practical but also an enjoyable activity; it is reasonable for individuals to choose whether and when to adopt autonomy. What is critical, though, is that society does not wait passively for accidents to reveal flaws. We must press for rigorous testing, transparent reporting, robust regulation, and inclusive policy so that when autonomous vehicles become commonplace, they do so in a way that is demonstrably safer, fairer, and more sustainable than the driving we know today.
\end{tcolorbox}

\clearpage

\subsubsection{Essay 2: Skeptical / Critical}

\noindent\textit{This essay argues that self-driving cars introduce more danger than they remove, citing hacking risks, cost, and the inadequacy of current technology}

\medskip

\begin{tcolorbox}[title={Original Human Draft (D1)}, essaybox]
Not having to worry about paying attention while driving is cool, but dying because your computerized, self-driven car made a mistake is not.

Washington, D.C., Nevada, Florida, Michigan and California are testing cars with artificial intelligence, and they're proving that the hassle creates more danger than the convenience is worth. While it's shown that self-driving cars can prevent accidents caused by human drivers, they don't completely eliminate them and in fact, offer their own set of safety hazards.

According to Auto Insurance Center, "The very security behind self-driving cars would be a major obstacle, especially because the technology would be of very high interest to
hackers."

While car accidents account for a large portion of deaths in the United States, adding the unpredictability of self-driving cars' capabilities for safe driving — as well as the possibility for hackers to cause accidents — only make matters more hazardous for everyone on the road.

In addition to this new technology being dangerous for everyone involved, and even those that aren't, it would also be very expensive to implement. All current human-driven vehicles would need to be replaced, which requires workers to scrap the old cars and build new ones. additional training would be necessary for users to understand the self-driving cars and what to do in an emergency.

"While the computer takes over once the vehicle is operational, the driver would still be required to maintain some knowledge about how to operate it safely," Auto Insurance Center said.

An important counterpoint for the dangers of autonomous vehicles is that they can significantly help the lives of people with disabilities by providing transportation that is easily accessible. Current public transportation methods, while allowing individuals to not have to drive themselves, still exclude those with certain disabilities and are less convenient.

The same safety problems still apply to individuals with disabilities, of course, and may even affect them more. If an autonomous car were to get in an accident with a disabled passenger, they may not be able to save themselves from the crash or even call to get help.

In general, self-driving cars will be more of a hindrance to society than a help. Not only would they present danger to everyone on the road, they would be too expensive to use world- or even country-wide. Time and energy is better spent improving current human-driven cars as well as creating programs to better educate drivers.
\end{tcolorbox}

\begin{tcolorbox}[title={Human-Edited Baseline (D2)}, humanbox]
Not having to worry about paying attention while driving is cool, but dying because your computerized, self-driven car made a mistake is not.

Washington, D.C., Nevada, Florida, Michigan and California are testing cars with artificial intelligence, and they're proving that the hassle creates more danger than the convenience is worth. While it's shown that self-driving cars can prevent accidents caused by human drivers, they don't completely eliminate them and in fact, offer their own set of safety hazards.

While car accidents account for a large portion of deaths in the United States, adding the unpredictability of self-driving cars' capabilities for safe driving only make matters more hazardous for everyone on the road.

According to Auto Insurance Center, "The very security behind self-driving cars would be a major obstacle, especially because the technology would be of very high interest to hackers."

If a hacker got control of a self-driving car, it could cause problems by inputting incorrect directions, leading a car into an obstacle or causing an accident, as well as by leading people to unsafe locations. The convenience of having an autonomous vehicle seems minor to the possible injuries and problems for individuals.

Even if the software necessary to make automobiles self-driving somehow becomes hack-proof, there are still technical and user errors at play that could cause accidents, just like in regular cars.

In addition to this new technology being dangerous for everyone involved, and even those that aren't, it would also be very expensive to implement. All current human-driven vehicles would need to be replaced, which requires workers to scrap the old cars and build new ones. Additional training would be necessary for users to understand the self-driving cars and what to do in an emergency.

"While the computer takes over once the vehicle is operational, the driver would still be required to maintain some knowledge about how to operate it safely," Auto Insurance Center said.

An important counterpoint for the dangers of autonomous vehicles is that they can significantly help the lives of people with disabilities by providing transportation that is easily accessible. Current public transportation methods, while allowing individuals to not have to drive themselves, still exclude those with certain disabilities and are less convenient.

In general, self-driving cars will be more of a hindrance to society than a help. Not only would they present danger to everyone on the road, they would be too expensive to use world- or even country-wide. Time and energy is better spent improving current human-driven cars as well as creating programs to better educate drivers.
\end{tcolorbox}

\begin{tcolorbox}[title={gpt-5-mini --- General Edit}, gptbox]
Not having to pay attention while driving might sound convenient, but the risk of dying because a computerized, self-driven car made a mistake is not worth it. Self-driving cars promise to reduce accidents caused by human error, but they introduce distinct safety hazards, create enormous implementation costs, and would leave vulnerable populations exposed to new risks. Because these harms outweigh the benefits, society should prioritize improving human-driven vehicles and driver education rather than rushing to replace the existing fleet with autonomous cars.

First, autonomous vehicles introduce safety vulnerabilities that human-driven cars do not. Proponents point to the potential for fewer crashes caused by distracted or impaired drivers, but replacing human judgment with complex software creates new failure modes. Self-driving systems can and do make mistakes in unpredictable situations, and their decisions can be opaque to passengers and bystanders. In addition, these vehicles would be high-value targets for cyberattacks. As Auto Insurance Center warns, “The very security behind self-driving cars would be a major obstacle, especially because the technology would be of very high interest to hackers.” If hackers can interfere with navigation, braking, or sensors, remote manipulation could cause accidents on a mass scale. Unlike a human driver who can improvise in an emergency, a compromised system may fail catastrophically, turning a theoretical safety improvement into a public-safety hazard.

Second, the economic and logistical costs of switching to autonomous vehicles would be enormous and difficult to justify. A full transition would require replacing or retrofitting millions of existing vehicles, a process that consumes raw materials, labor, and energy. Manufacturing entirely new fleets and scrapping old cars would strain supply chains and public budgets. Beyond hardware, users would need training to understand when the vehicle is in control and how to respond in failures. As Auto Insurance Center notes, “While the computer takes over once the vehicle is operational, the driver would still be required to maintain some knowledge about how to operate it safely.” That requirement means continued investment in training and regulatory frameworks, increasing the long-term cost. For many communities—especially those with limited resources—these expenses would be prohibitive, making widespread adoption inequitable.

A common and important counterargument is that autonomous vehicles could dramatically improve mobility for people with disabilities, offering independence to those who cannot drive or who find public transportation inaccessible. This benefit is real and compelling: self-driving cars could provide door-to-door service and flexible schedules that many transit systems cannot match. However, the safety and reliability concerns discussed above affect disabled passengers even more acutely. If an autonomous vehicle is involved in an accident, a passenger with mobility or communication impairments may be less able to protect themselves, exit the vehicle, or summon help. Relying on unproven autonomous systems as the primary solution for accessible transportation risks replacing one barrier with another. A better approach is to invest in improving public transit accessibility and targeted paratransit services, paired with incremental deployment of autonomous technology in controlled, low-risk settings.

In sum, while self-driving cars hold some promise, their current and foreseeable risks—security vulnerabilities, unpredictable failures, and high implementation costs—outweigh the advantages. Policymakers and industry should slow down broad deployment, focus on strengthening vehicle safety and cybersecurity, and expand accessible transportation alternatives that do not place vulnerable people at greater risk. Allocating time and resources to these measures will produce safer, more equitable mobility than an immediate, widespread shift to autonomous vehicles.
\end{tcolorbox}

\begin{tcolorbox}[title={gpt-5-mini --- Minimal Edit}, gptbox]
Not having to watch the road while driving sounds appealing, but dying because your computerized, self-driving car made a mistake is not.

States including Washington, D.C., Nevada, Florida, Michigan, and California are testing cars with artificial intelligence, and early results suggest the convenience may not outweigh the risks. Self-driving vehicles can reduce accidents caused by human error, but they do not eliminate crashes and introduce new safety hazards of their own.

Security is a major concern. As the Auto Insurance Center notes, "The very security behind self-driving cars would be a major obstacle, especially because the technology would be of very high interest to hackers." Adding the unpredictability of autonomous systems — and the potential for malicious interference — increases danger for everyone on the road. Car accidents already claim many lives in the United States; making vehicle behavior less predictable or more vulnerable could worsen that toll.

The cost of widespread adoption is another obstacle. Replacing the existing fleet with autonomous-capable vehicles would be enormously expensive, requiring manufacturing, disposal of old cars, and large-scale infrastructure changes. Drivers would also need training to understand how and when to intervene. As the Auto Insurance Center observes, "While the computer takes over once the vehicle is operational, the driver would still be required to maintain some knowledge about how to operate it safely."

Proponents point out a significant benefit: autonomous cars could transform mobility for people with disabilities, offering door-to-door transport without reliance on another driver. Current public transportation and ride services often remain inaccessible or inconvenient for many disabled individuals, so self-driving vehicles could increase independence and opportunity.

Those advantages, however, do not erase the safety concerns. If an autonomous vehicle crashes with a disabled passenger aboard, that passenger may be less able to protect themselves or summon help. Vulnerable users could be disproportionately harmed by system failures or security breaches.

Overall, self-driving cars present serious trade-offs. They offer real benefits for certain groups and the potential to reduce some types of accidents, but they create new risks, pose huge implementation costs, and demand ongoing human oversight. Rather than rushing to replace the entire fleet, it would be wiser to focus on improving the safety of human-driven cars and expanding education and programs that make existing transportation safer and more accessible.
\end{tcolorbox}

\begin{tcolorbox}[title={gpt-5-mini --- Grammar Edit}, gptbox]
Not having to pay attention while driving sounds appealing, but dying because your computerized, self-driven car made a mistake is not. Autonomous vehicles promise convenience, but their risks and costs outweigh the benefits.

Several states—Washington, D.C., Nevada, Florida, Michigan, and California—are testing cars with artificial intelligence, and those trials suggest the hassle may create more danger than the convenience is worth. Although self-driving cars can prevent accidents caused by human error, they do not eliminate crashes and introduce new safety hazards of their own. For example, Auto Insurance Center warns that “the very security behind self-driving cars would be a major obstacle, especially because the technology would be of very high interest to hackers.” If hackers can exploit a vehicle’s systems, they could cause collisions or disable safety features, producing outcomes more severe than typical human mistakes.

Even without malicious interference, autonomous systems remain unpredictable. Human drivers make split-second decisions based on complex cues—weather, sudden obstacles, and other drivers’ behaviors—that current AI still struggles to interpret in every situation. Adding this unpredictability to the existing rate of car accidents increases the overall hazard for everyone on the road. A disabled passenger, for example, might be less able to react or seek help if an autonomous vehicle malfunctions or is involved in a crash, compounding the danger for vulnerable riders.

Implementing self-driving technology would also be expensive and disruptive. Widespread adoption would require replacing current human-driven vehicles, scrapping old cars, and producing new autonomous models—an enormous economic and logistical undertaking. Users would need additional training to understand how to interact with these systems and what to do in emergencies. As Auto Insurance Center notes, “While the computer takes over once the vehicle is operational, the driver would still be required to maintain some knowledge about how to operate it safely.” That requirement undermines the convenience purported by proponents and creates further cost and complexity for drivers and employers alike.

Proponents argue that autonomous vehicles could greatly improve mobility for people with disabilities by providing more accessible transportation than many current public options. This is a persuasive point: self-driving cars have the potential to increase independence and access to services for many who cannot drive. However, the same safety and security problems apply and may affect disabled passengers even more severely. If an autonomous car is hacked or fails, a disabled occupant may not be able to take corrective action or summon help, which could turn a promising innovation into a new liability.

On balance, self-driving cars are likely to be more of a hindrance than a help in the near term. They present new and significant dangers to all road users, introduce costly implementation challenges, and require ongoing user involvement that diminishes their promised convenience. For now, society would be better served by improving human-driven vehicle safety and investing in programs that educate drivers, while continuing cautious, well-regulated testing of autonomous technologies.
\end{tcolorbox}

\begin{tcolorbox}[title={gpt-5-mini --- Completion Edit}, gptbox]
Not having to worry about paying attention while driving is cool, but dying because your computerized, self-driven car made a mistake is not.

Self-driving cars promise convenience, but their current risks outweigh their benefits. While autonomous technology can reduce accidents caused by human error, it does not eliminate accidents and introduces new, unique hazards. Testing in Washington, D.C., Nevada, Florida, Michigan, and California has shown that the technology is still imperfect; sensors fail in bad weather, software can misinterpret unusual road situations, and systems can behave unpredictably at the margins of their programming. These technical limitations mean that a machine’s failure can produce catastrophic outcomes just as a human mistake can—sometimes worse, because passengers may be less able to anticipate or correct the vehicle’s behavior.

A major safety concern is security. As Auto Insurance Center warns, "The very security behind self-driving cars would be a major obstacle, especially because the technology would be of very high interest to hackers." If attackers can gain control of a vehicle’s systems, they could cause crashes, disable brakes, or strand occupants in dangerous locations. Unlike a human driver who can take immediate evasive action, a hacked autonomous system might follow malicious commands or shut down at a critical moment. This potential for external exploitation turns every self-driving vehicle into a potential target, increasing systemic risk across the entire transportation network rather than leaving accidents as isolated human errors.

Beyond safety and security, the economic and logistical costs of widescale autonomous adoption are substantial. Converting an entire fleet of human-driven cars to self-driving models would require massive manufacturing changes, disposal or retrofitting of existing vehicles, and a workforce capable of building and maintaining sophisticated hardware and software. Consumers would face higher purchase prices and still need training: as Auto Insurance Center notes, "While the computer takes over once the vehicle is operational, the driver would still be required to maintain some knowledge about how to operate it safely." That means time and money spent teaching people when and how to intervene during malfunctions—an ongoing cost that could be especially burdensome for low-income drivers and communities.

Proponents rightly point out that autonomous vehicles could expand mobility for people with disabilities, offering independence to those who cannot drive. This is an important and sympathetic argument: better access to transportation can improve employment, healthcare access, and overall quality of life. However, the benefit does not negate the safety and security concerns. Disabled passengers could be more vulnerable in the event of an accident if they are unable to quickly exit a vehicle or assist emergency responders. Moreover, if self-driving systems are prone to hacks or software failures, relying on them as the primary transportation mode for vulnerable populations could concentrate risk precisely where it would cause the greatest harm.

Given these issues, society should be cautious about rushing to replace human-driven cars with autonomous ones. Rather than large-scale, immediate deployment, priority should be placed on improving vehicle safety features for human drivers, strengthening cybersecurity standards for automotive systems, and developing accessible public transportation options tailored to people with disabilities. Incremental approaches—such as limited, highly regulated autonomous services in controlled environments (e.g., dedicated shuttles with rigorous oversight)—allow us to gain benefits where they clearly outweigh risks while limiting exposure to the system-wide dangers of full adoption.

In sum, self-driving cars offer potential advantages, but current technological, security, and economic realities make them more hazardous and costly than many advocates admit. Careful, measured development combined with stronger protections for vulnerable populations will better serve public safety than wholesale replacement of human-driven vehicles.
\end{tcolorbox}

\begin{tcolorbox}[title={gpt-5-mini --- Expansion Edit}, gptbox]
Not having to pay attention while driving sounds appealing, but the promise of convenience from self-driving cars masks serious risks that outweigh their benefits. Autonomous vehicles introduce new and unique hazards—cybersecurity vulnerabilities, system failures in complex real-world conditions, and prohibitively high transition costs—that together make them a net danger rather than an improvement over well-regulated human-driven cars. While they could increase mobility for some disabled people, that advantage does not cancel out the significant safety, economic, and practical problems autonomous vehicles would create for society as a whole.

First, autonomous systems create fresh, consequential safety risks that human drivers do not face. Self-driving cars rely on networks of sensors, software, and machine learning models that can fail in ways humans do not. Sensors can be blinded by heavy rain, fog, or glare; cameras and lidar can misclassify uncommon obstacles; and software updates can introduce unforeseen bugs. These “edge cases” —rare situations not represented in training data—are precisely where machine-learning systems struggle, and it is these moments that can lead to catastrophic errors. Even where autonomous systems reduce certain types of human error (for example, distraction or impairment), they do not eliminate accidents and they introduce failure modes that can be harder to anticipate and mitigate. Because these failures can occur suddenly and across many vehicles at once, they pose a systemic risk that simple comparisons of human-versus-robot error rates can overlook.

Second, cybersecurity is a major and underappreciated threat to safety. As Auto Insurance Center warned, “The very security behind self-driving cars would be a major obstacle, especially because the technology would be of very high interest to hackers.” Connecting vehicles to networks and to each other increases the attack surface: a single vulnerability could allow an attacker to disable brakes, manipulate steering, or cause coordinated accidents across a fleet. Unlike a lone drunk driver, a malicious actor can act deliberately and on a large scale. The possibility of remote takeover creates not only immediate danger to occupants and other road users but also a new class of liability and policing challenges. Ensuring absolute, provable security in such complex, constantly updated systems is extremely difficult—more difficult than many proponents acknowledge—and until such guarantees are feasible, widespread deployment would expose millions to avoidable risk.

Third, the economic and logistical costs of replacing an entire vehicle fleet and ensuring safe operation are enormous. Moving from a predominantly human-driven vehicle environment to one dominated by autonomous vehicles would require not only manufacturing vast numbers of new vehicles but also retrofitting infrastructure, overhauling insurance systems, and training millions of drivers to interact safely with automated systems and to take over in emergencies. Even if gradual adoption reduces some upfront burdens, transitional periods create mixed traffic environments where human drivers, conventional cars, and autonomous vehicles share the road; those mixed conditions are especially dangerous because different actors behave according to different expectations and capabilities. Training drivers to “maintain some knowledge about how to operate [the vehicle] safely,” as Auto Insurance Center notes, is not a trivial task—many people already fail to master basic safe-driving practices, and adding complexities around when and how to intervene in an autonomous system raises the chance of human error at critical moments.

Proponents of autonomous vehicles point to improved mobility for people with disabilities, and that is a legitimate and important potential benefit. Self-driving cars could, in principle, provide door-to-door transportation for people who cannot drive and for whom current public transit is inaccessible or inconvenient. This is a compelling social good and should be pursued. However, the mobility advantage does not negate the safety and economic concerns discussed above. In fact, the very vulnerabilities that make autonomous vehicles risky for the general public could disproportionately endanger disabled passengers: if an autonomous vehicle fails or is targeted by malicious actors, a passenger with limited mobility may be unable to evacuate, protect themselves, or summon help quickly. Any program that seeks to use autonomy to expand accessibility must therefore include robust safeguards—redundant systems, rigorous cybersecurity standards, and emergency support mechanisms—to prevent turning a mobility solution into an additional hazard.

Given these problems, resources would be better spent on measures that improve safety and accessibility without introducing large new systemic risks. Strengthening vehicle safety standards for human-driven cars, investing in better driver education and enforcement, expanding and making public transit more accessible, and developing targeted technologies (for example, assisted driving features that enhance rather than replace human control) would address many of the harms autonomous vehicles aim to solve while avoiding the cascading new risks they create. Incremental automation—features that assist but do not fully replace human drivers—can capture safety benefits without placing full responsibility in the hands of fallible software.

In conclusion, while self-driving cars promise convenience and potential accessibility gains, they also bring unique and significant dangers: novel system failure modes, cybersecurity threats, and high transition costs. These factors make the widespread adoption of fully autonomous vehicles premature and potentially harmful. Policymakers, engineers, and the public should prioritize safer, more practical improvements to transportation that reduce risk and increase mobility without depending on technology that is not yet proven safe at scale.
\end{tcolorbox}

\clearpage

\subsubsection{Essay 3: Direct / Anecdotal}

\noindent\textit{This essay begins with a rhetorical question about distracted drivers and builds a broadly pro-AV case around safety statistics and accessibility.}

\medskip

\begin{tcolorbox}[title={Original Draft (D1)}, essaybox]
Have you ever seen a driver texting on their phones, putting on lipstick, eating a sandwich, or drinking a coffee?  If you have spent time in a moving vehicle and driving, you are bound to have encountered of one these distracted drivers.  Chances are, you may have been the distracted driver at least once in your life.

According to the Centers for Disease Control, there are upwards of 33,000 fatalities from traffic incidents that occur on an annual basis.  The main cause of these incidents are - you guessed it - distracted driving.  Distracted driving can influence poor driving behaviors of speeding, failing to follow road laws, and driving while tired, drunk, or under the influence of drugs.  However, there are new developments in technology that allow a driver to be distracted.  This technology is self-driving cars.

Self-driving cars present new possibilities for today's drivers.  As the cars are able to drive themselves, the driver would be able to fill wasted time in traffic or long commutes with productivity or leisurely activities.  Individuals with disabilities would have greater independence as those with physical disabilities or visual impairments would have greater mobility.  Additionally, elderly citizens would also be able to safely travel in their vehicles without worrying about any medical needs or memory loss.

In addition to productivity, independence, and mobility, self-driving cars would prove to be cost effective.  Maximizing on gasoline usage, self-driving cars would produce less air pollution while saving the consumer money.  There would also be savings from money not being spent on older mass transit projects such as trains.  Car insurance may eventually fade, as the computer in the vehicle would be the one to make the decisions and would prevent car incidents.

I understand, however, the technology in self-driving cars is still being developed.  The cost of implementing this new technology would be very expensive.  At up to \$100,000 a vehicle, it is unlikely that the majority of Americans could afford to make the switch to self-driving cars.  The only way for this trend to take off and be successful is if the majority people on the road are using this technology.

A computer in a self-driving car, with complicated algorithms and data that decreases the odds of a car accident, would not have the opportunity to become distracted.  As distracted driving is the leading factor of traffic incidents, traffic incidents could eventually cease because of self-driving cars.
\end{tcolorbox}

\begin{tcolorbox}[title={Human-Edited Baseline (D2)}, humanbox]
Have you ever seen a driver texting on their phones, putting on lipstick, eating a sandwich, or drinking a coffee?  If you have spent time driving, you are bound to have encountered of one these distracted drivers.  Chances are, you may have been the distracted driver at least once in your life.

According to the Centers for Disease Control, there are upwards of 33,000 fatalities from traffic incidents that occur on an annual basis.  The main cause of these incidents are - you guessed it - distracted driving.  Distracted driving can influence poor driving behaviors of speeding, failing to follow road laws, and driving while tired, drunk, or under the influence of drugs.  However, there are new developments in technology that allow a driver to be distracted.  This technology is self-driving cars.

Self-driving cars present new possibilities for today's drivers.  As the cars can drive themselves, the driver would not need to focus on the road.  Instead, the driver would be able to fill wasted time in traffic or long commutes with productivity or leisurely activities.  On long morning commutes, drivers could complete work or be social with friends or family, rather than sitting wasting time sitting in traffic.  Individuals with disabilities would have greater independence as those with physical disabilities or visual impairments would have greater mobility.  Instead of requiring transportation services to get to work or run errands, individuals with disabilities would be able to transport themselves with self-driving cars.  This would provide a new sense of freedom and choice for those with disabilities.  Additionally, elderly citizens would be able to safely travel in their vehicles without worrying about any medical needs or memory loss.

In addition to productivity, independence, and mobility, self-driving cars would prove to be cost effective for the consumer and society.  Maximizing on gasoline usage, self-driving cars would produce less air pollution while saving the consumer money.  There would also be savings from money not being spent on older mass transit projects such as trains.  Perhaps the consumer may refrain from using transportation services such as Uber or Lyft, as their personal car would be able to drive them to their desired destination.  Car insurance may eventually fade, as the computer in the vehicle would be the one to make the decisions and would prevent car incidents.  Therefore, car insurance would become irrelevant.

There may be opposition to self-driving cars, as the technology in self-driving cars is still being developed and tested.  Some drivers fear that the car may malfunction as they are driving which could lead to traffic incidents that the self-driving cars are trying to eliminate in the first place. Though this technology would be cost effective long-term, the cost of implementing this new technology would be very expensive to the consumer.  At up to \$100,000 a vehicle, it is unlikely that the majority of Americans could afford to make the switch to self-driving cars.  The only way for this trend to take off and be successful is if the majority of people on the road are using this technology.  Therefore, it might not be as successful as projected.

A computer in a self-driving car, with complicated algorithms and data that decreases the odds of a car accident, would not have the opportunity to become distracted.  As distracted driving is the leading factor of traffic incidents, traffic incidents could eventually cease because of self-driving cars.
\end{tcolorbox}

\begin{tcolorbox}[title={gpt-5-mini --- General Edit}, gptbox]
Have you ever seen a driver texting, putting on lipstick, eating a sandwich, or drinking coffee? If you have spent time in a moving vehicle, you have almost certainly encountered one of these distracted drivers — and you may have been one yourself. Distracted driving is a pervasive problem: according to the Centers for Disease Control and Prevention, tens of thousands of people are killed or seriously injured in traffic crashes each year, and driver inattention is a leading contributor to those incidents. Self-driving cars offer a promising technological response to this human problem by removing routine driving tasks from fallible humans and replacing them with automated systems designed to reduce error.

Self-driving cars could improve road safety in several concrete ways. Machines do not text, fall asleep at the wheel, or react emotionally to stress; an autonomous driving system continuously monitors its environment and can react faster than most human drivers when calibrated and maintained properly. For example, advanced sensors and algorithms can maintain optimal following distances, detect pedestrians in poor lighting, and apply emergency braking sooner than a distracted human might. By eliminating the most common sources of human error, these systems have the potential to lower crash rates, injuries, and fatalities — a direct response to the CDC data on crash-related deaths.

Beyond safety, autonomy promises social and economic benefits that follow directly from removing the driving burden. Commuters today waste hours per week attending to driving; if vehicles handle navigation and control reliably, riders can use that time productively — catching up on work, resting, or caring for family — which improves overall quality of life and economic efficiency. For people with disabilities, limited mobility, or visual impairments, self-driving cars could offer newfound independence: a person who cannot legally drive today could access jobs, medical care, and social opportunities without relying on others. Elderly drivers who give up manual control because of declining reflexes or memory could retain mobility without imposing safety risks, preserving dignity and reducing isolation.

Self-driving vehicles also have potential environmental and cost advantages when integrated into transportation systems. Automated driving can smooth traffic flow, reduce unnecessary acceleration and braking, and optimize routing, which together lower fuel consumption and emissions. Fleets of shared autonomous vehicles could reduce the need for private car ownership and for some costly mass-transit expansions in low-density areas, shifting capital from large infrastructure projects toward more flexible mobility services. Over time, fewer crashes would mean lower societal costs from medical care, lost productivity, and vehicle repair — savings that could be reflected in lower insurance premiums or different liability models.

Still, these benefits are not guaranteed and depend on several practical realities. The technology is expensive today: some prototype systems and high-level autonomous packages add tens of thousands of dollars to vehicle costs, which keeps them out of reach for many consumers. Widespread safety gains require high levels of market penetration and careful integration into mixed traffic: an autonomous vehicle performs best when surrounding traffic behaves predictably, and partial adoption can introduce new interaction challenges. There are also technical limitations — sensor failure modes, software bugs, and edge-case scenarios (extreme weather, unusual road layouts) that remain difficult to solve completely. Finally, regulatory frameworks, infrastructure adaptation, and standards for liability must evolve before the full benefits can be realized.

Opposition to self-driving cars arises from understandable concerns that the public, policymakers, and some experts hold. People worry about privacy and data collection, since autonomous vehicles gather detailed information about location and behavior. Workers in driving-related industries fear job losses if fleets become automated. Others distrust complex algorithms, preferring human judgment in life-and-death situations, especially after high-profile prototype failures receive intense media coverage. Acknowledging these concerns is important: addressing them requires transparent safety testing, clear rules for data governance, retraining programs for affected workers, and incremental deployment strategies that prioritize high-risk environments and public trust.

In short, self-driving cars represent a powerful tool for reducing the human causes of traffic incidents and for expanding mobility, productivity, and environmental efficiency. But realizing those benefits will require overcoming technical challenges, lowering costs, developing supportive regulation, and earning public trust by directly addressing privacy, employment, and safety concerns. If industry, government, and communities work together to meet these conditions, autonomous vehicles can move us toward safer, more accessible, and more efficient transportation systems.
\end{tcolorbox}

\begin{tcolorbox}[title={gpt-5-mini --- Minimal Edit}, gptbox]
Have you ever noticed a driver texting, applying makeup, eating, or sipping coffee behind the wheel? If you spend time in or around cars, you have likely seen one of these distracted drivers. Odds are you may have driven distracted at least once yourself.

The Centers for Disease Control report that roughly 33,000 people die each year in traffic incidents. A primary contributor to these fatalities is distracted driving, which can lead to speeding, failure to obey traffic laws, and impaired driving caused by fatigue, alcohol, or drugs. Now, however, a new technology promises to change the landscape of driving: self-driving cars.

Autonomous vehicles open up new possibilities. If cars can operate themselves, drivers could reclaim time wasted in traffic or on long commutes for work, study, or relaxation. People with disabilities—those with limited mobility or visual impairments—would gain greater independence and easier access to transportation. Elderly drivers who worry about medical conditions or memory decline could travel more safely and confidently.

Beyond personal convenience and accessibility, self-driving cars could yield economic and environmental benefits. More efficient driving patterns can reduce fuel consumption and emissions, saving money and helping the environment. Investments in expensive, large-scale transit projects might be reevaluated as personal autonomous vehicles become more common. Insurance models could also change: as vehicles rely on sensors and algorithms to avoid collisions, liability structures and premiums may shift substantially.

There are, of course, challenges. The technology is still evolving, and the initial cost of autonomous vehicles could be high—estimates sometimes reach tens of thousands of dollars per vehicle—putting them out of reach for many consumers. Widespread benefits will depend on broad adoption; the safety advantages of autonomy grow as more vehicles on the road are automated.

A well-designed autonomous driving system, driven by sophisticated algorithms and sensor data, would not be subject to human distractions. Because distracted driving is a leading cause of crashes, widespread deployment of self-driving cars could significantly reduce traffic incidents and save lives.
\end{tcolorbox}

\begin{tcolorbox}[title={gpt-5-mini --- Grammar Edit}, gptbox]
Have you ever seen a driver texting on their phone, putting on lipstick, eating a sandwich, or drinking coffee? If you have spent time in a moving vehicle, you are bound to have encountered one of these distracted drivers. Chances are you may even have been the distracted driver at least once in your life.

According to the Centers for Disease Control and Prevention, roughly 33,000 fatalities result from traffic incidents each year. The main cause of many of these incidents is distracted driving. Distraction can lead to speeding, ignoring traffic laws, and driving while drowsy or impaired. New developments in technology offer a potential solution: self-driving cars.

Self-driving cars present new possibilities for today’s drivers. Because the vehicle can operate without constant human control, drivers could use time spent in traffic or on long commutes for productive or leisure activities—working, studying, or resting—without increasing risk on the road. People with physical disabilities or visual impairments would gain independence and mobility, able to travel without relying on others. Elderly citizens who face medical challenges or memory loss could maintain safe, reliable transportation without the stress of manual driving.

Beyond convenience and accessibility, self-driving cars could be cost-effective. Autonomous vehicles can optimize acceleration, braking, and routing to use fuel or electricity more efficiently, reducing per-mile energy costs and air pollution. Fewer accidents would mean lower societal costs from emergency services, medical care, and lost productivity. If reliability improves, demand for some large-scale mass transit expansions could shift, allowing funds to be reallocated to other public needs. Additionally, as liability shifts toward manufacturers and software providers, personal auto insurance models and premiums would likely change, potentially reducing costs for individual drivers.

I understand, however, that self-driving technology is still developing and that many reasonable concerns exist. High initial costs—estimates for some fully autonomous models approach \$100,000—make early adoption difficult for most consumers. People are also worried about software failures, hacking, liability in mixed-traffic conditions, and the potential loss of driving-related jobs. Some distrust the idea of ceding control to a computer and fear that complex, rare situations might outstrip current algorithms. These are valid reasons for caution and explain why many people resist a rapid transition.

Despite these concerns, a well-designed autonomous system—built on extensive data, testing, and redundant safety mechanisms—would not be subject to human distractions like texting or drowsiness. Because distracted driving is a leading cause of traffic incidents, widespread adoption of reliable self-driving technology could substantially reduce the number and severity of crashes. Realizing that potential will require continued technical development, rigorous regulation and testing, affordable deployment strategies, and public education to build trust.

In summary, self-driving cars offer significant benefits: increased productivity and quality of life for drivers, greater independence for people with disabilities and the elderly, and potential cost and safety advantages. Legitimate hurdles remain, but with careful implementation and safeguards, autonomous vehicles could dramatically reduce traffic fatalities and transform transportation for the better.
\end{tcolorbox}

\begin{tcolorbox}[title={gpt-5-mini --- Completion Edit}, gptbox]
Have you ever seen a driver texting on their phone, putting on lipstick, eating a sandwich, or drinking a coffee? If you have spent time in a moving vehicle, you are bound to have encountered one of these distracted drivers. Chances are, you may have been the distracted driver at least once in your life.

According to the Centers for Disease Control, there are upwards of 33,000 fatalities from traffic incidents each year. The main cause of many of these incidents is—unsurprisingly—distracted driving. Distraction contributes to speeding, failure to obey road laws, and driving while tired, intoxicated, or under the influence of drugs. New developments in technology promise a dramatic reduction in these behaviors: self-driving cars. I argue that, despite current limitations and costs, widespread adoption of autonomous vehicles would improve public safety, increase mobility and independence for vulnerable populations, reduce environmental and economic costs, and ultimately save lives—provided the transition is carefully managed.

First and foremost, self-driving cars have the potential to greatly reduce crashes caused by human error. Human drivers are inconsistent: they react slowly when fatigued, make poor choices under stress, and are easily distracted by phones or other tasks. Autonomous systems, by contrast, can monitor the environment continuously, communicate with infrastructure and other vehicles, and execute split-second maneuvers without impairment. Real-world testing and simulation data from industry and research labs already suggest that well-designed autonomous systems can reduce certain types of collisions—rear-end crashes and collisions caused by inattention in particular. If machines take over the routine perceptual and decision-making tasks of driving, the number of fatal and debilitating crashes should fall substantially.

Beyond safety, self-driving cars expand mobility for people who currently have limited access to driving. Individuals with physical disabilities or visual impairments could gain independence from drivers or specialized transit; elderly people who can no longer meet the cognitive or physical demands of driving could retain mobility and social connection without risking their safety. These are not abstract benefits: improved access to jobs, healthcare, and social activities has direct quality-of-life and economic implications. Autonomous shuttles and on-demand robotaxi services could serve neighborhoods poorly connected by traditional public transit, complementing rather than replacing existing systems.

Self-driving technology can also yield environmental and economic benefits. Autonomous vehicles can be optimized to drive more efficiently—smoother acceleration, optimal routing, and platooning reduce fuel consumption and emissions per mile. When integrated into shared mobility models, they can decrease the number of vehicles needed overall, lowering manufacturing and maintenance costs. Over time, fewer crashes mean lower societal costs: less spending on emergency response, medical care, and property damage. While insurance models will evolve rather than disappear, the financial burden from human-caused accidents would decline significantly.

That said, the technology is not yet mature, and there are legitimate concerns that explain why many people remain skeptical. One major objection is cost: highly capable autonomous systems today can add tens of thousands of dollars to a vehicle’s price tag, putting them out of reach for many consumers. Another concern is safety in edge cases—rare or novel situations where machine perception or decision-making may fail. There are also worries about cybersecurity, privacy, and the potential for job loss in driving-related industries. Ethically fraught scenarios about how a vehicle should act in an unavoidable crash generate public discomfort as well.

These objections are important, but they do not negate the overall potential of the technology. First, costs tend to fall with scale and technological improvement; computing hardware, sensors, and software development become cheaper as adoption increases. Policies can accelerate equitable access: public investment in shared autonomous fleet pilots, subsidies for retrofitting or purchasing autonomous vehicles for mobility-impaired citizens, and incentives for electric autonomous fleets would spread benefits more fairly. Addressing edge-case failures requires rigorous testing, transparent reporting of performance, and strong regulatory standards for safety before scaling. Cybersecurity and privacy protections must be baked into design and enforced by law. Regarding jobs, history shows that technological change shifts employment patterns—well-designed transition programs, retraining, and social safety nets can help workers move into new roles in maintenance, fleet operations, or related sectors.

A careful, phased approach makes adoption safer and fairer. Start with limited use cases—geofenced shuttles, controlled campuses, and urban corridors—where conditions are predictable and oversight is feasible. Use these deployments to refine systems, build public trust, and craft regulatory frameworks. Require vehicle manufacturers and fleet operators to disclose safety metrics, incident reports, and software-update practices. Mandate strong cybersecurity standards and privacy protections. Invest in infrastructure improvements—connected traffic signals and better road markings—that enhance autonomous vehicle performance and benefit all road users.

In conclusion, self-driving cars are not a panacea and should not be adopted recklessly. But the evidence indicates that, if implemented responsibly with attention to equity, safety standards, and regulatory oversight, autonomous vehicles can meaningfully reduce the human toll of driving, expand independence for vulnerable populations, cut environmental and economic costs, and improve overall mobility. Policymakers should encourage continued development while insisting on transparency, rigorous testing, and policies that ensure benefits are widely shared. With careful stewardship, self-driving technology can help transform our roads from places of daily risk into safer, more accessible public spaces.
\end{tcolorbox}

\begin{tcolorbox}[title={gpt-5-mini --- Expansion Edit}, gptbox]
Have you ever seen a driver texting, applying makeup, eating, or sipping coffee while steering a vehicle? If you have spent any time on the road, you have probably encountered one of these distracted drivers—perhaps you have even been distracted at the wheel yourself. Distracted driving is a persistent and deadly problem: tens of thousands of people are killed or injured each year in traffic incidents that are frequently attributed to inattention behind the wheel. As automakers and technology companies accelerate development of autonomous vehicles, self-driving cars offer a compelling solution to reduce human error, expand mobility, and transform how we use travel time. Yet the promise of autonomy also raises practical, ethical, and economic questions that must be addressed before these vehicles can realize their full potential.

Distracted driving contributes directly to a large share of road deaths and injuries. When drivers take their eyes, hands, or attention off driving—whether to text, eat, adjust a stereo, or converse—they reduce their ability to respond to changing road conditions, sudden braking, or unpredictable behavior by other road users. This diminished situational awareness increases the likelihood of speeding, running red lights, unsafe lane changes, and failure to recognize hazards in time. Self-driving cars, by removing the need for human monitoring during routine operation, can substantially reduce these kinds of errors. Autonomous systems use cameras, radar, LIDAR, and integrated sensors to maintain continuous awareness of the surrounding environment, and their decision-making algorithms can react without the delays and cognitive failures that afflict human drivers. In principle, replacing human attention with reliable, continuously attentive systems can lower crash rates caused by distraction, fatigue, or impairment.

Beyond safety, self-driving technology promises to change the way people use travel time and to expand mobility for underserved populations. For commuters stuck in traffic, autonomy could transform otherwise wasted hours into time for work, family communication, or rest. For people with disabilities—those with physical limitations, visual impairments, or cognitive challenges—autonomous vehicles could restore independence, enabling errands, appointments, and social visits without reliance on others. Elderly individuals who can no longer safely drive because of declining eyesight, reaction time, or memory could retain mobility and autonomy, delaying or eliminating the need for others to provide transport. These changes would not only improve quality of life for individuals, but could also reduce burdens on family caregivers and public transportation systems by offering door-to-door flexibility that conventional transit often cannot match.

Cost-effectiveness is another potential benefit, though it is more complex than it first appears. On one hand, self-driving cars can be programmed to optimize routes, accelerate and brake smoothly, and coordinate with traffic systems to reduce fuel consumption and emissions per mile. Over time, more efficient driving patterns and vehicle platooning could lower aggregate energy use and air pollution in congested corridors. On the economic side, autonomy could reduce societal costs associated with crashes—medical care, lost productivity, emergency response, and property damage—if crash rates fall substantially. Moreover, a future with widely deployed autonomous ride-hailing fleets could shift some demand away from privately owned cars and reduce the need for expensive, large-scale mass transit projects in specific corridors, although it may not eliminate the need for public transit overall.

However, realizing these benefits requires careful attention to current limitations, costs, and systemic effects. Presently, the hardware and software required for high-level autonomy remain expensive. Retrofit and early-production models can cost tens of thousands of dollars more than comparable human-driven vehicles, putting them out of reach for many consumers. Economies of scale and technological maturation will likely lower prices over time, but widespread safety and societal gains depend on broad adoption: mixed fleets in which only a small percentage of vehicles are autonomous might not achieve the same safety improvements due to interactions between human drivers and autonomous systems. Additionally, autonomous systems must be robust across diverse weather, roadway, and lighting conditions; they must handle rare but critical edge cases—construction zones, erratic pedestrians, ambiguous right-of-way situations—without human intervention. Solving these technical challenges is essential before autonomy can fully substitute for human drivers.

It is also important to acknowledge and explore why many people remain skeptical or opposed to self-driving cars. Some concerns are practical: people worry about software reliability, cybersecurity vulnerabilities, and the legal accountability when a vehicle makes a harmful decision. Others are economic: workers in driving occupations—truck drivers, taxi and rideshare drivers, and public transit employees—fear job displacement. There are ethical objections as well, such as how an autonomous system should weigh competing harms in unavoidable crash scenarios and who should be liable for those choices. Cultural and psychological factors play a role too: many drivers enjoy control and resist ceding decision-making to machines, and some communities distrust new technologies due to historical disparities in how innovation has been deployed. These reasons are understandable and must be addressed through transparent regulation, public engagement, retraining programs for affected workers, and rigorous safety testing and accountability frameworks.

A further concern is equity: without deliberate policy interventions, autonomous vehicles could exacerbate existing inequalities. If early adoption is limited to wealthy individuals or neighborhoods, the safety and mobility benefits will be unevenly distributed. Policymakers and industry leaders must therefore consider subsidized pilot programs, public–private partnerships, and incentives that ensure equitable access to autonomy’s benefits. Insurance models will also need to evolve: while autonomous systems may reduce some individual driver liability, liability could shift towards manufacturers, software providers, or fleet operators, requiring new legal and insurance frameworks.

Despite these challenges, the central premise—that removing human distraction from routine driving can reduce many traffic incidents—remains powerful. Autonomous systems are designed to maintain continuous attention, to fuse multiple sensor inputs, and to follow optimized safety rules without the lapses that human cognition is susceptible to. If developers, regulators, and communities collaborate to refine technology, create transparent standards, and address social and economic disruptions proactively, self-driving cars can deliver the safety, mobility, and environmental gains they promise.

In conclusion, self-driving cars have the potential to dramatically decrease distracted-driving incidents, expand independence for people who cannot drive, and improve efficiency and environmental outcomes. Yet this future is not automatic: it depends on overcoming technical hurdles, lowering costs, ensuring equitable access, and responding thoughtfully to legitimate concerns about jobs, liability, and ethics. Addressing these challenges openly and deliberately will make it possible to move from the promise of autonomy to a safer, more inclusive transportation system for everyone.
\end{tcolorbox}

\clearpage

\subsection{Three Models, Expansion Mode}
\label{sec:app-models}

The three essays below were selected to represent a further range of student voices.
For each essay the human-edited baseline is followed by the expansion rewrites from all
three models, allowing direct comparison of how each system handles the same source text
under the most generative editing condition.

\subsubsection{Essay 1: Analytical}

\noindent\textit{This essay includes benefits of self-driving cars, including safety, productivity, law enforcement, fuel efficiency.}

\medskip

\begin{tcolorbox}[title={Original Human Draft (D1)}, essaybox]
With the increasing number of cars on the road and people becoming more careless about things, the number of fatalities due to accidents are also increasing. 
Even those who are the expert in driving, get into a false confidence and take it casually while driving. Moreover increasing amount of stress and getting engaged in alcohol or drugs among the young generation is common nowadays. All these results in distractions while driving and thus increasing the number of road accidents.

Self-driving cars can bring revolution in this direction by ensuring safe and fully focussed driving every time. A computer works same behind the wheel every time it drives. It is not affected by the external distractions and neither it is overconfident any time. Every time, it drives, it will take the same precautions and drive with as much safety as the first time. Moreover, it learns from the others mistake, keeps every data to take further decisions.

While commuting to work, most of the people have half of their mind over what deadlines they have to complete or what their boss may give him the break or not. In such cases, breaking traffic rules or getting distracted in between is common. With computer driving the car, this will no more be a problem and people can focus more on their important work even while traveling.
Even disabled people, who rely on others or public transport can commute from one place to other more easily.

Half of the time of the police officer is wasted on tracking and keeping traffic rules, giving tickets to people and managing traffics. With self-driving cars on road, traffic rules will be followed more strictly and even keeping a track of every vehicle will then be easier as everything will be connected by computer. The police can then divert their focus to more serious crimes.

The fuel consumption and hence the energy wastage can also be restricted by using self-driving cars. While people behave casually for turning off the engine while at lights or at stops, this can be easily controlled by computers.

In all, we can conclude that using self-driving cars can be worthy in terms of saving lives, saving energy, saving time and will let us divert our mind to more important work rather than driving. It can solve the problem of road accidents, traffics and will provide a safe commute to pedestrians as well as to the passengers.
\end{tcolorbox}

\begin{tcolorbox}[title={Human-Edited Baseline (D2)}, humanbox]
With the increasing number of cars on the road and the people becoming more and more careless about things, the number of fatalities due to accidents are also increasing.

Those who are the expert in driving sometimes get into false confidence and take it casually while driving. Moreover increasing amount of stress and getting engaged in alcohol, drugs among the young generation is common nowadays. All these results in distractions while driving and thus increases the number of road accidents.

Self-driving cars can bring revolution in this direction by ensuring safe and fully focussed driving each and every time. A computer works in the same way behind the steering, every time it drives. It is not affected by the external distractions nor it is overconfident at any point. Every time it drives, it will take the same precautions and drive with as much safety as the first time. Additionally, it learns from the mistake of others and keeps improving with every data to take further decisions.

While commuting to work, most of the people have their mind occupied over the deadlines they have to complete or their performance in corporate and social life. In such cases, breaking traffic rules or getting distracted while driving is quite common. With computers driving cars, this will no more be a problem and people can divert their focus on more important work even while traveling.

Even disabled people, who rely on others or public transport can commute from one place to other more easily.

A major portion of the time of the police officer is wasted on tracking and keeping traffic rules, giving tickets to people and managing traffics. With self-driving cars on road, traffic rules will be followed more strictly and even keeping track of vehicle will be easier with everything connected by computer. The police can then divert their focus to more serious crimes.

The fuel consumption and hence the energy wastage can also be restricted by using self-driving cars. While people behave casually for turning off the engine while at lights or at stops, this can be easily controlled by computers.

In all, we can conclude that using self-driving cars can be worthy in terms of saving lives, saving energy, saving time and will let us focus our mind to more important work rather than driving. It can solve the problem of road accidents, traffics and will provide a safe commute to pedestrians as well as to the passengers.
\end{tcolorbox}

\begin{tcolorbox}[title={Claude Haiku~4.5 --- Expansion Edit}, claudebox]
With the increasing number of vehicles on roads worldwide, traffic fatalities have reached alarming levels. This crisis stems from multiple interconnected factors that compromise driver safety and attention. Even experienced drivers, who possess extensive knowledge of traffic rules and driving techniques, often fall victim to overconfidence and complacency behind the wheel. They may underestimate risks or adopt casual attitudes toward safety protocols they have mastered over years of practice.

The problem is further exacerbated by contemporary societal challenges. Rising stress levels in professional and personal lives, combined with the growing prevalence of substance abuse—including alcohol and drug use—among younger generations, create a dangerous environment on roads. These factors significantly impair cognitive function, reaction time, and decision-making abilities. Additionally, the proliferation of smartphones and in-vehicle distractions has created a perfect storm of conditions that lead to preventable accidents, injuries, and loss of life.

Autonomous vehicles present a revolutionary approach to addressing these multifaceted road safety challenges. Unlike human drivers, computer-controlled vehicles operate with consistent, unwavering precision every single time they are activated. A computer system does not experience fatigue, overconfidence, or emotional fluctuations that impair judgment. It is not susceptible to external distractions such as phone notifications, passenger conversations, or environmental stimuli that typically divert human attention.

Moreover, autonomous vehicles demonstrate remarkable learning capabilities. These systems continuously collect and analyze data from every journey, learning from mistakes made by other vehicles on the road. This collective intelligence allows self-driving cars to progressively improve their decision-making algorithms, becoming safer with each passing day. Every autonomous vehicle benefits from the accumulated experience of the entire network, creating an exponentially safer driving ecosystem.

For working professionals, the advent of self-driving cars offers unprecedented advantages. During their daily commutes, most workers struggle to maintain focus on the road while their minds preoccupy themselves with pressing work deadlines, professional anxieties, or concerns about their career advancement. This divided attention inevitably leads to traffic violations, dangerous lane changes, and momentary lapses in concentration that can have catastrophic consequences.

With autonomous vehicles handling the driving responsibilities, commuters can productively utilize their travel time for work-related tasks, personal development, or simply mental relaxation. This transformation converts what is currently a stressful and potentially dangerous activity into valuable, reclaimed time.

Furthermore, self-driving cars would dramatically improve mobility for disabled and elderly populations. Currently, individuals with visual impairments, physical disabilities, or age-related limitations depend heavily on family members, caregivers, or expensive public transportation services. Autonomous vehicles would grant them unprecedented independence and freedom of movement, enabling them to travel safely and conveniently without relying on others.

The implementation of widespread autonomous vehicle adoption would fundamentally reshape law enforcement priorities. Currently, police officers dedicate approximately half their working hours to traffic management, monitoring vehicle compliance, issuing citations, and managing congestion. These resources, while necessary, divert attention from serious criminal investigations and community safety initiatives.

With self-driving cars strictly adhering to all traffic regulations by design, the need for traffic enforcement would diminish substantially. Additionally, since all autonomous vehicles would be interconnected through centralized computer systems, tracking and monitoring traffic flow would become seamless and automated. This technological infrastructure would free law enforcement personnel to concentrate on violent crimes, criminal investigations, and genuine public safety threats that require human judgment and intervention.

Beyond safety improvements, autonomous vehicles offer significant environmental advantages. Self-driving cars can optimize fuel consumption through intelligent acceleration, deceleration, and engine management. While human drivers frequently leave engines idling at traffic lights or during traffic stops—wasting fuel and increasing emissions—autonomous systems automatically shut down engines during unnecessary operation and coordinate movement with surrounding vehicles to minimize fuel consumption.

Additionally, autonomous vehicles can be programmed to select the most fuel-efficient routes and driving patterns, reducing overall energy waste across transportation networks and contributing to broader environmental sustainability goals.

The adoption of self-driving cars represents a comprehensive solution to multiple interconnected problems facing modern transportation systems. By eliminating human error, distraction, and impaired judgment, autonomous vehicles have the potential to dramatically reduce road fatalities and injuries. Simultaneously, they would enhance quality of life for commuters, provide mobility solutions for disabled populations, optimize law enforcement resource allocation, and reduce environmental impact through improved fuel efficiency.

While challenges remain in implementation and regulation, the transformative benefits of autonomous vehicles—in terms of saving lives, conserving energy, reclaiming time, and improving overall public safety—make this technology a worthy investment in humanity's future. Self-driving cars promise not merely incremental improvements, but a fundamental reimagining of safe, efficient, and equitable transportation for all members of society.
\end{tcolorbox}

\begin{tcolorbox}[title={gpt-5-mini --- Expansion Edit}, gptbox]
With increasing numbers of vehicles on the road and growing carelessness among some drivers, traffic fatalities and serious accidents continue to rise. Inattention, overconfidence, substance use, and stress all contribute to dangerous driving behavior. Self-driving cars offer a promising solution: by replacing human inconsistency with consistent, data-driven control, autonomous vehicles can reduce crashes, save time and energy, improve mobility for people with disabilities, and allow law enforcement to focus on higher-priority tasks.

Human drivers are fallible in predictable ways. Even experienced drivers sometimes become overconfident, take unnecessary risks, or succumb to distraction when they are tired or preoccupied with work, family, or digital devices. Young drivers, in particular, are more likely to engage in risky behavior such as driving under the influence of alcohol or drugs. Self-driving systems do not suffer from fatigue, distraction, or emotional fluctuations. They follow programmed safety rules consistently, maintain appropriate following distances, and monitor the environment without lapses in attention. Because autonomous vehicles can process inputs from multiple sensors—cameras, radar, lidar—and aggregate real-time traffic data, they can react faster and more reliably than a distracted human, reducing the likelihood of collisions.

Autonomous vehicles also have the potential to improve access and independence for people with disabilities and those who cannot drive. For many disabled individuals, lack of reliable personal transportation limits employment opportunities, social participation, and access to healthcare. Self-driving cars could provide door-to-door mobility on demand without requiring a human driver, helping to integrate disabled passengers into daily life more fully and reducing dependence on paratransit services and caregivers.

Widespread adoption of autonomous vehicles could also make traffic enforcement and management more efficient. If vehicles are designed to obey traffic laws and exchange information with infrastructure and other vehicles, traffic flow can be optimized to reduce congestion, idling, and the inefficiencies that cause delays. Police resources now devoted to directing traffic or issuing citations could be reallocated to investigating serious crimes or community policing. Moreover, connected vehicle networks can provide anonymized data to planners to identify and address systemic safety problems.

Fuel consumption and emissions can be reduced through more consistent, optimized driving behaviors. Human drivers often accelerate and brake inefficiently, and they may leave engines idling at stops. Autonomous systems can employ smoother acceleration profiles, maintain steady speeds, and coordinate with traffic signals and other vehicles to minimize stop-and-go driving. These efficiencies translate into lower fuel use or longer electric vehicle range, reduced greenhouse gas emissions, and decreased urban pollution.

A credible critique of self-driving technology concerns safety, cybersecurity, liability, and employment impacts. High-profile accidents involving autonomous prototypes have shown that the technology is not infallible; sensor limitations, unpredictable human behavior, and edge-case scenarios can still produce failures. Additionally, connected vehicles present cybersecurity risks: malicious actors could attempt to interfere with navigation or control systems. There are also legitimate concerns about job displacement for professional drivers and the legal framework that determines liability after a crash.

These concerns, while serious, can be addressed through regulation, design, and phased deployment. Rigorous safety standards, transparent testing protocols, and third-party audits can raise reliability before large-scale rollouts. Cybersecurity can be strengthened by building multi-layered defenses, encryption, and fail-safe modes that transition control to a safe state if tampering is detected. Policymakers can develop liability rules that allocate responsibility fairly among manufacturers, software developers, and fleet operators while supporting workers in transition through retraining programs and social safety nets. Finally, a gradual implementation—starting in controlled environments such as dedicated lanes, campuses, or geofenced urban corridors—can allow systems to mature and public trust to grow.

In conclusion, while self-driving cars are not a panacea and require careful oversight, their potential benefits are substantial. By reducing human error, improving accessibility, optimizing traffic flow, and lowering energy consumption, autonomous vehicles can make transportation safer, more efficient, and more inclusive. With thoughtful regulation, robust engineering, and social policies to manage the transition, self-driving cars can become a valuable component of safer and more sustainable mobility systems.
\end{tcolorbox}

\begin{tcolorbox}[title={Gemini~2.5 Flash --- Expansion Edit}, geminibox]
With the increasing number of vehicles on the road, coupled with a growing human tendency towards carelessness and distraction, the global incidence of traffic fatalities and accidents continues to climb at an alarming rate. Even seasoned drivers can develop a false sense of confidence, leading them to drive carelessly. Moreover, the increasing levels of stress and engagement with alcohol or drugs among younger generations are increasingly common, further exacerbating the problem. All these factors contribute significantly to driver distractions, directly resulting in a rise in road accidents.

Self-driving cars offer a revolutionary solution to this pervasive issue by ensuring consistently safe and fully focused driving. Unlike human drivers, a computer operates with unwavering consistency and precision behind the wheel every single time. It is immune to external distractions such as mobile phones or passenger conversations, and it never experiences fatigue, stress, or overconfidence. Each journey is executed with the same rigorous application of safety protocols and precautions as the very first. Furthermore, autonomous systems continuously learn from accumulated data and collective driving experiences, allowing them to proactively identify potential hazards and refine their decision-making processes in real-time, far surpassing the individual learning capacity of human drivers. This data-driven approach promises a dramatic reduction in human error, the leading cause of accidents.

Beyond safety, self-driving technology promises to reclaim valuable human attention and time. While commuting to work, many individuals are mentally preoccupied with work deadlines, upcoming meetings, or personal anxieties. In such states, mental distractions or even unintentional traffic violations become common. With autonomous vehicles taking over the driving task, these cognitive burdens are lifted, allowing occupants to fully engage in productive work, relax, or simply enjoy their journey without the stress of navigating traffic. This shift transforms commute time from a lost period into an opportunity for personal or professional growth. Moreover, self-driving cars offer unprecedented independence and mobility for disabled individuals, the elderly, or those who cannot drive, granting them easier access to employment, social activities, and essential services that were previously limited by reliance on others or public transport.

The efficiency gains extend to public services as well. A significant portion of law enforcement officers' time is currently spent monitoring traffic rules, issuing citations for minor infractions, and actively managing congestion. With self-driving cars on the road, traffic laws will be adhered to rigorously, as vehicles are programmed to follow regulations precisely. Furthermore, vehicle tracking and traffic flow management will be streamlined through interconnected digital systems, allowing for predictive traffic management and dynamic routing. This shift would free up valuable police resources, enabling them to divert their focus to more serious crimes and community safety initiatives, thereby enhancing overall public security.

Environmental and economic benefits are also substantial. Fuel consumption and, consequently, energy wastage can be significantly reduced through the widespread adoption of self-driving cars. While human drivers often exhibit casual habits regarding engine idling at traffic lights or during brief stops, autonomous systems can precisely manage engine operation, implementing intelligent stop-start functions and optimizing acceleration and braking for peak fuel efficiency. Furthermore, these systems can identify the most efficient routes, minimize unnecessary detours, and even facilitate vehicle platooning, where cars travel in close succession to reduce aerodynamic drag, leading to substantial savings in fuel and a corresponding decrease in carbon emissions.

Of course, the transition to autonomous vehicles is not without its challenges. Concerns regarding cybersecurity risks, the potential for job displacement in the transportation sector, and complex ethical dilemmas in unavoidable accident scenarios are frequently raised. However, these challenges, while significant, are being actively addressed through rigorous research and development. Robust cybersecurity protocols are paramount in the development of self-driving systems, with continuous advancements aimed at safeguarding against external threats, much like the security measures protecting other critical digital infrastructures. While job roles may shift, the emergence of new opportunities in vehicle maintenance, software development, data analysis, and fleet management is anticipated. Regarding ethical dilemmas, autonomous systems are designed with safety as their paramount objective, programmed to minimize harm in all circumstances, often making more rational and consistent decisions than humans under extreme pressure. The benefits of drastically reduced accidents, increased efficiency, and enhanced accessibility far outweigh these resolvable challenges.

In conclusion, the widespread adoption of self-driving cars promises profound and multifaceted benefits, encompassing the preservation of human lives, substantial energy savings, increased productivity, and the liberation of human attention for more meaningful pursuits. They offer a compelling solution to pervasive issues like road accidents and traffic congestion, while simultaneously ensuring a safer, more reliable, and more accessible commute for both passengers and pedestrians alike. The integration of self-driving technology marks a pivotal step towards a smarter, safer, and more efficient future for transportation.
\end{tcolorbox}

\clearpage

\subsubsection{Essay 2: Questioning Feasibility}

\noindent\textit{acknowledges the technology's potential but concludes that adoption is premature given cost, security, and economic disruption}

\medskip

\begin{tcolorbox}[title={Original Draft (D1)}, essaybox]
The self-driving car is a groundbreaking piece of technology. There is no doubt about that. However, I really question how feasible it is to transition them into every day life in America. The potential advantages of self-driving cars are incredible: safer driving via decreased accidents, the inability for a computer to get distracted, identifying traffic problems and no longer needing a designated driver, saving money, and saving time. On the other hand, the disadvantages of self-driving cars are quite frightening as they are weather dependent, they can't detect human traffic signals, they could malfunction and put the driver at even more risk, they rely on other pieces of technology we haven't yet mastered and jobs in fields connected to transportation would suffer. In other words, there is no solid evidence that self-driving cars are qualitatively safer.

In my opinion, I don't think it is the time to bring them out and try to adopt them into society. We need time to develop our technology in general to a point where if we were to start using them, they'd have a smooth transition into the market and outside parties would suffer as little as possible. I think they have the potential to be incredibly beneficial to society but not right now.

If we were to start using them in today's society, it wouldn't work out in many ways. Only a small percentage of the population would be able to afford them as they cost more than \$100,000 and require training to operate in a safe manner. Furthermore, self-driving cars rely on being adopted by everybody and this will certainly not happen if only the richest people are able to buy them. Security risks are also a relevant issue as hackers would be very interested in the technology and specialized computers could be able to collect personal data. Even if self-driving cars were cheap enough to be used by the general public, many jobs would be eliminated. Workers in the transportation sector would no longer be needed, the gasoline industry would suffer if self-driving cars are electric, driver's ed courses wouldn't be needed anymore and personal injury lawyers would have reduced earnings due to less accidents on the road. This would be detrimental for the lives of many people and also harm the economy. Because of these reasons, 1) I don' think it is feasible to get the general public to use self-driving cars and 2) in the unlikely case that everyone uses them, it would cause damage to the economy.
\end{tcolorbox}

\begin{tcolorbox}[title={Human-Edited Baseline (D2)}, humanbox]
The self-driving car is a groundbreaking piece of technology. There is no doubt about that. However, I really question how feasible it is to transition them into every day life in America. The potential advantages of self-driving cars are incredible. Safer driving would be accomplished by decreasing accidents due to the inability for a computer to get distracted and no longer needing a designated driver. The identification of traffic problems, faster speed limits and the ability to be dropped off would save people time. It would also save many individuals money due to lower insurance and healthcare costs.

On the other hand, the disadvantages of self-driving cars are quite frightening:  they are weather dependent, they can't detect human traffic signals, they could malfunction and put the driver at even more risk, they rely on other pieces of technology we haven't yet mastered  and they would cause jobs in fields connected to transportation would suffer. In other words, there is no solid evidence that self-driving cars would be a true benefit to society.

In my opinion, I don't think it is the time to bring them out and try to adopt them into society. We need time to develop our technology to a point where if we were to start using them, they'd have a smooth transition into the market and outside parties would suffer as little as possible. I think they have the potential to be incredibly beneficial to society but not right now.

If we were to start using them in today's society, it wouldn't work out in many ways. Only a small percentage of the population would be able to afford them as they would initially cost more than \$100,000 and require time and resource-heavy training to operate in a safe manner. Furthermore, self-driving cars rely on being adopted by everybody and this will certainly not happen if only the richest people are able to buy them. This is because their safety is created only when the road consists of an entire network of driverless cars communicating with one another.  Security risks are also a relevant issue as hackers would be very interested in the technology and specialized computers could be able to collect personal data.

Even if self-driving cars were cheap enough to be used by the general public, many jobs would be eliminated. Workers in the transportation sector would no longer be needed, the gasoline industry would suffer if self-driving cars are electric, driver's ed courses wouldn't be needed anymore and personal injury lawyers would have reduced earnings due to less accidents on the road. This would be detrimental for the lives of many people and also harm the economy. Because of these reasons, 1) I don' think it is feasible to get the general public to use self-driving cars and 2) in the unlikely case that everyone uses them, it would cause damage to the economy.

Like many technological advances throughout history, self-driving cars have a thick silver lining.  In many cases, the technology is beneficial to outweigh all of its downsides, but in the case of driverless cars, we may not want to commit to it without further investigation.
\end{tcolorbox}

\begin{tcolorbox}[title={Claude Haiku~4.5 --- Expansion Edit}, claudebox]
The self-driving car represents one of the most groundbreaking technological innovations of our time. There is no doubt about that. However, I genuinely question how feasible it truly is to transition autonomous vehicles into everyday life across America. While the potential advantages are undeniably compelling—including enhanced safety through dramatically decreased accident rates, the elimination of human distraction, improved traffic management, reduced need for designated drivers, significant cost savings, and substantial time savings—the disadvantages are equally concerning and deserve serious consideration.

The challenges facing self-driving car adoption are multifaceted and profound. These vehicles remain heavily weather-dependent, struggling to function optimally during heavy rain, snow, or fog when their sensors cannot accurately read road conditions. Additionally, current autonomous systems cannot reliably interpret human traffic signals, such as a police officer directing traffic at an intersection, which creates dangerous gaps in their decision-making capabilities. Beyond these operational limitations, the technology is vulnerable to malfunctions that could put drivers at greater risk than traditional vehicles would. Furthermore, self-driving cars depend on complementary technologies—such as sophisticated GPS systems, 5G networks, and cloud computing infrastructure—that we have not yet fully mastered. Perhaps most troublingly, there is no solid empirical evidence demonstrating that self-driving cars are qualitatively safer than human drivers, despite industry claims to the contrary.

In my opinion, this is not the appropriate time to introduce self-driving cars into mainstream society. We need considerably more time to develop our technological infrastructure to a point where autonomous vehicles could achieve a smooth market transition with minimal disruption to society. While self-driving cars undoubtedly possess the potential to be incredibly beneficial to society in the future, that future is not now.

If we were to begin deploying self-driving cars in today's society, the transition would fail in numerous critical ways. First, there is a severe affordability crisis. Currently, self-driving cars cost well over \$100,000 and require specialized training to operate safely—expenses that place them far beyond the reach of average Americans. Consequently, only a small, wealthy percentage of the population could initially afford these vehicles. This creates a fundamental problem: self-driving cars require near-universal adoption to function effectively within our transportation infrastructure. If only the richest members of society can purchase them, widespread adoption becomes virtually impossible, and the technology cannot fulfill its intended purpose of revolutionizing transportation for all Americans.

Security concerns present another formidable obstacle. Hackers and cybercriminals would undoubtedly be intensely interested in targeting self-driving car technology, viewing autonomous vehicles as valuable sources of personal data and potential vectors for attack. Specialized computer systems embedded in these vehicles could be compromised to collect sensitive information about drivers' locations, habits, and personal details. The security infrastructure required to protect millions of autonomous vehicles from coordinated cyberattacks has not yet been adequately developed.

Perhaps the most devastating consequence of premature self-driving car adoption would be the catastrophic job losses across multiple industries. The transportation sector, which employs millions of Americans as taxi drivers, truck drivers, delivery drivers, and rideshare operators, would face near-complete obsolescence. Simultaneously, the entire gasoline industry would suffer severe damage if self-driving cars transition to electric power, eliminating demand for petroleum products and destabilizing an industry that supports hundreds of thousands of jobs. Additionally, driver's education courses would no longer be necessary, eliminating employment for driving instructors. Personal injury lawyers would experience significantly reduced earnings as accident rates plummeted, affecting an entire legal profession.

These job losses would create profound economic and social consequences. Millions of workers would face unemployment without adequate retraining programs or social safety nets. Communities dependent on transportation-related industries would experience economic collapse. The resulting unemployment would strain government resources, reduce consumer spending, and potentially trigger broader economic recession.

For these interconnected reasons, I maintain two fundamental conclusions: First, it is simply not feasible to convince the general public to adopt self-driving cars under current conditions, given their prohibitive cost and unproven safety advantages. Second, in the unlikely scenario that universal adoption somehow occurred, the resulting economic damage from massive job displacement would be severely detrimental to both individual lives and the broader American economy. Self-driving cars represent a promising future technology, but that future must wait until we have developed adequate supporting infrastructure, proven safety records, affordable pricing, and comprehensive economic transition plans.
\end{tcolorbox}

\begin{tcolorbox}[title={gpt-5-mini --- Expansion Edit}, gptbox]
The advent of self-driving cars represents a transformative technological development with the potential to reshape transportation, urban planning, and daily life. Proponents argue that autonomous vehicles (AVs) will reduce accidents, increase mobility for those who cannot drive, and improve traffic flow. While these claims are promising, I contend that broad adoption of fully autonomous vehicles in the United States is premature. Significant gaps remain in technology reliability, safety validation, economic feasibility, workforce impact, and regulatory and ethical frameworks. For these reasons, a cautious, staged approach that emphasizes further research, robust regulation, and mitigation of social harms is preferable to rapid deployment.

Potential benefits are frequently highlighted in discussions about AVs and deserve careful consideration. Human error is a factor in a large majority of traffic collisions; in principle, computer systems that do not suffer from distraction, fatigue, or impairment could reduce accident rates. Autonomous systems can also be programmed to optimize routing and spacing, potentially reducing congestion and fuel consumption. Additionally, properly designed AVs could expand mobility for elderly, disabled, and otherwise transportation-disadvantaged populations. In controlled environments—such as limited-access highways, dedicated lanes, or geofenced urban districts—self-driving technology has demonstrated meaningful progress and could arguably deliver early safety and efficiency gains.

Despite these promising features, several substantial obstacles make immediate, widespread adoption inadvisable. First, the technology is not yet robust across the wide range of real-world conditions. Many AV systems rely heavily on sensors—LiDAR, radar, cameras—and on software trained with vast datasets. Adverse weather (heavy rain, snow, dense fog), unusual road configurations, poorly marked lanes, and unpredictable human behavior (pedestrians, cyclists, drivers violating rules) continue to challenge perception and decision-making algorithms. These edge cases are rare but consequential; solving them requires either near-perfect prediction or conservative behaviors that could impair traffic flow and usability.

Second, claims about safety remain unproven at scale. Some companies report favorable safety metrics for limited operations, but these reports are often based on test miles driven under constrained conditions and are not directly comparable to the tens of billions of human-driven miles annually. Demonstrating that AVs are qualitatively safer across the entire driving environment requires transparent, peer-reviewed evidence, independent oversight, and long-term post-deployment monitoring. Until such standards and data exist, blanket assertions of superior safety are premature.

Third, economic and equity considerations complicate a wholesale transition. Current AV prototypes and early production models are expensive; even with anticipated cost reductions, initial deployments will likely be accessible primarily to affluent consumers or to well-funded fleets. If AV adoption remains uneven, a mixed road environment—where human drivers and autonomous vehicles share the road—could create complex interactions that degrade safety and efficiency. Moreover, widespread automation in transportation threatens substantial job displacement: long-haul truck drivers, taxi and rideshare drivers, delivery couriers, and related service workers could face significant unemployment without adequate retraining programs and social supports. The ripple effects would extend to industries that currently rely on human-driven vehicle patterns, from fuel suppliers (if electrification of AVs accelerates fuel demand shifts) to legal and insurance services scaled around human-driver liability.

Fourth, cybersecurity and privacy pose real, underappreciated risks. Autonomous vehicles will be networked devices—a convergence of sensors, onboard compute, and cloud services. Hackers exploiting vulnerabilities could endanger passengers or use vehicles as a vector for broader attacks. Additionally, the detailed telemetry and location data generated by AVs raise privacy concerns; without strong legal protections and data governance, manufacturers and third parties could collect and monetize sensitive personal information.

Finally, regulatory, legal, and ethical frameworks remain unsettled. Questions about liability in collisions involving autonomous systems, standards for validation and certification, and ethical decision-making in unavoidable accident scenarios lack consensus. Fragmented federal-state regulatory landscapes could further complicate deployment and create legal uncertainty for manufacturers and consumers alike.

A fair counterargument is that continued delay also imposes costs. If AVs can substantially reduce fatalities, expand mobility, and lower long-term transportation costs, earlier adoption could save lives and provide social benefits sooner. Pilots and phased rollouts, advocates argue, can capture benefits while allowing iterative improvement.

This is a persuasive point, and it supports a middle path: aggressive, targeted pilots combined with rigorous oversight. However, the rebuttal to the push for rapid mass deployment is that pilots must be carefully scoped and transparent, not used to justify premature scaling. Pilots should focus on controlled settings where edge cases are minimized (e.g., dedicated lanes, well-mapped urban zones, commercial hubs) and must include independent safety audits, mandated reporting of incidents, and community impact assessments. Complementary policies should require manufacturers to demonstrate safety equivalence or superiority under independently verifiable metrics before broader approval; fund retraining and income-transition programs for displaced workers; institute strong cybersecurity standards and data-privacy protections; and coordinate federal and state regulations to avoid fragmented rules that could hinder safe adoption.

In conclusion, self-driving cars hold considerable promise, but the transition to a society relying broadly on autonomous vehicles is not yet feasible without unacceptable risks and social costs. A prudent course combines continued technical development, transparent safety validation, phased pilot deployments, and proactive policy measures to mitigate economic and privacy harms. By pausing before full-scale adoption and investing in governance and support systems now, society can increase the likelihood that autonomous vehicles—when widely introduced—will be safe, equitable, and genuinely beneficial.
\end{tcolorbox}

\begin{tcolorbox}[title={Gemini~2.5 Flash --- Expansion Edit}, geminibox]
The advent of self-driving car technology represents a profound leap in automotive engineering, promising to redefine personal transportation. While the potential for innovation is undeniable, a critical examination reveals significant hurdles that question the immediate feasibility of integrating these autonomous vehicles into the fabric of everyday American life. This essay argues that despite their compelling advantages, the current limitations in technology, infrastructure, and societal readiness, coupled with substantial economic and ethical challenges, render widespread adoption premature.

The theoretical benefits of self-driving cars are indeed transformative. Proponents often highlight the promise of enhanced safety, envisioning a future with drastically reduced accidents. Autonomous systems, free from human errors such as distraction, fatigue, or impairment, could theoretically navigate roads with unparalleled precision and consistency. Furthermore, these vehicles hold the potential to optimize traffic flow through vehicle-to-vehicle (V2V) and vehicle-to-infrastructure (V2I) communication, alleviating congestion and significantly reducing commute times. The convenience of not needing a designated driver, coupled with increased mobility for the elderly and disabled, presents a compelling vision for a more accessible and efficient transportation system.

However, the disadvantages and current limitations of autonomous vehicle technology are substantial and often underestimated. Self-driving cars remain highly dependent on ideal environmental conditions; adverse weather such as heavy rain, snow, or dense fog can severely impair the functionality of their crucial sensors, including LIDAR, radar, and cameras, compromising their ability to perceive and respond to their surroundings safely. Moreover, these systems struggle with the nuanced and often unpredictable nature of human behavior, failing to reliably interpret ad-hoc human traffic signals, pedestrian gestures, or the spontaneous actions of other drivers. The risk of software malfunctions or hardware failures also looms large, potentially placing occupants and other road users in unforeseen danger. Crucially, the technology relies on a robust digital infrastructure, including ubiquitous 5G connectivity and highly detailed, constantly updated digital maps, which are not yet fully developed or universally available. Consequently, there is no definitive, long-term real-world evidence to qualitatively demonstrate that self-driving cars are unequivocally safer than human-driven vehicles across all conditions.

Beyond technological readiness, the societal and economic implications of widespread self-driving car adoption present formidable challenges. The initial cost of these advanced vehicles, often exceeding \$100,000, places them far beyond the reach of the average consumer. This prohibitive price point creates a significant barrier to the widespread adoption necessary for many of their touted benefits, such as optimized traffic flow, to materialize fully. Furthermore, security risks are a pressing concern. The sophisticated computer systems that power autonomous vehicles are attractive targets for hackers, who could potentially exploit vulnerabilities to gain remote control, manipulate data, or breach personal privacy by collecting sensitive user information.

Perhaps the most significant societal disruption would be the widespread job displacement across various sectors. If self-driving cars become commonplace, millions of jobs in the transportation industry—including truck drivers, taxi and ride-share operators, and bus drivers—would be rendered obsolete. The economic ripple effect would extend to ancillary industries: gasoline stations would suffer if autonomous fleets are predominantly electric, driving schools would see a drastic reduction in demand, and even personal injury lawyers and insurance adjusters could experience reduced earnings due to a projected decrease in accidents. Such a massive shift would be detrimental to the livelihoods of countless individuals and could inflict substantial harm on the broader economy without comprehensive retraining and social support programs.

While proponents argue that technological advancements will inevitably overcome current limitations, and that the ethical imperative to save lives outweighs economic concerns, this perspective often overlooks the sheer scale and complexity of the necessary transitions. Developing reliable AI for all driving conditions, establishing a robust and secure digital infrastructure, and creating effective societal safety nets for displaced workers are not trivial tasks. These are multi-decade endeavors requiring immense investment and careful planning, not simply a matter of incremental software updates. Rushing deployment without adequately addressing these foundational issues could lead to a chaotic and inequitable transition, exacerbating existing societal inequalities rather than solving them.

In conclusion, while the vision of self-driving cars offers tantalizing prospects for enhanced safety and efficiency, the current state of technology, infrastructure, and societal preparedness renders their widespread integration into American life unfeasible at this juncture. The significant hurdles—ranging from technological limitations in adverse conditions and security vulnerabilities to the prohibitive costs and profound economic disruption—demand a cautious and methodical approach. Until these complex challenges are comprehensively addressed through further technological refinement, robust regulatory frameworks, and thoughtful societal planning, the promise of autonomous vehicles must remain a future aspiration rather than an immediate reality.
\end{tcolorbox}

\clearpage

\subsubsection{Essay 3: Strong Opinion}

\noindent\textit{This essay takes a strong stance for self-driving cars, tying autonomous vehicles to the elimination of the gasoline industry and framing opposition as standing in the way of progress.}

\medskip

\begin{tcolorbox}[title={Original Draft (D1)}, essaybox]
Up with Self-Driving Cars and Down with Gasoline Industry
     Self-driving cars will eliminate many accidents and the gasoline industry which causes too much pollution and whose time has run out. The computer running the car won’t be distracted like humans. The average attention span is only eight seconds so a person might not respond fast enough to a deer in the headlights or a bicyclist going past. Human error causes 94\% of traffic accidents. Most traffic accidents are caused by drunk and distracted driving.
     A self-driving car will prevent many accidents especially ones caused by distracted drivers. Distracted driving is a common cause of traffic accidents. Often people like to talk or text on their cell phones, eat and drink, talk to people in their vehicle, or play with the radio/CD player while driving which is highly dangerous. Children, pets, conversations, and preoccupations such as applying makeup can distract a driver.
    Other causes of accidents such as speeding, drunk or reckless driving, running red lights, running stop signs, unsafe lane changes, wrong-way driving, road rage, and drowsy driving will be eliminated. 
     A self-driving car will probably be electric and electric cars will be better for the environment. Electric cars are more efficient and generate less pollution. An electric car is very quiet and very smooth compared to gas operated cars which are clunky and outdated. Electric cars are cheaper to operate so have a big cost advantage over gas. Electric cars are also more efficient so cost about one-third to one-quarter the cost of gasoline. Maintenance costs are less since electric cars don’t need oil changes. Since electric cars don’t have a tailpipe they will improve the air quality.
     Gas cars pollute the atmosphere so it’s time we eliminate or lessen the gasoline industry so we can breathe healthier air. High levels of chemicals taken in from gasoline exposure like benzene may cause cancer. The inhalation of fumes can lead to confusion, dizziness, headaches and breathing difficulties. Using gasoline is bad since it relies on fossil fuels which are in short supply and dwindling. Therefore it’s time we switched to electric cars.
     The mechanics in gas garages can be trained to work on electric cars instead or for other jobs. When we switched from the horse and buggy to cars we didn’t worry about the people who raised horses losing jobs. We can’t let that stop inevitable progress. People who don’t own cars will still ride the buses and take taxis so those operating the vehicles will still have jobs.
\end{tcolorbox}

\begin{tcolorbox}[title={Human-Edited Baseline (D2)}, humanbox]
Up with Self-Driving Cars and Down with Gasoline Industry

Self-driving cars are beneficial to society since they will eliminate many accidents and the gasoline industry which causes too much pollution and whose time has run out. The computer running the car won’t be distracted like humans. The average attention span is only eight seconds so a person might not respond fast enough to a deer in the headlights or a bicyclist going past. Human error causes 94\% of traffic accidents. Most traffic accidents are caused by drunk and distracted driving.

A self-driving car will prevent many accidents especially ones caused by distracted drivers. Distracted driving is a common cause of traffic accidents. Often people like to talk or text on their cell phones, eat and drink, talk to people in their vehicle, or play with the radio/CD player while driving which is highly dangerous. Children, pets, conversations, and preoccupations such as applying makeup can distract a driver.

Other causes of accidents such as speeding, drunk or reckless driving, running red lights, running stop signs, unsafe lane changes, wrong-way driving, road rage, and drowsy driving will be eliminated.

A self-driving car will probably be electric and electric cars will be better for the environment. Electric cars are more efficient and generate less pollution. An electric car is very quiet and very smooth compared to gas operated cars which are clunky and outdated. Electric cars are cheaper to operate so have a big cost advantage over gas. Electric cars are also more efficient so cost about one-third to one-quarter the cost of gasoline. Maintenance costs are less since electric cars don’t need oil changes. Since electric cars don’t have a tailpipe they will improve the air quality.

Gas cars pollute the atmosphere so it’s time we eliminate or lessen the gasoline industry so we can breathe healthier air. High levels of chemicals taken in from gasoline exposure like benzene may cause cancer. The inhalation of fumes can lead to confusion, dizziness, headaches and breathing difficulties. Using gasoline is bad since it relies on fossil fuels which are in short supply and dwindling. Therefore it’s time we switched to electric cars.

The mechanics in gas garages can be trained to work on electric cars instead or for other jobs. When we switched from the horse and buggy to cars we didn’t worry about the people who raised horses losing jobs. We can’t let that stop inevitable progress. People who don’t own cars will still ride the buses and take taxis so those operating the vehicles will still have jobs.

Although self-driving cars are still in our future, it's inevitable that they are bound to replace our existing cars when the price comes down. Hybrid cars are the first step to electric cars which have been around for over a hundred years. Electric cars are finally entering the mainstream thanks to improved battery technology, faster charging, greater range and lower costs. They are available in virtually every size and style, from compact two-seaters to seven-passenger SUVs. Modern electric cars can travel between 100 and 400 miles on a single charge, depending on type and driving conditions, so range anxiety is still a concern for some.
\end{tcolorbox}

\begin{tcolorbox}[title={Claude Haiku~4.5 --- Expansion Edit}, claudebox]
Self-driving cars represent a transformative technology that will dramatically reduce traffic accidents while simultaneously phasing out the gasoline industry, which has become environmentally unsustainable. The artificial intelligence systems governing autonomous vehicles operate without the cognitive limitations that plague human drivers. The average human attention span has declined to merely eight seconds in our digital age, making it nearly impossible for drivers to react appropriately to unexpected hazards such as a deer suddenly crossing the road or a bicyclist appearing in a blind spot. According to the National Highway Traffic Safety Administration, human error is responsible for approximately 94\% of traffic accidents, with distracted driving and impaired driving being the two most significant contributing factors.

Distracted driving has become an epidemic on modern roadways, creating unnecessary dangers for all travelers. Contemporary drivers face an unprecedented array of distractions competing for their attention. Cell phone usage—whether texting, calling, or browsing social media—remains the most prevalent distraction, but drivers also engage in numerous other risky behaviors including eating and drinking, conversing with passengers, adjusting climate controls and entertainment systems, and applying cosmetics while operating a vehicle. Additionally, internal distractions such as daydreaming, worry about personal matters, and the presence of children or pets in the vehicle further compromise driving ability. Self-driving cars would eliminate these dangers entirely by removing human attention as a factor in vehicle operation.

Beyond distracted driving, autonomous vehicles would prevent accidents caused by numerous other human failings. Speeding, drunk driving, reckless driving, running traffic signals, unsafe lane changes, driving in the wrong direction, road rage incidents, and drowsy driving—all major contributors to traffic fatalities—would become impossible when human drivers are removed from active vehicle control. Autonomous systems maintain consistent speeds, never become intoxicated, and operate continuously without fatigue, making them inherently safer than human drivers.

The transition to self-driving cars will almost certainly coincide with a shift toward electric propulsion, offering profound environmental benefits. Electric vehicles demonstrate superior efficiency compared to traditional gasoline engines, generating significantly less pollution throughout their operational lifetime. Beyond environmental considerations, electric vehicles provide substantial economic advantages. Operating costs are substantially lower, with electric vehicles costing approximately one-third to one-quarter as much to fuel as gasoline-powered vehicles. Maintenance expenses are dramatically reduced since electric motors require no oil changes, spark plug replacements, or transmission fluid servicing. Additionally, electric vehicles operate nearly silently and deliver smooth acceleration, providing a more refined driving experience than the mechanical vibrations and engine noise characteristic of internal combustion vehicles.

The environmental case against gasoline-powered vehicles is compelling and urgent. Gasoline combustion releases numerous harmful chemicals into the atmosphere, including benzene, a known carcinogen that has been linked to various cancers with prolonged exposure. The inhalation of gasoline fumes causes acute health effects including confusion, dizziness, severe headaches, and respiratory difficulties. Beyond immediate health concerns, gasoline dependency perpetuates reliance on fossil fuels, finite resources that are rapidly depleting and becoming increasingly difficult and expensive to extract. The transition to electric vehicles powered by renewable energy sources represents not merely an improvement but an environmental imperative.

Concerns about employment displacement in the automotive service industry should not impede this necessary technological advancement. Mechanics currently employed in gasoline vehicle repair facilities can transition to maintaining and servicing electric vehicles, which require different but equally skilled technical expertise. Alternatively, these skilled workers can retrain for other positions in expanding sectors of the economy. History demonstrates that technological transitions inevitably create new opportunities; when society transitioned from horse-drawn carriages to automobiles, we did not halt progress due to concerns about stable workers and horse breeders. Similarly, workers in transportation sectors—whether bus operators, taxi drivers, or ride-sharing services—will continue to find employment as autonomous vehicle technology is implemented, as these vehicles will still require operators, maintenance personnel, and fleet management professionals. Progress cannot be indefinitely delayed by legitimate but ultimately manageable workforce concerns.
\end{tcolorbox}

\begin{tcolorbox}[title={gpt-5-mini --- Expansion Edit}, gptbox]
Up with Self-Driving Cars and Down with the Gasoline Industry

I strongly support the transition to self-driving electric vehicles because they promise to drastically reduce traffic injuries and fatalities, improve air quality, and accelerate society’s move away from a finite, polluting fossil-fuel economy. While legitimate concerns exist about employment disruption, cybersecurity, and ethical decision-making by machines, these risks can be managed through regulation, retraining programs, and robust technological safeguards. On balance, the public benefits of autonomous electric vehicles outweigh their drawbacks.

Self-driving cars will sharply reduce the vast majority of traffic accidents that result from human error. Human lapses—distraction, impairment, fatigue, poor judgment—are responsible for roughly 94\% of traffic crashes, a striking figure reported by safety researchers. Unlike human drivers, an autonomous driving system does not suffer from inattention, impaired reaction time, or emotional distraction. It can monitor multiple sensors continuously, detect hazards earlier, and react more consistently than people do. For example, advanced lidar, radar, and camera fusion can detect a bicyclist emerging from between parked cars in low light and apply controlled braking well before a distracted driver would perceive the danger. Because reaction time and decision consistency are core contributors to crash risk, the predictable, round-the-clock performance of well-designed autonomous systems should prevent many collisions now caused by texting, talking, drowsiness, drink-driving, and road rage.

Beyond distracted driving, self-driving vehicles can reduce accidents caused by speeding, improper lane changes, running red lights, and other risky behaviors. Automated systems can be programmed to obey speed limits, maintain safe following distances, and execute lane changes only when sensors confirm it is safe. Intersection safety can be improved by vehicle-to-infrastructure communication that coordinates traffic signal timing with vehicle approaches, reducing the likelihood of dangerous red-light running. Collectively, these improvements point to large potential reductions in property damage, injury, and loss of life.

Most self-driving vehicles under development are also electric, and electrification compounds the environmental benefits. Battery-electric drivetrains are more energy-efficient than internal combustion engines, converting a higher percentage of stored energy into motion. Electric vehicles (EVs) have no tailpipe emissions, which immediately reduces local pollutants—nitrogen oxides, particulate matter, and volatile organic compounds—that contribute to respiratory and cardiovascular disease. Over their lifecycle, EVs powered increasingly by low-carbon grids produce fewer greenhouse-gas emissions than gasoline cars, helping address climate change. In addition, EVs have lower operating and maintenance costs: fewer moving parts, no oil changes, and regenerative braking that reduces wear on brake systems. Reduced noise pollution from quieter electric motors also improves urban livability.

The environmental and safety gains argue for reducing reliance on gasoline and the broader fossil-fuel infrastructure that sustains it. Gasoline combustion emits known carcinogens and pollutants—benzene among them—that pose public-health risks, and continued dependence on fossil fuels perpetuates supply vulnerabilities and geopolitical exposure. Transitioning to electric, autonomous fleets contributes to cleaner air, fewer pollution-related illnesses, and a more resilient, low-carbon energy system.

Common objections should be addressed rather than ignored. Critics worry about job losses in driving and automotive repair sectors. It is true that automation historically displaces some jobs; however, it also creates new opportunities. Mechanics can be retrained to service electric drivetrains and the increasingly software-defined subsystems in modern vehicles. Vehicle manufacturing, battery production, charging infrastructure deployment, and fleet operations will generate employment. Policy can and should smooth this transition through workforce development grants, apprenticeships, and targeted support for communities disproportionately affected by industrial change. The historical shift from horse-drawn transportation to motor vehicles did disrupt livelihoods, yet society adapted and new industries emerged—this precedent suggests mitigation is possible.

Safety concerns extend to system failures, edge-case scenarios, and malicious hacking. No technology is risk-free, but risk can be minimized. Rigorous testing standards, independent certification, redundant hardware architectures, and transparent incident reporting are essential. Cybersecurity protocols—encryption, intrusion detection, frequent software updates, and hardware isolation of critical control systems—can reduce the threat of remote takeover. Ethical questions, such as how a vehicle should behave in unavoidable crash scenarios, require public deliberation and clear regulatory guidance so that manufacturers implement systems aligned with societal values rather than opaque proprietary choices.

Finally, questions of accessibility and equity must guide deployment. Self-driving electric vehicles can expand mobility for older adults, people with disabilities, and non-drivers, reducing dependence on private car ownership and improving access to jobs, education, and healthcare. Public policy should ensure that advances in autonomous mobility are not restricted to affluent users but benefit underserved communities—through subsidized transit-on-demand services, shared autonomous vehicle pilots in transit deserts, and equitable charging infrastructure placement.

In conclusion, self-driving electric vehicles represent a transformative opportunity: they can dramatically reduce crashes caused by predictable human failings, cut local air pollution, and lessen dependence on finite fossil fuels. While challenges around employment, safety, cybersecurity, and fairness require proactive policy, investment, and oversight, these are solvable problems. The balance of evidence and moral imperative favor embracing autonomous electrification as a central component of safer, cleaner, and more inclusive transportation systems.
\end{tcolorbox}

\begin{tcolorbox}[title={Gemini~2.5 Flash --- Expansion Edit}, geminibox]
Up with Self-Driving Cars and Down with the Gasoline Industry

The advent of self-driving cars represents a pivotal moment in human history, promising not only a dramatic reduction in traffic fatalities but also the critical displacement of the environmentally destructive gasoline industry. I firmly advocate for the widespread adoption of autonomous vehicles and the swift transition away from fossil fuels, believing that this technological evolution will usher in an era of unprecedented safety, environmental health, and economic efficiency. The current paradigm of human-operated, gasoline-powered vehicles is inherently flawed, marked by preventable accidents and unsustainable pollution, making the shift to intelligent, electric mobility an imperative for a healthier, safer future.

A primary driver for embracing self-driving technology is its profound potential to eliminate human error, which accounts for a staggering 94\% of all traffic accidents. Unlike human drivers, who are susceptible to distractions, fatigue, and emotional volatility, the advanced computer systems governing autonomous vehicles operate with unwavering focus and precision. The human attention span, often cited as merely eight seconds, is a stark contrast to the constant vigilance of an AI, which can instantaneously process vast amounts of data from multiple sensors to react to sudden hazards like a deer darting into headlights or a cyclist appearing unexpectedly. This consistent, data-driven responsiveness will virtually eradicate accidents caused by the inherent fallibility of human perception and reaction.

Self-driving cars are uniquely positioned to prevent the myriad accidents stemming from distracted driving, a pervasive and deadly issue in contemporary society. While humans frequently succumb to the temptation of talking or texting on cell phones, eating, engaging in animated conversations with passengers, or adjusting entertainment systems, autonomous vehicles remain impervious to such diversions. The subtle yet dangerous allure of applying makeup, tending to children or pets, or simply getting lost in thought—all common human preoccupations—will cease to be factors in vehicle operation. By removing the human element, self-driving cars guarantee a level of undivided attention and situational awareness that no human driver, no matter how conscientious, can consistently maintain.

Beyond mere distraction, self-driving cars promise to eliminate a host of other dangerous driving behaviors that contribute to the current epidemic of traffic accidents. Speeding, often born of impatience or recklessness, will be mitigated by vehicles programmed to adhere strictly to legal limits and adapt to real-time traffic conditions. Drunk or impaired driving, a leading cause of fatalities, becomes obsolete when the "driver" is an AI incapable of intoxication. Reckless maneuvers, running red lights and stop signs, unsafe lane changes, wrong-way driving, and road rage—all manifestations of human judgment errors or emotional impulses—will be systematically eradicated. Even drowsy driving, a silent killer on highways, will be a concern of the past, as autonomous systems are perpetually alert and unburdened by fatigue. This comprehensive elimination of dangerous human behaviors will transform our roads into far safer corridors.

The environmental benefits of self-driving cars are intrinsically linked to their inevitable electric propulsion systems. Electric vehicles (EVs) are demonstrably more efficient and generate significantly less pollution than their gasoline counterparts. Their superior energy conversion efficiency, coupled with the absence of a tailpipe, means they emit no greenhouse gases or harmful particulate matter directly into the atmosphere, leading to a substantial improvement in urban air quality. Beyond their environmental advantages, electric cars offer a quieter, smoother, and more refined driving experience compared to the often clunky and outdated mechanics of internal combustion engines. Economically, EVs present a compelling case: they are cheaper to operate, costing roughly one-third to one-quarter the price of gasoline for equivalent mileage, and require substantially less maintenance due to fewer moving parts, eliminating the need for oil changes and complex exhaust system repairs.

The transition to self-driving electric vehicles necessitates a decisive reduction in, or even elimination of, the gasoline industry, which has long been a major contributor to atmospheric pollution and public health crises. The combustion of gasoline releases a cocktail of harmful chemicals, including benzene, carbon monoxide, nitrogen oxides, and particulate matter, which contribute to smog formation, acid rain, and global warming. Chronic exposure to these pollutants has been unequivocally linked to severe health issues, including various cancers, respiratory diseases like asthma and bronchitis, cardiovascular problems, and neurological damage, manifesting as confusion, dizziness, and persistent headaches. Furthermore, the gasoline industry's reliance on finite fossil fuels is unsustainable, leading to geopolitical instability, resource depletion, and environmental degradation from extraction to refining. Phasing out gasoline is not merely an environmental preference but a crucial step towards safeguarding public health and ensuring long-term planetary sustainability.

While some express legitimate concerns about job displacement within the traditional automotive and fossil fuel industries, this apprehension, though understandable, should not impede inevitable progress. Critics often point to mechanics and gas station attendants as professions at risk. However, history provides a powerful precedent: when society transitioned from horse-drawn carriages to automobiles, the blacksmiths and stable hands of that era adapted, retrained, or found new opportunities. Today, mechanics can be retrained to specialize in electric vehicle maintenance, battery technology, and autonomous system diagnostics—fields that are rapidly expanding and demanding skilled labor. The self-driving and electric vehicle revolution will create a wealth of new jobs in software development, artificial intelligence engineering, data analysis, cybersecurity, charging infrastructure installation and maintenance, and advanced manufacturing. Moreover, the need for human operators in public transport, logistics, and specialized delivery services will persist, albeit potentially in different forms. Progress has always brought disruption, but it has also consistently led to the creation of new industries, new job categories, and an overall improvement in living standards. To halt innovation for fear of change is to deny society the immense benefits that self-driving electric cars promise.
\end{tcolorbox}

\clearpage